# *Using Machine Learning to Discover Parsimonious and Physically-Interpretable Representations of Catchment-Scale Rainfall-Runoff Dynamics*

Yuan-Heng Wang[1,2] and Hoshin V. Gupta[1]

[1] Department of Hydrology and Atmospheric Science, The University of Arizona, Tucson, AZ
[2] Earth and Environmental Science Area, Lawrence Berkeley National Laboratory, Berkeley, CA

Corresponding Author: yhwang0730@gmail.com | YuanHengWang@lbl.gov | yhwang0730@arizona.edu

## Abstract

Due largely to challenges associated with physical interpretability of machine learning (ML) methods, and because model interpretability is key to credibility in management applications, many scientists and practitioners are hesitant to discard traditional physical-conceptual (PC) modeling approaches despite their poorer predictive performance. Here, we examine how to develop parsimonious minimally-optimal representations that can facilitate better insight regarding system functioning. The term "*minimally-optimal*" indicates that the desired outcome can be achieved with the smallest possible effort and resources, while "*parsimony*" is widely held to support understanding. Accordingly, we suggest that ML-based modeling should use computational units that are inherently physically-interpretable, and explore how generic network architectures comprised of Mass-Conserving-Perceptron can be used to model dynamical systems in a physically-interpretable manner.

In the context of spatially-lumped catchment-scale modeling, we find that both physical interpretability and good predictive performance can be achieved using a "*distributed-state*" network with context-dependent gating and "*information sharing*" across nodes. The distributed-state mechanism ensures a sufficient number of temporally-evolving properties of system storage while information-sharing ensures proper synchronization of such properties. The results indicate that MCP-based ML models with only a few layers (up to two) and relativity few physical flow pathways (up to three) can play a significant role in ML-based streamflow modelling.

## Plain Language Summary

To be credible during decision-making, the predictive performance of *"purely data-based models"* is not sufficient. Models must be interpretable in a physical-conceptual sense. Consequently, scientists and practitioners often hesitate to adopt models developed via machine learning (ML). To better support scientific understanding, there is a need to create representations that provide good predictive performance while providing insight into system functioning. We argue, therefore, that ML-based modeling should be based in the use of computational units that are easy to interpret, and pursue such an approach using the Mass-Conserving-Perceptron (MCP) as the fundamental computational unit. Doing so enables us to explore several issues related to the development of dynamical models for catchment-scale Rainfall-Runoff (RR) systems. We show that excellent predictive performance can be achieved using a *"distributed-state"* model with "*context-dependent*" gating and *"information sharing"* across nodes. Our results suggest that MCP based modeling can play a significant role in geoscientific investigation.

## Key Points

- The mass-conserving-perceptron (MCP) supports development of minimally-optimal representations of catchment-scale rainfall-runoff dynamics.
- MCP-based models are interpretable by design and can therefore provide insight into system functioning.
- MCP-based modeling has the potential to facilitate scientific investigation in support of improved geoscientific understanding.

# 1. Introduction, Background and Scope

## 1.1. Introduction

[1] Modern data-based machine/deep learning (ML/DL) provides a viable alternative to knowledge-based Physical/Conceptual (PC) modeling for simulating catchment-scale Rainfall-Runoff (RR) processes, by focusing primarily on high predictive accuracy (*Sorooshian, 1983; Nearing et al., 2024*), and without incorporating prior knowledge about their internal form and functioning (*Bunge, 1963*). However, commonly used ML models often rely on generic architectures and computational units that are difficult to interpret, many scientists and practitioners remain hesitant to discard traditional PC-based approaches, due to their relative ***interpretability*** — that is, their ability to conceptually link model structure and parameters to underlying physical processes (in contrast to the largely opaque internal representations of purely data-driven ML models), which enhances credibility during decision-making (*Rudin, 2019*).

[2] Attempts are being made to enhance the interpretability of ML-based approaches (see *Althoff et al., 2021; Yang & Chui; 2021; Jiang et al., 2022, 2024*). Knowledge Guided Machine Learning (KGML; *Willard et al., 2022*) provides an approach to doing so (see detailed summary in *Varadharajan et al., 2022*) by transforming "*black box*" models into "*glass (clear) box*" models (*Rai, 2020*). Practical implementations that explicitly address challenges such as interpretability, predictive uncertainty estimation, and ease of use have observed in operational river forecasting systems used by government agencies (*Fleming et al., 2015; Fleming et al., 2021*).

[3] In general, ML-based predictive accuracy is typically achieved at the expense of high network complexity, which contributes to poor interpretability. Given that an important goal of science is the pursuit of parsimonious explanations that support reasoning, there is a critical need to explore "*minimally optimal*" network architectures, defined here as those of the minimal level of complexity necessary for adequate predictive performance, that can provide deeper insight into system functioning. Parsimonious representations can provide more effective support for scientific insight (*Weijs & Ruddell, 2020*), and can be progressively augmented, as necessary and appropriate, during training (*Hsu et al., 1995; Chen & Chang, 2009*). One general strategy is to apply pruning or compression techniques, which remove redundant connections or parameters and compress the network representation, to identify simplified sub-networks that retain performance while reducing complexity (*Tishby & Zaslavsky, 2015; Shwartz-Ziv & Tishby, 2017; Schotthöfer et al., 2022; Zangrando et al., 2023*). A complementary strategy is to begin with simple models and gradually increase complexity as needed to achieve sufficient predictive performance (*Lan et al., 2022; Stanley & Miikkulainen, 2002*). Our view is that the most productive strategy is to begin with computational units that are interpretable by design (*Wang & Gupta, 2024a,b*).

[4] This paper is third in a series that explores how ML can be exploited to achieve better understanding via scientific investigation. In Section 1.2, we provide some foundational background, after which we discuss the goals and scope of the studies reported here (Section 1.3), and the organization of this paper (Section 1.4).

## 1.2. Background

[5] Hydrological model development has a long history (see *Singh, 1988; Clark et al., 2008; Fenicia et al., 2011; Gupta et al., 2012*). Since the 1990's, ANNs used for hydrologic prediction have consistently outperformed PC-based models as evaluated by a variety of metrics (*Daniell, 1991; Halff et al., 1993; Hsu et al., 1995; Smith & Eli, 1995*). Such performance can be attributed to the fact that the "*learned*" representations of ML models can extract relevant information from data (*Xu & Liang, 2021*), while the pre-specified architectures of knowledge-based models must be constrained to be consistent with physical principles/laws (such as mass and energy conservation).

[6] The hydro-geo-scientific community is currently witnessing rapid adoption, development, and application of machine learning and deep learning (ML/DL) methods (*Shen, 2018*). Among these, the *Long Short-Term*

*Memory* (LSTM; *Hochreiter & Schmidhuber, 1997*) network has received considerable attention for its ability to represent the storage effects that characterize soil-moisture and snowpack accumulation and release (*Kratzert et al., 2018, 2019a,b, 2021; Feng et al., 2020; Gauch et al., 2021*; *Mai et al., 2022; Nearing et al., 2022; Arsenault et al., 2023*). This capability makes LSTM models particularly attractive for operational hydrologic forecasting (*Harrigan et al., 2023*; *Sabzipour et al., 2023*), especially at the global scale (*Nearing et al., 2024; Kratzert et al., 2024*). More recently, the transformer neural network (TNN; *Vaswani et al., 2017*) also shows promise for hydrological applications (*Liu et al., 2024b; Koya & Roy, 2024; Li et al., 2025; Liu et al., 2025*). In parallel, substantial progress has been made in integrating convolutional neural network (CNN)-based methods for spatial-pattern extraction with LSTM-based architectures for dynamical prediction (*Anderson & Radić, 2022; Ehsani et al., 2022; Xu et al., 2022; Longyang et al., 2024; Ambika et al., 2025*). Recent attempts have also combined spatiotemporal information using Vision Transformers for time-series prediction (*Tayal et al., 2024*).

[7] The exponential rate of development and widespread access to ML technology (and especially the power of differentiable programming; *Baydin et al., 2018*), has spawned considerable innovation in the development of modeling strategies and tools that combine physical principles with ML (*Shen et al., 2023*). Some major research threads include; 1) The use of data-driven methods to post-process the outputs of PC-based models (*Nearing et al., 2020b; Frame et al., 2021*) thereby improving their predictive accuracy, 2) Adding functional complexity to PC-based models to enhance their behavioral expressivity (i.e., to increase the model's ability to reproduce diverse hydrologic responses and capture nonlinear system behaviors; *Jiang et al., 2020; Tsai et al., 2021; Bennett & Nijssen, 2021; Feng et al., 2022; Höge et al., 2022; Feng et al., 2023*), and 3) Modifying the internal architectures of ML models to improve generalization ability (*Nourani, 2017; Kratzert et al., 2019b*). Examples of the latter include incorporating physical principles such as mass conservation (*Hoedt et al., 2021; Frame et al., 2022*), adding constraints to regularize evapotranspiration loss (which serves to guide the model toward physically consistent estimates and avoid unrealistic values; *Wi & Steinschneider, 2024*), or introducing attention mechanisms (which enable the model to focus on the most relevant variables or time periods when learning from data to identify important system-specific features; *De la Fuente et al., 2024a*). Overall, such investigations have enlivened the community and contributed to rapid advances in the Earth, Environmental, and Geosciences (*Fleming et al., 2021*).

### 1.3. Goals and Scope

[8] As stated above, our preferred approach to model development is to use computational units that are fundamentally (physically-conceptually) interpretable by design, while *also* being sufficiently flexible that they can readily be used to "*learn*" from data. Accordingly, this paper is third in a series that progressively develops and explores different aspects of that approach.

[9] First, in *Wang & Gupta (2024a)* we proposed a generic physically-interpretable computational unit called the *Mass-Conserving-Perceptron* (MCP; ***Figure 1***) that can be used as a component in NNs to directly learn the functional nature of physical processes from available data using off-the-shelf ML technologies, while being regularized to obey conservation principles at the nodal level. In particular, we explored the behavioral expressivity, physical interpretability, and performance achievable by a *single* MCP node (single cell-state model), and demonstrated how different hypothesis regarding the internal functioning of the catchment can be tested by encoding prior knowledge regarding system dynamics into the model.

[10] Next, in *Wang & Gupta (2024b)* we showed how the MCP unit can be used as a building block to construct more complex, but parsimonious, directed-graph node (state variable) and link (flow path) PC-based architectures (*Gupta & Nearing, 2014*) that obey conservation principles and are conceptually-interpretable in the traditional PC sense, while achieving performance comparable to purely data-*based* models. In particular, we showed how ML can be used to effectively combine theory-based prior information with novel information extracted from data, thereby enabling hypothesis testing regarding appropriate system architecture (numbers of dynamical state variables and their flow-path interconnectedness) and an examination of how information

is increased, decreased, or altered during stagewise model development (*Fenicia et al., 2008; Kavetski & Fenicia, 2011*; *Nearing & Gupta, 2015; Gharari et al., 2021*).

[11] Here, we build upon what was previously learned. Specifically, instead of using traditional PC theory to guide the form of the network architecture, where the nodes and links are pre-emptively assigned conceptual meaning, we follow the function approximation paradigm employed by modern ML wherein a generic network architecture is implemented consisting of "*basis function nodes*" (here specified to be MCP units; **Figure 1**) arranged in series and parallel. This approach enables us to explore (in the context of a lumped catchment system) several generic and important modeling issues that are important to understand if model interpretability is our goal.

### *1.3.1 Key Questions*

[12] This study explores several generic yet critical modeling issues that arise when pursuing physically interpretable, mass-conserving neural network formulations. These include the following five topics.

*(1) Design and complexity of the model/network architecture*

[13] Dynamical models (whether ML- or PC-based) consist of directed-graph node and link architectures where the nodes (cell-states) act as memory units, and the links (flow-paths) correspond to processes that transfer fluxes of information (ML models) and/or mass/energy (PC models) into and out of the system and between nodes. In ML-based models, it is typical to use generic network architectures consisting of forward-feeding layers of nodes, with the nodes at each layer being fully-interconnected with those of the next layer, and where the numbers of layers and nodes-per-layer are "*hyper-parameters*" to be determined (along with the network parameters) during model training. In contrast, PC-based models typically consist of a parsimonious and sparsely-connected network architecture that is pre-specified by the model developer, where the nodes represent physical state variables of the system and the links correspond to hypothesized physical processes and pathways via which mass and/or energy flow through the system. In either case, a fundamental question that arises is:

> How many network layers, nodes (cell-states) and links (flow pathways) are potentially necessary/sufficient to accurately model the input-state-output dynamics of a given system?

*(2) Water balance closure at the overall catchment-scale*

[14] A further question arises in the context of model architectures that are intended to be "*mass-conserving*", such as those based on the MCP unit. In such cases, either (i) the overall system input can be fractionally-partitioned so that each portion flows (in parallel) to a different part of the network, and the contributions of all those parts are then summed together to generate the system output (herein called the "*distributed-input*" representation), or (ii) the entire system input is sent (in parallel) to each of the different parts of the network, and the weighted contributions (with weights summing to 1.0) of all those parts are then summed together to generate the system output (herein called the "*distributed-state*" representation), or (iii) some combination of the two. For example, in catchment scale spatially lumped modeling, the *distributed-input* representation might be applicable when fixed fractions of the total precipitation are assumed to fall on different parts of the watershed (impermeable, permeable soil without vegetation cover, permeable soil with vegetation cover, etc.). Conversely, the *distributed-state* representation might be applicable when it is not reasonable to assume that a single number (a catchment-scale mean) is sufficient to represent the statistics of the distribution of soil moisture (or snow) storage within the system, analogous to the *Probability Distributed Store* concept (*Moore, 2007*) encoded into versions of the HyMod (*Boyle et al., 2000*), VIC (*Liang et al., 1994*) and other conceptual hydrological models. In this case, a generic question that arises is:

> Is a "*distributed-input*" or a "*distributed-state*" representation (or some hybrid combination of the two) a more suitable approach to network regularization, in terms of achieving *improved predictive accuracy, enhanced generalization, and physically consistent internal representations*?

[15] It is clear that our current level of physical understanding and available data makes it typically impossible to assume (with any reasonable degree of confidence) that water/energy balance is closed at the overall catchment-scale. So while it is reasonable to assume mass/energy balance closure at each individual node of the system network (i.e., local mass/energy balance), as is done when using the MCP as a node, it might not be reasonable to assume such closure at the overall system (catchment) level (see example in Wang & Gupta, 2024a, where incorporating a mass-relaxation gate into a single-node MCP improved low-flow predictive accuracy). In this case, a generic question that arises is:

> Is there benefit to relaxing mass conservation at the overall network level while maintaining mass-conservation at the nodal level?

*(3) Internal sharing of information between nodes*

[16] It is common in PC-based models for the time-varying strength of the fluxes leaving a given node to depend primarily on the current state of the node. For example the flux rate of water exiting a soil-moisture node via drainage at any time is typically assumed to depend only on the magnitude of water stored within that node, independent of how wet (or dry) the other parts of the system are. In contrast, in the generic LSTM network architecture (see **Text S1** and **Figure 1c** in **Supplementary Material**), the degree of open-or-closedness of the gating mechanisms (nonlinear mechanisms that control the flow of information) at a node (cell-state) can depend on the cell-state values at every other node in the same layer of the network. We will refer to this as "*information-sharing*". Accordingly, a generic question for both ML- and PC-based modeling is:

> How much benefit might there be to allowing each of the nodes to have full access to information regarding the entire distribution of moisture across the system when determining the behaviors of the flux-computational mechanisms at each time step?

*(4) Degree of achievable interpretability*

[17] Regarding the fourth issue, the Data Processing Inequality of Information Theory (*Cover & Thomas, 2006*) makes it clear that no degree of structural regularization to the architecture of a generic neural network can improve its performance on the training data (see discussion in *Nearing & Gupta 2015*), although regularization can help to prevent overfitting and to therefore maintain optimal performance when applied to the independent testing data. Accordingly, any structural regularization that attempts to "*improve the degree of achievable interpretability*" of a model should not typically enable it to have superior predictive performance over a suitably designed generic ML-based model trained on the same data. In other words, properly designed and trained ML-based models can be expected to provide approximate upper-bounds on achievable predictive performance (*Gong et al., 2013; Kumar & Gupta 2020; Nearing et al., 2020a*). Accordingly, two question that arise in the context of attempts to improve the interpretability of ML-based modeling via structural regularization are:

> How much interpretability can be achieved/maintained by a physically-interpretable network based in MCP computational units, while permitting it to pursue the aim of optimal predictive performance?
>
> How does the predictive performance of such MCP-based networks compare to that obtained using pure ML modeling, and/or using conventional PC-based models?

*(5) Strategies for determining optimal network complexity*

[18] Finally, it has become established in the ML literature that there can be considerable benefits, during initial training, to using larger network architectures than are functionally necessary, and then pruning them down to smaller sizes once a reasonable degree of functional performance has been achieved (*Han et al., 2015; Frankle, J. & Carbin, 2018*), thereby achieving better network parsimony and resulting in enhanced computational, and even performance, benefits. Accordingly, an interesting question in the context of MCP-based modeling is:

Are there benefits to training and then pruning such MCP-based machine-learning networks?

[19] In this work, we conduct several experiments that enable us to examine the aforementioned questions. To our knowledge, no similar efforts have been made in this regard. As mentioned previously, we seek to develop relatively parsimonious, and therefore more easily interpretable, ML-based representations of catchment-scale RR processes, using MCP-based network components that are fundamentally interpretable by design, while achieving close-to-optimal levels of predictive performance (as indicated approximately by generic LSTM-based networks; *Nearing et al. (2021); Kratzert et al. (2024)*). Following Occam's Second Razor (*Domingos, 1998*) we consider it sensible to not abandon interpretability in favor of non-understandable complexity.

### 1.4. Organization of the Paper

[20] The paper is organized as follow. Section 2 briefly presents the MCP and the various MCP-based network architectures examined in this study. Section 3 outlines the data and methods used for model development, training, and benchmarking of performance. Experimental results are reported in two parts, in Sections 4 and 5. Part 1 (Section 4) consists of a detailed investigation of several network architectures at a single "*rainfall-dominated*" catchment in the southeastern US. In particular, Section 4.2 explores the value of "*information sharing*" across nodes of the network, a feature that is built-in to the standard LSTM architecture but is not common in PC-based representations of geoscientific (e.g., hydrological) systems. Section 4.3 then presents a benchmark comparison against both the physical-conceptual model architectures introduced in *Wang & Gupta (2024b)* and a purely data-based LSTM network model, and Section 4.4 examines the interpretability of the MCP-based neural networks. Part 2 (Section 5) reports on a preliminary multi-catchment investigation conducted using 513 catchments across the continental US (CONUS), as well as detailed investigation of network interpretability at another rainfall-dominated catchment, but one that is in the western US and experiences a small amount of snow. Section 6 concludes with a summary of our findings, discussion of its implications, and directions for future work.

## 2. Network Architectures Based on the Mass-Conserving Perceptron

### 2.1 The Mass-Conserving-Perceptron (MCP)

[21] Extending upon the linear-reservoir type bucket model (**Figure 1a**), the physically interpretable Mass-Conserving Perceptron (MCP; **Figure 1b**), proposed by *Wang and Gupta (2024a),* is an ML-based computational unit that is isomorphically similar to a single node in a generic gated recurrent neural network (**Figure 1**). As such, it offers a means to incorporate physical constraints into data-driven architectures (see **Figure S1** for the full set of gate options discussed in Wang and Gupta, 2024a). The node represents mass-conservative system dynamics at the nodal level via a discrete-time update equation:

$$X_{t+1} = X_t + U_t - O_t - L_t \qquad (1)$$

whereby the mass state $X_{t+1}$ of the system (node) at time step $t+1$ is computed by adding the mass of input flux $U_t$ that enters the node, and subtracting the masses of output fluxes $O_t$ and $L_t$ that leave the node, during the time interval from $t$ to $t+1$ (**Figure 1b**).

[22] In the context of spatially-lumped catchment-scale RR modeling, $U_t$ can represent the precipitation mass input flux, and $L_t$ and $O_t$ can represent the evapotranspirative and streamflow mass output fluxes from the system control volume represented by the node. The MCP assumes that $O_t$ and $L_t$ depend on the value of the state $X_t$ through the process parameterization equations $O_t = G_t^O \cdot X_t$ and $L_t = G_t^L \cdot X_t$, where $G_t^O$ and $G_t^L$ are context-dependent time-varying "*output*" and "*loss*" conductivity gating functions respectively, so that Eqn (1) can be rewritten as:

$$X_{t+1} = X_t + U_t - G_t^O \cdot X_t - G_t^L \cdot X_t \qquad (2a)$$

$$X_{t+1} = G_t^R \cdot X_t + U_t \qquad (2b)$$

[23] where $G_t^R$ is the "*remember*" gate that represents the fraction of the state $X_t$ that is retained by the system from one time step to the next. Notice that by *context-dependent gating*, we refer to the idea that nodal gate conductivities (e.g., output and loss gates) are dynamically modulated at each time step based on the overall hydrologic state of the system — for example, the spatial distribution of soil moisture or other hydro-meteorological and geophysical variables — rather than being determined solely by local node inputs or fixed parameters.

[24] To ensure physical realism, the time-evolving values of each of these gates ($G_t^O$, $G_t^L$ and $G_t^R$) are constrained to remain both non-negative and less than or equal to 1.0 at all times. In addition, to ensure mass conservation the sum of the three gates must equal 1.0 (i.e., $G_t^R + G_t^O + G_t^L = 1$), reflecting the assumption that all of the state mass at a given time step is either retained, lost, or output. Accordingly, the value of the remember gate can be determined from the other two as $G_t^R = 1 - G_t^O - G_t^L$ which imposes a strict interdependence among the three gating functions.

[25] Now, assuming knowledge of the initial mass state $X_0$, and given the time history of inputs $U_1, \ldots, U_t$, Eqn (2) can be used to sequentially update the state $X_t$ of the system if the time-evolving values of the gating functions $G_t^O$ and $G_t^L$ (and therefore $G_t^R$) are also provided.

[26] Given this nodal architecture, it is clear that the time evolving nature of the gating functions $G_t^O$ and $G_t^L$ determines the dynamical response of the system to inputs. Since the precise forms of these gating functions are not, in general, known a priori, *Wang & Gupta (2024a)* proposed to parameterize these functions using machine learning (ML) architectures so that their functional forms can be learned directly from available data.

[27] In this study, we begin by parameterizing the output and loss gates as $G_t^O = f_{ML}^O(X_t)$ and $G_t^L = f_{ML}^L(PE_t)$ respectively, where $f_{ML}^O(\cdot)$ and $f_{ML}^L(\cdot)$ are implemented as ML architectures that can learn the functional forms of the dependence of $G_t^O$ on the system state $X_t$ and the dependence of $G_t^L$ on potential evapotranspiration $PE_t$. Consistent with physical theory, we assume that these functional forms are monotonic non-decreasing functions, and for simplicity approximate their functional forms as being sigmoid. However more complex functional forms can also be learned by increasing functional flexibility, such as through the use of a multi-layer perceptron, as discussed by *Wang & Gupta (2024a)*.

[28] Specifically, the output and loss gates are initially represented as $G_t^O = \kappa_O \cdot \sigma(S_t^O)$ and $G_t^L = \kappa_L \cdot \sigma(S_t^L)$ respectively, where $S_t^O = a_O + b_O\, x_t$ and $S_t^L = a_L + b_L\, PE_t$, where $\sigma$ can be any appropriate ML activation function (here chosen to be the sigmoid activation function $\sigma(S) = 1/(1 + \exp(-S))$), and $\kappa_O$, $a_O$, $b_O$, $\kappa_L$, $a_L$, and $b_L$ are trainable parameters. In other words, the output gate is assumed to be a function of the cell state at the current time step, and the loss gate is assumed to be a function of the PET at the current time step, with both gates exhibiting a nonlinear, non-decreasing relationship implemented using the sigmoid activation function.

[29] Further, to ensure that $G_t^O$, $G_t^L$ and $G_t^R$ each remain on $[0,1]$ and also that $G_t^O + G_t^L + G_t^R = 1$, we actually set $\kappa_O = \exp(c_O)/\Psi$ and $\kappa_L = \exp(c_L)/\Psi$ where $\Psi = \exp(c_O) + \exp(c_L) + \exp(c_R)$ and instead train on the set of seven parameters $\{a_O, b_O, c_O, a_L, b_L, c_L, c_R\}$ all of which can vary on $(-\infty, +\infty)$. This is equivalent to implementing the SoftMax function on the gates (where SoftMax is a normalized exponential transformation that converts a vector of arbitrary real-valued inputs into a set of non-negative, normalized weights that sum to one; *Bridle, 1990*). Further, we constrain the computed evapotranspirative loss ($L_t = G_t^L \cdot x_t$) to be less than or equal to the potential evapotranspirative loss $D_t$ by replacing the unconstrained loss gate $G_t^L$ described above with the physically-constrained loss gate defined as $G_t^{L^{con}} = G_t^L - ReLU(G_t^L - \frac{D_t}{X_t})$. Namely, $D_t = G_t^{L^{con}} \cdot X_t$. The remember gate needs to update as $G_t^R = 1 - G_t^O - G_t^{L^{con}}$. For more details, please see *Wang & Gupta (2024a)*.

[30] As discussed in our previous work (*Wang & Gupta, 2024a,b*) the MCP computational shares a structurally similar (i.e., isomorphic) form with data-driven LSTM, except that the MCP is designed to (i) conserve mass, (ii) have an additional gating function called the "*loss*" gate, and (iii) be physically-interpretable. For a more

detailed discussion of how MCP-based and LSTM-based computational units (and directed graph networks based on them) are isomorphically related shown in **Figure 1b** and **Figure 1c**.

## 2.2 Desirable Properties of the MCP Unit

[31] Overall, this basic MCP unit has several features that make it suitable for interpretable physical-conceptual modeling of dynamical systems such as are of interest in hydrology (i.e., rainfall-runoff modeling). These include:

1) **Recurrence,** enabling the dynamical evolution of the system state (memory) to be represented. This feature captures the Markovian nature of the system, where the current nodal states summarize all relevant historical information needed to predict future dynamics.
2) **Ability to impose conservation principles at the nodal level**, as constraints on system evolution. *This ensures that the learned system behavior adheres to physically consistent, conservative dynamics—providing a strong regularizing effect compared to unconstrained ML models such as LSTMs.*
3) **Ability to represent and learn the dynamics of unobserved gains/losses of mass from the system**. This feature allows exploration of mass-balance closure by accommodating unobserved inflows or outflows while maintaining local (nodal) conservation, thereby enabling hypothesis testing regarding their causes and dynamics (*Wang & Gupta, 2024a,b*).
4) **Ability to learn the forms of the process parameterization equations** (gating functions) that govern the dynamical behaviors of the system based on context (current and past conditions). This feature allows the model to infer the effective process equations—analogous to physically prescribed flux relationships—through learnable gating functions that capture time-varying conductivities, while remaining consistent with thermodynamic principles such as monotonic flux–force relationships (Wang & Gupta, 2024a).
5) **Ease of model implementation and training via automatic backpropagation**, enabled by off-the-shelf machine learning libraries such as PyTorch in Python (*Paszke et al., 2019*), providing a modern alternative to the manual or global optimization approaches (e.g., *Hsu et al., 1995*) previously used for parameter estimation.

[32] The first and second features, recurrence and nodal-level conservation, reflect the Markovian and conservative nature of PC-based models, in which the nodal cell-states summarize all information required to describe the historical evolution of the system so that its future dynamics can be predicted using only the current states and future inputs. ML-based recurrent neural network models, such as LSTMs, are similarly Markovian in structure, although the system states are not necessarily constrained to be "conservative". Given that physical-systems are generally considered to obey conservation principles, nodal-level conservation imposes a strong regularizing constraint on the behaviors that can be learned by an ML-based model.

[33] The third feature concerns the treatment of mass-balance closure. Although physical hydrology often assumes overall system closure at the catchment level, it is necessary to consider possible unobserved gains or losses of mass while still enforcing conservation at the local (nodal) level. As demonstrated by previous work (*Wang & Gupta, 2024a,b)*, use of the MCP enables the modeler to propose and test different hypotheses regarding the existence, causes, and behavioral dynamics of such unobserved gains/losses.

[34] The fourth feature relates to the specification of the process parameterization equations that govern the fluxes that feed and deplete the nodal cell-states. In PC-based models, these equations are typically prescribed by the modeler based on physical theory. As such they constitute interpretable scientific "*hypotheses*" whose validity must be tested against data. In contrast, the analogous functionality in ML-based models is achieved by layers of nodal "*activation functions*" arranged in parallel and series, exploiting the Kolmogorov neural network existence theorem, which states that a three-layer feed-forward Artificial Neural Network (ANN) with a finite number of nodes in the hidden layer can closely approximate any multivariate function (*Kolmogorov,*

*1957; Hecht-Nielsen, 1987*). In MCP-based networks, as in LSTM-based ones, these process equations are implemented via the dynamic (time-varying conductivity) "*gating*" functions (Section 2.1). As discussed in *Wang & Gupta (2024a),* the precise form of these gating functions can be learned by using sufficiently flexible sub-networks that are constrained to obey thermodynamic principles such as the monotonic relationship of a flux to its driving force.

[35] Finally, the fifth feature makes it possible to conduct all of these investigations while exploiting the full power and rapid developments that are characteristic of modern ML. In particular, modern frameworks such as PyTorch (*Paszke et al., 2019*) provide built-in backpropagation capabilities that greatly simplify the implementation process by automatically handling the optimization step, thereby eliminating the need for users to manually code gradient updates. This advancement has substantially facilitated modeling efforts within the hydrologic science community, providing a practical alternative to the global optimization approaches that were commonly used for parameter estimation in the late 1990s. A well-known example is *Hsu et al.* (1995), where the implementation of an artificial neural network for the Leaf River basin relied on a combination of optimal linear least squares estimation and a multistart version of the simplex nonlinear optimization algorithm.

## 2.3 Basic MCP-Based Network Architectures Investigated in this Study

[36] We use the MCP node (*Wang & Gupta, 2024a*) as the basic computational unit for constructing various interpretable neural networks architectures (**Figure 1b**). While *Wang & Gupta (2024b)* used traditional PC theory to guide the design of network architecture, for instance, the HyMod-Like $MA_5$ architecture is composed of three MCP units with $MC\{O_\sigma L_\sigma^{con}\}$ serving as soil moisture tank and a simplified version $MC\{O_\sigma L_\sigma^{con}\}$ without loss gate being used as surface routing and groundwater tanks—the corresponding single-node case with a complete suite of gates, which can be used to infer how the simplified architectures are constructed, is illustrated in Figure S1. Note that HYMOD is a simple conceptual rainfall–runoff model developed by *Boyle et al. (2000),* where the rainfall–runoff dynamics are represented by two flow paths and three storage tanks, namely the soil-moisture, surface-routing, and subsurface-groundwater tanks. Here, we instead follow the function approximation paradigm employed by modern ML wherein generic network architectures, consisting of nodes arranged in series and parallel, are implemented.

[37] Consistent with *Wang & Gupta (2024a)* the basic MCP node is notated as $MC\{O_\sigma L_\sigma^{con}\}$. The subscript $\sigma$ indicates that the unit uses simple sigmoid activation functions for construction of the output and loss gates ($O_\sigma$ and $L_\sigma$ respectively), while the superscript $con$ indicates that evapotranspirative loss is constrained to not exceed PET. To achieve additional expressive power, we modified the loss gate $L_\sigma^{con}$ to be informed by both PET and the value of the cell-state, and refer to this '*augmented*' unit as $MC\{O_\sigma L_{\sigma+}^{con}\}$, where the + symbol indicates this augmentation. Unless otherwise mentioned, all nodes in the MCP-based networks discussed below are of the $MC\{O_\sigma L_{\sigma+}^{con}\}$ type. In Section 4.1.1, we show that this modification results in a notable performance improvement.

[38] The (interpretable) MCP-based network architectures investigated here consist of one or more layers of augmented MCP nodes arranged in parallel. We use $\ell$ to refer to the layer number ($\ell = 1$ indicates the hidden layer closest to the system inputs) and $N_\ell$ to refer to the number of nodes in hidden layer $\ell$. All nodes in layer $\ell$ are fully-connected to all nodes in the previous layer ($\ell - 1$) in the sense that they receive inputs from all of those nodes, except for the first layer which is only connected to the system inputs.

[39] We notate such architectures as $MN(N_1, N_2, \ldots)$, so that $MN\{3,3,2\}$ indicates a three-layer network of 3 nodes in each of the first and second layers and 2 nodes in the third layer, while the simpler $MN(3)$ indicates a single-layer network of 3 nodes. Note that $MN(1) \equiv MC\{O_\sigma L_{\sigma+}^{con}\}$. Accordingly, the numbers of layers and the numbers of nodes in each layer are network hyperparameters that must be tuned during training.

### *2.3.1 Methods for Distributing the System Inputs across Nodes*

[40] To ensure network-level mass conservation of such an architecture, we need a consistent method for distributing/allocating the system inputs (e.g., precipitation) across the nodes of the first layer, and subsequently for aggregating the system outputs from the final layer to obtain the system output (e.g., streamflow). The two basic kinds of such network-level mass-conserving methods are the "*Distributed-Input (DI)*" and "*Distributed-State (DS)*" mechanisms explained below. Network-level mass conservation can then be relaxed in three different ways, resulting in the "*Distributed-Input Relaxed (DIR)*", "*Distributed-State Relaxed (DSR)*", and "*ML Benchmark (MLB)*" types of networks, also explained below.

[41] Accordingly there are 5 possible types of networks (***Figure 2***), two of which (DI & DS) are designed to ensure conservation of mass at the overall network level, while the other three (DIR, DSR & MLB) relax the requirement for network-level mass conservation while retaining mass conservation at the level of each individual node. To identify the specific type of network-level conservation mechanism, we modify the notation to $MN^{NType}(N_1, N_2, \dots)$ where superscript $NType$ indicates the network type.

### 2.3.2 Distributed-Input Network

[42] In a Distributed-Input (DI) network, the input weights $w_j^{in}$ are constrained to be positive and sum to one ($\sum_j w_j^{in} = 1, w_j^{in} \geq 0$ for all $j$), while the output weights are all set to $1.0$ ($w_k^{out} = 1$ for all $k$). This mechanism distributes different *fractions* of the total input mass along different flow pathways through the network. Because the MCP nodes are mass conserving, setting all the output weights to 'one' ensures that the overall network is also mass conserving. In the RR modeling context, this can be interpreted as allocating different fractions ($P_j = w_j^{in} \cdot P$) of incoming precipitation ($P$) along different flow paths (***Figure 2a***). For example, different fractions of rain may fall on impermeable ground, grassland, and forested portions of the catchment. Accordingly, the nodal cell-states in the first layer represent different degrees of surface wetness (moisture storage) associated with each of these fractional portions of the catchment. By setting the output weights $w_k^{out}$ to $1.0$, we simply aggregate together the outputs (fractional flow components) generated by each flow path. Such networks are notated as $MN^{DI}(\dots)$.

### 2.3.3 Distributed-State Network

[43] Conversely, in a Distributed-State (DS) network, the input distribution weights $w_j^{in}$ are all set to $1.0$ ($w_j^{in} = 1$ for all $j$), while the output aggregation weights are constrained to be positive and sum to one ($\sum_k w_k^{out} = 1, w_k^{out} \geq 0$ for all $k$). This mechanism distributes *all* of the total input mass along each of the different flow pathways through the network. Because the MCP nodes are mass conserving, setting the output weights to *fractional* values (that sum to 'one') ensures that the overall network is also mass conserving. In the RR modeling context, this can be interpreted as a way of modeling the system in a probabilistic manner (***Figure 2b***), where the lumped overall average (single statistic) value of storage is inadequate to describe the state of the system at any point in time and instead a "*distributional description*" of storage is required to better represent the system dynamics. The concept is analogous (but not identical) to that encoded by the Probability Distributed Store component (*Moore, 2007*) used in versions of the HyMod and other conceptual hydrological models (*Liang et al., 1994; Boyle et al., 2000*). Accordingly, this can be interpreted as simulating a discrete distribution of different magnitudes of catchment soil moisture states (where each node represents a different dynamically changing magnitude) that gives rise to a discrete distribution of *flow* magnitudes leaving those nodes, for further processing by subsequent layers. By setting the output aggregation weights $w_k^{out}$ to be positive and sum to one, we integrate the distribution of outputs flow magnitudes generated by the network and thereby conserve mass. Such networks are notated as $MN^{DS}(\dots)$.

### 2.3.4 Distributed-Input-Relaxed Network

[44] The Distributed-Input-Relaxed (DIR) network is similar to the DI type, but with the input distribution weights $w_j^{in}$ *not* constrained to sum to one, while still being required to be positive ($\sum_j w_j^{in} \neq 1, w_j^{in} \geq 0$ for all $j$). This can be interpreted as allowing for bias correction in the magnitude of incoming precipitation, effectively reducing the strictness of the mass conservation constraint at the network level. Specifically, instead

of enforcing the weighted sum of outputs to equal the input at every time step, this formulation allows for slight discrepancies that can be learned as part of the model's optimization—thus enabling more flexibility in capturing systematic input-output biases. Such networks are notated as $MN^{DIR}(\dots)$.

#### 2.3.5 Distributed-State-Relaxed Network

[45] The Distributed-State-Relaxed (DSR) network is similar to the DS type, but with the output aggregation weights $w_k^{out}$ *not* constrained to sum to one, while still being required to be positive ($\sum_k w_k^{out} \neq 1$, $w_k^{out} \geq 0$ for all $k$). Since each node conserves mass streamflow, requiring the associated weights in the output layer to sum to 1 would maintain overall mass balance at the network level. Not enforcing this criterion can be interpreted as allowing for bias correction of the magnitude of outgoing flow, thereby relaxing the requirement for mass conservation at the network level. Such networks are notated as $MN^{DSR}(\dots)$.

#### 2.3.6 ML-Benchmark Network

[46] Finally, in the Machine-Learning-Benchmark (MLB) network, *neither* the input distribution weights $w_j^{in}$ or the output aggregation weights $w_k^{out}$ are constrained to sum to one, but are only required to be positive ($\sum_j w_j^{in} \neq 1$, $w_j^{in} \geq 0$ for all $j$ and $\sum_k w_k^{out} \neq 1$, $w_k^{out} \geq 0$ for all $k$). In addition, learnable bias weights $w_0^{in}$ are included at the input distribution level (so that the input $u_j$ to node $j$ of the first hidden layer is given by $u_j = w_j^{in} \cdot u + w_0^{in}$ where $u$ is the system input) and a learnable bias weight $w_0^{out}$ is included at the output aggregation level (which means that the system output $o$ is given by $o = w_k^{out} \cdot o_k + w_0^{out}$). In such a network, conservation still occurs at each node, but where the conservation requirement is completely relaxed at the overall network-level. Such networks are notated as $MN^{MLB}(\dots)$.

#### 2.3.7 Summary of Network Architectures

[47] Overall, we proposed five different network architectures that all obey mass conservation at both the nodal and network levels, including DI and DS, while relaxing this constraint at the network level for cases such as DIR, DSR, and MLB. The relationships among these five cases are determined by the weights, bias values, and associated constraints at both the input-distribution gate and the output layer. The basic hypotheses for these networks are summarized below:

**DI & DIR Hypotheses:** Assumes the river basin is represented by a set of mass-conserving nodes ($\mathrm{MC}\{\mathrm{O}_\sigma \mathrm{L}_{\sigma+}^{\mathrm{con}}\}$),), where the input precipitation is distributed among these nodes (flow paths) in a mass-conserving manner, given that $\sum_j w_j^{in} = 1$ ($w_j^{in} \geq 0$ for all $j$). In contrast, the DIR case relaxes the mass-conservation constraint at the network level by allowing $\sum_j w_j^{in} \neq 1$ ($w_j^{in} \geq 0$ for all $j$). For both cases, the total output streamflow is computed by summing the contributions from all flow paths, i.e., $w_k^{out} = 1$.

**DS & DSR Hypotheses:** Assumes the mapping from input precipitation to output streamflow can be approximated functionally. Here, all nodes in parallel receive the same input precipitation amount, i.e., $w_j^{in} = 1$. The mass-conserving DS case satisfies $\sum_k w_k^{out} = 1$ ($w_k^{out} \geq 0$ for all $k$), while the DSR case relaxes this network-level constraint with $\sum_k w_k^{out} \neq 1$ ($w_k^{out} \geq 0$ for all $k$).

**MLB Hypothesis:** Combines features of both DI and DS cases by allowing network-level mass relaxation to occur at both the input-distribution gate and the output layers. In other words, either $\sum_j w_j^{in} \neq 1$ or $\sum_k w_k^{out} \neq 1$ (or both) and with additional unconstraint bias terms $w_0^{in}$ and $w_0^{out}$ at both the input and output layers.

### 2.4 Enhancements to the Basic Network Architectures to Enable Info-Sharing

[48] In all of the afore-mentioned MCP-based architectures (Section 2.3 above), each nodal output gate $G_t^O = f_{ML}^O(X_t)$ and loss gate $G_t^L = f_{ML}^L(PET_t, X_t)$ is only provided information regarding its own cell-state $X_t$, and not regarding the cell-states of other nodes in the network (Wang & Gupta, 2024b). In contrast, LSTMs typically provide the gating functions at each node with information regarding the cell-states of all of the nodes in the

same layer; we refer to this as layer-wise "sharing" of cell-state information. Accordingly, the MCP-based architectures describe above can be similarly enhanced to facilitate information sharing among the nodes.

[49] To be clear, we should distinguish between 'nodes' and their 'gates': nodes refer to functional units that store and evolve internal states (e.g., $X_t$), while gates are parametric functions—ranging from a relatively simple nonlinear function such as a sigmoid to a full-blown neural network—that control how these states are updated, released, or dissipated. Given that each MCP node has both an output and a loss gate, we can consider four types of layer-wise cell-state-information-sharing. To identify the specific type of network-level conservation mechanism, we further modify the notation to $MN_{SType}^{NType}(N_1, N_2, \dots)$ where subscript $SType$ indicates the type of information sharing.

1) ***No Sharing:*** The output and loss gates at each node are <u>not</u> provided with information regarding cell-state magnitudes of other nodes. Such networks are notated as $MN_{None}^{NType}(\dots)$.
2) ***Sharing-Augmented Loss Gates:*** The loss gates at each node in a layer are provided with information regarding cell-state magnitudes of other nodes in that layer (along with potential evapotranspiration), while each output gate is only provided with information regarding the magnitude of its own cell-state. Such networks are notated as $MN_{SAL}^{NType}(\dots)$.
3) ***Sharing-Augmented Output Gates:*** The output gates at each node in a layer are provided with information regarding cell-state magnitudes of other nodes in that layer, while each loss gate is only provided with information regarding the magnitude of its own cell-state (along with potential evapotranspiration); Such networks are notated as $MN_{SAO}^{NType}(\dots)$.
4) ***Sharing-Augmented Loss & Output Gates:*** The loss and output gates at each node in a layer are <u>both</u> provided with information regarding cell-state magnitudes of other nodes in that layer. Such networks are notated as $MN_{SALO}^{NType}(\dots)$.

[50] While it is possible, in theory, to allow cell-state information sharing between layers, or other possible variations such as providing access to time-lagged cell state information, such configurations would complicate conceptual interpretation and potentially obscure the connection with the actual dynamical system. In this study, we restrict our scope to only testing cell-state-information-sharing on a layer-wise basis, consistent with the assumptions embedded in the LSTM architecture.

## 2.5 Summary

### 2.5.1 Note on Network Architecture Type tested

[51] With five network types (Section 2.3) and four information-sharing types (Section 2.4) we have a total of $5x4 = 20$ MCP-based network architectures to be tested (Table 1). In all cases, the weights associated with information sharing across nodes of a layer are permitted to be either positive or negative and will typically be close to zero, where a zero value means that the associated information is not being used (is not useful) to determine the value of the gate-state.

### 2.5.2 Note on the Use of the Terms Mass Conservation and Mass Relaxation

[52] It is important to note that "*mass relaxation*", as referred to in the present network architectures, specifically concerns the degree to which the weights in the linear output layer are constrained, thereby determining the extent to which mass adjustment is permitted at the network level. This is distinct from the concept introduced in *Wang & Gupta (2024a),* where the *mass-relaxation gate* operates at the nodal level within a single MCP node to represent potential mass variation under unseen environmental conditions. Hence, the relaxation discussed here pertains to the network-level formulation rather than the nodal-level mechanism.

[53] Whether mass conservation is *maintained* or *relaxed* refers to the internal input–storage–output relationships within the model architecture (both nodal and network levels). During model evaluation (Section 3.2.2), we compute the ratio of the mean simulated to observed streamflow as a diagnostic of mass balance performance. The "*mass balance error*" discussed in the results section is therefore unrelated to the architectural terminology of mass conservation or relaxation used to describe the model design.

# 3. Experimental Methods

## 3.1. Data Used

[54] This study consists of two main parts. In Part 1, we conduct detailed testing of the aforementioned 20 different MCP-based network architectures at a single location, the Leaf River basin in Mississippi. Based on those results, in Part 2 we conduct further testing on a large sample of catchments, representing multiple hydroclimatic regimes across the continental US.

### 3.1.1. Data Used for Part 1 – Detailed Investigation at a Single Catchment

[55] First, building upon our previous work, we explore the aforementioned questions in detail via experiments conducted at a single location, the Leaf River basin in Mississippi. In this part of the study, all of the experiments use the Leaf River data set (compiled by the US National Weather Service), consisting of 40 years (WY 1949-1988) of daily data from the humid, 1944 $km^2$, Leaf River Basin (LRB) located near Collins in southern Mississippi, USA. This dataset consists of cumulative daily values of observed mean areal precipitation ($PP$; mm/day), potential evapotranspiration ($PET$; mm/day), and watershed outlet streamflow (O; mm/day). The Leaf River dataset was chosen because it has been widely used for model development and testing by the hydrological science community since the 1980s (*Sorooshian et al., 1983*).

### 3.1.2. Data Used for Part 2 – Multi-Catchment Investigation

[56] For the second part of this study, guided by the results of Part 1, we conduct a preliminary investigation of the kind of spatially extensive predictive performance achievable by MCP-based networks. For this, we used a total of 513 basins across the CONUS, selected from the US Catchment Attributes and Meteorological for Large-sample Studies (CAMELS-US) dataset (*Addor et al., 2017*) based on the availability of continuous streamflow records from WY 1982 to WY 2008.

[57] In this second part, we also selected two of the 513 catchments for a closer examination of internal architectural dynamics (similar to that reported in part 1), aiming to gain insight into the interpretability and potential limitations of the physical representations learned by the MCP-based model.

[58] For model forcings, we used the Maurer dataset, with potential evapotranspiration (PET) computed using the Penman-Monteith equation (*Allen et al., 1998*), using the open-source library developed by *Ahani & Nadoushani (2023)*. The resulting precipitation and PET time series were used as meteorological inputs to drive the models. Because most of the CAMELS-US catchments have snowpack, we used the 4-km University of Arizona (UA) snow product (*Broxton et al., 2016*) to derive basin-average snow water equivalent (SWE) at the daily timescale. This product has passed rigorous tests and been benchmarked against many other well-established approaches (see *Zeng at al., 2018* and *Wang et al., 2022*) for more information.

## 3.2. Methods used for Model Development and Training

### 3.2.1. Data Splitting

[59] As discussed by *Shen et al. (2022)*, it is important to use only a portion of the available data $\mathcal{D}$ for making decisions about the model representation (choices regarding model structure and parameter values), while retaining a separate portion for testing the validity of those choices. Here, we follow the data-splitting procedure reported in *Wang & Gupta (2024ab)* and adopt the robust data allocation method developed by *Zheng et al., (2022)* that partitions the data ($\mathcal{D}$) to ensure distributional consistency of the observational streamflow records across three subsets of the data to be used for training ($\mathcal{D}_{train}$), selection ($\mathcal{D}_{select}$), and

testing ($\mathcal{D}_{test}$). Note that we replace the commonly used term 'validation' with the word 'selection' to better reflect the specific purpose of each dataset. Detailed information regarding the data splitting methodology is provided in the Supplementary Materials (**Text S2**).

### 3.2.2. Metrics Used for Training and Performance Assessment

[60] The metric used for model training was the *Kling-Gupta Efficiency* ($KGE$; Eqn. 3; *Gupta et al., 2009)*. As in *Wang & Gupta (2024a,b)*, each model architecture was trained 10 times with random initialization of the parameters, from which the one having the highest scaled $KGE$ score ($KGE_{ss}$; Eqn. 4; *Knoben et al., 2019*) computed on the *selection* set was retained. Performance assessment was conducted using $KGE_{ss}$ and the components of $KGE$ (Eqns 5-7):

$$KGE = 1 - \sqrt{(\rho^{KGE} - 1)^2 + (\beta^{KGE} - 1)^2 + (\alpha^{KGE} - 1)^2} \tag{3}$$

$$KGE_{ss} = 1 - \frac{(1-KGE)}{\sqrt{2}} \tag{4}$$

$$\alpha^{KGE} = \frac{\sigma_s}{\sigma_o} \tag{5}$$

$$\beta^{KGE} = \frac{\mu_s}{\mu_o} \tag{6}$$

$$r^{KGE} = \frac{Cov_{so}}{\sigma_s \sigma_o} \tag{7}$$

where $\sigma_s$ and $\sigma_o$ are the standard deviations, and $\mu_s$ and $\mu_o$ are the means, of the corresponding data-period simulated and observed streamflow hydrographs respectively and, similarly, $Cov_{so}$ is the covariance between the simulated and observed values. Note that $KGE$ (and therefore $KGE_{ss}$) is maximized when $\alpha^{KGE}$, $\beta^{KGE}$ and $r^{KGE}$ are all 1.0.

[61] Note that both $KGE$ and $KGE_{ss}$ use exactly the same three components—correlation, variability ratio, and bias ratio. The difference is that $KGE_{ss}$ rescales the distance from the ideal score of 1, so as to share with $NSE$ the desirable property of being zero when the mean observed value is used as the simulation. Although $NSE$ and $KGE$ are fundamentally different criteria, using $KGE_{ss}$ provides a relatively objective benchmark because it simultaneously accounts for linear correlation ($r^{KGE}$), flow variability ($\alpha^{KGE}$), and overall mass balance ($\beta^{KGE}$), thereby offering a more physically-interpretable basis when comparing model performance across different studies. Further, while $\alpha^{KGE}$ and $\beta^{KGE}$ are both optimal at 1.0, their values can be larger or smaller than this optimal value which lead to ambiguity when conducting model comparisons using these metrics. Therefore, we define $\alpha_*^{KGE}$ and $\beta_*^{KGE}$ as shown in Eqns (8-9) to circumvent this problem through allowing 1.0 to be the upper bound value.

$$\alpha_*^{KGE} = 1 - |1 - \alpha^{KGE}| \tag{8}$$

$$\beta_*^{KGE} = 1 - |1 - \beta^{KGE}| \tag{9}$$

[62] Note that the models are trained to maximize $KGE$ over all the data points in the training set, while model selection is based on the best $KGE$ obtained on the selection set when trained using 10 different random seeds. The training, selection, and testing sets are determined using the data-splitting algorithm proposed by *Zheng et al. (2022),* which accounts for the representativeness of flow distributions to ensure that the flow magnitude characteristics are consistent across the three sets, a necessary condition for ML-based model development (see Text S2 for more information). For the purpose of model evaluation, the $KGE$ and other evaluation metrics reported here are recomputed on an annual basis. The reason for this will become apparent when discussing the results.

[63] Although we have primarily used KGE-based metrics and their component scores (*Gupta et al., 2009*) as the foundation for model development and evaluation, we recognize the inherently multi-objective nature of the optimization problem (*Gupta et al., 1998*). Because there is no single perfect criterion, once the experimental setup is sufficiently well-developed it will be prudent to emphasize training and assessment

across multiple metrics (*Gupta et al., 1999; Vrugt, 2024*). Arguably the design of such metrics should be rooted in a theoretical foundation such as information theory (*Vrugt et al., 2024, Hodson et al., 2024*).

#### 3.2.3. Methods Used for Model Training

[64] The methods used for model (network) training were adopted from *Wang & Gupta (2024a,b)* and are fairly standard, so mentioned only briefly here. Each network architecture was trained $10$ times, from different random initializations of the weights, using $KGE$ as the objective function. Results are reported for the model that obtained the highest selection-period $KGE_{ss}$. A three-year "*spin-up*" period was used to initialize the model cell-states and thereby minimize the potential effects of state initialization errors (*De la Fuente et al., 2023*). The gradient-based ADAM optimization algorithm (*Kingma & Ba, 2014*) was used for model training. The training metric and its gradient were computed using the streamflow values/time steps assigned to the training subset. Detailed information regarding how the network parameters were initialized and trained, and the protocol used to specify and adjust learning rates, is provided in the Supplementary Materials (**Text S2**).

### 3.3. Benchmarking of Model Performance

[65] To assess relative performance of the MCP-based networks, we compare against some traditional interpretable and mass-conserving PC-based models (as a lower performance benchmark) and also a data-driven ML benchmark (*Gong et al., 2013*; *Nearing et al., 2020a*) that is not constrained to be interpretable or obey principles of conservation (as an upper performance benchmark). For the latter, we use the LSTM network adapted from code provided by *Kratzert et al. (2019b)*.

## 4. Experimental Results Part 1 – Detailed Investigation at a Single Catchment

[66] We first conducted a detailed investigation of the system identification issues raised in Section 1, using experiments conducted at the Leaf River basin in Mississippi. **Table 1** summarizes the single-layer model architectures, and **Table 2** reports their $KGE_{ss}$ percentile scores, as these experiments were found to be most informative. The corresponding results *for* $\alpha_*^{KGE}$, $\beta_*^{KGE}$, and $r^{KGE}$ are presented in **Tables S1 to S3**. Also, results based on flow separation into five groups for $KGE_{ss}$, $\alpha^{KGE}$, $\beta^{KGE}$ , and $r^{KGE}$ are shown in **Tables S4 to S7**. For completeness we provide the annual root mean square error ($RMSE$) and mean absolute error ($MAE$) at various percentiles in **Table S8** and **Table S9** as additional information to assess model performance. The architectural hypotheses for the multi-layer models are summarized in **Table S10,** respectively. Detailed information on the training setups for both single-layer and multi-layer cases is provided in **Table S11** and **Table S12**. Finally, we provide working examples of the single-node case, the distributed-state single-layer network, and the multi-layer distributed-state network in the supplementary materials (**Text S3**), to illustrate the entire calculation process for streamflow.

### 4.1 Results for MCP-Based Architectures without Information Sharing

[67] To begin, we restrict our attention to a single-hidden-layer network, and compare the five basic architectures (*DI*, *DS*, *DIR*, *DSR*, and *MLB*) described in Section 2.3 – where information-sharing between nodes is not permitted, and where network complexity is varied by progressively increasing the number $N_1$ of nodes in the layer. Recall that the *DI* and *DS* architectures conserve mass at both nodal and network level, while the *DIR*, *DSR*, and *MLB* architectures conserve mass only at the nodal level. We varied $N_1$ from 1 to 5.

#### 4.1.1 Single-Layer MCP-Based Architectures

[68] Results are presented in ***Figures 3a-f*** as distributions (over forty-years) of annual $KGE_{ss}$ performance. The leftmost sub-plot (***Figure 3a***) shows results for the original $MC\{O_\sigma L_\sigma^{con}\}$ (dark grey) and the augmented loss gate $MC\{O_\sigma L_{\sigma+}^{con}\}$ (blue) versions of a single MCP node. Distributional performance of the augmented node, overall, outperforms that of the original. Although median year performance improves only slightly, the interquartile range is significantly reduced, mainly due to performance improvements in the lower-quantile (drier) years. Further, $KGE_{ss}$ performance for the worst (driest) year improves dramatically from $0.3 \rightarrow 0.57$

(yellow dot in ***Figure 3a***). Accordingly, all subsequent network architectures are based on use of the augmented MCP node.

[69] ***Figure 3b*** shows results for the distributed input (DI) case, with number of nodes ($N_1$) varying from 1-5. This case corresponds to the situation where an adequate description of the dynamics by which precipitation is partitioned into storage, evapotranspirative loss, and streamflow output requires that different *fractions* of incoming precipitation be routed along different flow paths. Increasing $N_1$ corresponds to increasing numbers of nodes in parallel, corresponding to different fractional distributions of precipitation to different flow paths. In the DI case, increasing the number of nodes does not improve overall performance, and tends to degrade performance during the worst years.

[70] ***Figure 3c*** shows results for the distributed state (DS) case, which corresponds to the situation where an adequate description of the soil moisture state of the system requires more than one summary statistic (the values of the cell-states). In this case, the total incoming precipitation provided to each node of the hidden layer is the same. Note that the DI and DS cases with only one node ($N_1 = 1$; colored blue) are equivalent to the case of a single augmented MCP node. In the DS case, increasing the number of nodes does not improve overall performance, but leads to better performance during the worst years.

[71] ***Figures 3d&e*** correspond to the above-mentioned two cases, but where mass-conservation is relaxed at the network level. ***Figure 3d*** shows results for the distributed input relaxed (DIR) case, where total input mass can be adjusted, while ***Figure 3e*** shows results for the distributed state relaxed (DSR) case, where total output mass can be adjusted. Finally ***Figure 3f*** corresponds to the MLB case where the network-level constraint on mass-conservation is relaxed both at the input layer and the output layer. Among them, in the DIR case, increasing the number of nodes does not improve overall performance, nor does it enhance the model behavior during the worst years.

[72] For all of these subplots (***Figures 3b-f***), the dashed blue line indicates median performance of a single (augmented) MCP node, while the dashed red line indicates best median performance over all the architectures being compared at this stage. As can be seen, while all of the networks with more than one node obtain better median performance (above the blue dashed line), that improvement is not large. For the DI case (***Figures 3b***), no median improvement is obtained beyond two nodes, whereas for the DS case there are fewer outlier years (indicated by the dark red $+$ symbols) and worst year performance (indicated by the yellow-filled circle symbols) is significantly improved with increasing $N_1$.

[73] In contrast, for the DIR case (where bias correction of the input is permitted), we see considerable improvement in performance for the drier years in the catchment history. The overall distribution boxplot for the DIR case (***Figures 3d***) is narrower than that for the DI case (***Figures 3b***). For the driest year (identified based on the annual flow peak value), the median $KGE_{ss}$ ranges between $0.77$ and $0.79$ for the DI case, but increases to 0.82–0.84 for the DIR case when the number of nodes varies between $2$ and $5$. Similarly, for the driest year, identified based on the mean flow volume, the most likely value of $KGE_{ss}$ increases from $0.62$ to $0.69$ for cases with 2 to 5 nodes.

[74] And for the DSR case the performance distributions become less skewed (fewer outlier years, indicated by the dark red $+$ symbols) than the DIR and worst year performance (indicated by the yellow-filled circle symbols) is overall improved compared to DIR cases. Notably, worst year performance generally improves as the number of nodes is increased. Overall, these dry-year improvements persist into the MLB case, where both the input and output can be bias corrected, and best median performance is obtained by the MLB case when the number of nodes $N_1 = 3$ to $5$. Note, however, that removing the input and output mass conservation constraints on the single MCP node (case $MN_{None}^{MLB}(1)$) results in a considerable deterioration in performance.

[75] Overall, these results suggest that the distributed-state (DS) architecture should be preferred over the distributed-input (DI) architecture, but that network-level mass relaxation at both input and output layers (MLB) is beneficial to overall model performance. Of course, when adopting the MLB architecture, the ability to interpret what is happening is diminished, and the machine-learning benchmark can be interpreted as

incorporating both input and output bias corrections along with a hybrid "*distributed-input-state*" architecture in which the distinction between these notions is made fuzzy. Arguably, this might make sense since it simultaneously allows for both spatial distribution (and bias correction) of input mass and a distributional notion for the state of moisture in the system.

[76] ***Figure 4a*** shows the best cases from each of these architectures selected based on median year, below median year, and worst year performance, compared with the baseline single-augmented-node case (blue). Overall, these cases show some improvement compared to the baseline case. However, drawing upon notions of parsimony (minimum description length) to impose an inductive bias on model selection, we might select the MLB architecture with three nodes ($MN_{None}^{MLB}(3)$; the case shown on the far right) as displaying the "best" overall distributional $KGE_{ss}$ performance, with $KGE_{ss}^{min} = 0.67$ and $KGE_{ss}^{50\%} = 0.87$.

[77] ***Figures 4b-d*** show dry, medium and wet year (note the differing y-axis scaling) hydrograph comparisons against observations (red circles) for the DS case with 5 nodes ($MN_{None}^{DS}(5)$; orange lines) and MLB case with 3 nodes ($MN_{None}^{MLB}(3)$; yellow lines), selected for their better performance under dry year conditions (most ML approaches tend to demonstrate robust performance under wet year conditions). The selection of dry, median, and wet years is determined based on the results of annual flow peak, with the discussion further supported by the selection of years based on annual mean flow volume, since the two selection criteria yield a similar ranking of the corresponding years. The dry and wet years are defined as the years with the lowest and highest peak flow values, respectively, while the median year is the one that ranks 20th in peak flow among all 40 years. Note that both models (and particularly $MN_{None}^{MLB}(3)$) demonstrate improved ability to capture high flow peaks compared to the baseline single-node model (blue lines), while the DS model ($MN_{None}^{DS}(5)$) tends to perform better under low-flow conditions, such as during the recession period from January to February of 1952 (see also log-y-axis plots shown in ***Figure S2*** in the supplementary materials). These results suggest that future model development may benefit from a more flexible combination of the two strategies. Specifically, the distributed-state (DS) representation contributes to better simulation of low flows by improving the representation of subsurface storage and recession processes, whereas the mass-relaxation (MLB) formulation—with its capacity to adjust input precipitation and output streamflow—enhances the model's responsiveness to high-flow events. Integrating these two mechanisms could therefore yield a model that maintains mass consistency while adapting dynamically to different wetness regimes. This hypothesis is also physically plausible, as precipitation measurement and input data errors typically increase with rainfall intensity, implying that a moderate degree of mass relaxation could compensate for such uncertainties during wet years.

### 4.1.2 Multi-Layer MCP-Based Architectures

[78] Results from the previous section suggest that the distributed-state (DS) approach is better suited than the distributed-input (DI) approach for representing the hydrologic system of the Leaf River catchment. For this next part of the study, we therefore only pursued the DS approach (input weights all equal to 1.0) when investigating the performance of a multi-layer network.

[79] One reason for not pursuing the DI approach further is that adding more layers could be considered analogous to creating a more complex representation of flow routing, whereas our experience with the Leaf River Basin, supported by decades of model development and experimentation (*Boyle, 2000*), suggests that a fairly simple representation of routing is sufficient. In contrast, the DS approach aligns better with the idea of learning a more complex representation for functional approximation. For now, we choose to leave further exploration of such complexities for future work.

[80] In pursuing the DS approach, we next considered three possible configurations for the linear output layer – the $MN_{None}^{DS}$ case where the (positive valued) output weights sum to 1.0, the $MN_{None}^{DSR}$ case where the (positive valued) output weights are not constrained to sum to 1.0, and the $MN_{None}^{DS-MLB}$ case where the output weights are neither constrained to be positive or sum to 1.0, and where an additional bias term is included in both the transformation and output layers. In theory, the added flexibility permitted by the latter architecture

should increase the learning capability of the network and allow it to better capture the complex dynamics of the system. For each of these cases, we examined 36 networks by varying the number of layers from 1 to 3 and the number of nodes per layer from 1 to 3 (***Table S10***).

[81] ***Figure 5*** presents box plots summarizing all of these cases, along with the corresponding single-layer cases where node counts ranged up to 5. The leftmost set of figures corresponds to single layer networks, the middle set corresponds to networks with two hidden layers, and the rightmost set corresponds to networks with three hidden layers. The top row corresponds to the mass-conserving DS architectures ($MN_{None}^{DS}$), the middle row to the DS-relaxed architectures ($MN_{None}^{DSR}$), and the bottom row to the least constrained $MN_{None}^{DS-MLB}$ architectures. To facilitate comparison of network complexity, across the top of the Figure we report the numbers of cell-states for each network.

[82] For the single-layer networks, all three cases, each employing different linear transformation layers (***Figures 5a–c***), lead to the same conclusion, which is that there is marginal utility in increasing the number of nodes beyond 2, with best performance (median $KGE_{ss} = 0.84$ ) achieved by the $MN_{None}^{DS-MLB}(2,0,0)$ architecture. For the two-layer networks, performance noticeably improves when the first layer contains more than one node. Of these, the $(2,1,0)$ architectures with three cell-states can be viewed as analogous to the physical-conceptual HYMOD-like architecture where the subsurface flow path merges with the surface flow path before channel routing occurs. This observation supports previous findings (see *Wang & Gupta (2024b)*) that an adequate representation of Leaf River Basin rainfall-runoff dynamics requires more than one flow path (quick and slow). Overall, the (3,3,0) architectures—with three cell states in each layer—achieve best overall performance among the two-layer networks, particularly for the DS case, while also ranking among the top-performing configurations for the DSR and DS-MLB cases. This supports the notion that representing three distributional statistics in each layer is necessary to adequately capture the system behavior. Finally, the three-layer network results show a clear and significant drop in performance. Overall, considering all combinations of layer configurations and node numbers shown in ***Figure 5***, the models with one or two layers perform better, indicating that adding a third layer is unnecessary.

[83] Dry, medium and wet year hydrographs for the different $(3,3,0)$ architectures are included in the supplementary materials (***Figure S3***). Small improvements were found in reproduction of timing and flow peak. These multi-layer cases were also found to be less biased in their representations of low-flow. In general, the MLB case can be removed from consideration due to the fact that it can permit the flow values to become negative.

[84] It is important to note that our network approach relies on functional approximation, which differs from hydrological models such as HYMOD that are built upon physical process understanding. Nevertheless, the results presented here provide an alternative perspective for inferring necessary model complexity and the underlying system dynamics of the catchment, thereby offering guidance on how to appropriately incorporate gating complexity when selecting a fixed system architecture with a given number of links and nodes. They also raise an interesting discussion on the differences between neural networks and hydrologic models, both of which can be represented as directed-graph-like architectures consisting of various nodes (states) and links (flow paths).

## 4.2 Results for MCP-Based Architectures with Information Sharing

### 4.2.1 Single-Layer MCP-Based Architectures with Information Sharing

[85] In the *DI, DS, DIR, DSR* and *MLB* architectures tested above, the value of the output gate (at each node in a layer) depends only on the magnitude of its own cell-state. And similarly, the value of the loss gate (at each node in a layer) depends on cell-state and the magnitude of PET. Because there is no direct communication of cell-state information between nodes, the dynamic behavior of a node cannot be *directly* influenced by the dynamic behavioral of other nodes, and each node operates independently when computing output and loss fluxes. In principle, however, the behavior of a node could be influenced, at least in part, by processes occurring

at (some or all) other nodes within the network. In other words, the nodes would _not_ operate independently when computing output and loss fluxes, which would require some form of sharing of information between nodes. For example, in the case of the output gates this conceptually corresponds to the output conductivities of the system being dependent (in some complex fashion) on the overall description of the state of the system (as characterized by the cell-state values of all the nodes). Just as the set of cell-states can be thought of as representing an approximate discrete representation of the distribution of storage in the catchment (a set of "*almost sufficient*" statistics), the set of output conductivities can be considered to represent a corresponding approximate discrete representation of the (cell-state conditional) distribution of mass-flow conductivities. A similar idea applies to the loss gates.

[86] To investigate the potential benefits of such information sharing, we modified the output and loss gating functions of all the nodes to provide access to the cell state values of all other nodes in the same layer (in addition to their own cell-state value). As with the original gating functions (see Section 2.1), this cell-state information is combined in a linear weighted fashion, such that the output gate for the $i^{\text{th}}$ node in layer $\ell$ can be written as $G_t^{O_{i\ell}} = \kappa_{O_{i\ell}} \cdot \sigma\left(b_{O_{i\ell}} + \sum_{j=1}^{N} a_{ij\ell}^{O} \cdot \tilde{X}_t^{j\ell}\right)$, and the loss gate is written as $G_t^{L_{i\ell}} = \kappa_{L_{i\ell}} \cdot \sigma\left(b_{L_{i\ell}} + \sum_{j=1}^{N} a_{ij\ell}^{L} \cdot \tilde{X}_t^{j\ell}\right)$ where $a_{ij\ell}^{O}$, $b_{O_{i\ell}}$, $\kappa_{O_{i\ell}}$, and $a_{ij\ell}^{L}$, $b_{L_{i\ell}}$, $\kappa_{L_{i\ell}}$ are all trainable parameters.

[87] In the rest of this section, we test the benefits of information sharing for the five basic network single-layer architectures (*DI, DS, DIR, DSR* and *MLB*) discussed in Section 2.3. Since we have two kinds of gates – output and loss – we consider three cases: (i) *SAO*: information sharing at the output gates only, (ii) *SAL*: information sharing at the loss gates only, and (iii) *SALO*: information sharing at both output and loss gates. The results, for different numbers of nodes in the hidden layer, are presented in ***Figures 6-8***.

#### 4.2.1.1 Information Sharing at the Output Gate Only

[88] ***Figure 6*** shows results for the *SAO* case where information sharing is permitted at the output gate only. Regardless of architecture type (*DI, DS, DIR, DSR* and *MLB*), information sharing at the output gate improves performance (compare with ***Figure 3***), with overall best median and above median (wet) year $KGE_{ss}$ performance achieved by the *DSR* architecture ($N_1 = 5$) and overall best distributional and below median (dry) year performance achieved by the *MLB* architecture ($N_1 = 5$).

[89] Performance improvement for the distributed input (DI) architecture (***Figure 6a***) saturates at about $N_1 = 2$, in the sense that the performance during the worst year is higher than that obtained with node numbers between 3 and 5. The median and minimum $KGE_{ss}$ values reach 0.87 and 0.65 respectively. Permitting input-mass-adjustments to the DI architecture (DIR) does tend to help a bit (***Figure 6c***), with narrowing of the distribution and performance generally improving for above-median years. We next examine the driest and wettest years more closely. When comparing both flow peak and annual mean flow, three years overlap among the driest and wettest years, respectively, resulting in a total of 13 years identified for each category. For the DI cases, the median $KGE_{ss}$ for the driest years decreases from 0.83 for the two-node network to 0.78–0.79 for networks with three to five nodes, while remaining between 0.88 and 0.89 for the wettest years across the same node range. For the DIR cases, the median $KGE_{ss}$ for the driest years remains around 0.84–0.87, showing only slight improvement with increasing node number. In contrast, the wettest years exhibit a clear upward trend, with $KGE_{ss}$ increasing from 0.87 to 0.91 as the number of nodes increases from two to five.

[90] Performance for the distributed state (DS) architecture (***Figure 6b***) continues to improve till $N_1 = 4$ with the median and minimum $KGE_{ss}$ values reaching 0.91 and 0.72 respectively. But when output-mass-adjustments are permitted (*DSR*: ***Figure 6d***), we see progressive improvement till $N_1 = 5$, (median $KGE_{ss} = 0.92$) with significant improvements extending into below median years with minimum $KGE_{ss}$ reaching 0.70.

[91] For the MLB architecture (***Figure 6e***) where mass-adjustments are permitted to both inputs and outputs, although single node performance – where the concept of information sharing does not exist – is rather poor (see ***Figure 3***), increasing $N_1$ up till 5 results in progressive overall improvement with the _minimum_ (driest year) $KGE_{ss}$ reaching 0.76, the best achieved by any of the architectures tested so far. This highlights the importance

of allowing each output gate to learn from the cell states of other nodes when the number of nodes increases sufficiently to enable overall performance improvement.

#### 4.2.1.2 Information Sharing at the Loss Gate Only

[92] ***Figure 7*** shows similar results for the *SAL* case where information sharing is permitted only at the loss gate. In this case, we see almost no improvement, which is perhaps not surprisingly given that runoff fluxes strongly dominate over evapotranspirative fluxes in this basin. The main improvement is when input-mass adjustments are allowed (*DIR*; ***Figure 7c***), when we get noticeable improvements in the minimum $KGE_{ss}$ (reaches 0.77) especially for $N_1 = 5$.

#### 4.2.1.3 Information Sharing at both Output and Loss Gates

[93] ***Figure 8*** shows results for the *SALO* case where information sharing is permitted at both the output and loss gate. For all architectural cases (*DI, DS, DIR, DSR* and *MLB*), we see performance improvements (particularly on the below median years) compared with *SAO*, where sharing is only at the output gates. Once again, the best overall $KGE_{ss}$ performance is achieved for the distributed state (DS; median $KGE_{ss} = 0.92$) and mass-relaxed distributed state (DSR; median $KGE_{ss} = 0.92$) cases with $N_1 = 5$, with the latter being slightly better (similar median, but tighter distribution, better below median performance, and overall best worst year performance $KGE_{ss} = 0.76$).

### 4.2.2 Multi-Layer MCP-Based Architectures with Information Sharing

[94] Next, we repeated the multi-layer MCP network experiments summarized in Section 4.1.2 but allowed the cell states to share information at the gates (see ***Figure 9***). For brevity, we report only results for the SALO case where cell-state information is shared at both the output and loss gates. Note that this is analogous to modern recurrent neural networks, such as LSTMs, where sharing occurs across all gates.

[95] Unlike in the non-sharing case (***Figure 5***), overall network performance declines as the number of layers are increased, in the sense that no 2-layer network outperforms the best single-layer network, and no 3-layer network outperforms the best 2-layer network. Overall, best performance is obtained by the single-layer 5 node case ($N_1 = 5$) with median $KGE_{ss} = 0.92$ for the distributed state *DS*, *DSR* and *MLB* architectures. This suggests that information-sharing helps the model achieve greater architectural parsimony and is more important than increasing network depth. However, we suggest that future large-sample studies, possibly incorporating alternative model development and evaluation frameworks, could further revisit and test this conclusion.

## 4.3 Comparison to Physical-Conceptual Models and LSTM Network Benchmarks

### 4.3.1 PC Model Benchmarks

[96] For benchmark comparison against traditional mass-conserving PC model architectures, we use the four progressively more complex architectural hypotheses presented by *Wang & Gupta (2024b)*, referred to as $MA_3$, $MA_4$, $MA_5$, and $MA_6$ (see **Figure S4**). The $MA_3$ model consists of one flow-path with two cell-states in series, the first interpreted as a lumped-average catchment soil moisture storage and the second interpreted as a routing store. The $MA_4$ model consists of two flow-path with two cell-states in parallel, with one interpreted as a lumped-average catchment soil moisture storage that generates (near)surface flow, and the second interpreted as a subsurface (groundwater) store that can sustain baseflow. The $MA_5$ model consists of three cell-states and two flow-paths. It essentially extends the $MA_4$ architecture to include surface/channel routing. The $MA_6$ model extends the $MA_5$ architecture to include an overland flow path that routes overland flow directly to the catchment outlet. All of these PC model architectures conserve mass at both the nodal and overall architectural levels.

[97] Further, given the finding (reported above) that providing access to information about <u>*all*</u> of the cell states can improve the estimation of output conductivities computed by the gates, and thereby the overall performance of the model, we investigate two versions of the aforementioned physical-conceptual $MA_3 -$

$MA_6$ models. In the first, we retain the traditional strategy used in catchment-scale hydrological modeling where the output fluxes at each moisture tank (cell-state) *do not draw upon* on knowledge of the overall distribution of soil moisture *throughout* the system – i.e., information remains local. In the second, we *do allow* the output fluxes at each moisture tank to depend on the overall distribution of soil moisture throughout the system – i.e., information is shared globally.

#### 4.3.2 ML Model Benchmarks

[98] For benchmark comparison against the purely data-driven LSTM-based models, we examine LSTM networks (*Hochreiter & Schmidhuber, 1997*) with up to 5 nodes per layer and up to three layers (***Figure S5***). Overall, the numbers of trainable parameters for the most complex (three-layer 15-cell-state) LSTM network reaches 486. Note that, due to the lack of any physically-conceptual regularizing restrictions on its architectural form, the performance of the *LSTM* network represents an upper benchmark on precipitation-to-streamflow conversion performance achievable using the Leaf River data set (*Nearing et al., 2020a*).

#### 4.3.3 Benchmarking Results

[99] Results of the benchmarking study are summarized by ***Figure 10*** (see also ***Table 3 and 4***). The results are grouped into five sets, where the pink color represents the traditional PC modeling approach (architectures $MA_3 - MA_6$) *without* global sharing of cell-state information, the red color represents the same architectures but *with* global sharing of here cell-state information, the light green color represents various selected MCP-based network architectures *without* sharing of cell-state information, the dark green color represents selected MCP-based network architectures *with* sharing of cell-state information, and the blue color represents selected LSTM-based networks. The MCP-based networks reported here were selected based on overall best performance as reported in the previous sections, and the LSTM-based networks were based on overall best performance as reported in the supplementary materials (***Table S13-14***).

[100] In summary, we notice the following:

1) Best overall distributional, median year, and worst year performance is achieved by the single-layer, 5-node MCP-based mass-conserving networks with cell-state information sharing (dark green).
2) For the physical-conceptual modeling approach (pink and red), cell-state information sharing results in some degree of performance improvement, suggesting that this strategy might be beneficial to adopt more generally in physically-based modeling.
3) The distributed-state MCP-based mass-conserving networks (light green in ***Figure 10c***) tend to provide better below-median and worst-year performance, than the traditional physical-conceptual architectures (pink in ***Figure 10a***), meaning that the MCP-based mass-conserving network architecture is capable of handling diverse wetness conditions in terms of performance. This suggests that we might be able to learn something from the network-based approach about how to construct better physical-conceptual hypotheses. In particular, the notion of a distributed-state (multiple-statistic) surface moisture storage might be worth exploring – this would be an extension of the probability distributed moisture (PDM; *Moore, 2007*) store concept already successfully adopted by models such as HyMod (*Boyle et al., 2000*) and the variable infiltration capacity model (VIC; *Liang et al., 1994*), etc.
4) The single-layer, 5-node MCP-based mass-conserving network with cell-state information sharing (dark green) performs comparably to the purely data-driven LSTM-based models (which do not incorporate any mass-conservation restrictions) tested here. This is encouraging given that the *LSTM* network can be assumed to represents an approximate upper performance benchmark. Interestingly, the more regularized architecture of the MCP-based network tends to result in better performance on the below-median and drier years, suggesting that the mass-conserving regularization tends to result in more stable model performance.

### 4.4 Physical Interpretability of Networks

[101] A major premise of this work is that catchment-scale models constructed using MCP-based networks are more readily interpretable than ones constructed using conventional ML strategies. ***Figures 11 & 12*** illustrate the interpretability of the single-layer distributed-state ($DS$) models reported in Section 4.1.1.

[102] ***Figures 11a-e*** show the forms of the *output-gate* conductivity functions learned by the single-layer DS models as the number of MCP nodes is progressively increased from 1 to 5. Notice that each gate—for example, in the single-node case— remains "*closed"* when the cell-state is below some level (e.g., $< \sim 700\ mm$ in ***Figure 11a***), and then rapidly "*opens*" till it reaches some maximum conductivity level (e.g., $\sim 0.045\ mm/mm$ in ***Figure 11a***); recall that the minimum and maximum possible conductivity levels are $0.0$ and $1.0\ mm/mm$ respectively. This also applies to output gates in other cases with a larger number of nodes, where different maximum conductivity levels are reached and the cell-state thresholds for gate activation (opening) may vary (***Figure 11b-e***). To facilitate interpretability, the gating functions are color coded according to their *maximum* conductivity values (y-axis) such that blue < orange < yellow < purple < green.

[103] Meanwhile ***Figure 12*** shows plots of dynamically evolving time series values for a dry-year (WY 1952; left column), median-year (WY 1953; middle column), and wet-year (WY 1974; right column). Here, for brevity, we focus our discussion primarily on the specific case of the three-node DS architecture ($MN_{None}^{DS}(3)$). Corresponding results for all five cases ($MN_{None}^{DS}(1)$ to $MN_{None}^{DS}(5)$) are presented in the supplementary materials (***Figure S6-S13***).

[104] Each row of ***Figure 12*** shows a different variable. The top row shows the observed streamflow hydrograph. The second, third and fouth rows show the corresponding cell states, output gate states, and the *accumulated* components of streamflow, color coded to correspond to ***Figure 11c*** (blue, orange and yellow for the nodes with lowest, intermediate and largest maximum output conductivity values respectively). Meanwhile, the fifth row shows the same values as in rows four but plotted as fractions of its total values.

[105] *Wang & Gupta (2024a)* previously discussed in detail the behavioral expressivity of the single-node case ($MN_{None}^{DS}(1)$); its cell state value varies between $\sim 375 - 700\ mm$ (***Figures S6d-f***), and the output gate activates when the cell state reaches $\sim 600\ mm$. In contrast, the two-node $MN_{None}^{DS}(2)$ case has two cell states and two flow-paths. From ***Figure 11b*** we see that the output gate for the first node (blue) activates at a smaller cell state value ($\sim 577\ mm$) than for the second node (orange; $\sim 707\ mm$), and is therefore a more active contributor to streamflow during the dry periods. In contrast, the second node activates during wetter periods, when the cell-state value is larger, whereupon it becomes the more dominant contributor to total streamflow (***Figures S13g-i***).

[106] In several ways, however, the $MN_{None}^{DS}(3)$ case is more interesting (***Figures 11c & 12***) because the three flow paths can be loosely interpreted as representing a slowly responding groundwater flow path (blue color; lower maximum conductivity), moderately responding interflow path (orange color; intermediate maximum conductivity), and quickly responding surface flow path (yellow color; higher maximum conductivity). Node 1 (blue) can be interpreted as representing the behavior of the groundwater system and storing the largest fraction of water in the system (***Figures 12d-f***). Its output gate activates at $\sim 585\ mm$ but reaches a relatively low maximum output conductivity value of $0.055\ mm/mm$ (***Figure 11c***), so that it contributes the smallest total volume to streamflow ($\sim 12\%$; ***Figures 12m-o***), but remains active during the non-raining periods and thereby sustains baseflow throughout the year.

[107] Meanwhile, given that the LRB is a perennial stream, the second (orange) and third (yellow) nodes can be interpreted as both generating significant rapid contributions to streamflow in response to pulses of rainfall. The second node (orange) starts to generate contributions to streamflow when its cell-state becomes larger than $\sim 496\ mm$, while the third node (yellow) activates when its cell-state becomes larger than $\sim 226\ mm$ (***Figure 11c***). Overall, these two nodes correspond to smaller cell-state values but generate the largest proportions ($\sim 44\%$ each) of overall streamflow by becoming active/responsive during the rainfall-driven periods, consistent with an interpretation of corresponding to shallower depths of soil-moisture storage located closer to the surface.

[108] Finally, it is interesting to note that Node 1 of the four-node $MN_{None}^{DS}(4)$ case also behaves in a manner representing a groundwater flow path (***Figure S6m-o***) in that it contributes to streamflow only during very dry periods (***Figure S12m-o***), with its cell state exhibiting only very mild variations. Meanwhile, the other three output gates behave very similarly to those of the three-node $MN_{None}^{DS}(3)$ case. This suggests the possibility that *two* groundwater flow components – faster and slower – may be required in the Leaf River Basin (a feature that is built into the SACSMA conceptual rainfall-runoff model used by the National Weather Service for streamflow forecasting). However, when investigating the five-node case (subplots ***p*** to ***r*** in ***Figure S6-13***), although metric-based performance clearly improves, interpretability becomes increasingly more ambiguous/difficult, with the results allowing for multiple possible explanations.

## 4.5 Results of Network Pruning

[109] The analysis conducted in Section 4.1.1 indicates that the five-node $MN_{None}^{DS}(5)$ network may contain redundant flow paths. This hypothesis is supported by two observations: (1) multiple gating functions (yellow and orange in ***Figure 11e***) exhibit very similar functional — both activate when the cell state exceeds approximately 571 mm and 582 mm, respectively, and saturate at a value of 0.056, and (2) several flow paths (especially orange and green) make very similar contributions to total streamflow (***Figure S12p–r***). To gain further insight into how these internal mechanisms may be functioning, we investigated how the predictive performance of the "*five-node*" models are affected by successive pruning (removing) of nodes from the networks. We do this for the single-layer distributed state $MN_{None}^{DS}(5)$ network, as well as the single-layer distributed state $MN_{SALO}^{DS}(5)$ network that has been enhanced to incorporate *Sharing-Augmented Loss and Output* (SALO) into the gates (***Table S15***).

[110] The strategy for removing nodes that we followed was to systematically prune different numbers of nodes (1, 2, 3 and 4) from the optimized 5-node configuration and identify (for each pruning case) the one having the best possible performance. Note that, at this stage, no further re-training (fine-tuning) of the networks was conducted. So, to be clear, pruning *one* node results in *five* different cases, from which the case with the best median $KGE_{ss}$ performance was selected. Similarly, pruning two, three, and four nodes yields ten, ten, and five possible cases, respectively. For each pruning scenario, we select and report only the case having the best median $KGE_{ss}$ performance (without re-training).

[111] For ease of comparison, ***Figure 13a*** presents the results (boxplots of annual $KGE_{ss}$) showing how performance of the "*non-information sharing*" $MN_{None}^{DS}(N)$ networks varies for $N = 1 \rightarrow 5$, while ***Figure 13b*** shows the (best) results obtained as successive numbers of nodes (from one to four) were pruned from the $MN_{None}^{DS}(5)$ architecture. So, the results labeled as $MN_{None}^{DS}(5) - D1(P)$ correspond to removing *one* node from $MN_{None}^{DS}(5)$, and so on, where the notation *"P"* indicates that the corresponding flow *"paths"* have been removed from the output layer of the network.

[112] It is interesting to note that some aspects of the distribution of $KGE_{ss}$ skill have improved when comparing the "*pruned*" four-cell-state $MN_{None}^{DS}(5) - D1(P)$ case to its parent "*un-pruned*" five-cell-state $MN_{None}^{DS}(5)$ case. Specifically, we note a small increase in $KGE_{SS}^{max}$ (from $0.93 \rightarrow 0.94$), and a larger increase in $KGE_{SS}^{25\%}$ (from $0.78 \rightarrow 0.82$) and in the lower quartiles, while $KGE_{SS}^{50\%}$ remains essentially unchanged (at $0.86$). This improvement is also observed when compared to the four-cell-state $MN_{None}^{DS}(4)$ case. Meanwhile, a small loss in performance is observed for the very driest (outlier) year. However, when more than one node is removed from $MN_{None}^{DS}(5)$, performance significantly declines. This suggests that the most useful information in the data has been encoded into four independent flow paths, implying that further improvement in single-node architectural adequacy could help reduce the number of flow paths used**,** and potentially provide guidelines for constructing a parsimonious and interpretable model architecture**.**

[113] Similarly, ***Figures 13c & d*** present corresponding results for the "*information sharing*" cases. Note, as reported before (Section 4.2), that overall performance improves (***Figure 13c***) compared to the non-sharing case (***Figure 13a***) for $N \geq 3$. However, as flow paths (not nodes, see discussion below) are progressively pruned (***Figure 13d***), removing 1 or 2 paths has little or no apparent impact on distributional performance,

while removing 3 paths has a small impact and removing 4 paths has a significant impact. Specifically, removing 1 and 2 paths (cases $MN_{SALO}^{DS}(5) - D1(P)$ and $MN_{SALO}^{DS}(5) - D2(P)$) results in only small declines of $KGE_{SS}^{50\%} = 0.92 \rightarrow 0.91$ and $KGE_{SS}^{min} = 0.73 \rightarrow 0.71$ compared to the un-pruned $MN_{SALO}^{DS}(5)$ network. Meanwhile the pruned two-path $MN_{SALO}^{DS}(5) - D3(P)$ network maintains very similar predictive accuracy to the unpruned $MN_{SALO}^{DS}(5)$ network.

[114] To be clear, in these "*information-sharing*" cases (and unlike in the "*non-sharing*" cases), removal of *output flow paths* does *not* correspond to removal of *nodes*. In fact, the corresponding cell-states remain active and continue to provide (discrete distributional) cell-state information that is used by the output and loss gates of the nodes corresponding to the non-pruned flow pathways. Again, these results support the hypothesis that three flow paths are required to model the dynamics of the LRB, even though the desirable number of (information sharing) cell-states may be larger (as many as 5). Since there is no obvious need to account for snow accumulation and melt dynamics in the LRB, these three flow paths can again be interpreted as corresponding to slow, intermediate and fast flow pathways.

[115] For completeness, we finally examine an alternative approach to pruning the five-node/pathway $MN_{SALO}^{DS}(5)$ network, wherein we effectively remove *both* the pathway and its associated node, so that the information regarding the cell-state of node corresponding to the removed pathway is not shared with the gating functions of the other (non-pruned) nodes. This is easily done by setting the associated weights to zero. Results are shown in ***Figure 13e***, where *"F"* refers to full (as opposed to partial) removal of shared information. Clearly, performance declines significantly as flow paths and associated cell-states (nodes) are removed.

[116] Overall, from these experiments, we can conclude that maintaining sufficiently many cell-states (here $3 - 5$) to represent the distributional properties (minimal sufficient statistics) of the storage dynamics of the catchment is critical, while also ensuring that the fast, intermediate and slow flow pathways are properly represented.

## 4.6 Overall Summary and Discussion of Part 1

[117] In summary, our detailed testing of various MCP-based network architectures on the Leaf River basin indicates that:

1) ***Methods for Distributing the System Inputs across Nodes***: Overall, the distributed state (DS) hypothesis performs better than the corresponding distributed input (DI) hypothesis, supporting the idea that more than one soil-moisture statistic is required to adequately model the input-state-output dynamics of the Leaf River Basin.

2) ***Relaxing Mass Conservation at the Network Level:*** There appears to be some (albeit small) benefit to allowing mass-relaxation at the overall network level, with the most stable performance being achieved by the mass-relaxed distributed state *DSR* architecture.

3) ***Information Sharing Across Nodes:*** Consistent with conceptual understanding, cell-state information sharing at the output gates improves performance on above median years (which tend to be wetter) while information sharing at the loss gates improves performance on the below median years (which tend to be drier). Interestingly, PC models also seem to benefit somewhat from cell-state information sharing, suggesting that physically-based modeling should also consider adopting such a strategy (see example in Figure 10b, where four mass-conserving hydrologic model architectures described in *Wang & Gupta (2024b)* are tested).

4) ***Network Depth:*** Network performance tends to decline as the number of layers is increased. Overall, best performance is obtained by a relatively parsimonious single-layer distributed state (*DS*) or distributed state with mass-relaxation (*DSR*) architecture when information sharing across the nodes is permitted.

5) ***Benchmarking Results:*** A single-layer MCP-based mass-conserving network with cell-state information sharing performs comparably to the upper benchmark data-driven LSTM-based model which does not incorporate mass-conservation restrictions. Further, the MCP-based model provides better performance on the below-median and drier years, suggesting that mass-conservation tends to result in more stable model performance.

6) ***Interpretability:*** The single-layer three-node MCP-based mass-conserving network shows physically interpretable behavior consistent with the perennial nature of flow in the LRB. One node can be interpreted as representing the groundwater system with large storage capacity, low output conductivity, and a slowly responding groundwater flow path that sustains baseflow throughout the year. A second node corresponds to a moderately responding interflow path that generates contributions to streamflow in response to pulses of rainfall when its cell-state capacity is exceeded, while a third node corresponds to a quickly responding surface flow path that becomes active during periods of more intense rainfall. When a fourth node is included, it contributes to streamflow only during very dry periods, suggesting that *two* groundwater flow components (faster and slower) may actually be required for longer-term hydroclimatic simulations.

7) ***Network Pruning:*** With increasing numbers of nodes, the single layer MCP-based networks appear to contain redundant flow paths. While pruning *nodes* results in declining performance, pruning *connections* actually improves some aspects of model skill. Overall, the results indicate that sufficiently many cell-states (here $3-5$) must be present to represent the distributional properties of storage, while maintaining three flow pathways (fast, intermediate and slow) to properly model the streamflow dynamics of the LRB.

# 5. Experimental Results Part 2 – Multi-Catchment Investigation

## 5.1. Modeling Approaches Tested

[118] Based on the insights obtained via the detailed study conducted in Part 1, we chose the four MCP-based model architectures listed below to test for this multi-catchment investigation. Note that all of these models were trained individually for each basin, using the same procedure reported for Part 1.

1) A simple MCP-based single-node mass-conserving model ($MC\{O_\sigma L_{\sigma+}^{con}\}$)

2) A single-layer three-node '*distributed-state*' network *without* cell-state info-sharing ($MN_{None}^{DS}(3)$).

3) A single-layer five-node '*distributed-state*' network *with* cell-state info-sharing ($MN_{SALO}^{DS}(5)$).

4) A variant of $MN_{SALO}^{DS}(5)$ network in which the evapotranspirative constraint at the loss gate was relaxed (notated as $MN_{SALO}^{DS-R_{con}}(5)$).

[119] For a data-based ML upper benchmark, we use a 5-node LSTM, where the number of nodes was determined based on both current and previous research (*Wang & Gupta, 2024a,b*). To ensure apples-to-apples comparison, the LSTM models were also developed separately for each basin (see details in Text S2 in the Supplementary Materials). Given the local catchment-by-catchment nature of network training, these results should be treated as preliminary.

## 5.2. Spatial Distribution of Performance Across the CONUS

[120] Testing period results are summarized in ***Figure 14 & Figure 15***. The cumulative density plots for $KGE_{ss}$ (***Figure 14a***) show that distributional performance improves progressively with network architectural complexity, with median performance increasing from 0.72 for the simplest single-MCP-node model ($MC\{O_\sigma L_{\sigma+}^{con}\}$; blue), to 0.78 for the three-MCP-node model ($MN_{None}^{DS}(3)$; orange), 0.81 for the five-MCP-node model ($MN_{SALO}^{DS}(5)$; black), and finally 0.86 for the five-MCP-node PET-relaxed model ($MN_{SALO}^{DS-R_{con}}(5)$; yellow). This compares with 0.89 for the benchmark five-node generic LSTM model ($LSTM(5)$); purple. ***Figures 14b-d***

show a generally similar trend in skill for the three $KGE$ components – linear-correlation/timing ($r^{KGE}$), mass balance error ($\beta^{KGE}$) and flow variability error ($\alpha^{KGE}$).

[121] Spatial distribution maps of skill are shown in ***Figures 15a-t***. As can be seen, performance of the physically-interpretable MCP-based models improve progressively with increasing network complexity (row one through four). Performance of the simplest (most parsimonious) single-MCP-node $MC\{O_\sigma L_{\sigma+}^{con}\}$ is shown in ***Figures 15a-d***. Relative performance is generally better (more orange-yellow as opposed to light and dark purple) in the eastern US, with $\mathrm{KGE}_{ss}$ skill tending to decrease at higher latitudes. Comparable performance in the western US is only found in the northwestern region, around the Cascade Range, and skill declines southward toward the Sierra Nevada. In particular, poorest performance is seen in the mountainous regions of the western US, where large snowpacks exist at high elevations and there are dry flow conditions at lower latitudes. This pattern of performance seems reasonable given that a single cell state can be insufficient to represent the more complex dynamics associated with both snow-dominant and very dry regions (see ongoing effort reported in *Wang et al., 2025*). A very similar spatial pattern is seen for linear-correlation/timing ($r^{KGE}$; ***Figure 15b***). Flow variability tends to be overestimated (red) nearly everywhere, with underestimation (blue) occurring primarily in the mountainous areas of the western US ($\alpha^{KGE}$; ***Figure 15c***). Meanwhile, mass-balance error ($\beta^{KGE}$; ***Figure 15d***) is relatively mild, with overestimation (red) tending to occur east of 100°W and in the mountainous regions of the western US.

[122] With increasing network complexity, performance improves significantly. The three-node '*distributed-state*' network without cell-state info-sharing ($MN_{None}^{DS}(3)$; row two). ) has clearly better $KGE_{ss}$ performance across the majority of basins in both the eastern (89.3%; 291 out of 326) and western (83.4%; 156 out of 187) CONUS regions (***Figure 15e***). Improvement in median $KGE_{ss}$ skill is similar (~0.03) for both the East and West regions, supporting the hypothesis that multiple flow paths are generally required to adequately represent precipitation–runoff dynamics. As before, spatial patterns in $r^{KGE}$ (***Figure 15f***) align closely with those of $KGE_{ss}$. Overall, we see a significant reduction in high bias, as indicated by fewer red dots in both flow variability and mass balance error (***Figures 15g-h***; white is perfect).

[123] The five-node configuration with cell-state information sharing ($MN_{SALO}^{DS}(5)$) shows further overall improvement (***Figures 15i-l***; row three). Similar improvements are again observed for $r^{KGE}$ (***Figure 14n***) while mass balance error (***Figure 15j***) remains roughly comparable to the simpler $MN_{None}^{DS}(3)$ architecture, and flow variability error tends to worsen in the southern CONUS near Texas and in parts of the Appalachian Mountain range (***Figure 15k***). Note, however that in the mountainous US, we observe both increases and decreases in skill (***Figure 15i***). While performance in several basins improves (shift towards yellow), that of others declines (shift towards purple). Careful examination reveals that the basins located between 105°W and 115°W that have $KGE_{ss} < 0$ correspond to median snow fractions of $63\%$ (snowfall fraction based on the feature reported in CAMELS US dataset) and average elevations of $2.579$ meters. Under these snow-dominated conditions the current MCP-based architecture seems unable to properly capture the transformation of precipitation and PET into snow-related processes. This suggests that future work should explicitly incorporate an interpretable snow component into the MCP architecture (see also *Wang et al., 2024 & Wang et al., 2025*).

[124] When the PET constraint at the loss gate is relaxed ($MN_{SALO}^{DS}(5)$) we achieve the best-performing of the physically-interpretable MCP-based architectures (***Figure 15m-p***; row four). It is particularly interesting to observe that basins in the mountainous western US exhibit significant improvements in $KGE_{ss}$ and all its components. Relaxation of the PET constraint enables a re-adjustment of the overall mass balance, which appears to compensate for the absence of explicit representation of snow accumulation and melt processes in the current MCP-based architectures. Again, these results suggest the importance of incorporating a snowpack component when modeling basins heavily dominated by snow processes.

[125] Finally, as can be expected, overall best performance is seen for the benchmark non-mass-conserving locally trained LSTM(5) network models (***Figures 15q-t***; bottom row). These locally trained LSTM models easily achieve $KGE_{ss} > 0.90$ and exhibit very small mass-balance errors, meaning that the ratio between the mean

of simulated and observed flows ($\beta^{KGE}$) is close to one (***Figure 15t***), particularly in the mountainous regions, with all three KGE components exhibiting relatively small errors (***Figures 15r-t***). As might be expected, the purely "*information-driven*" nature of the LSTM without physical constraints enables it to achieve superior performance compared to the MCP-based network approach, particularly in locations where key physically based input information may be missing.

[126] In summary, these preliminary results demonstrate the general applicability of our MCP-based network approach for different hydroclimatic regimes across the U.S. ***Table S16*** of the supplementary materials provides a summary of representative percentiles of the cumulative distribution functions (CDFs) for several metrics, including $KGE_{ss}$ and its three components, root mean squared error ($RMSE$), and mean absolute error ($MAE$). It is clear that while some locations can be parsimoniously modeled using an MCP-based model with only a single cell-state, other locations have more complex dynamics that require more complex multi-cell-state representations. Of course, it is not surprising that there is a need to explicitly incorporate a representation of snow dynamics, particularly in the mountainous western US. Given that performance of a 5-cell-state model with cell-state info-sharing but with removal of the PET constraint approaches that of a 5-cell-state LSTM, it seems likely that integrating info-sharing between soil moisture and snowpack states is a promising direction for future exploration.

## 5.3. Detailed Analysis for a Selected Catchment

### 5.3.1. Catchment Selected

[127] Since there are too many catchments (500+) for us to reasonably examine their inner functional workings in individual detail, we selected basin 11473900 for an in-depth analysis of both performance and internal architectural functioning, due to it being located in the western US (the Middle Fork Eel River in northern California). Note that this basin has been classified as "*rainfall-dominated*" based on the LSTM gradient analysis conducted by *Jiang et al. (2022).* Being in the western US, it represents a spatially different part of the US from the Leaf River. Further, the three-node $MN_{None}^{DS}(3)$ configuration at this location exhibited the largest improvement in performance compared to the single-node $MC\{O_\sigma L_{\sigma+}^{con}\}$ model, suggesting that the multi-state representation is particularly useful in this basin, and prompting our curiosity regarding interpretability of its internal model architecture.

### 5.3.2. Comparison of Model Performance

[128] Boxplots of the distribution of annual $KGE_{ss}$ performance across 27 climatically disparate water years (WY 1982 – WY 2008) are shown in ***Figure 16a***. Note that median year performance ($KGE_{ss}^{50\%}$) improves from 0.66 for the single-cell-state $MC\{O_\sigma L_{\sigma+}^{con}\}$ model to 0.82 for the $MN_{SALO}^{DS-R_{con}}(5)$ model with relaxed PET constraint, but declines for the $MN_{SALO}^{DS}(5)$ model when forced to obey the PET constraint. Ignoring this poorer performing variant, we note that worst year performance ($KGE_{ss}^{worst}$) similarly improves from 0.38 to 0.53. Comparing the $MN_{SALO}^{DS-R_{con}}(5)$ model with the $LSTM(5)$ model ($KGE_{ss}^{50\%} = 0.84$ and $KGE_{ss}^{worst} = 0.32$) we see that while the distribution of annual performance is comparatively slightly worse, the significant improvement in worst-year performance (0.53 compared to 0.32) suggests that the $MN_{SALO}^{DS-R_{con}}(5)$ model is arguably more stable.

[129] While skill scores (metrics) can be informative in an overall (aggregate) sense, a more intuitive sense for how the models are performing can only be obtained by actual examination of their streamflow hydrograph time series. Accordingly, we show hydrograph comparisons (***Figure 16b-g***) for a dry year (WY 1994; top row), median year (WY 1988; middle row) and a wet year (WY 1997 bottom row), corresponding to the water years with smallest, median, and largest annual flow peaks ( $KGE_{ss}$, $RMSE$, and $MAE$ metric values are reported in the inset tables). Here, in the left column of plots we focus on comparing the MCP-based single-cell-state model ($MC\{O_\sigma L_{\sigma+}^{con}\}$; blue lines) with the three-cell-state model ($MN_{None}^{DS}(3)$; orange lines), while in the right column of plots we focus on comparing the MCP-based five-cell-state model ($MN_{SALO}^{DS-R_{con}}(5)$; yellow lines) with the

benchmark model ($LSTM(5)$; purple lines). Each figure shows streamflow observations using red circles and model simulations using solid lines.

[130] From the left column of plots we see that the three-cell-state model ($MN_{None}^{DS}(3)$; orange) shows improvement over the single-cell-state model ($MC\{O_\sigma L_{\sigma+}^{con}\}$; blue) in the median and wet years respectively (***Figures 15c,d***), but fails to reproduce some of the flow peaks during the dry year (***Figure 16b***). Further, it provides better simulations during the flow recessions – May to September of the median year, 1988 (***Figure 16c***), and April to September of the wet year, 1997 (***Figure 16d***).

[131] From the right column of plots we see that the five-cell-state model ($MN_{SALO}^{DS-R_{con}}(5)$; yellow) shows comparable performance to that of the $LSTM(5)$ during the median year (***Figure 16f***), but performs overall somewhat worse in both the dry (***Figure 16e***) and wet years (***Figure 16g***). Overall, these results suggest that the improved performance of $MN_{SALO}^{DS-R_{con}}(5)$ over $MN_{None}^{DS}(3)$ is likely achieved by means that are physically inconsistent, namely that relaxation of the constraint on evapotranspirative loss compensates for the architectural inability to represent water inputs and outputs from snowpack melting and ablation. Meanwhile, the purely data-driven LSTM network, that is not constrained by physical laws, allows for greater flexibility in redistributing (adjusting) water among precipitation inputs, internal storage, and streamflow outputs**,** rather than enforcing strict mass conservation.

#### *5.3.3. Analysis of Internal Functioning*

[132] To investigate internal functioning, we examine the forms of the MCP gating functions. The output gate functions for $MN_{None}^{DS}(3)$ are shown in ***Figure 17***. Here the nodes are shown in descending order of output gate peak value — the node with the largest peak value is treated as the first node, and the one with the smallest as the third. This ordering is the opposite of that presented in Section 4.4 for the Leaf River, and is done to facilitate interpretation of the results for this selected catchment.

[133] To examine the internal architecture we focus on the aforementioned dry (1994), median (1988), and wet (1997) years (three columns of plots shown in ***Figure 18***). The top row shows the precipitation time series, while the second row shows corresponding observed streamflow and UA-SWE. The remaining rows show corresponding time series plots for simulated streamflow, SWE, cell-states, output-gate-states, stacked streamflow, and associated ratios. Where relevant, the color coding reflects the order of output gate saturation values, from highest to lowest.

[134] From ***Figures 18d-f*** we see that the basin experiences accumulation of snowpack in all three of the years, with the original snow fraction value reported in the CAMELS dataset as $26\%$. From ***Figure 17g-i*** and ***Figures 18j-l*** we see that larger peak values for the output gates correspond to smaller magnitudes of the internal cell state. Further, the output gate of the first node (blue line) consistently exhibits the highest activation among the three nodes, dominating the output gate ratio from around August to roughly January in each of the years (***Figures 18m-o***). From ***Figure 17b***, we see that the output gate of the first node is triggered whenever the corresponding cell state exceeds 76 mm. This is consistent with the accumulative (stacked) streamflow plot (***Figures 18p-r***) where the 1st node provides the largest portion of streamflow output.

[135] Note that the stacked streamflow is computed by first multiplying the streamflow generated along each flow path with its corresponding output layer weight and then summing the contributions from all three flow paths in an accumulative manner, following the order of their output gate peak values. In our demonstration, the linear output layer is implemented using a SoftMax function, which ensures that the weights sum to 1. Given that the assigned weights are 0.44 for the first two nodes and 0.12 for the third node, the resulting streamflow components—also summing to 1—highlight the dominant contribution of the first node to the overall streamflow prediction (***Figures 18v-x***).

[136] Therefore, the streamflow output from the first node likely represents the dominant response to precipitation input. In contrast, ***Figures 18v-x*** indicate that the second node (orange) provides almost no flow contribution during the period from July to January, and instead contributes primarily during January through

July. This pattern suggests that the cell state of the second node primarily represents snow water equivalent (SWE), with the contribution to streamflow corresponding to early-spring snowmelt (snowpack exists primarily between January and March, as indicated by the blue line in ***Figures 18d-f***). Further support for this interpretation comes from the fact that the output gate of the second node activates only when the internal cell state (interpreted to be SWE) exceeds 434 mm (***Figure 17c***). During this period, the cell states across all three flow regimes (***Figure 18g-i***) also exhibit a gradual decline.

[137] Finally, we observe that although the third node exhibits a very small output gate peak value (***Figure 17d***), it maintains the largest overall cell state value among all the nodes. This correspondence of a small conductivity (output gate value) with a large cell state, and the fact that it contributes constantly to streamflow (see log-scale ***Figure 18s-u***), suggests that it primarily represents the slow-draining groundwater and baseflow system (except for a very short period around December of the median and wet years).

[138] In summary, the analysis supports a hypothesis that any representation of precipitation–runoff catchment dynamics at this location (11473900) must be capable of tracking at least three key processes—surface runoff, snowmelt, and groundwater baseflow.

# 6. Discussion, Related Work, Future Directions and Conclusions

## 6.1. Discussion

[139] To recap, this study adopted the function approximation network paradigm employed by modern ML wherein a generic network architecture is implemented consisting of basis function nodes arranged in series and parallel. By adopting the physically-interpretable MCP as the fundamental computational unit, we were able to explore a variety of generic and important modeling issues, as listed in Section 1.3.

### 6.1.1. *Part 1 - Proof-of-Concept Study*

[140] For Overall, for the humid Leaf River Basin, the results support a single-layer, three-to-five-nodes-in-parallel, *distributed-state* (DS) multiple-flow-path architecture with *information sharing* across the nodes. While mass-relaxation at the network level did slightly improve overall performance, it came with some loss in overall conceptual interpretability. Overall, adding depth (more than one layer) did not significantly improve performance, as long as nodal-information-sharing was permitted. Further, a suitable balance between performance and overall conceptual interpretability was achieved by a network having three context-dependent-gating-controlled flow pathways (conceptually interpretable as slow, intermediate and fast) that "*activate*" at different times during the water year (with the slow pathway remaining active all year round) in response to different *cell-state* and *hydroclimatic* conditions. Meanwhile the results suggest that three-to-five cell states are required to adequately track the time-varying dynamics of moisture accumulation and release within the system.

[141] An interesting finding is that, as in data-based LSTM architectures, sharing of information across the gating mechanisms of the cell-states (rather than allowing them to operate in local isolation) helps to improve performance. Information-sharing augmentation of the output gate (***Figure 6***), loss gate (***Figure 7***), and both gates together (***Figure 8***) consistently outperforms the default configuration (***Figure 3***). Given that the distribution of annual $KGE_{ss}$ values spans a range of dry and wet conditions, introducing information sharing across gates generally enhances model skill under a variety of conditions. Overall, this feature leads to quantitative and qualitative improvements across the full range of hydroclimatic regimes (dry, medium, and wet years). Sharing at the output gates enhances above-median (wetter) year performance, while sharing at the loss gates improves below-median (drier) year performance; see Section 4.2 for a more nuanced discussion.

[142] It is important to use the same evaluation metric that was employed during model development in order to ensure a fair assessment of model performance. In this regard, by tracking, reporting and examining the distributions of annual performance metrics (rather than the conventional approach of reporting only the average, or median, performance over the entire testing period) we were able to explore better understand

which aspects of model performance were sensitive to variations in network architectural design – a practice that we strongly recommend be adopted by other model development and evaluation studies.

[143] Finally, we showed that information sharing can also be potentially beneficial if incorporated in the context of traditional physics-based models (see the Leaf River example in ***Figure 10***, where the time constant parameters of the physical-conceptual model are replaced with time-variable gating, leading to improved performance). Overall, this finding is consistent with recent advances in understanding regarding the modular processing of information (*Boyd et al., 2018*).

#### 6.1.2. Part 2 - Preliminary CONUS-Wide Investigation

[144] Our very preliminary evaluation of the performance of several MCP-based model architectures across 500+ CAMELS US basins within the CONUS, found that the single-layer three-cell-state $MN_{None}^{DS}(3)$ architecture without cell-state info-sharing achieves median $KGE_{ss}^{50\%}$performance of 0.78 (compared to 0.72 for the single cell-state $MC\{O_{\sigma}L_{\sigma+}^{con}\}$ model), and reasonably good performance across the CONUS (except in catchments where snow accumulation and melt dynamics dominate). This supports the validity of a general multi-flow-path hypothesis for representation of catchment dynamics.

[145] While our preliminary results indicate that cell-state sharing does not necessarily lead to performance improvement at all catchments (an issue that should receive more detailed in future work), it also highlighted a model limitation – namely the lack of a dedicated module to account for snowpack dynamics.

[146] A more detailed examination of physical interpretability was conducted for the three-MCP-node representation at a selected catchment in the Western US where (in contrast to the Leaf River basin) snow accumulation and melt processes clearly exist (but are not dominant). The analysis revealed that even through the MCP-based network model is arguably not architecturally adequate for this location, it still displays a valuable degree of interpretability. Future explorations would benefit from an analysis of how closely those MCP cell-states and processes align with real world observations thereby informing future model development (*Lees et al., 2021*).

### 6.2. Related Work

[147] The MCP node can be essentially considered an enhanced version of the *Nash* linear reservoir (*Nash, 1957*), with the time-fixed conductivity function being replaced by time-variable gating. Alternatively, it can be viewed as a simplified form of gated recurrent unit, as forms the basis for LSTM networks (*Hochreiter & Schmidhuber, 1997*). Our results suggest that the distributed-state (DS) form of the network architectures examined here can potentially serve the role of a universal function approximator (*Hornik et al., 1989*) for geoscientific time series prediction. In contrast, the distributed-input (DI) network architecture more closely resembles conventional semi-distributed hydrological modeling wherein the incoming precipitation is partitioned into alternative flow pathways by means of an input-distribution function, in a manner analogous to the *Nash Cascade* network proposed by *Frame et al. (2024)*. Although not explored here, MCP-based neural network architectures have the flexibility to incorporate spatially-distributed hydrologic data, allowing different aspects of the watershed to be represented with varying levels of complexity. In other words, the watershed can be treated as an interconnected system, leveraging downstream flow information to constrain and improve upstream flow estimates (*Molina et al., 2024*).

### 6.3. Future Directions

[148] Several directions towards enhanced modeling of mass/energy conservative geoscientific systems suggest themselves. Future work could benefit from drawing upon ideas encoded by the recently proposed *Hydro-LSTM* network architecture (*De la Fuente et al., 2024a*) which demonstrates the value of using time-series information provided by recent hydroclimatic history to improve the performance of context-dependent gating, and by the complementary idea of latent space encoding recently proposed by *Yang & Chui, (2023a,b)* to facilitate the development of interpretable ***regional*** models (*De la Fuente et al., 2024b*). Another possibility

is the use of *Kolmogorov-Arnold Network* theory combined with symbolic regression to construct/learn the forms of gating functions so as to gain additional physical interpretability (*Klotz et al., 2017; Feigl et al., 2020; Udrescu & Tegmark, 2020; Liu et al., 2024b; Liu et al., 2024c; Liu et al., 2024d*).

[149] In this regard, an underutilized potential approach to improving physical interpretability of ML-based models (explored only briefly in Section 4.5) is through carefully designed strategies for network pruning (*Blalock et al., 2020*). In contrast to neural architecture search, as suggested in our previous work (*Wang & Gupta, 2024b*), the pruning strategy would begin with large networks and then progressively apply penalty regularization to the loss function with the goal of discovering more parsimonious optimal architectures (in the sense of balancing both performance and interpretability). It is interesting to note that in the context of image segmentation, the field known as TinyML (where deep compression is applied to large neural networks through pruning, along with trained quantization and Huffman coding) has shown that similar task performance can be achieved with a much more parsimonious representation (*Han et al., 2015*), and that functionally optimal architectures can be achieved simply by building extensions upon a small fundamental unit (*Cai et al., 2021*).

[150] Note that, in contrast with the well-recognized approach of simultaneously training LSTM networks on large heterogeneous datasets with basin static attributes (*Kratzert et al., 2019b; Feng et al., 2022; Kratzert et al., 2024),* our study focuses on benchmarking by training the LSTM network locally for each basin. In *Kratzert et al., (2019b)* the best-performing CONUS-wide LSTM was obtained by training using a weighted NSE metric based on flow magnitude at each location to account for the heteroscedastic nature of flow series, and where the inputs used for network training included meteorological variables and 27 static features (using only meteorological variables for training did not yield comparable performance). As such, future work should properly account for heteroscedasticity within the KGE metric before a balanced and fair comparison can be made. We leave such an apples-to-apples comparison for future work. Our main objective here was to obtain insights into potential behaviors and performance of the MCP-based architecture, noting that different levels of (interpretable) architectural complexity may be necessary when applying the approach over large areas with varying hydro-climatological process complexity.

## 6.4. Conclusions

[151] This paper is third in a series that explores how machine learning can be leveraged to enhance scientific understanding through the use of the MCP-based computational units developed in *Wang & Gupta (2024a)*. As demonstrated in *Wang & Gupta (2024b)*, the MCP can serve as a foundation for constructing physically interpretable conceptual hydrologic models. However, in this study instead of imposing strong physical regularization we instead used the MCP as a basis to construct standard neural network architectures. As proof of concept, we first tested our approach on the "*rainfall dominated*" (non-snowy) Leaf River basin in Mississippi and then apply it to 513 CAMELS US basins as a preliminary evaluation of its general applicability. To examine potential transferability of our approach across diverse hydrologic regimes, we then selected a second rainfall dominated catchment—the Middle Fork Eel River in California (which has a small degree of snow accumulation and melt)—to perform a similar in depth analysis as was conducted for the Leaf River. The successful application of our method to both basins supports its potential broader adoption for hydrologic model interpretation and data-driven scientific inference.

[152] We have proposed two main conceptual ways for representing the spatially lumped dynamics of hydrologic catchment systems: the "*distributed-state*" (DS) and the "*distributed-input*" (DI) representations. The DS case, which inherits its idea from the Kolmogorov neural network existence theorem, enables estimation of precipitation-runoff dynamics via multiple parallel flow paths. In contrast, the DI case may be more useful for modeling systems where spatially distributed measurements of input fluxes are available. Together, this can enable flexible representation of processes such as spatially varying rainfall distribution and upstream–downstream network routing (*Ramirez Molina et al., 2025*).

[153] A key idea underlying the success of our approach is the incorporation of cell-state info-sharing within the gating functions—an idea drawn from LSTMs, but also reminiscent of hypernetworks, where one network

generates the weights for another (*Ha et al., 2016*). We believe that the proposed MCP-based physically-interpretable neural network approach can support both the traditional local-scale development of multilayer perceptron-based hydrologic models (*Hsu et al., 1995*) and the current paradigm of employing a single network to represent diverse hydrologic processes across continental and global scales as is popular in modern deep learning applications (*Kratzert et al., 2019; Feng et al., 2020; Kratzert et al., 2024*). The former can be viewed as analogous to the use of Kolmogorov–Arnold network theory combined with recurrence (*Liu et al., 2024c,d*), while the latter aligns with the recently developed CONUS-wide differentiable hydrologic model paradigm (*Feng et al., 2022*) and the regionalization of differentiable modeling units (*De la Fuente et al., 2024b*).

[154] Finally, owing to the relative parsimony and physical interpretability of the Mass-Conserving Perceptron (MCP)—rooted in the principles and language of modern recurrent neural network theory—we believe that the MCP can effectively support scientific inference and discovery. We further anticipate that MCP-based modeling will play a significant role in the future of machine learning and its applications to geoscientific investigation.

## Acknowledgments

The authors would like to thank the WRR editorial team including Chiyuan Miao (Editor), an anonymous Associate Editor and three anonymous reviewers for taking the time to provide constructive comments. The first author (YHW) would like to thank the late *Thomas Meixner*, as well as *Jennifer Mcintosh*, *Martha Whitaker*, *Eyad Atallah*, *Dale Ward*, *Ty Ferré*, *Jim Yeh*, and *Chris Castro* for their support, and to acknowledge the teaching and outreach assistantship support provided by the *Department of Hydrology and Atmospheric Sciences* and the *University of Arizona Data Science Institute* during the final two years of his Ph.D. study, which made the finalization of this work possible. We also thank the *University of Arizona Data Science Institute* for providing HPC computation resources. We particularly thank *Yang Yang* and *Patrick Broxton* for their assistance, guidance, and insightful discussions regarding data preparation for the large-sample study, *Derrick Zwickl, Sara Willis* and *Ethan Jahn* for their invaluable consultation regarding high-performance computing at the *University of Arizona, Frederik Kratzert* for the open-source LSTM implementation that serves as important learning material, and *Chaopeng Shen, Dipankar Dwivedi, Xingyuan Chen*, and *Jasper Vrugt* for organizing the *Hydro-ML Conferences* that made the finalization of this series of studies possible. The second author (HVG) acknowledges partial support by the *Australian Centre of Excellence for Climate System Science* (CE110001028), the inspiration and encouragement provided by members of the *Information Theory in the Geosciences group* (geoinfotheory.org), and support for a 4-month research visit to the *Karlsruhe Institute of Technology*, Germany provided by the *KIT International Excellence Fellowship Award* program.

## Conflict of Interest

The authors declare no conflicts of interest relevant to this study.

## Open Research

The code and the dataset used in this study are available in the Zenodo repository (*Wang & Gupta, 2025*).

# Table & Figure in the Main Text

## Table Captions

**Table 1.** Summary of Single-Layer Mass-Conserving Neural Network ( $MN_s$ ) architectural hypothesis in this study

**Table 2.** $KGE_{ss}$ scores for the Single-Layer Mass-Conserving Neural Networks (MNs)

**Table 3.** $KGE_{ss}$ Statistics and Numbers of Parameter for the single-layer Mass-Conserving Networks and the single-layer LSTM Networks

**Table 4.** $KGE_{ss}$ Statistics and Numbers of Parameter for cases used for the benchmark comparison

## Figure Captions

**Figure 1:** Directed-graph representation and mathematical formulation of the modeling network used in this study, including (a) linear reservoir, (b) mass-conserving perceptron, and (c) long short-term memory network. The green arrow reflects a shift toward greater model complexity and reduced interpretability, whereas the purple arrow reflects the reverse trend—toward simpler models with improved interpretability. The text within the green and purple boxes elaborates on the architectural differences associated with each direction. Although notation varies across frameworks, there exists a functional isomorphism between components of RNN-based LSTM models and physics-informed models such as LR or MCP. The input, forget, and output gates— $i[t]$,$f[t]$, $o[t]$ —correspond functionally to $G_t^U$, $G_t^R$, $G_t^O$ respectively. The $l[t]$ is the newly introduced loss gate denoted as $G_t^L$. The LSTM cell state $c[t]$ is analogous to the internal state variable $X[t]$ in LR/MCP. Similarly, the external input $u[t]$ aligns with the LSTM input $x[t]$, and the internal update $g[t]$ resembles the candidate state update. The final outputs— $h[t]$ in LSTM and $O[t]$ in LR/MCP—also serve functionally equivalent roles. Note that the remaining symbols— $\kappa_o, a_o, b_o, a_L, b_L b_L^*, \kappa_L, a_i, b_i, w_i, a_f, b_f, w_f, a_g, b_g, w_g$ —represent trainable parameters associated with the model architecture. $\odot$ is element-wise multiplication.

**Figure 2:** Visualization of the various types of Mass-Conserving Networks (MNs) utilized in this study, including: (a) A conceptual (four-node) interpretation of the single-layer (mass-conserving) distributed-input (DI) and (b) (mass-conserving) distributed-state (DS) MCP-based representations of catchment-scale rainfall-storage-runoff dynamics. The input ($U_t$) at the current timestep (day) includes precipitation ($P_t$) and potential evapotranspiration ($PET_t$). The output ($O_t$) represents streamflow at the current timestep (day). For a single-layer four-node DI case, precipitation is distributed across the different nodes of the system using the Softmax function applied at the input layer, where the weights satisfy $\sum w_j^{in} = 1$. The total output is obtained by summing the streamflow components from all flow paths ($w_k^{out} = 1$). For the single-layer four-node DS case, precipitation is equally distributed to all nodes ($w_j^{in} = 1$), with input layer weights all set to 1. The total output is a weighted sum of streamflow components from all flow paths, where the weights are determined by the Softmax function ( $\sum w_k^{out} = 1$ ). (c) An example mathematical formulation, illustrated using a single-layer two-node configuration, summarizes the five different cases tested. The differences among these cases stem from the use of linear weights at both the input ($w_j^{in}$) and output ($w_k^{out}$) layers. These include the mass-conserving

configurations—distributed-input (DI) and distributed-state (DS)—as shown in part (a). The distributed-input-relaxed (DIR) and distributed-state-relaxed (DSR) cases allow mass relaxation by relaxing the values of weights at the input and output layers, respectively. The machine learning benchmark (MLB) case combines both DIR and DSR, with an additional bias term $w_0^{out}$ allowing mass relaxation at both the input and output layers respectively.

**Figure 3:** Box and whisker plots showing medians, interquartile ranges and 95% confidence intervals of the distributions of annual $\mathrm{KGE_{ss}}$ scores for network performance. From left to right the subplots show (a) single MCP nodes, and the single-layer network cases for (b) Distributed Input (DI), (c) Distributed State (DS), (d) Distributed Input Relaxed (DIR), (e) Distributed State Relaxed (DSR), and (f) Machine Learning Benchmark, with the number of nodes varying from 1 to 5. The gray shading indicates the number of parallel nodes used: lighter shading indicates larger numbers of nodes. Blue shading indicates the baseline $MC\{O_\sigma L_{\sigma+}^{con}\}$ case, while green shading represents the "best" case selected based on the overall performance of the $\mathrm{KGE_{ss}}$ distribution. The solid line inside the box indicates the median, and the box edges mark the interquartile range (IQR; Range between 25th to 75th percentile). The red "+" symbol denotes outliers. Dashed whiskers represent the 5th and 95th percentiles when outliers are present, or the minimum and maximum values if there are no outliers. The lower whisker extends to the smallest data point greater than or equal to $\mathrm{Q_1} - 1.5 \times \mathrm{IQR}$ and the upper whisker extends to the largest data point less than or equal to $\mathrm{Q_3} + 1.5 \times \mathrm{IQR}$. These properties apply to all the box-whisker plots shown throughout this study. *Baseline Median Year (blue dash line)*: the median-year (among all 40 years) $\mathrm{KGE_{ss}}$ score achieved by $MC\{O_\sigma L_{\sigma+}^{con}\}$ networks; *Baseline Worst Year (red dot line)*: the worst-year (among all 40 years) $\mathrm{KGE_{ss}}$ score achieved by $MC\{O_\sigma L_{\sigma+}^{con}\}$ networks; *Best Median Year (green dash line)*: the highest median-year (among all 40 years) $\mathrm{KGE_{ss}}$ score achievable by any of the single-layer MN networks; *Best of Best Year (black dash line)*: the highest single-year $\mathrm{KGE_{ss}}$ score achievable by any of the single-layer MN networks. Yellow-Filled Red Circles: the worst-year (among all 40 years) $\mathrm{KGE_{ss}}$ score derived from each single-layer MN network.

**Figure 4:** Performance comparison of selected single-layer Mass-Conserving Networks, showing (a) box and whisker plots of the distributions of annual KGE scores, and (b - d) associated hydrographs for dry (WY 1952), median (WY 1953), and wet years (WY 1974). In subplot (a), the blue dashed line indicates the median of baseline case, green dashed line indicates the median of the best case, and red dashed line indicates the worst year of the worst case. The skills of annual $KGE_{ss}$, root mean square error ($RMSE$), and mean absolute error ($MAE$) for each model are summarized for each year. The annual performance metrics—$KGE_{ss}$, root mean square error ($MSE$), and mean absolute error ($MAE$)—are summarized for each model by dry, median, and wet year. In subplot (a), *Best Median Year (green dash line)*: the highest median-year (among all 40 years) $\mathrm{KGE_{ss}}$ score achievable by any of the single-layer MN networks; *Best of Best Year (black dash line)*: the highest single-year $\mathrm{KGE_{ss}}$ score achievable by any of the single-layer MN networks. The blue dashed line and red dashed line represent the baseline median year and the worst year, respectively. Yellow-Filled Red Circles: the worst-year (among all 40 years) $\mathrm{KGE_{ss}}$ score derived from each single-layer MN network.

**Figure 5:** Box and whisker plots of the distributions of annual $\mathrm{KGE_{ss}}$ performance values for the Single- and Multi-Layer Networks. (a) Distributed-State (DS), (b) Distributed-State Relaxed (DSR), and (c) Distributed-State Machine Learning Benchmark. Each network architecture includes 41

cases, with the number of nodes in the first layer varying from 1 to 3, and in the second and third layers from 0 to 3. Single-layer cases with 4 and 5 nodes are also included for comparison. Note that cases using the same number of cell states are shown using the same color scheme, and the best-performing case—with two layers and three nodes in each layer—is highlighted with a red dashed box.

**Figure 6:** Box and whisker plots of the distributions of annual $\mathrm{KGE_{ss}}$ performance values for single-layer Sharing-Augmented Output Gates (SAO) networks. (a) Distributed Input (DI), (b) Distributed State (DS), (c) Distributed Input Relaxed (DIR), (d) Distributed State Relaxed (DSR), and (e) Machine-Learning Benchmark (MLB). The number of nodes varies from 1 to 5. *Baseline Median Year (blue dash line)*: the median-year (among all 40 years) $\mathrm{KGE_{ss}}$ score achieved by $MC\{O_\sigma L_{\sigma+}^{con}\}$ networks; *Baseline Worst Year (red dot line)*: the worst-year (among all 40 years) $\mathrm{KGE_{ss}}$ score achieved by $MC\{O_\sigma L_{\sigma+}^{con}\}$ networks; *Best Median Year (green dash line)*: the highest median-year (among all 40 years) $\mathrm{KGE_{ss}}$ score achieved by any of the single-layer MN networks shown here. Yellow-Filled Red Circles: The worst-year (among all 40 years) $\mathrm{KGE_{ss}}$ score is derived from each single-layer MN network.

**Figure 7:** Box and whisker plots of the distributions of annual $\mathrm{KGE_{ss}}$ performance values for single-layer Sharing-Augmented Loss Gates (SAL) networks. (a) Distributed Input (DI), (b) Distributed State (DS), (c) Distributed Input Relaxed (DIR), (d) Distributed State Relaxed (DSR), and (e) Machine-Learning Benchmark (MLB). The number of nodes varies from 1 to 5. *Baseline Median Year (blue dash line)*: the median-year (among all 40 years) $\mathrm{KGE_{ss}}$ score achieved by $MC\{O_\sigma L_{\sigma+}^{con}\}$ networks; *Baseline Worst Year (red dot line)*: the worst-year (among all 40 years) $\mathrm{KGE_{ss}}$ score achieved by $MC\{O_\sigma L_{\sigma+}^{con}\}$ networks; *Best Median Year (green dash line)*: the highest median-year (among all 40 years) $\mathrm{KGE_{ss}}$ score achieved by any of the single-layer MN networks shown here.

**Figure 8:** Box and whisker plots of the distributions of annual $\mathrm{KGE_{ss}}$ performance values for single-layer Sharing-Augmented Loss & Output Gates (SALO) networks. (a) Distributed Input (DI), (b) Distributed State (DS), (c) Distributed Input Relaxed (DIR), (d) Distributed State Relaxed (DSR), and (e) Machine-Learning Benchmark (MLB). The number of nodes varies from 1 to 5. *Baseline Median Year (blue dash line)*: the median-year (among all 40 years) $\mathrm{KGE_{ss}}$ score achieved by $MC\{O_\sigma L_{\sigma+}^{con}\}$ networks; *Baseline Worst Year (red dot line)*: the worst-year (among all 40 years) $\mathrm{KGE_{ss}}$ score achieved by $MC\{O_\sigma L_{\sigma+}^{con}\}$ networks; *Best Median Year (green dash line)*: the highest median-year (among all 40 years) $\mathrm{KGE_{ss}}$ score achieved by any of the single-layer MN networks shown here.

**Figure 9:** Box and whisker plots of the distributions of annual $\mathrm{KGE_{ss}}$ performance values for Single- and Multi-Layer Sharing-Augmented Loss & Output Gates Networks. (a) Distributed-State (DS), (b) Distributed-State Relaxation (DSR), and (c) Distributed-State Machine Learning Benchmark Case. Each network architecture includes 41 cases, with the number of nodes in the first layer varying from 1 to 3, and in the second and third layers from 0 to 3. Cases with a single layer and 4 or 5 nodes are also included for comparison. Note that cases using the same number of cell states are shown using the same color scheme, and the best-performing case—with single layers and five nodes—is highlighted with a red dashed box.

**Figure 10:** Box and whisker plots of the distributions of annual $KGE_{ss}$ scores, comparing (a) Mass-Conserving Architectures (MAs) reported in *Wang & Gupta (2024b)*, (b) associated MAs with Sharing-Augmented Output Gates, (c) several selected mass-conserving neural networks (MNs) developed in the current study, and (d) the associated architectures with Sharing-Augmented output or loss gates (or both) developed in this study, and (e) LSTM networks with varying numbers of layers and nodes.

**Figure 11:** Illustration of output gate functions for the distributed-state (DS) case $MN_{None}^{DS}(N)$ for number of nodes $(N)$ varying from 1 to 5 (subplots (a) through (e). The color coding represents the peak value of each gate (blue is lowest and green is highest).

**Figure 12:** (a-c) Time series plots of streamflow observations for dry, median, and wet years; (d-f) Corresponding cell-state plots for the single-layer three-node distributed case ($MN_{None}^{DS}(3)$); (g-i) Corresponding output gate series; (j-l) Corresponding accumulated nodal streamflow trajectories; (m-o) Corresponding ratios for accumulated nodal streamflow. Q = streamflow. The nodes are ordered according to the magnitude of the peak values (from small to large) of the output gate, as shown in Figure 11 of the main text.

**Figure 13:** Box and whisker plots of the distributions of annual $\mathrm{KGE}_{\mathrm{ss}}$performance for: (a) single-layer distributed state case $MN_{None}^{DS}(N)$ with number of nodes $N$ varying from 1 to 5; (b) the $MN_{None}^{DS}(5)$ network when pruning 1 to 4 flow paths ($MN_{None}^{DS}(5) - D1(P)$ to $MN_{None}^{DS}(5) - D4(P)$); (c) the single-layer distributed state Sharing-Augmented Loss & Output Gates (SALO) networks with $N$ varying from 1 to 5; (d) the $MN_{SALO}^{DS}(5)$ network when pruning 1 to 4 flow paths ($MN_{SALO}^{DS}(5) - D1(P)$ to $MN_{SALO}^{DS}(5) - D4(P)$); and (e) the $MN_{SALO}^{DS}(5)$ network when pruning 1 to 4 flow paths ($MN_{SALO}^{DS}(5) - D1(F)$ to $MN_{SALO}^{DS}(5) - D4(F)$). *Baseline Median Year (blue dash line)*: the median-year (among all 40 years) $\mathrm{KGE}_{\mathrm{ss}}$ score achieved by $MC\{O_\sigma L_{\sigma+}^{con}\}$ networks; *Baseline Worst Year (red dot line)*: the worst-year (among all 40 years) $\mathrm{KGE}_{\mathrm{ss}}$ score achieved by $MC\{O_\sigma L_{\sigma+}^{con}\}$ networks; *Best Median Year (green dash line)*: the highest median-year (among all 40 years) $\mathrm{KGE}_{\mathrm{ss}}$ score achieved by any of the single-layer MN networks shown here.

**Figure 14:** The skill metrics, including $KGE_{ss}$ and the three components of $KGE$ ($r^{KGE}$, $\alpha^{KGE}$, $\beta^{KGE}$), during the testing period are evaluated for the following models: the single-node $MC\{O_\sigma L_{\sigma+}^{con}\}$, the 3-node distributed-state $MN_{None}^{DS}(3)$, the 5-node distributed-state with cell-state information sharing $MN_{SALO}^{DS}(5)$, the 5-node distributed-state with shared cell-state information and a relaxed evapotranspirative loss constraint $MN_{SALO}^{DS-R_{con}}(5)$, and the single-layer 5-node LSTM network $LSTM(5)$, across the 513 CAMELS US catchments. Subplots (a) to (d) show the cumulative density plots for the $KGE_{ss}$, $r^{KGE}$, $\alpha^{KGE}$, and $\beta^{KGE}$ skill metrics across the five models.

**Figure 15:** The skill metrics, including $KGE_{ss}$ and the three components of $KGE$ ($r^{KGE}$, $\alpha^{KGE}$, $\beta^{KGE}$), during the testing period are evaluated for the following models: the single-node $MC\{O_\sigma L_{\sigma+}^{con}\}$, the 3-node distributed-state $MN_{None}^{DS}(3)$, the 5-node distributed-state with cell-state information sharing $MN_{SALO}^{DS}(5)$, the 5-node distributed-state with shared cell-state information and a relaxed evapotranspirative loss constraint $MN_{SALO}^{DS-R_{con}}(5)$, and the single-layer 5-node LSTM network $LSTM(5)$, across the 513 CAMELS US catchments. Subplots (a) to (d) present the spatial maps of these four metrics for the $MC\{O_\sigma L_{\sigma+}^{con}\}$ model, followed by subplots (e) to (h) for $MN_{None}^{DS}(3)$, subplots (i) to (l) for $MN_{SALO}^{DS}(5)$, subplots (m) to (p) for $MN_{SALO}^{DS-R_{con}}(5)$,

and subplots (q) to (t) for $LSTM(5)$. Triangles, diamonds, squares, and pentagons represent basins in the Sierra Nevada, Cascade, Appalachian, and Rocky Mountain ranges, respectively.

**Figure 16:** Comparison of selected single-layer Mass-Conserving Network performance, showing (a) box and whisker plots of the distributions of annual KGE scores, and (b - d) associated hydrographs for dry (WY 1994), median (WY 1988), and wet years (WY 1997) for $MC\{O_\sigma L_{\sigma+}^{con}\}$ and $MN_{None}^{DS}(3)$, and (f - h) for $MN_{SALO}^{DS-R_{con}}(5)$ and $LSTM(5)$. The skill in terms of annual $KGE_{ss}$, root mean square error ($RMSE$), and mean absolute error ($MAE$) for each model are summarized for each year. The annual performance metrics—$KGE_{ss}$, root mean square error ($MSE$), and mean absolute error ($MAE$)—are summarized for each model by dry, median, and wet year. In subplot (a), *Best Median Year (green dash line)*: the highest median-year (among all 27 years) $\mathrm{KGE_{ss}}$ score achieved by any of the single-layer MN networks; *Best of Best Year (black dash line)*: the highest single-year $\mathrm{KGE_{ss}}$ score achieved by any of the single-layer MN networks. *Baseline (blue dashed line):* median single-year $KGE_{ss}$score of the baseline case. *Best Case (green dashed line):* median single-year $KGE_{ss}$ score of the best case. *Worst Case (red dashed line):* lowest single-year $KGE_{ss}$score of the worst case. *Yellow-Filled Red Circles:* the worst-year (among all 27 years) $\mathrm{KGE_{ss}}$ score derived from each single-layer MN network. Note that the box-and-whisker plots in subplot (a) for the LSTM model are reported only for the last 25 years due to the maximum sequence length of 390 days. This limitation does not affect the three selected years shown in subplots (b)–(g).

**Figure 17:** Output gate functions for the three-node distributed-state (DS) case $MN_{None}^{DS}(3)$. Subplot (a) shows the output gates from each of the three nodes, while subplots (b)–(d) display the individual gate activations Color coding indicates the peak gate value, with blue representing the largest and yellow the smallest.

**Figure 18:** (a-c) Time series plots of precipitation for dry, median, and wet years, and streamflow observations and UA SWE (d-f) for; (g-i) Corresponding cell-state plots for the single-layer three-node distributed case ($MN_{None}^{DS}(3)$); (j-l) Corresponding output gate series; (m-o) Corresponding output gate ratios; (p-r) Corresponding accumulated nodal streamflow; (s-u) Corresponding accumulated nodal streamflow in logarithmic scale; (v-x) Corresponding ratios for accumulated nodal streamflow. Q = streamflow. The nodes are ordered according to the magnitude of the peak values (from large to small) of the output gate, as shown in Figure 16 of the main text.

**Reference in the captions:**

Table 1. Summary of Single-Layer Mass-Conserving Neural Network ($MN_s$) architectural hypothesis in this study

| Model Name | MCP Basic Kernel | Input Distribution Gate | Linear Output Layer | Information (Internal State) Sharing |
|---|---|---|---|---|
| $MN^{DI}_{None}(N_1)$<br>$MN^{DI}_{SAO}(N_1)$<br>$MN^{DI}_{SAL}(N_1)$<br>$MN^{DI}_{SALO}(N_1)$ | | Linear layer with weights sums equal to 1 | Sum operation (combining all the output from each node) | No Sharing<br>Output gate only<br>Loss gate only<br>Output and Loss gate |
| $MN^{DS}_{None}(N_1)$<br>$MN^{DS}_{SAO}(N_1)$<br>$MN^{DS}_{SAL}(N_1)$<br>$MN^{DS}_{SALO}(N_1)$ | | Unity | Linear layer with weights sums equal to 1 | No Sharing<br>Output gate only<br>Loss gate only<br>Output and Loss gate |
| $MN^{DIR}_{None}(N_1)$<br>$MN^{DIR}_{SAO}(N_1)$<br>$MN^{DIR}_{SAL}(N_1)$<br>$MN^{DIR}_{SALO}(N_1)$ | $MC\{O_\sigma L^{con}_{\sigma+}\}$ | Linear layer (without bias) with weights to be positive | Sum operation (combining all the output from each node) | No Sharing<br>Output gate only<br>Loss gate only<br>Output and Loss gate |
| $MN^{DSR}_{None}(N_1)$<br>$MN^{DSR}_{SAO}(N_1)$<br>$MN^{DSR}_{SAL}(N_1)$<br>$MN^{DSR}_{SALO}(N_1)$ | | Unity | Linear layer (without bias) with weights to be positive | No Sharing<br>Output gate only<br>Loss gate only<br>Output and Loss gate |
| $MN^{MLB}_{None}(N_1)$<br>$MN^{MLB}_{SAO}(N_1)$<br>$MN^{MLB}_{SAL}(N_1)$<br>$MN^{MLB}_{SALO}(N_1)$ | | Linear layer (w/o bias term) without constraints on the weights | Linear layer (with bias term) without constraints on weights and bias | No Sharing<br>Output gate only<br>Loss gate only<br>Output and Loss gate |

Table 2. $KGE_{ss}$ scores for the Single-Layer Mass-Conserving Neural Networks (MNs)

**1-Node**

| | $MN_{None}^{DI}$ | $MN_{None}^{DS}$ | $MN_{None}^{DIR}$ | $MN_{None}^{DSR}$ | $MN_{None}^{MLB}$ | | $MN_{SAL}^{DI}$ | $MN_{SAL}^{DS}$ | $MN_{SAL}^{DIR}$ | $MN_{SAL}^{DSR}$ | $MN_{SAL}^{MLB}$ | | $MN_{SAO}^{DI}$ | $MN_{SAO}^{DS}$ | $MN_{SAO}^{DIR}$ | $MN_{SAO}^{DSR}$ | $MN_{SAO}^{MLB}$ | | $MN_{SALO}^{DI}$ | $MN_{SALO}^{DS}$ | $MN_{SALO}^{DIR}$ | $MN_{SALO}^{DSR}$ | $MN_{SALO}^{MLB}$ |
|---|---|---|---|---|---|---|---|---|---|---|---|---|---|---|---|---|---|---|---|---|---|---|---|
| $KGE_{ss}^{min}$ | 0.57 | 0.57 | 0.57 | 0.55 | 0.36 | $KGE_{ss}^{min}$ | | | | | | $KGE_{ss}^{min}$ | | | | | | $KGE_{ss}^{min}$ | | | | | |
| $KGE_{ss}^{5\%}$ | 0.70 | 0.70 | 0.68 | 0.66 | 0.42 | $KGE_{ss}^{5\%}$ | | | | | | $KGE_{ss}^{5\%}$ | | | | | | $KGE_{ss}^{5\%}$ | | | | | |
| $KGE_{ss}^{25\%}$ | 0.81 | 0.81 | 0.81 | 0.80 | 0.64 | $KGE_{ss}^{25\%}$ | | | | | | $KGE_{ss}^{25\%}$ | | | | | | $KGE_{ss}^{25\%}$ | | | | | |
| $KGE_{ss}^{50\%}$ | 0.84 | 0.84 | 0.84 | 0.85 | 0.83 | $KGE_{ss}^{50\%}$ | | | | | | $KGE_{ss}^{50\%}$ | | | | | | $KGE_{ss}^{50\%}$ | | | | | |
| $KGE_{ss}^{75\%}$ | 0.87 | 0.87 | 0.87 | 0.88 | 0.85 | $KGE_{ss}^{75\%}$ | | | | | | $KGE_{ss}^{75\%}$ | | | | | | $KGE_{ss}^{75\%}$ | | | | | |
| $KGE_{ss}^{95\%}$ | 0.91 | 0.91 | 0.91 | 0.91 | 0.91 | $KGE_{ss}^{95\%}$ | | | | | | $KGE_{ss}^{95\%}$ | | | | | | $KGE_{ss}^{95\%}$ | | | | | |
| PNs | 8 | 8 | 9 | 9 | 11 | PNs | | | | | | PNs | | | | | | PNs | | | | | |

**2-Node**

| | $MN_{None}^{DI}$ | $MN_{None}^{DS}$ | $MN_{None}^{DIR}$ | $MN_{None}^{DSR}$ | $MN_{None}^{MLB}$ | | $MN_{SAL}^{DI}$ | $MN_{SAL}^{DS}$ | $MN_{SAL}^{DIR}$ | $MN_{SAL}^{DSR}$ | $MN_{SAL}^{MLB}$ | | $MN_{SAO}^{DI}$ | $MN_{SAO}^{DS}$ | $MN_{SAO}^{DIR}$ | $MN_{SAO}^{DSR}$ | $MN_{SAO}^{MLB}$ | | $MN_{SALO}^{DI}$ | $MN_{SALO}^{DS}$ | $MN_{SALO}^{DIR}$ | $MN_{SALO}^{DSR}$ | $MN_{SALO}^{MLB}$ |
|---|---|---|---|---|---|---|---|---|---|---|---|---|---|---|---|---|---|---|---|---|---|---|---|
| $KGE_{ss}^{min}$ | 0.53 | 0.60 | 0.55 | 0.57 | 0.63 | $KGE_{ss}^{min}$ | 0.48 | 0.63 | 0.56 | 0.54 | 0.65 | $KGE_{ss}^{min}$ | 0.65 | 0.58 | 0.59 | 0.67 | 0.61 | $KGE_{ss}^{min}$ | 0.61 | 0.56 | 0.63 | 0.63 | 0.68 |
| $KGE_{ss}^{5\%}$ | 0.60 | 0.67 | 0.62 | 0.64 | 0.69 | $KGE_{ss}^{5\%}$ | 0.55 | 0.66 | 0.67 | 0.63 | 0.69 | $KGE_{ss}^{5\%}$ | 0.68 | 0.63 | 0.65 | 0.69 | 0.70 | $KGE_{ss}^{5\%}$ | 0.62 | 0.63 | 0.66 | 0.68 | 0.71 |
| $KGE_{ss}^{25\%}$ | 0.79 | 0.79 | 0.82 | 0.77 | 0.82 | $KGE_{ss}^{25\%}$ | 0.78 | 0.79 | 0.83 | 0.76 | 0.82 | $KGE_{ss}^{25\%}$ | 0.80 | 0.76 | 0.81 | 0.81 | 0.81 | $KGE_{ss}^{25\%}$ | 0.81 | 0.77 | 0.82 | 0.80 | 0.82 |
| $KGE_{ss}^{50\%}$ | 0.86 | 0.86 | 0.87 | 0.86 | 0.86 | $KGE_{ss}^{50\%}$ | 0.86 | 0.86 | 0.87 | 0.85 | 0.88 | $KGE_{ss}^{50\%}$ | 0.87 | 0.87 | 0.87 | 0.86 | 0.87 | $KGE_{ss}^{50\%}$ | 0.87 | 0.86 | 0.88 | 0.86 | 0.88 |
| $KGE_{ss}^{75\%}$ | 0.88 | 0.89 | 0.88 | 0.89 | 0.89 | $KGE_{ss}^{75\%}$ | 0.88 | 0.89 | 0.89 | 0.89 | 0.89 | $KGE_{ss}^{75\%}$ | 0.90 | 0.90 | 0.89 | 0.89 | 0.89 | $KGE_{ss}^{75\%}$ | 0.89 | 0.90 | 0.90 | 0.89 | 0.89 |
| $KGE_{ss}^{95\%}$ | 0.90 | 0.92 | 0.92 | 0.92 | 0.93 | $KGE_{ss}^{95\%}$ | 0.90 | 0.92 | 0.92 | 0.92 | 0.93 | $KGE_{ss}^{95\%}$ | 0.92 | 0.94 | 0.92 | 0.94 | 0.93 | $KGE_{ss}^{95\%}$ | 0.93 | 0.94 | 0.92 | 0.95 | 0.93 |
| PNs | 18 | 18 | 18 | 18 | 21 | PNs | 20 | 20 | 20 | 20 | 23 | PNs | 20 | 20 | 20 | 20 | 23 | PNs | 22 | 22 | 22 | 22 | 25 |

**3-Node**

| | $MN_{None}^{DI}$ | $MN_{None}^{DS}$ | $MN_{None}^{DIR}$ | $MN_{None}^{DSR}$ | $MN_{None}^{MLB}$ | | $MN_{SAL}^{DI}$ | $MN_{SAL}^{DS}$ | $MN_{SAL}^{DIR}$ | $MN_{SAL}^{DSR}$ | $MN_{SAL}^{MLB}$ | | $MN_{SAO}^{DI}$ | $MN_{SAO}^{DS}$ | $MN_{SAO}^{DIR}$ | $MN_{SAO}^{DSR}$ | $MN_{SAO}^{MLB}$ | | $MN_{SALO}^{DI}$ | $MN_{SALO}^{DS}$ | $MN_{SALO}^{DIR}$ | $MN_{SALO}^{DSR}$ | $MN_{SALO}^{MLB}$ |
|---|---|---|---|---|---|---|---|---|---|---|---|---|---|---|---|---|---|---|---|---|---|---|---|
| $KGE_{ss}^{min}$ | 0.52 | 0.62 | 0.55 | 0.58 | 0.67 | $KGE_{ss}^{min}$ | 0.47 | 0.49 | 0.53 | 0.49 | 0.62 | $KGE_{ss}^{min}$ | 0.62 | 0.68 | 0.58 | 0.71 | 0.63 | $KGE_{ss}^{min}$ | 0.72 | 0.50 | 0.74 | 0.72 | 0.68 |
| $KGE_{ss}^{5\%}$ | 0.59 | 0.69 | 0.62 | 0.62 | 0.69 | $KGE_{ss}^{5\%}$ | 0.55 | 0.64 | 0.57 | 0.56 | 0.69 | $KGE_{ss}^{5\%}$ | 0.66 | 0.75 | 0.64 | 0.73 | 0.67 | $KGE_{ss}^{5\%}$ | 0.75 | 0.66 | 0.76 | 0.75 | 0.71 |
| $KGE_{ss}^{25\%}$ | 0.78 | 0.79 | 0.82 | 0.78 | 0.82 | $KGE_{ss}^{25\%}$ | 0.78 | 0.82 | 0.78 | 0.80 | 0.82 | $KGE_{ss}^{25\%}$ | 0.81 | 0.86 | 0.81 | 0.85 | 0.82 | $KGE_{ss}^{25\%}$ | 0.84 | 0.82 | 0.84 | 0.83 | 0.85 |
| $KGE_{ss}^{50\%}$ | 0.86 | 0.86 | 0.87 | 0.86 | 0.87 | $KGE_{ss}^{50\%}$ | 0.86 | 0.87 | 0.87 | 0.86 | 0.87 | $KGE_{ss}^{50\%}$ | 0.85 | 0.90 | 0.87 | 0.89 | 0.87 | $KGE_{ss}^{50\%}$ | 0.88 | 0.87 | 0.88 | 0.88 | 0.89 |
| $KGE_{ss}^{75\%}$ | 0.88 | 0.88 | 0.88 | 0.89 | 0.90 | $KGE_{ss}^{75\%}$ | 0.88 | 0.90 | 0.89 | 0.89 | 0.90 | $KGE_{ss}^{75\%}$ | 0.90 | 0.93 | 0.90 | 0.93 | 0.92 | $KGE_{ss}^{75\%}$ | 0.90 | 0.92 | 0.90 | 0.91 | 0.92 |
| $KGE_{ss}^{95\%}$ | 0.90 | 0.92 | 0.92 | 0.92 | 0.92 | $KGE_{ss}^{95\%}$ | 0.90 | 0.92 | 0.92 | 0.92 | 0.92 | $KGE_{ss}^{95\%}$ | 0.94 | 0.96 | 0.93 | 0.96 | 0.94 | $KGE_{ss}^{95\%}$ | 0.94 | 0.95 | 0.94 | 0.94 | 0.95 |
| PNs | 27 | 27 | 27 | 27 | 31 | PNs | 33 | 33 | 33 | 33 | 37 | PNs | 33 | 33 | 33 | 33 | 37 | PNs | 39 | 39 | 39 | 39 | 43 |

**4-Node**

| | $MN_{None}^{DI}$ | $MN_{None}^{DS}$ | $MN_{None}^{DIR}$ | $MN_{None}^{DSR}$ | $MN_{None}^{MLB}$ | | $MN_{SAL}^{DI}$ | $MN_{SAL}^{DS}$ | $MN_{SAL}^{DIR}$ | $MN_{SAL}^{DSR}$ | $MN_{SAL}^{MLB}$ | | $MN_{SAO}^{DI}$ | $MN_{SAO}^{DS}$ | $MN_{SAO}^{DIR}$ | $MN_{SAO}^{DSR}$ | $MN_{SAO}^{MLB}$ | | $MN_{SALO}^{DI}$ | $MN_{SALO}^{DS}$ | $MN_{SALO}^{DIR}$ | $MN_{SALO}^{DSR}$ | $MN_{SALO}^{MLB}$ |
|---|---|---|---|---|---|---|---|---|---|---|---|---|---|---|---|---|---|---|---|---|---|---|---|
| $KGE_{ss}^{min}$ | 0.59 | 0.62 | 0.55 | 0.58 | 0.65 | $KGE_{ss}^{min}$ | 0.53 | 0.63 | 0.57 | 0.46 | 0.64 | $KGE_{ss}^{min}$ | 0.62 | 0.72 | 0.55 | 0.69 | 0.73 | $KGE_{ss}^{min}$ | 0.71 | 0.49 | 0.71 | 0.74 | 0.62 |
| $KGE_{ss}^{5\%}$ | 0.61 | 0.67 | 0.63 | 0.63 | 0.69 | $KGE_{ss}^{5\%}$ | 0.56 | 0.71 | 0.60 | 0.62 | 0.71 | $KGE_{ss}^{5\%}$ | 0.64 | 0.74 | 0.64 | 0.72 | 0.79 | $KGE_{ss}^{5\%}$ | 0.73 | 0.70 | 0.73 | 0.82 | 0.68 |
| $KGE_{ss}^{25\%}$ | 0.80 | 0.79 | 0.82 | 0.77 | 0.82 | $KGE_{ss}^{25\%}$ | 0.81 | 0.83 | 0.80 | 0.81 | 0.83 | $KGE_{ss}^{25\%}$ | 0.81 | 0.87 | 0.83 | 0.86 | 0.85 | $KGE_{ss}^{25\%}$ | 0.84 | 0.85 | 0.84 | 0.88 | 0.84 |
| $KGE_{ss}^{50\%}$ | 0.84 | 0.86 | 0.86 | 0.86 | 0.88 | $KGE_{ss}^{50\%}$ | 0.86 | 0.87 | 0.86 | 0.87 | 0.88 | $KGE_{ss}^{50\%}$ | 0.85 | 0.91 | 0.89 | 0.90 | 0.91 | $KGE_{ss}^{50\%}$ | 0.88 | 0.89 | 0.88 | 0.91 | 0.89 |
| $KGE_{ss}^{75\%}$ | 0.87 | 0.89 | 0.88 | 0.89 | 0.89 | $KGE_{ss}^{75\%}$ | 0.88 | 0.89 | 0.89 | 0.89 | 0.90 | $KGE_{ss}^{75\%}$ | 0.90 | 0.93 | 0.90 | 0.93 | 0.94 | $KGE_{ss}^{75\%}$ | 0.90 | 0.93 | 0.90 | 0.93 | 0.91 |
| $KGE_{ss}^{95\%}$ | 0.91 | 0.92 | 0.92 | 0.92 | 0.93 | $KGE_{ss}^{95\%}$ | 0.91 | 0.92 | 0.92 | 0.92 | 0.93 | $KGE_{ss}^{95\%}$ | 0.94 | 0.97 | 0.93 | 0.96 | 0.96 | $KGE_{ss}^{95\%}$ | 0.94 | 0.96 | 0.94 | 0.97 | 0.95 |
| PNs | 36 | 36 | 36 | 36 | 41 | PNs | 48 | 48 | 48 | 48 | 53 | PNs | 48 | 48 | 48 | 48 | 53 | PNs | 60 | 60 | 60 | 60 | 65 |

**5-Node**

| | $MN_{None}^{DI}$ | $MN_{None}^{DS}$ | $MN_{None}^{DIR}$ | $MN_{None}^{DSR}$ | $MN_{None}^{MLB}$ | | $MN_{SAL}^{DI}$ | $MN_{SAL}^{DS}$ | $MN_{SAL}^{DIR}$ | $MN_{SAL}^{DSR}$ | $MN_{SAL}^{MLB}$ | | $MN_{SAO}^{DI}$ | $MN_{SAO}^{DS}$ | $MN_{SAO}^{DIR}$ | $MN_{SAO}^{DSR}$ | $MN_{SAO}^{MLB}$ | | $MN_{SALO}^{DI}$ | $MN_{SALO}^{DS}$ | $MN_{SALO}^{DIR}$ | $MN_{SALO}^{DSR}$ | $MN_{SALO}^{MLB}$ |
|---|---|---|---|---|---|---|---|---|---|---|---|---|---|---|---|---|---|---|---|---|---|---|---|
| $KGE_{ss}^{min}$ | 0.53 | 0.63 | 0.57 | 0.58 | 0.63 | $KGE_{ss}^{min}$ | 0.55 | 0.59 | 0.77 | 0.55 | 0.64 | $KGE_{ss}^{min}$ | 0.61 | 0.61 | 0.57 | 0.71 | 0.76 | $KGE_{ss}^{min}$ | 0.62 | 0.73 | 0.76 | 0.76 | 0.70 |
| $KGE_{ss}^{5\%}$ | 0.59 | 0.66 | 0.59 | 0.62 | 0.68 | $KGE_{ss}^{5\%}$ | 0.63 | 0.72 | 0.79 | 0.67 | 0.68 | $KGE_{ss}^{5\%}$ | 0.64 | 0.70 | 0.67 | 0.77 | 0.78 | $KGE_{ss}^{5\%}$ | 0.74 | 0.78 | 0.79 | 0.80 | 0.77 |
| $KGE_{ss}^{25\%}$ | 0.78 | 0.78 | 0.80 | 0.77 | 0.82 | $KGE_{ss}^{25\%}$ | 0.80 | 0.81 | 0.84 | 0.79 | 0.83 | $KGE_{ss}^{25\%}$ | 0.81 | 0.85 | 0.83 | 0.87 | 0.88 | $KGE_{ss}^{25\%}$ | 0.85 | 0.87 | 0.85 | 0.88 | 0.87 |
| $KGE_{ss}^{50\%}$ | 0.86 | 0.86 | 0.87 | 0.86 | 0.88 | $KGE_{ss}^{50\%}$ | 0.85 | 0.87 | 0.89 | 0.85 | 0.87 | $KGE_{ss}^{50\%}$ | 0.87 | 0.91 | 0.88 | 0.92 | 0.90 | $KGE_{ss}^{50\%}$ | 0.89 | 0.92 | 0.89 | 0.92 | 0.90 |
| $KGE_{ss}^{75\%}$ | 0.88 | 0.88 | 0.89 | 0.89 | 0.90 | $KGE_{ss}^{75\%}$ | 0.88 | 0.89 | 0.90 | 0.89 | 0.90 | $KGE_{ss}^{75\%}$ | 0.90 | 0.93 | 0.91 | 0.94 | 0.93 | $KGE_{ss}^{75\%}$ | 0.91 | 0.95 | 0.92 | 0.94 | 0.94 |
| $KGE_{ss}^{95\%}$ | 0.90 | 0.92 | 0.93 | 0.92 | 0.93 | $KGE_{ss}^{95\%}$ | 0.91 | 0.92 | 0.93 | 0.92 | 0.92 | $KGE_{ss}^{95\%}$ | 0.94 | 0.96 | 0.93 | 0.96 | 0.96 | $KGE_{ss}^{95\%}$ | 0.94 | 0.97 | 0.94 | 0.96 | 0.96 |
| PNs | 45 | 45 | 45 | 45 | 51 | PNs | 65 | 65 | 65 | 65 | 71 | PNs | 65 | 65 | 65 | 65 | 71 | PNs | 85 | 85 | 85 | 85 | 91 |

*PNs=Parameter Numbers

Table 3. $KGE_{ss}$ Statistics and Numbers of Parameter for the single-layer Mass-Conserving Networks and the single-layer LSTM Networks

| Model Names | $MN^{DS}_{None}(2)$ | $MN^{DS}_{SALO}(2)$ | $MN^{DS}_{None}(3)$ | $MN^{DS}_{SALO}(3)$ | $MN^{DS}_{None}(4)$ | $MN^{DS}_{SALO}(4)$ | $MN^{DS}_{None}(5)$ | $MN^{DS}_{SALO}(5)$ |
|---|---|---|---|---|---|---|---|---|
| $KGE^{min}_{ss}$ | 0.60 | 0.56 | 0.62 | 0.50 | 0.62 | 0.49 | 0.63 | 0.73 |
| $KGE^{5\%}_{ss}$ | 0.67 | 0.63 | 0.69 | 0.66 | 0.66 | 0.70 | 0.66 | 0.78 |
| $KGE^{25\%}_{ss}$ | 0.78 | 0.75 | 0.78 | 0.83 | 0.78 | 0.84 | 0.77 | 0.88 |
| $KGE^{median}_{ss}$ | 0.86 | 0.86 | 0.86 | 0.87 | 0.86 | 0.89 | 0.86 | 0.92 |
| $KGE^{75\%}_{ss}$ | 0.89 | 0.90 | 0.89 | 0.92 | 0.89 | 0.92 | 0.89 | 0.95 |
| $KGE^{95\%}_{ss}$ | 0.92 | 0.94 | 0.92 | 0.95 | 0.92 | 0.96 | 0.92 | 0.96 |
| Par No. | 18 | 22 | 27 | 39 | 36 | 60 | 45 | 85 |
| Model Names | $MN^{DSR}_{None}(2)$ | $MN^{DSR}_{SALO}(2)$ | $MN^{DSR}_{None}(3)$ | $MN^{DSR}_{SALO}(3)$ | $MN^{DSR}_{None}(4)$ | $MN^{DSR}_{SALO}(4)$ | $MN^{DSR}_{None}(5)$ | $MN^{DSR}_{SALO}(5)$ |
| $KGE^{min}_{ss}$ | 0.57 | 0.63 | 0.58 | 0.72 | 0.58 | 0.74 | 0.58 | 0.76 |
| $KGE^{5\%}_{ss}$ | 0.64 | 0.68 | 0.62 | 0.75 | 0.62 | 0.81 | 0.62 | 0.80 |
| $KGE^{25\%}_{ss}$ | 0.77 | 0.80 | 0.76 | 0.82 | 0.76 | 0.88 | 0.76 | 0.88 |
| $KGE^{median}_{ss}$ | 0.86 | 0.86 | 0.86 | 0.88 | 0.86 | 0.91 | 0.86 | 0.92 |
| $KGE^{75\%}_{ss}$ | 0.89 | 0.89 | 0.89 | 0.92 | 0.89 | 0.93 | 0.89 | 0.94 |
| $KGE^{95\%}_{ss}$ | 0.92 | 0.94 | 0.92 | 0.94 | 0.92 | 0.96 | 0.92 | 0.95 |
| Par No. | 18 | 22 | 27 | 39 | 36 | 60 | 45 | 85 |
| Model Names | $MN^{MLB}_{None}(2)$ | $MN^{MLB}_{SALO}(2)$ | $MN^{MLB}_{None}(3)$ | $MN^{MLB}_{SALO}(3)$ | $MN^{MLB}_{None}(4)$ | $MN^{MLB}_{SALO}(4)$ | $MN^{MLB}_{None}(5)$ | $MN^{MLB}_{SALO}(5)$ |
| $KGE^{min}_{ss}$ | 0.57 | 0.70 | 0.56 | 0.69 | 0.55 | 0.71 | 0.57 | 0.76 |
| $KGE^{5\%}_{ss}$ | 0.61 | 0.74 | 0.61 | 0.73 | 0.58 | 0.80 | 0.59 | 0.80 |
| $KGE^{25\%}_{ss}$ | 0.80 | 0.84 | 0.80 | 0.84 | 0.79 | 0.88 | 0.79 | 0.88 |
| $KGE^{median}_{ss}$ | 0.86 | 0.87 | 0.85 | 0.89 | 0.86 | 0.92 | 0.86 | 0.91 |
| $KGE^{75\%}_{ss}$ | 0.90 | 0.89 | 0.89 | 0.93 | 0.90 | 0.94 | 0.90 | 0.93 |
| $KGE^{95\%}_{ss}$ | 0.92 | 0.95 | 0.91 | 0.95 | 0.91 | 0.96 | 0.91 | 0.97 |
| Par No. | 19 | 23 | 28 | 40 | 37 | 61 | 46 | 86 |
| Model Names | $LSTM(2)$ | | $LSTM(3)$ | | $LSTM(4)$ | | $LSTM(5)$ | |
| $KGE^{min}_{ss}$ | 0.68 | | 0.66 | | 0.69 | | 0.66 | |
| $KGE^{5\%}_{ss}$ | 0.72 | | 0.75 | | 0.76 | | 0.78 | |
| $KGE^{25\%}_{ss}$ | 0.83 | | 0.84 | | 0.84 | | 0.85 | |
| $KGE^{median}_{ss}$ | 0.87 | | 0.88 | | 0.89 | | 0.90 | |
| $KGE^{75\%}_{ss}$ | 0.92 | | 0.93 | | 0.93 | | 0.93 | |
| $KGE^{95\%}_{ss}$ | 0.95 | | 0.96 | | 0.96 | | 0.98 | |
| Par No. | 43 | | 76 | | 117 | | 166 | |

- Note: Par No.=Parameter Number

Table 4. $KGE_{ss}$ Statistics and Numbers of Parameter for cases used for the benchmark comparison

| Model Names | $MA_3$ | $MA_4$ | $MA_5$ | $MA_6$ | $MA_{3-SAO}$ | $MA_{4-SAO}$ | $MA_{5-SAO}$ | $MA_{6-SAO}$ | $MN_{None}^{DS}(2,0,0)$ | $MN_{None}^{DS}(3,0,0)$ |
|---|---|---|---|---|---|---|---|---|---|---|
| $KGE_{ss}^{min}$ | 0.30 | 0.58 | 0.58 | 0.57 | 0.57 | 0.59 | 0.65 | 0.57 | 0.60 | 0.62 |
| $KGE_{ss}^{5\%}$ | 0.46 | 0.60 | 0.59 | 0.62 | 0.59 | 0.61 | 0.66 | 0.59 | 0.67 | 0.69 |
| $KGE_{ss}^{25\%}$ | 0.75 | 0.79 | 0.77 | 0.81 | 0.80 | 0.81 | 0.82 | 0.83 | 0.78 | 0.78 |
| $KGE_{ss}^{median}$ | 0.82 | 0.86 | 0.84 | 0.86 | 0.86 | 0.86 | 0.87 | 0.88 | 0.86 | 0.86 |
| $KGE_{ss}^{75\%}$ | 0.87 | 0.90 | 0.88 | 0.91 | 0.90 | 0.91 | 0.90 | 0.92 | 0.89 | 0.89 |
| $KGE_{ss}^{95\%}$ | 0.91 | 0.94 | 0.92 | 0.94 | 0.94 | 0.95 | 0.95 | 0.96 | 0.92 | 0.92 |
| Par No. | 11 | 14 | 18 | 21 | 13 | 16 | 24 | 27 | 18 | 27 |
| Model Names | $MN_{None}^{DS}(3,3,0)$ | $MN_{None}^{DSR}(2,1,0)$ | $MN_{None}^{DSR}(3,1,0)$ | $MN_{None}^{DS-MLB}(2,3,0)$ | $MN_{SALO}^{DS}(5,0,0)$ | $MN_{SALO}^{DSR}(5,0,0)$ | $MN_{SALO}^{DS-MLB}(5,0,0)$ | $LSTM(5)$ | $LSTM(6)$ | $LSTM(5,5,5)$ |
| $KGE_{ss}^{min}$ | 0.72 | 0.72 | 0.65 | 0.75 | 0.73 | 0.76 | 0.76 | 0.66 | 0.68 | 0.59 |
| $KGE_{ss}^{5\%}$ | 0.74 | 0.73 | 0.69 | 0.77 | 0.78 | 0.80 | 0.80 | 0.78 | 0.78 | 0.66 |
| $KGE_{ss}^{25\%}$ | 0.84 | 0.82 | 0.82 | 0.80 | 0.88 | 0.88 | 0.88 | 0.85 | 0.84 | 0.87 |
| $KGE_{ss}^{median}$ | 0.87 | 0.85 | 0.86 | 0.84 | 0.92 | 0.92 | 0.91 | 0.90 | 0.90 | 0.91 |
| $KGE_{ss}^{75\%}$ | 0.90 | 0.90 | 0.91 | 0.89 | 0.95 | 0.94 | 0.93 | 0.93 | 0.93 | 0.93 |
| $KGE_{ss}^{95\%}$ | 0.94 | 0.93 | 0.94 | 0.92 | 0.96 | 0.95 | 0.97 | 0.98 | 0.96 | 0.97 |
| Par No. | 60 | 27 | 36 | 50 | 85 | 85 | 86 | 166 | 223 | 486 |

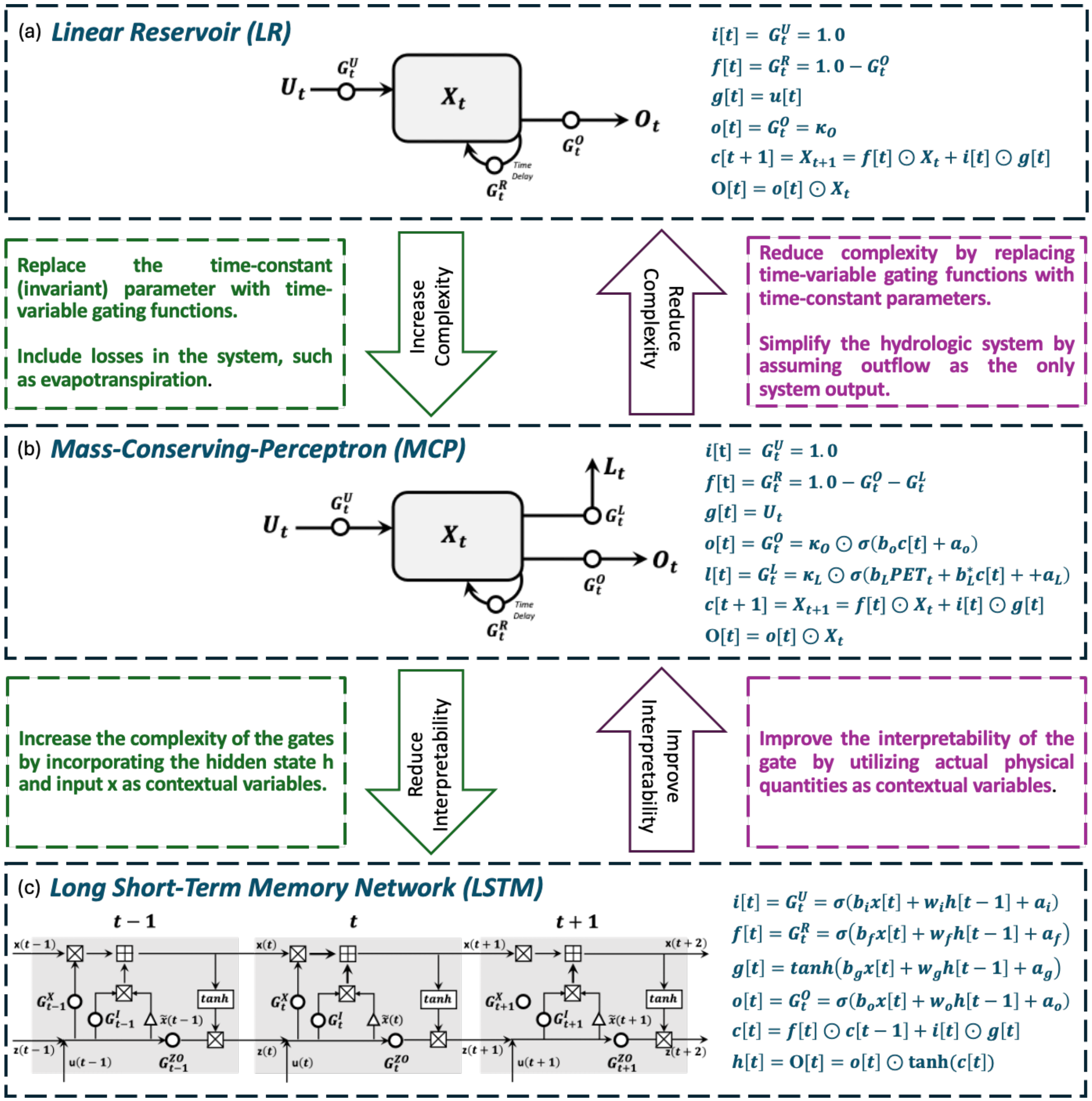

**Figure 1:** Directed-graph representation and mathematical formulation of the modeling network used in this study, including (a) linear reservoir, (b) mass-conserving perceptron, and (c) long short-term memory network. The green arrow reflects a shift toward greater model complexity and reduced interpretability, whereas the purple arrow reflects the reverse trend—toward simpler models with improved interpretability. The text within the green and purple boxes elaborates on the architectural differences associated with each direction. Although notation varies across frameworks, there exists a functional isomorphism between components of RNN-based LSTM models and physics-informed models such as LR or MCP. The input, forget, and output gates— $i[t]$,$f[t]$, $o[t]$ —correspond functionally to $G_t^U$, $G_t^R$, $G_t^O$ respectively. The $l[t]$ is the newly introduced loss gate denoted as $G_t^L$. The LSTM cell state $c[t]$ is analogous to the internal state variable $X[t]$ in LR/MCP. Similarly, the external input $u[t]$ aligns with the LSTM input $x[t]$, and the internal update $g[t]$ resembles the candidate state update. The final outputs— $h[t]$ in LSTM and $O[t]$ in LR/MCP—also serve functionally equivalent roles. Note that the remaining symbols— $\kappa_o, a_o, b_o, a_L, b_L b_L^*, \kappa_L, a_i, b_i, w_i, a_f, b_f, w_f, a_g, b_g, w_g$ — represent trainable parameters associated with the model architecture. $\odot$ is element-wise multiplication.

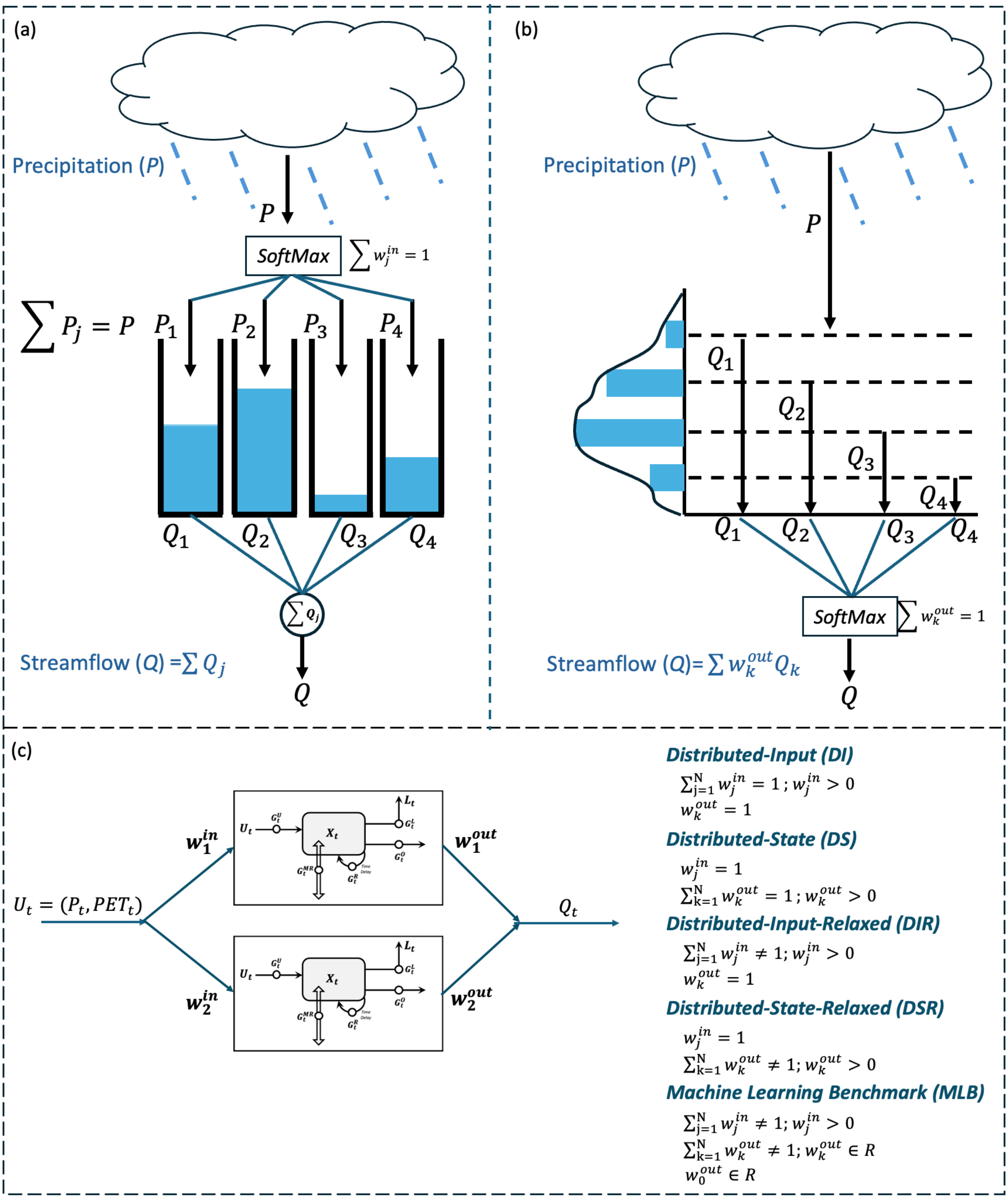

**Figure 2:** Visualization of the various types of Mass-Conserving Networks (MNs) utilized in this study, including: (a) A conceptual (four-node) interpretation of the single-layer (mass-conserving) *distributed-input* (DI) and (b) (mass-conserving) *distributed-state* (DS) MCP-based representations of catchment-scale rainfall-storage-runoff dynamics. The input ($U_t$) at the current timestep (day) includes precipitation ($P_t$) and potential evapotranspiration ($PET_t$). The output ($O_t$) represents streamflow at the current timestep (day). For a single-layer four-node DI case, precipitation is distributed across the different nodes of the system using the *Softmax* function applied at the input layer, where the weights satisfy $\sum w_j^{in} = 1$. The total output is obtained by summing the streamflow components from all flow paths ($w_k^{out} = 1$). For the single-layer four-node DS case, precipitation is equally distributed to all nodes ($w_j^{in} = 1$), with input layer weights all set to 1. The total output is a weighted sum of streamflow components from all flow paths, where the weights are determined by the *Softmax* function ($\sum w_k^{out} = 1$). (c) An example mathematical formulation, illustrated using a single-layer two-node configuration, summarizes the five different cases tested. The differences among these cases stem from the use of linear weights at both the input ($w_j^{in}$) and output ($w_k^{out}$) layers. These include the mass-conserving configurations—distributed-input (DI) and distributed-state (DS)—as shown in part (a). The distributed-input-relaxed (DIR) and distributed-state-relaxed (DSR) cases allow mass relaxation by relaxing the values of weights at the input and output layers, respectively. The machine learning benchmark (MLB) case combines both DIR and DSR, with an additional bias term $w_0^{out}$ allowing mass relaxation at both the input and output layers respectively.

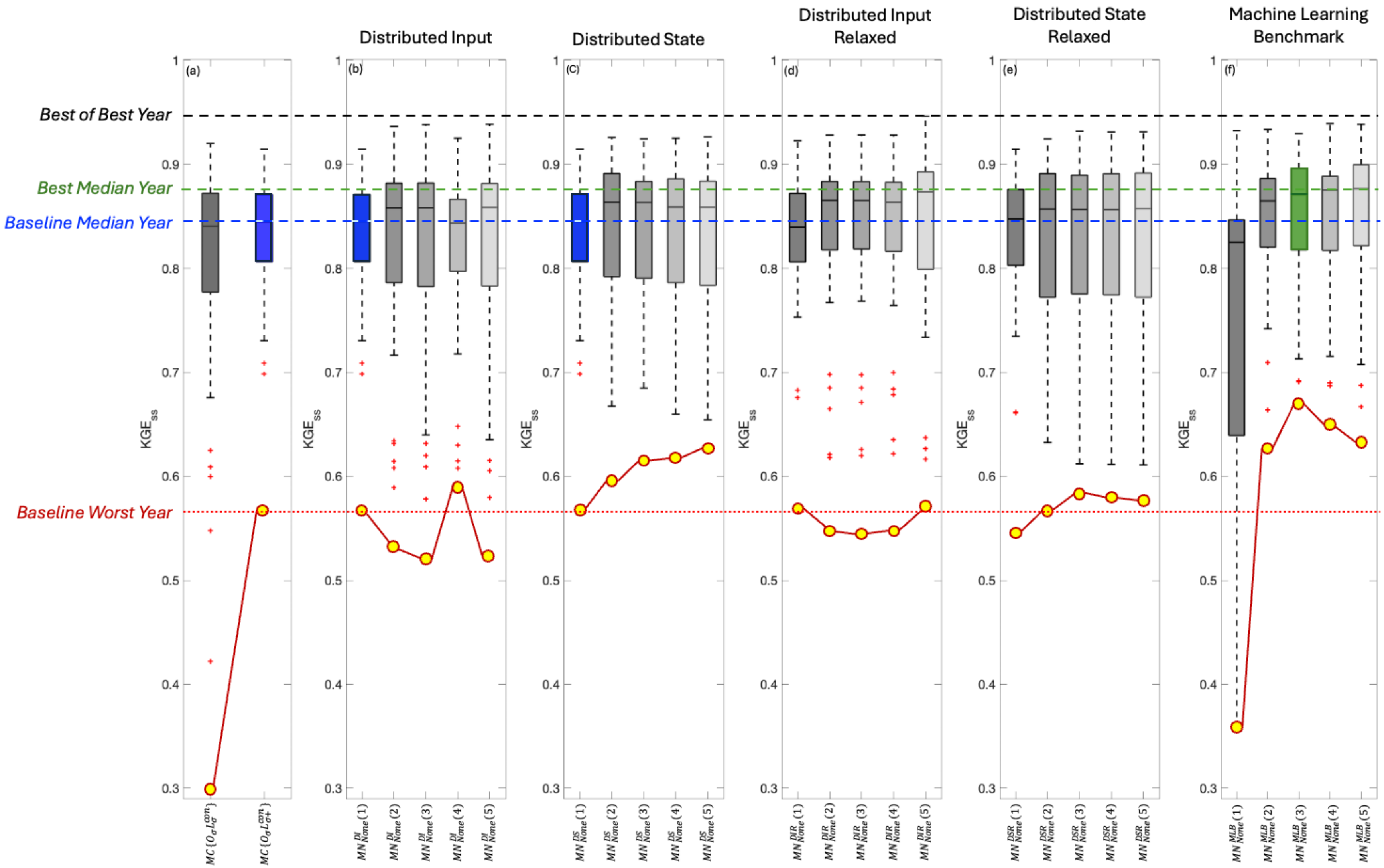

**Figure 3:** Box and whisker plots showing medians, interquartile ranges and 95% confidence intervals of the distributions of annual $\mathrm{KGE_{ss}}$ scores for network performance. From left to right the subplots show (a) single MCP nodes, and the single-layer network cases for (b) Distributed Input (DI), (c) Distributed State (DS), (d) Distributed Input Relaxed (DIR), (e) Distributed State Relaxed (DSR), and (f) Machine Learning Benchmark, with the number of nodes varying from 1 to 5. The gray shading indicates the number of parallel nodes used: lighter shading indicates larger numbers of nodes. Blue shading indicates the baseline $MC\{O_\sigma L_{\sigma+}^{con}\}$ case, while green shading represents the "best" case selected based on the overall performance of the $\mathrm{KGE_{ss}}$ distribution. The solid line inside the box indicates the median, and the box edges mark the interquartile range (IQR; Range between 25th to 75th percentile). The red "+" symbol denotes outliers. Dashed whiskers represent the 5th and 95th percentiles when outliers are present, or the minimum and maximum values if there are no outliers. The lower whisker extends to the smallest data point greater than or equal to $\mathrm{Q_1} - 1.5 \times \mathrm{IQR}$ and the upper whisker extends to the largest data point less than or equal to $\mathrm{Q_3} + 1.5 \times \mathrm{IQR}$. These properties apply to all the box-whisker plots shown throughout this study. *Baseline Median Year (blue dash line)*: the median-year (among all 40 years) $\mathrm{KGE_{ss}}$ score achieved by $MC\{O_\sigma L_{\sigma+}^{con}\}$ networks; *Baseline Worst Year (red dot line)*: the worst-year (among all 40 years) $\mathrm{KGE_{ss}}$ score achieved by $MC\{O_\sigma L_{\sigma+}^{con}\}$ networks; *Best Median Year (green dash line)*: the highest median-year (among all 40 years) $\mathrm{KGE_{ss}}$ score achievable by any of the single-layer MN networks; *Best of Best Year (black dash line)*: the highest single-year $\mathrm{KGE_{ss}}$ score achievable by any of the single-layer MN networks. Yellow-Filled Red Circles: the worst-year (among all 40 years) $\mathrm{KGE_{ss}}$ score derived from each single-layer MN network.

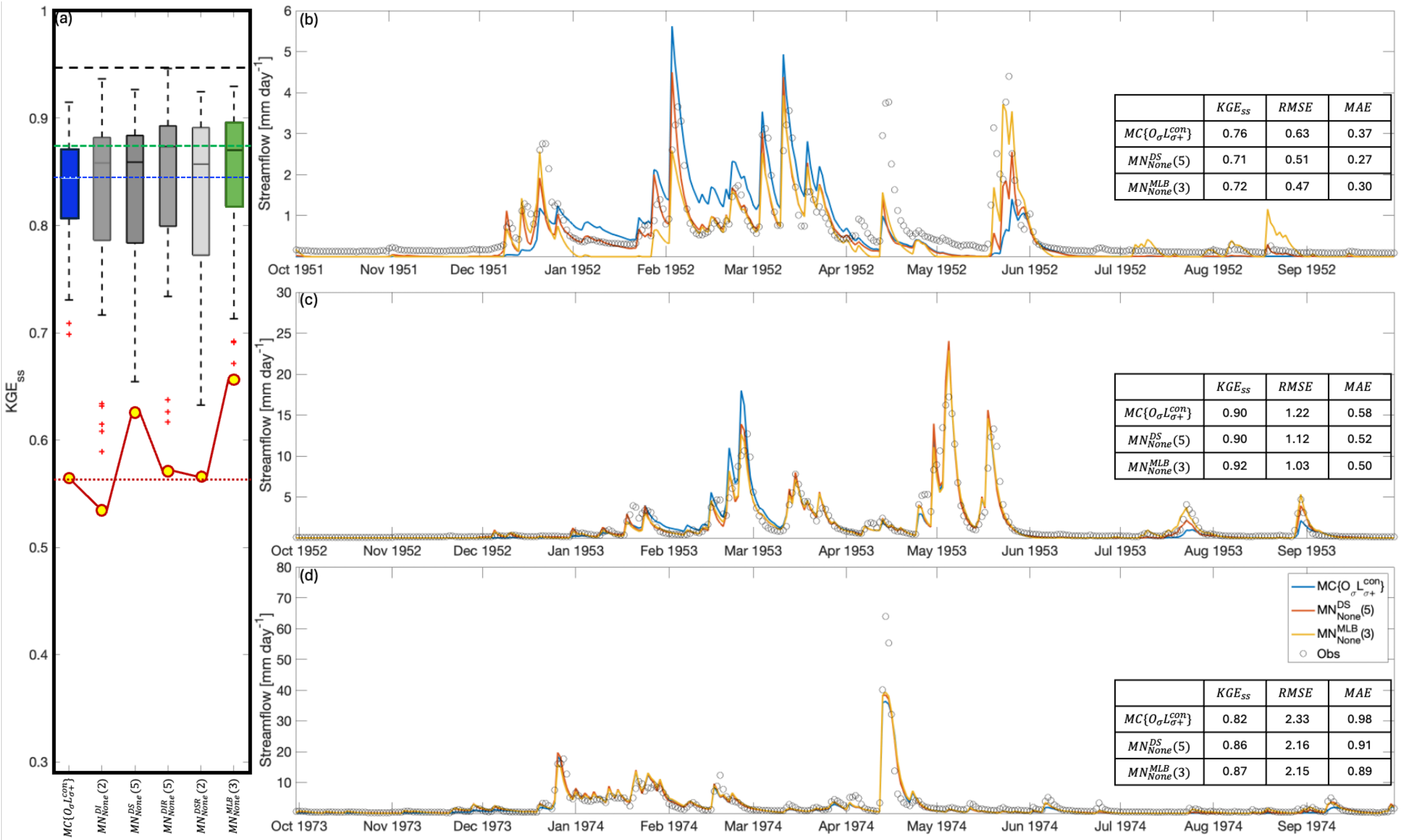

**Figure 4:** Performance comparison of selected single-layer Mass-Conserving Networks, showing (a) box and whisker plots of the distributions of annual KGE scores, and (b - d) associated hydrographs for dry (WY 1952), median (WY 1953), and wet years (WY 1974). In subplot (a), the blue dashed line indicates the median of baseline case, green dashed line indicates the median of the best case, and red dashed line indicates the worst year of the worst case. The skills of annual $KGE_{ss}$, root mean square error ($RMSE$), and mean absolute error ($MAE$) for each model are summarized for each year. The annual performance metrics—$KGE_{ss}$, root mean square error ($MSE$), and mean absolute error ($MAE$)—are summarized for each model by dry, median, and wet year. In subplot (a), *Best Median Year (green dash line)*: the highest median-year (among all 40 years) $\mathrm{KGE}_{ss}$ score achievable by any of the single-layer MN networks; *Best of Best Year (black dash line)*: the highest single-year $\mathrm{KGE}_{ss}$ score achievable by any of the single-layer MN networks. The blue dashed line and red dashed line represent the baseline median year and the worst year, respectively. Yellow-Filled Red Circles: the worst-year (among all 40 years) $\mathrm{KGE}_{ss}$ score derived from each single-layer MN network.

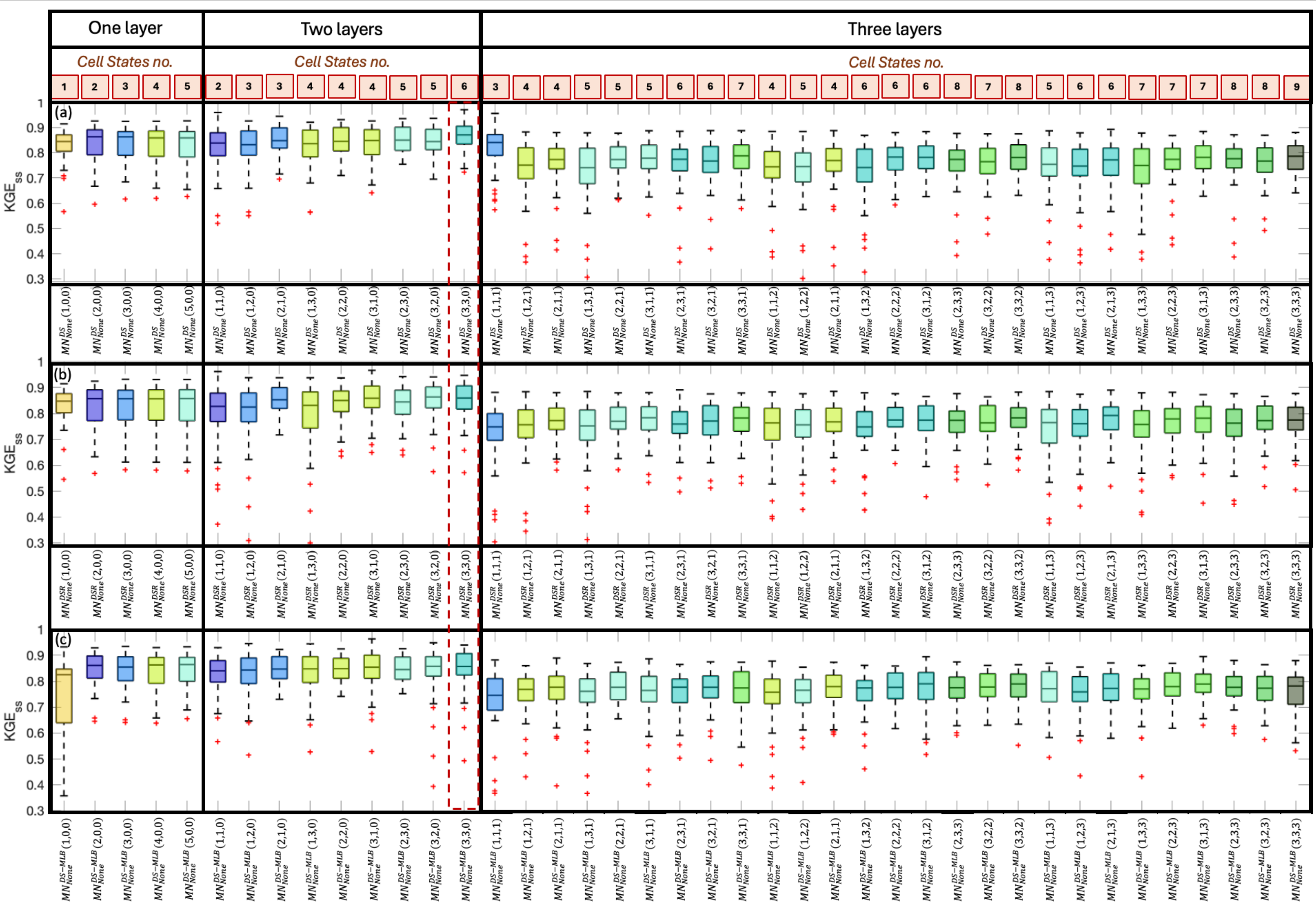

**Figure 5:** Box and whisker plots of the distributions of annual $\mathrm{KGE_{ss}}$ performance values for the Single- and Multi-Layer Networks. (a) Distributed-State (DS), (b) Distributed-State Relaxed (DSR), and (c) Distributed-State Machine Learning Benchmark. Each network architecture includes 41 cases, with the number of nodes in the first layer varying from 1 to 3, and in the second and third layers from 0 to 3. Single-layer cases with 4 and 5 nodes are also included for comparison. Note that cases using the same number of cell states are shown using the same color scheme, and the best-performing case—with two layers and three nodes in each layer—is highlighted with a red dashed box.

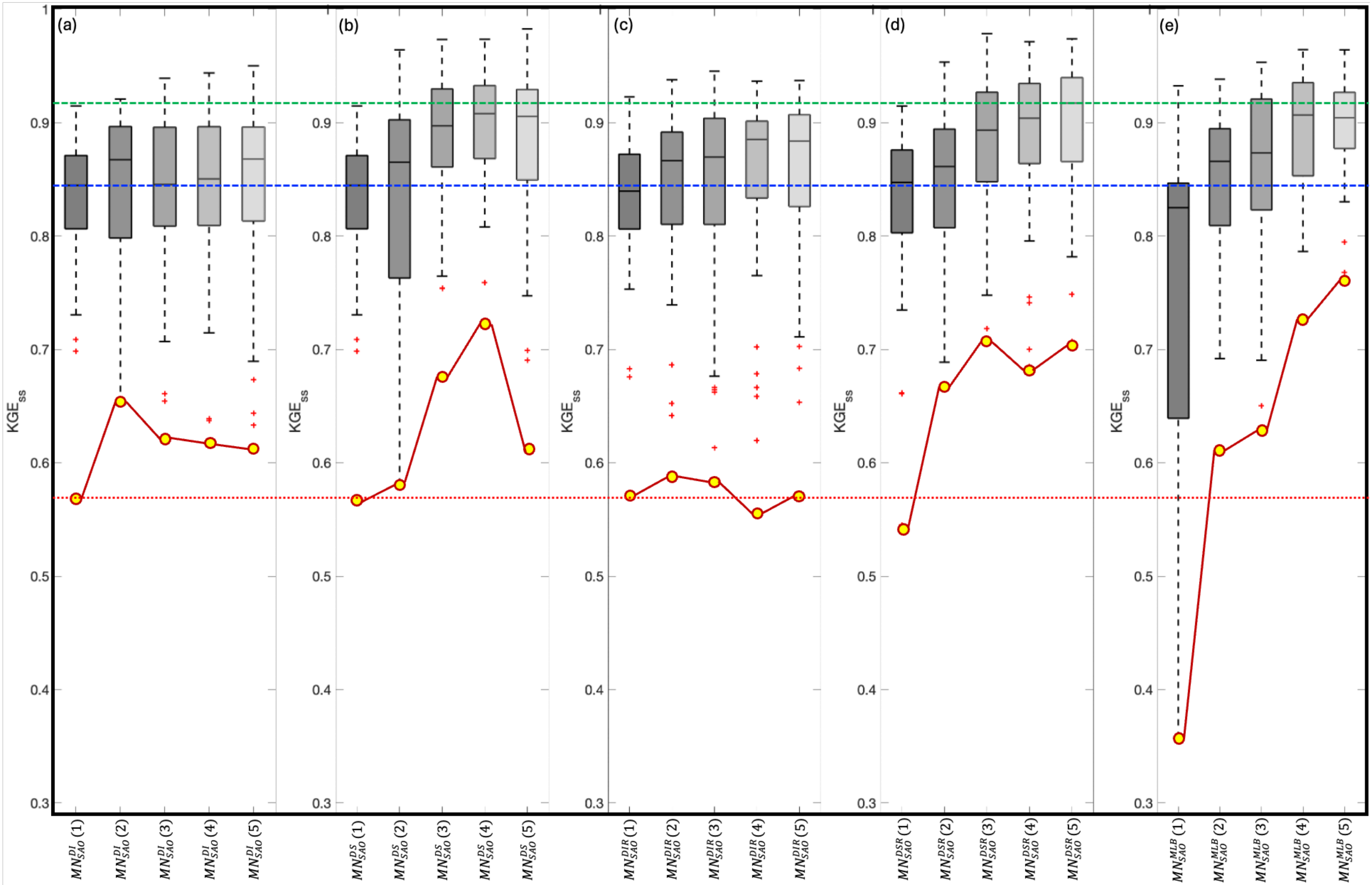

**Figure 6:** Box and whisker plots of the distributions of annual $\mathrm{KGE_{ss}}$ performance values for single-layer Sharing-Augmented Output Gates (SAO) networks. (a) Distributed Input (DI), (b) Distributed State (DS), (c) Distributed Input Relaxed (DIR), (d) Distributed State Relaxed (DSR), and (e) Machine-Learning Benchmark (MLB). The number of nodes varies from 1 to 5. *Baseline Median Year (blue dash line)*: the median-year (among all 40 years) $\mathrm{KGE_{ss}}$ score achieved by $MC\{O_\sigma L_{\sigma+}^{con}\}$ networks; *Baseline Worst Year (red dot line)*: the worst-year (among all 40 years) $\mathrm{KGE_{ss}}$ score achieved by $MC\{O_\sigma L_{\sigma+}^{con}\}$ networks; *Best Median Year (green dash line)*: the highest median-year (among all 40 years) $\mathrm{KGE_{ss}}$ score achieved by any of the single-layer MN networks shown here. Yellow-Filled Red Circles: The worst-year (among all 40 years) $\mathrm{KGE_{ss}}$ score is derived from each single-layer MN network.

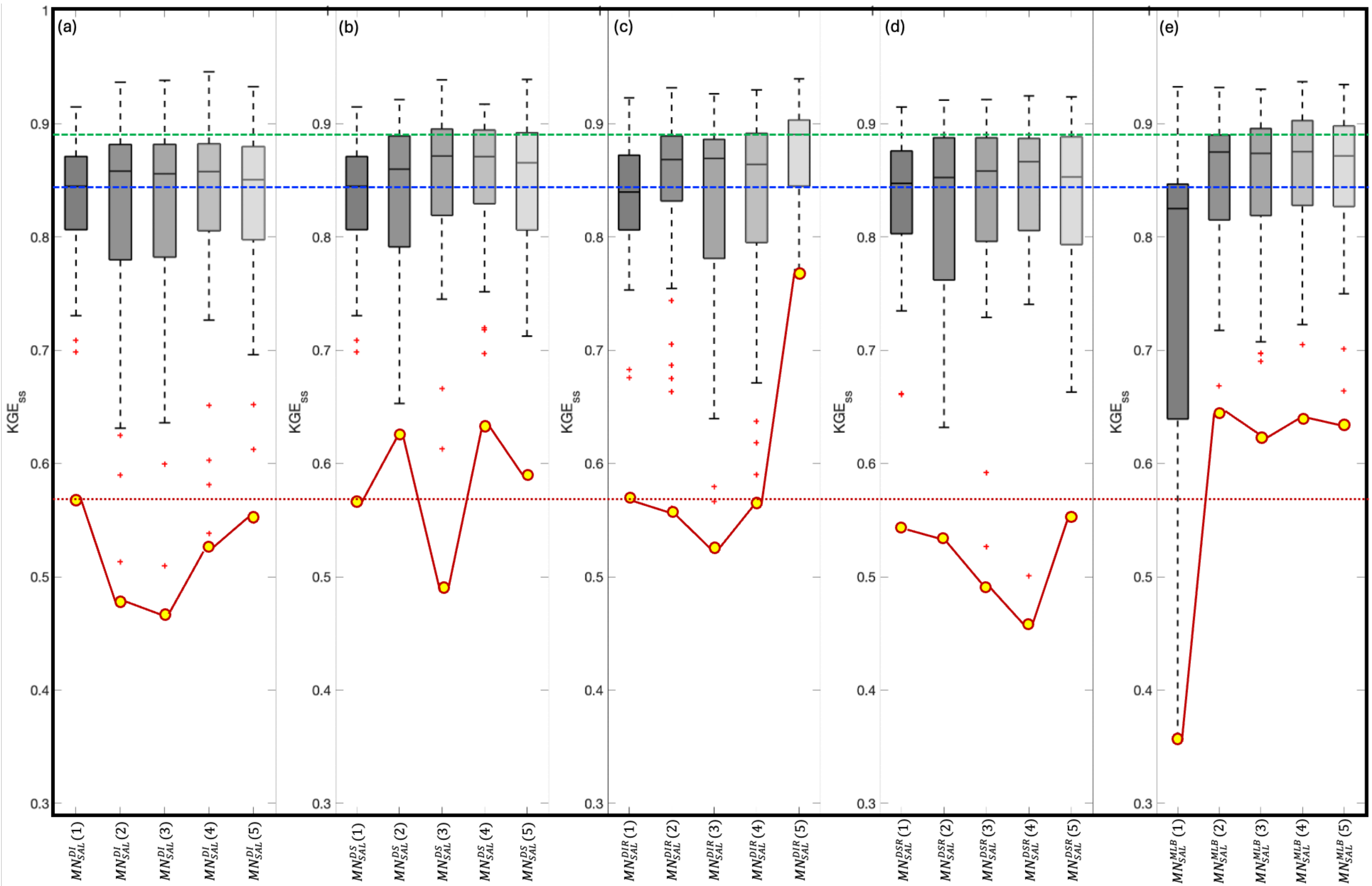

**Figure 7:** Box and whisker plots of the distributions of annual $\mathrm{KGE_{ss}}$ performance values for single-layer Sharing-Augmented Loss Gates (SAL) networks. (a) Distributed Input (DI), (b) Distributed State (DS), (c) Distributed Input Relaxed (DIR), (d) Distributed State Relaxed (DSR), and (e) Machine-Learning Benchmark (MLB). The number of nodes varies from 1 to 5. *Baseline Median Year (blue dash line)*: the median-year (among all 40 years) $\mathrm{KGE_{ss}}$ score achieved by $MC\{O_\sigma L^{con}_{\sigma+}\}$ networks; *Baseline Worst Year (red dot line)*: the worst-year (among all 40 years) $\mathrm{KGE_{ss}}$ score achieved by $MC\{O_\sigma L^{con}_{\sigma+}\}$ networks; *Best Median Year (green dash line)*: the highest median-year (among all 40 years) $\mathrm{KGE_{ss}}$ score achieved by any of the single-layer MN networks shown here.

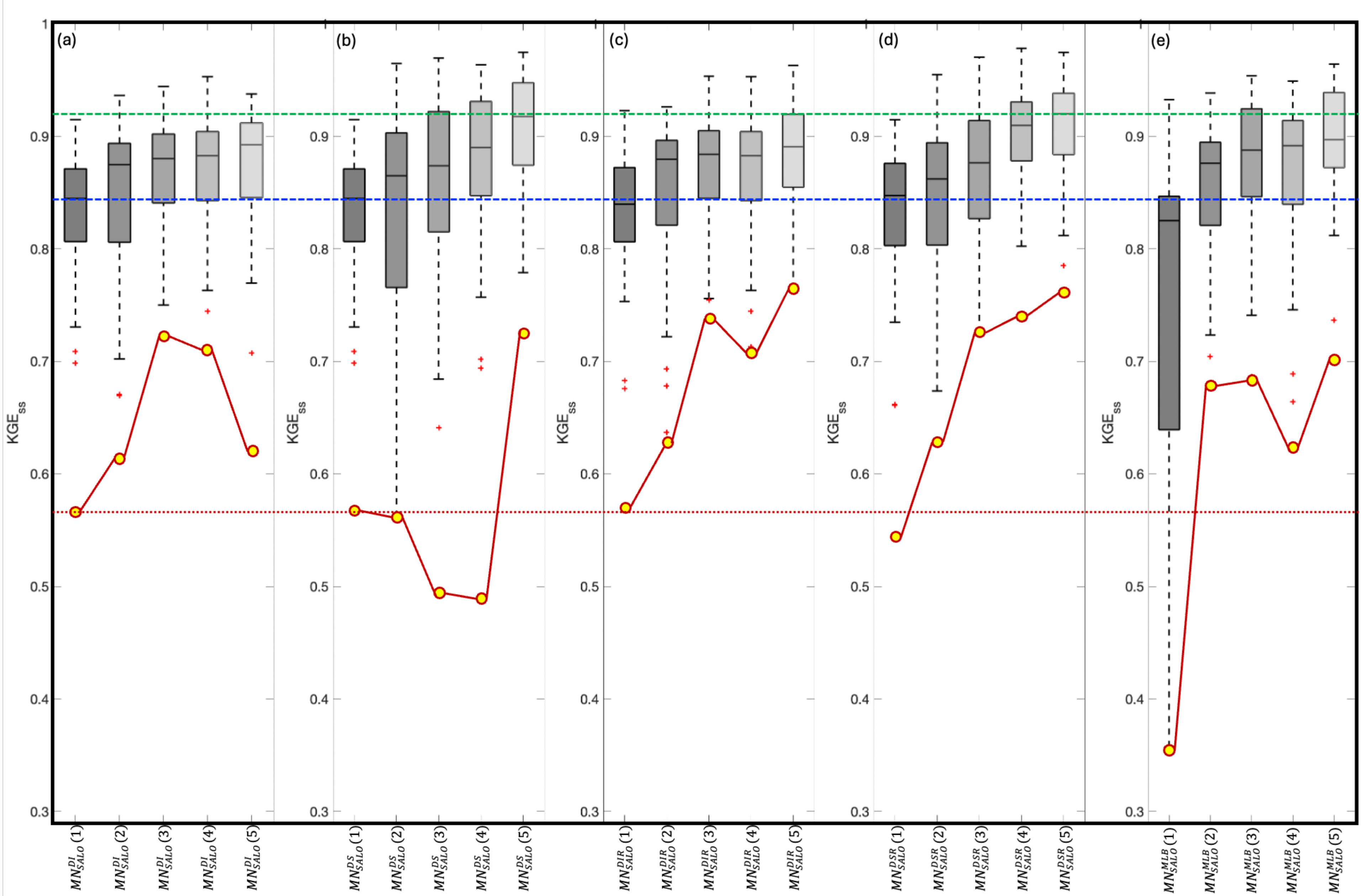

**Figure 8:** Box and whisker plots of the distributions of annual $\mathrm{KGE_{ss}}$ performance values for single-layer Sharing-Augmented Loss & Output Gates (SALO) networks. (a) Distributed Input (DI), (b) Distributed State (DS), (c) Distributed Input Relaxed (DIR), (d) Distributed State Relaxed (DSR), and (e) Machine-Learning Benchmark (MLB). The number of nodes varies from 1 to 5. *Baseline Median Year (blue dash line)*: the median-year (among all 40 years) $\mathrm{KGE_{ss}}$ score achieved by $MC\{O_\sigma L_{\sigma+}^{con}\}$ networks; *Baseline Worst Year (red dot line)*: the worst-year (among all 40 years) $\mathrm{KGE_{ss}}$ score achieved by $MC\{O_\sigma L_{\sigma+}^{con}\}$ networks; *Best Median Year (green dash line)*: the highest median-year (among all 40 years) $\mathrm{KGE_{ss}}$ score achieved by any of the single-layer MN networks shown here.

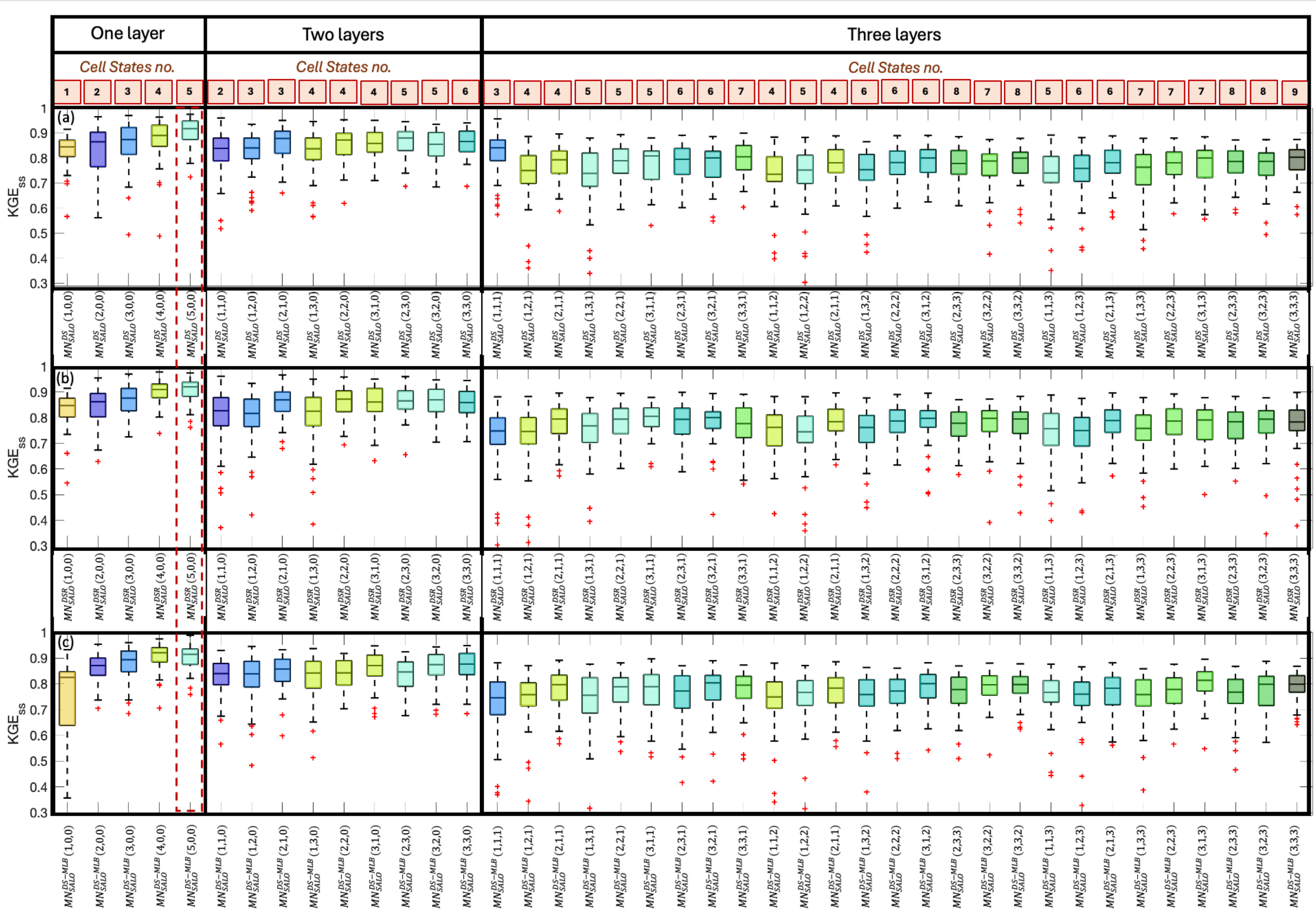

**Figure 9:** Box and whisker plots of the distributions of annual $KGE_{ss}$ performance values for Single- and Multi-Layer Sharing-Augmented Loss & Output Gates Networks. (a) Distributed-State (DS), (b) Distributed-State Relaxation (DSR), and (c) Distributed-State Machine Learning Benchmark Case. Each network architecture includes 41 cases, with the number of nodes in the first layer varying from 1 to 3, and in the second and third layers from 0 to 3. Cases with a single layer and 4 or 5 nodes are also included for comparison. Note that cases using the same number of cell states are shown using the same color scheme, and the best-performing case—with single layers and five nodes—is highlighted with a red dashed box.

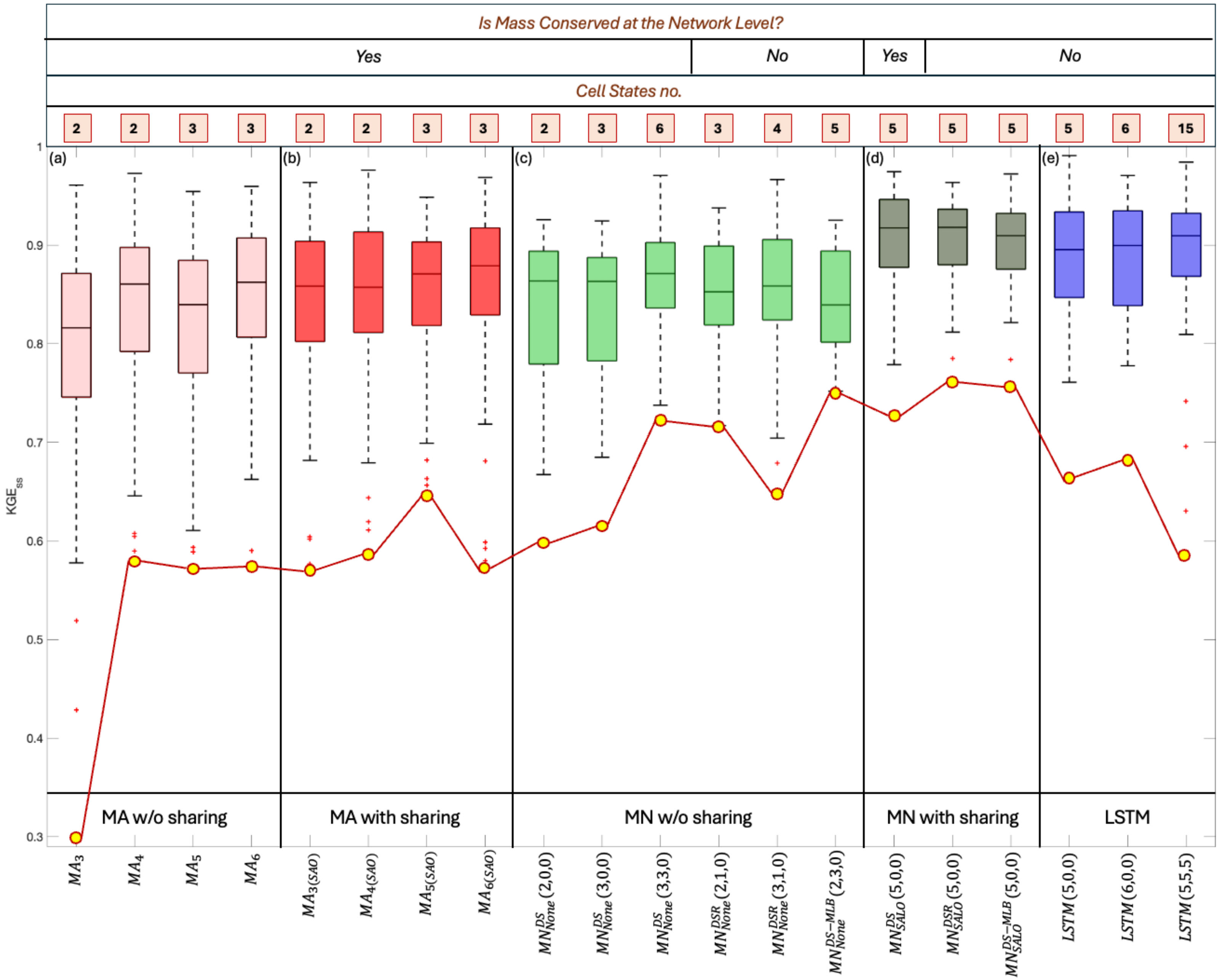

**Figure 10**: Box and whisker plots of the distributions of annual $KGE_{ss}$ scores, comparing (a) Mass-Conserving Architectures (MAs) reported in *Wang & Gupta (2024b)*, (b) associated MAs with Sharing-Augmented Output Gates, (c) several selected mass-conserving neural networks (MNs) developed in the current study, and (d) the associated architectures with Sharing-Augmented output or loss gates (or both) developed in this study, and (e) LSTM networks with varying numbers of layers and nodes.

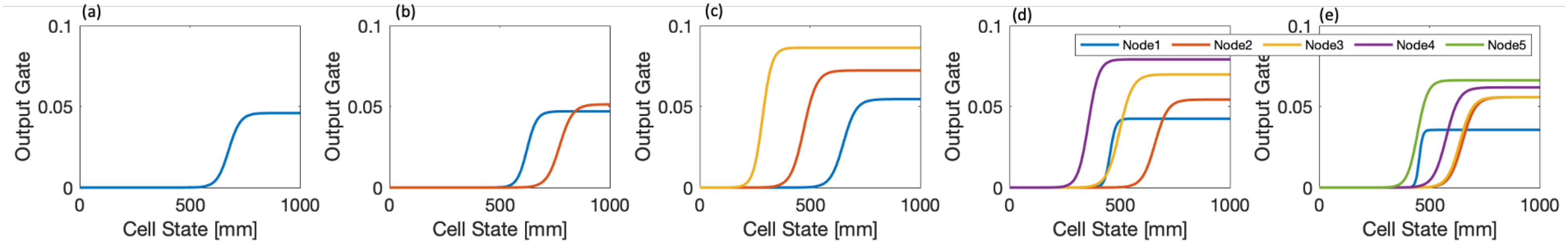

**Figure 11:** Illustration of output gate functions for the distributed-state (DS) case $MN_{None}^{DS}(N)$ for number of nodes $(N)$ varying from 1 to 5 (subplots (a) through (e). The color coding represents the peak value of each gate (blue is lowest and green is highest).

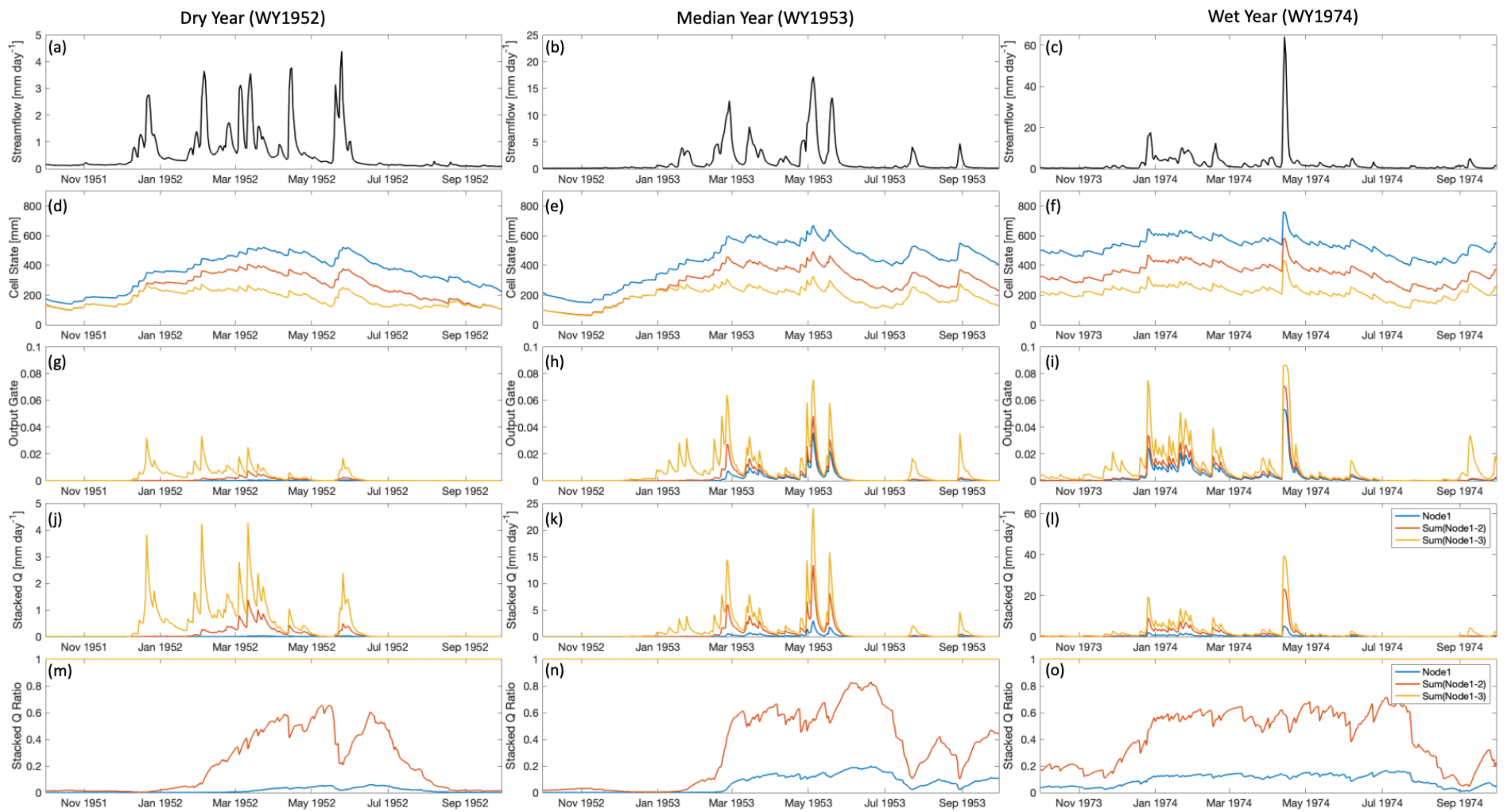

**Figure 12:** (a-c) Time series plots of streamflow observations for dry, median, and wet years; (d-f) Corresponding cell-state plots for the single-layer three-node distributed case ($MN_{None}^{DS}(3)$); (g-i) Corresponding output gate series; (j-l) Corresponding accumulated nodal streamflow trajectories; (m-o) Corresponding ratios for accumulated nodal streamflow. Q = streamflow. The nodes are ordered according to the magnitude of the peak values (from small to large) of the output gate, as shown in Figure 11 of the main text.

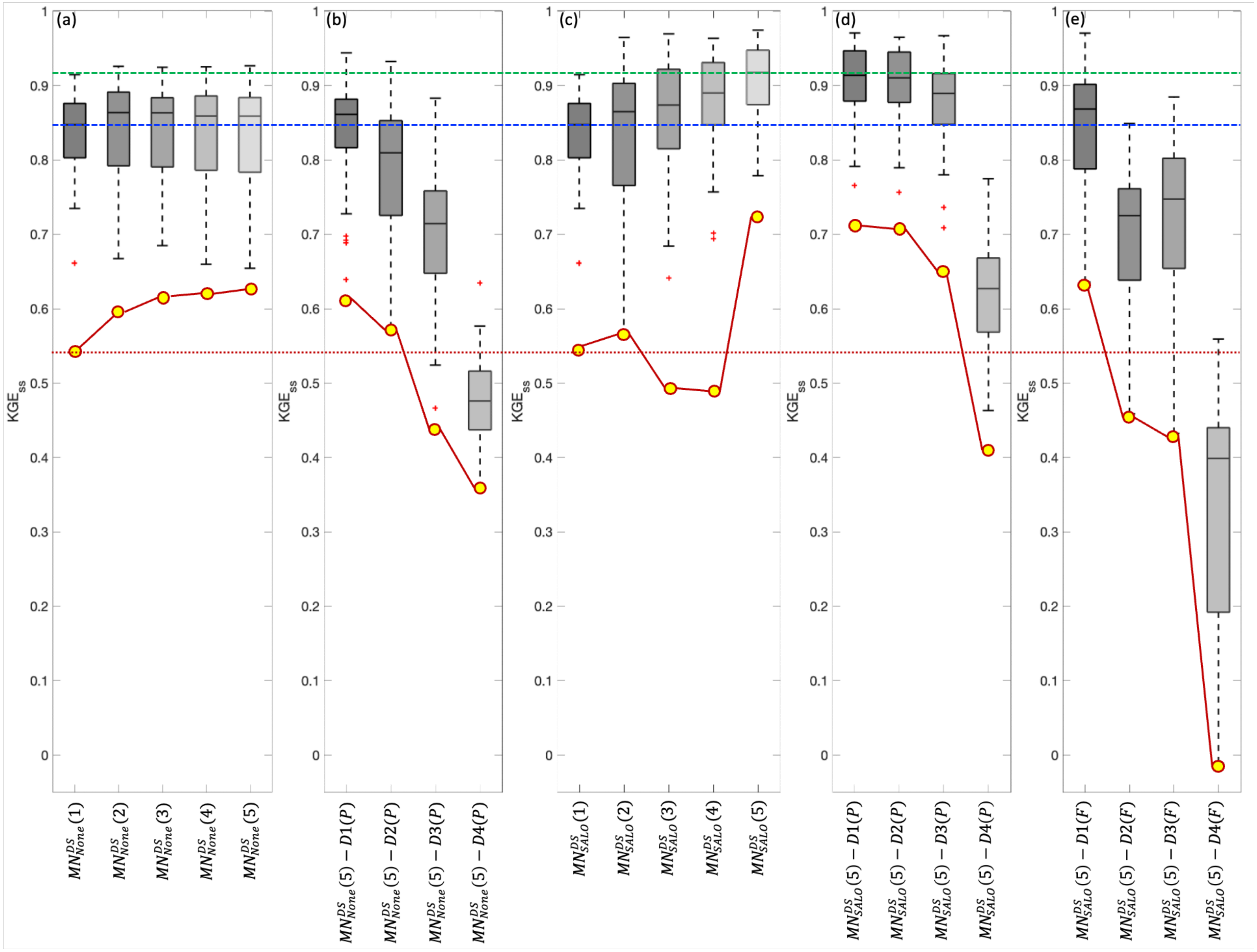

**Figure 13:** Box and whisker plots of the distributions of annual $\mathrm{KGE_{ss}}$performance for: (a) single-layer distributed state case $MN^{DS}_{None}(N)$ with number of nodes $N$ varying from 1 to 5; (b) the $MN^{DS}_{None}(5)$ network when pruning 1 to 4 flow paths ($MN^{DS}_{None}(5)-D1(P)$ to $MN^{DS}_{None}(5)-D4(P)$); (c) the single-layer distributed state Sharing-Augmented Loss & Output Gates (SALO) networks with $N$ varying from 1 to 5; (d) the $MN^{DS}_{SALO}(5)$ network when pruning 1 to 4 flow paths ($MN^{DS}_{SALO}(5)-D1(P)$ to $MN^{DS}_{SALO}(5)-D4(P)$); and (e) the $MN^{DS}_{SALO}(5)$ network when pruning 1 to 4 flow paths ($MN^{DS}_{SALO}(5)-D1(F)$ to $MN^{DS}_{SALO}(5)-D4(F)$). *Baseline Median Year (blue dash line)*: the median-year (among all 40 years) $\mathrm{KGE_{ss}}$ score achieved by $MC\{O_\sigma L^{con}_{\sigma+}\}$ networks; *Baseline Worst Year (red dot line)*: the worst-year (among all 40 years) $\mathrm{KGE_{ss}}$ score achieved by $MC\{O_\sigma L^{con}_{\sigma+}\}$ networks; *Best Median Year (green dash line)*: the highest median-year (among all 40 years) $\mathrm{KGE_{ss}}$ score achieved by any of the single-layer MN networks shown here.

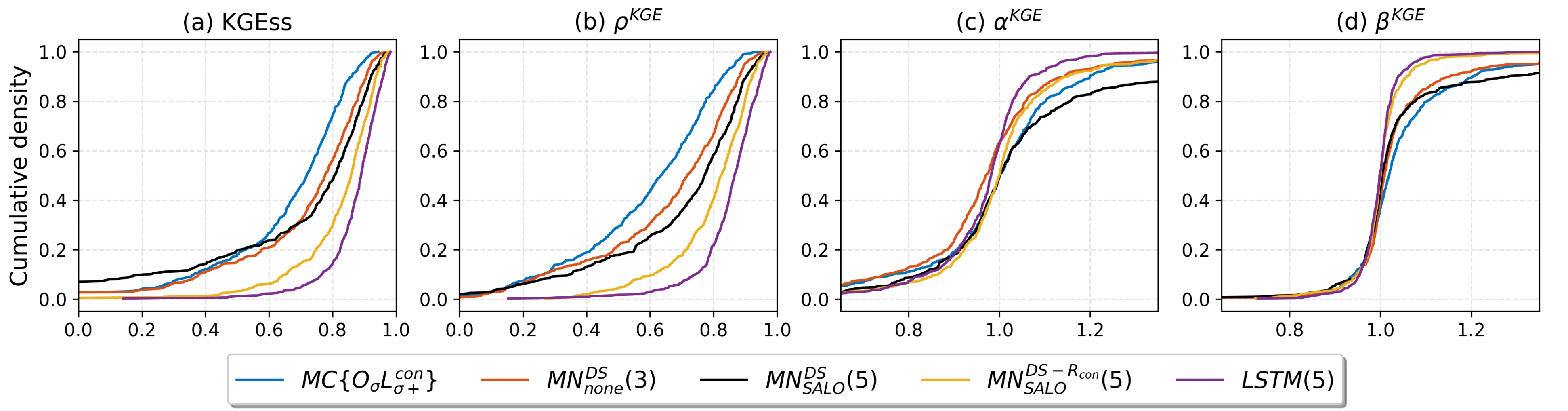

**Figure 14:** The skill metrics, including $KGE_{ss}$ and the three components of $KGE$ ($r^{KGE}, \alpha^{KGE}, \beta^{KGE}$), during the testing period are evaluated for the following models: the single-node $MC\{O_\sigma L_{\sigma+}^{con}\}$, the 3-node distributed-state $MN_{None}^{DS}(3)$, the 5-node distributed-state with cell-state information sharing $MN_{SALO}^{DS}(5)$, the 5-node distributed-state with shared cell-state information and a relaxed evapotranspirative loss constraint $MN_{SALO}^{DS-R_{con}}(5)$, and the single-layer 5-node LSTM network $LSTM(5)$, across the 513 CAMELS US catchments. Subplots (a) to (d) show the cumulative density plots for the $KGE_{ss}$, $r^{KGE}$, $\alpha^{KGE}$, and $\beta^{KGE}$ skill metrics across the five models.

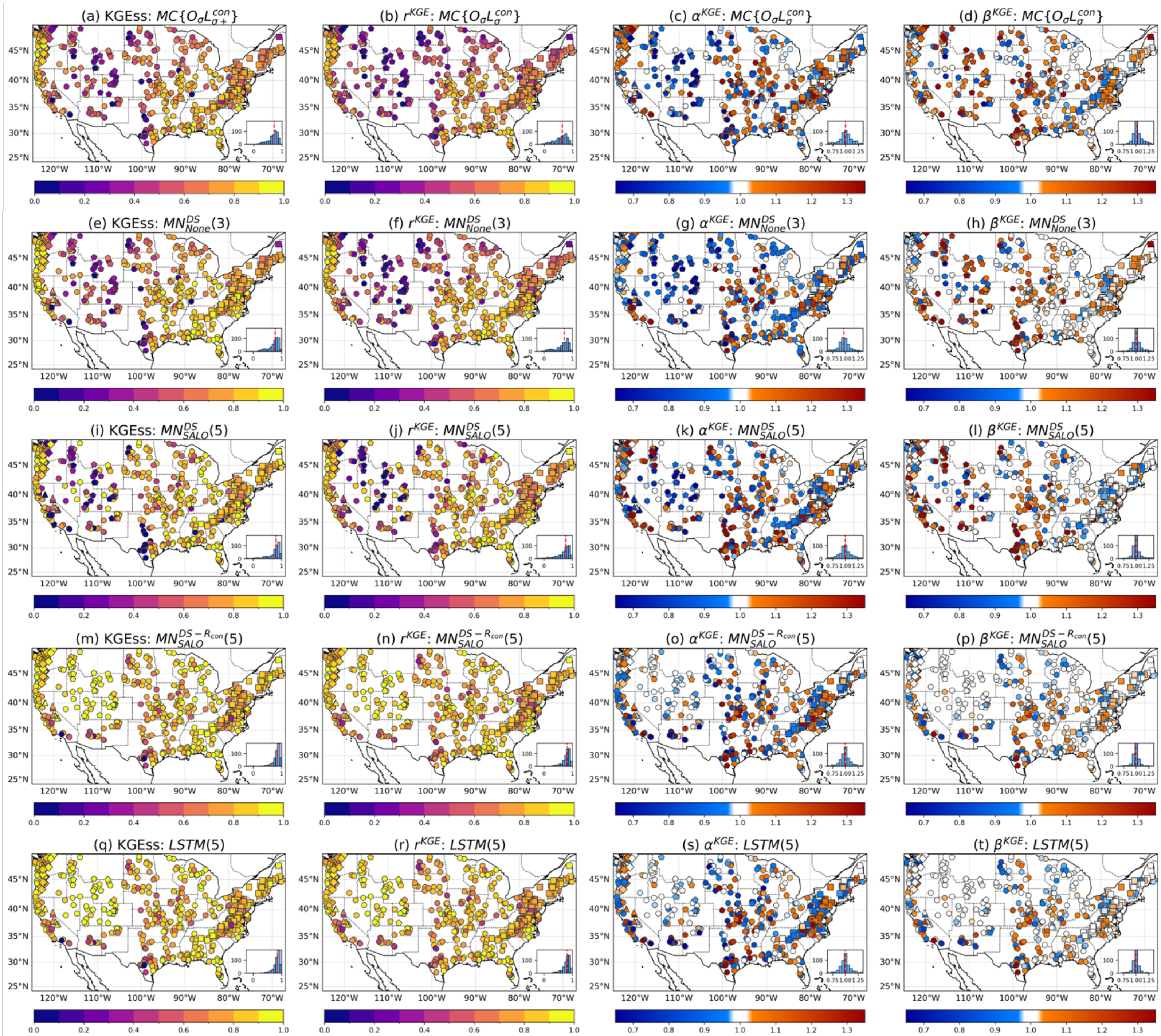

**Figure 15:** The skill metrics, including $KGE_{ss}$ and the three components of $KGE$ ($r^{KGE}, \alpha^{KGE}, \beta^{KGE}$), during the testing period are evaluated for the following models: the single-node $MC\{O_\sigma L_{\sigma+}^{con}\}$, the 3-node distributed-state $MN_{None}^{DS}(3)$, the 5-node distributed-state with cell-state information sharing $MN_{SALO}^{DS}(5)$, the 5-node distributed-state with shared cell-state information and a relaxed evapotranspirative loss constraint $MN_{SALO}^{DS-R_{con}}(5)$, and the single-layer 5-node LSTM network $LSTM(5)$, across the 513 CAMELS US catchments. Subplots (a) to (d) present the spatial maps of these four metrics for the $MC\{O_\sigma L_{\sigma+}^{con}\}$ model, followed by subplots (e) to (h) for $MN_{None}^{DS}(3)$, subplots (i) to (l) for $MN_{SALO}^{DS}(5)$, subplots (m) to (p) for $MN_{SALO}^{DS-R_{con}}(5)$, and subplots (q) to (t) for $LSTM(5)$. Triangles, diamonds, squares, and pentagons represent basins in the Sierra Nevada, Cascade, Appalachian, and Rocky Mountain ranges, respectively.

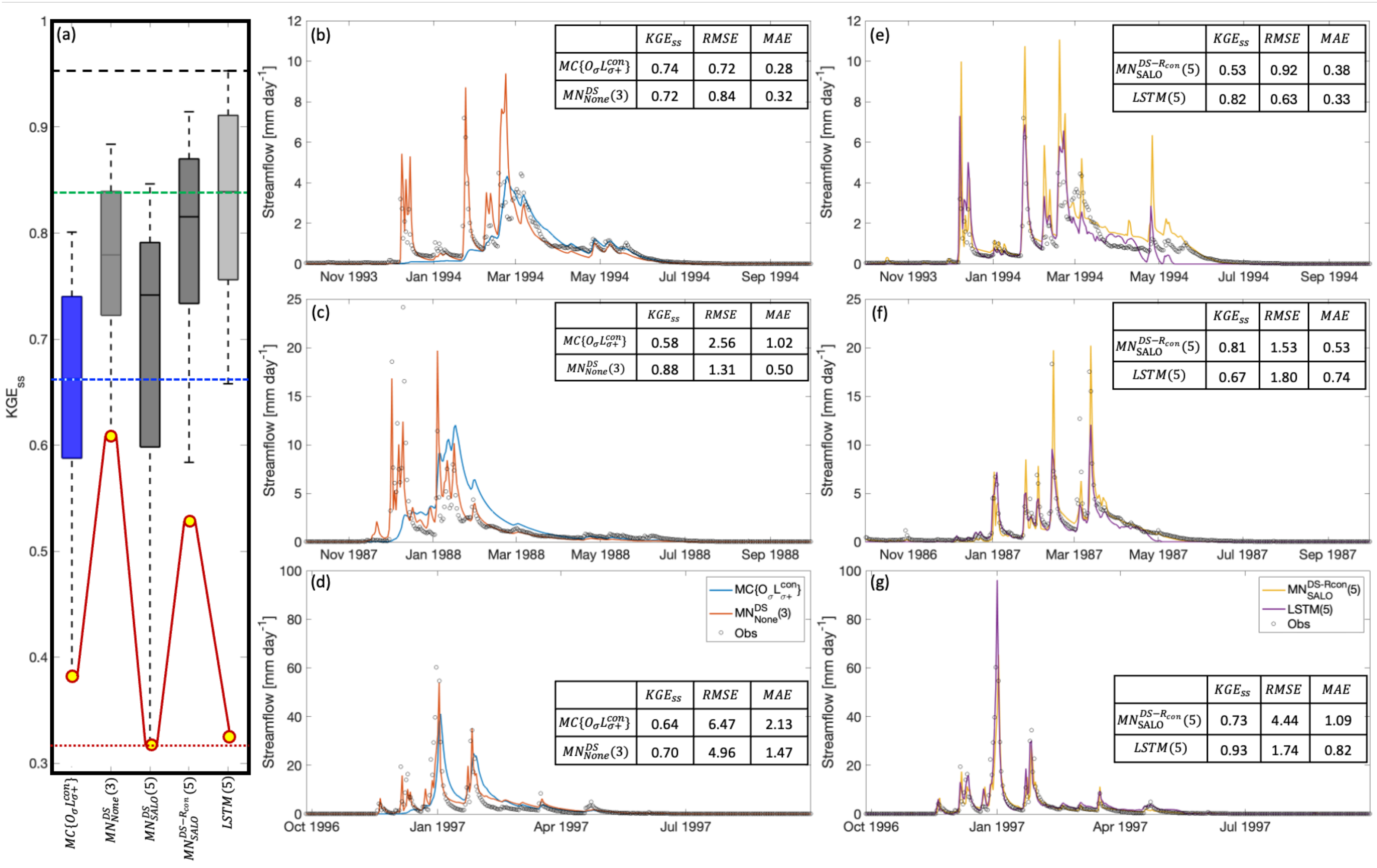

**Figure 16:** Comparison of selected single-layer Mass-Conserving Network performance, showing (a) box and whisker plots of the distributions of annual KGE scores, and (b - d) associated hydrographs for dry (WY 1994), median (WY 1988), and wet years (WY 1997) for $MC\{O_\sigma L_{\sigma+}^{con}\}$ and $MN_{None}^{DS}(3)$, and (f - h) for $MN_{SALO}^{DS-R_{con}}(5)$ and $LSTM(5)$. The skill in terms of annual $KGE_{ss}$, root mean square error ($RMSE$), and mean absolute error ($MAE$) for each model are summarized for each year. The annual performance metrics—$KGE_{ss}$, root mean square error ($MSE$), and mean absolute error ($MAE$)—are summarized for each model by dry, median, and wet year. In subplot (a), *Best Median Year (green dash line)*: the highest median-year (among all 27 years) KGE$_{ss}$ score achieved by any of the single-layer MN networks; *Best of Best Year (black dash line)*: the highest single-year KGE$_{ss}$ score achieved by any of the single-layer MN networks. *Baseline (blue dashed line):* median single-year $KGE_{ss}$score of the baseline case.*Best Case (green dashed line):* median single-year $KGE_{ss}$score of the best case. *Worst Case (red dashed line):* lowest single-year $KGE_{ss}$score of the worst case. *Yellow-Filled Red Circles:* the worst-year (among all 27 years) KGE$_{ss}$ score derived from each single-layer MN network. Note that the box-and-whisker plots in subplot (a) for the LSTM model are reported only for the last 25 years due to the maximum sequence length of 390 days. This limitation does not affect the three selected years shown in subplots (b)–(g).

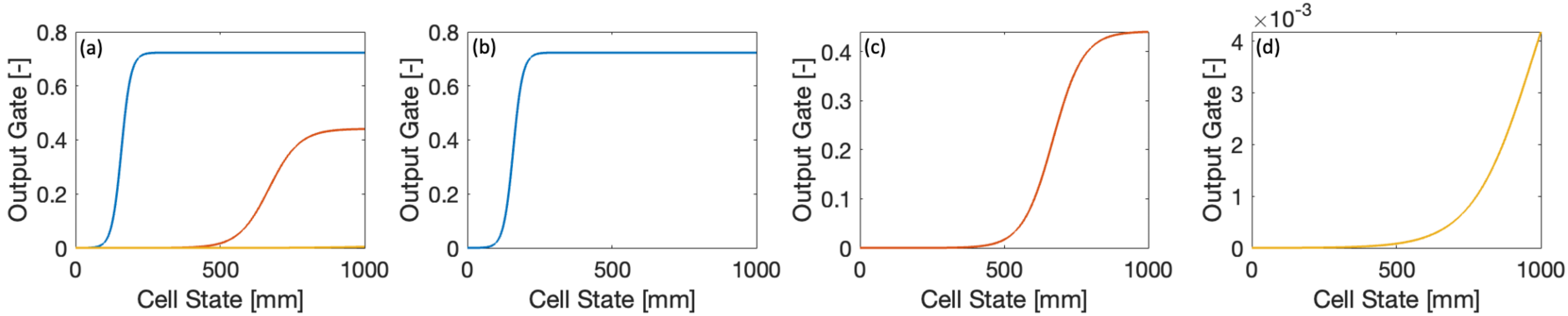

**Figure 17:** Output gate functions for the three-node distributed-state (DS) case $MN_{None}^{DS}(3)$. Subplot (a) shows the output gates from each of the three nodes, while subplots (b)–(d) display the individual gate activations Color coding indicates the peak gate value, with blue representing the largest and yellow the smallest.

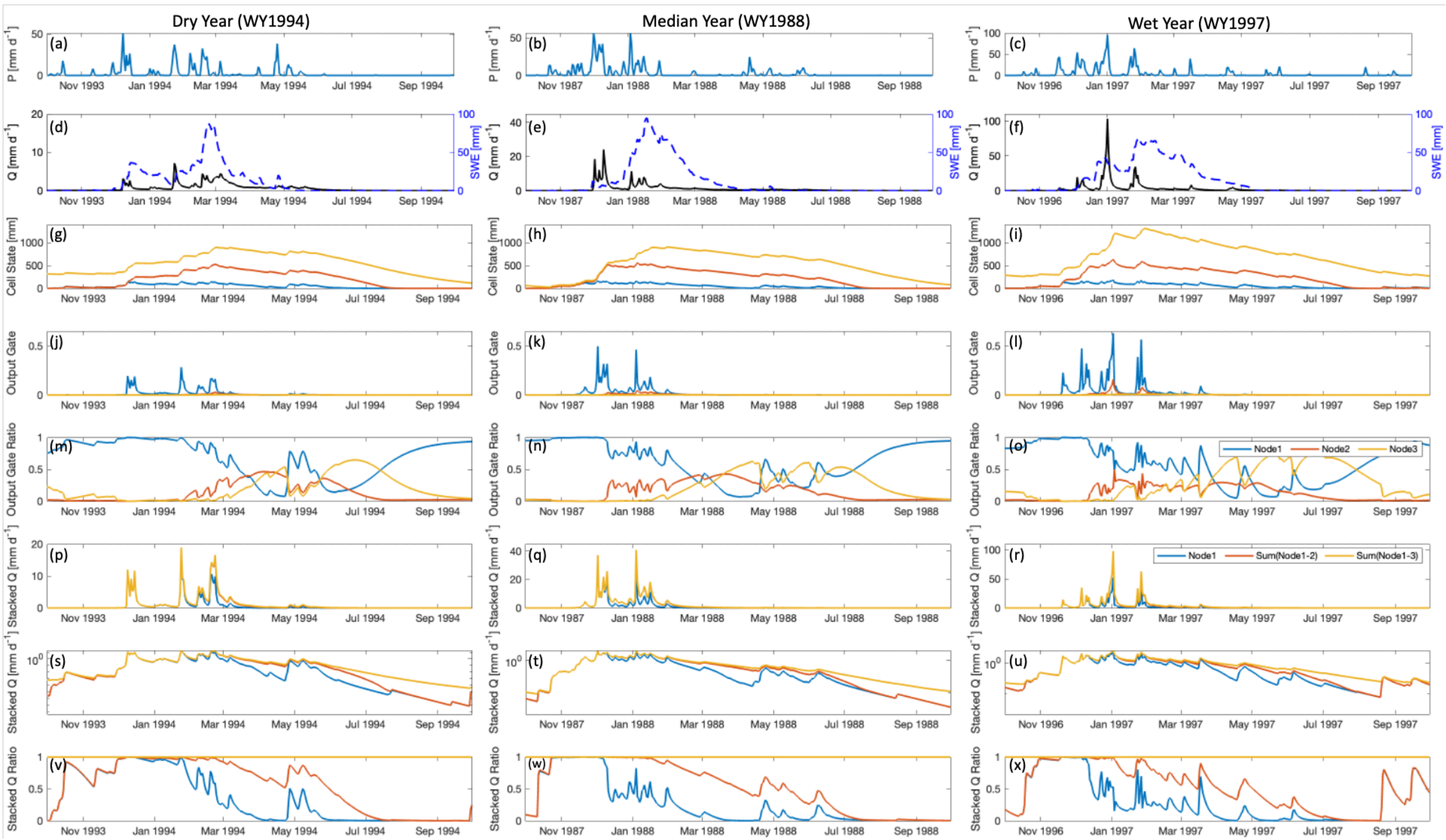

**Figure 18:** (a-c) Time series plots of precipitation for dry, median, and wet years, and streamflow observations and UA SWE (d-f) for; (g-i) Corresponding cell-state plots for the single-layer three-node distributed case ($MN^{DS}_{None}(3)$); (j-l) Corresponding output gate series; (m-o) Corresponding output gate ratios; (p-r) Corresponding accumulated nodal streamflow; (s-u) Corresponding accumulated nodal streamflow in logarithmic scale; (v-x) Corresponding ratios for accumulated nodal streamflow. Q = streamflow. The nodes are ordered according to the magnitude of the peak values (from large to small) of the output gate, as shown in Figure 16 of the main text.

Supporting Information for

# Using Machine Learning to Discover Parsimonious and Physically-Interpretable Representations of Catchment-Scale Rainfall-Runoff Dynamics

Yuan-Heng Wang[1,2], Hoshin V. Gupta[1]

[1] Department of Hydrology and Atmospheric Science, The University of Arizona, Tucson, AZ
[2] Environmental Sciences Area, Lawrence Berkeley National Laboratory, Berkeley, CA

Corresponding Author: Yuan-Heng Wang, Ph.D.
Email: yhwang0730@gmail.com | YuanHengWang@lbl.gov | yhwang0730@arizona.edu

Contents of this supplementary material:

## Introduction

This Supporting Information provides a preface for the entire supplementary material, 3 supplementary texts, 18 supplementary tables, and 13 supplementary figures to support the discussions in the main text. The contents are as follows.

**Text S1.** Mathematical formulation of Long Short-Term Memory (LSTM) Network

**Text S2.** Summary of the Training Procedure

**Text S3.** Working examples of MCP-based model

**Table S1.** $\alpha_*^{KGE}$ scores for the Single-Layer Mass-Conserving Neural Networks (MNs)

**Table S2.** $\beta_*^{KGE}$ scores for the Single-Layer Mass-Conserving Neural Networks (MNs)

**Table S3.** $\gamma^{KGE}$ scores for the Single-Layer Mass-Conserving Neural Networks (MNs)

**Table S4.** $KGE_{ss}$ scores of flow magnitude for the Single-Layer Mass-Conserving Neural Networks (MNs)

**Table S5.** $\alpha^{KGE}$ scores of flow magnitude for the Single-Layer Mass-Conserving Neural Networks (MNs)

**Table S6.** $\beta^{KGE}$ scores of flow magnitude for the Single-Layer Mass-Conserving Neural Networks (MNs)

**Table S7.** $r^{KGE}$ scores of flow magnitude for the Single-Layer Mass-Conserving Neural Networks (MNs)

**Table S8.** Root Mean Squared Error ($RMSE$) scores for the Single-Layer Mass-Conserving Neural Networks (MNs)

**Table S9.** Mean Absolute Error ($MAE$) scores for the Single-Layer Mass-Conserving Neural Networks (MNs)

**Table S10.** Summary of Multi-Layer Mass-Conserving Neural Network ($MN_s$) architectural hypothesis tested in this study

**Table S11.** Detailed information on the training setup for single-layer cases

**Table S12.** Detailed information on the training setup for multi-layer cases

**Table S13.** Summary of benchmark models in this study

**Table S14.** $KGE_{ss}$ statistics and numbers of parameter for the benchmark LSTM network used in this study

**Table S15.** Summary of the pruned network models in this study

**Table S16**. Performance percentiles across the 513 CAMELS US basins using the scaled Kling–Gupta Efficiency ($KGE_{ss}$), its three components—$\alpha^{KGE}$ (flow variability), $\beta^{KGE}$ (mass balance), and $r^{KGE}$: (Pearson correlation)—as well as the root-mean-squared error (RMSE) and mean absolute error (MAE).

**Table S17**. Background Information on the selected CAMELS US catchment

**Table S18.** Skill Metrics during testing period for on the selected CAMELS US catchment

**Figure S1:** Directed-graph representation and mathematical formulation of the mass-conserving-perceptron (MCP) modeling network used in this study.

**Figure S2**. Comparison of selected single-layer Mass-Conserving Network performance using hydrographs in log-scale for selected dry, median, and wet years in subplots (a) to (c).

**Figure S3**. Comparison of selected single-layer Mass-Conserving Network performance using the hydrographs for selected dry, median, and wet years in subplots (a) to (c).

**Figure S4**. Mass-Conserving Architecture (MA) proposed by Wang & Gupta (2024b), including: (a) one-flow-path, two-state MA3 hypothesis; (b) two-flow-path, two-state MA4 hypothesis; (c) two-flow-path, three-state MA5 hypothesis; and (d) three-flow-path, three-state MA6 hypothesis.

**Figure S5**. Box and whisker plots of the distributions of annual KGE scores, comparing the (a) 1-layer, (b) 2-layer, and (c) 3-layer long short-term memory (LSTM) networks, with the number of nodes varying from 1 to 5.

**Figure S6**. Time series plots of streamflow observations for dry, median, and wet years are shown in subplots (a) to (c). The cell states for the distributed case (DS) are presented with 1-node in subplots (d) to (f) (dry, median, and wet years), 2-node in subplots (g) to (i), 3-node in subplots (j) to (l), 4-node in subplots (m) to (o), and 5-node in subplots (p) to (r).

**Figure S7**. Time series plots of streamflow observations for dry, median, and wet years are shown in subplots (a) to (c). The output gate series for the distributed case (DS) are presented with 1-node in subplots (d) to (f) (dry, median, and wet years), 2-node in subplots (g) to (i), 3-node in subplots (j) to (l), 4-node in subplots (m) to (o), and 5-node in subplots (p) to (r).

**Figure S8**. Time series plots of streamflow observations for dry, median, and wet years are shown in subplots (a) to (c). The streamflow components for the distributed case (DS) are presented with 1-node in subplots (d) to (f) (dry, median, and wet years), 2-node in subplots (g) to (i), 3-node in subplots (j) to (l), 4-node in subplots (m) to (o), and 5-node in subplots (p) to (r).

**Figure S9**. Time series plots of streamflow observations for dry, median, and wet years are shown in subplots (a) to (c). The cumulative streamflow components for the distributed case (DS) are presented with 1-node in subplots (d) to (f) (dry, median, and wet years), 2-node in subplots (g) to (i), 3-node in subplots (j) to (l), 4-node in subplots (m) to (o), and 5-node in subplots (p) to (r).

**Figure S10**. Time series plots of streamflow observations for dry, median, and wet years are shown in subplots (a) to (c). The cell states ratio for the distributed case (DS) are presented with 1-node in subplots (d) to (f) (dry, median, and wet years), 2-node in subplots (g) to (i), 3-node in subplots (j) to (l), 4-node in subplots (m) to (o), and 5-node in subplots (p) to (r).

**Figure S11**. Time series plots of streamflow observations for dry, median, and wet years are shown in subplots (a) to (c). The ratio of output gate series for the distributed case (DS) are presented with 1-node in subplots (d) to (f) (dry, median, and wet years), 2-node in subplots (g) to (i), 3-node in subplots (j) to (l), 4-node in subplots (m) to (o), and 5-node in subplots (p) to (r).

**Figure S12**. Time series plots of streamflow observations for dry, median, and wet years are shown in subplots (a) to (c). The ratio of streamflow components for the distributed case (DS) are presented with 1-node in subplots (d) to (f) (dry, median, and wet years), 2-node in subplots (g) to (i), 3-node in subplots (j) to (l), 4-node in subplots (m) to (o), and 5-node in subplots (p) to (r).

**Figure S13**. Time series plots of streamflow observations for dry, median, and wet years are shown in subplots (a) to (c). The ratio of cumulative streamflow components for the distributed case (DS) are presented with 1-node in subplots (d) to (f) (dry, median, and wet years), 2-node in subplots (g) to (i), 3-node in subplots (j) to (l), 4-node in subplots (m) to (o), and 5-node in subplots (p) to (r).

## Preface

This supplementary material aims to enhance readability and understanding by providing training details for the MCP-based network, an illustrative example of its internal functioning, and detailed information about the performance metrics and internal functioning plots.

The first three text sections provide (1) mathematical formulation of long short-term memory (LSTM) network, (2) detailed information on model training and development using both the Leaf River and CAMELS US datasets, and (3) a working example explaining the internal functioning of the basic $MC\{O_{\sigma}L_{\sigma+}^{con}\}$ unit, as well as the mathematical formulation for single-layer and multi-layer distributed-state networks.

The 13 supplementary figures and 18 supplementary tables are intended, primarily, to provide comprehensive information on all modeling exercises. They mainly support the modeling exercise in the Leaf River Basin and the 513 CAMELS US basins. Both datasets use precipitation and potential evapotranspiration as model inputs and streamflow as the output at the daily timestep. The Leaf River dataset contains 40 years of data, covering water years (WY) 1949 to 1988, while all 513 selected CAMELS US basins have 27 years of continuous data from WY 1982 to 2008.

Note that three years with different hydrologic regimes are selected as representative dry, median, and wet years to illustrate the hydrograph and overall performance. These years are determined based on annual flow peaks, with the smallest, median, and largest peak-flow years classified as dry, median, and wet years, respectively. For the Leaf River, WY 1952, WY 1953, and WY 1974 are selected as the dry, median, and wet years, whereas WY 1994, WY 1988, and WY 1997 are chosen for the CAMELS US basins.

Additionally, one key message of this paper is to demonstrate the isomorphic relationship between a physical modeling system (such as a linear reservoir) and a modern deep learning–based gated recurrent neural network (such as an LSTM). We provide details in Figure S1 to help readers properly understand the rationale behind the design of the mass-conserving-perceptron-based computational unit.

In Leaf River Basin, most information is associated with the single-layer performance:

a) Figure S2 and Figure S3 show the hydrographs of the selected single-layer network at the Leaf River. Figure S4 illustrates the performance of the locally trained LSTM (the machine-learning upper benchmark) in the Leaf River basin. Figure S5 shows the mass-conserving architectures (MAs) proposed by Wang & Gupta (2024b), which are used as the physics-based model benchmark in the Leaf River basin.
b) Figures S6 to S12 present the internal functioning of the single-layer distributed-state network, including the use of cell states, output gate series, output gate ratios, streamflow, and the contributions from each flow path.
c) Tables S1 to S3 summarize the performance metrics of $\alpha_{*}^{KGE}$, $\beta_{*}^{KGE}$, and $r^{KGE}$, respectively, for all the single-layer networks tested.
d) Similar to our previous work (Wang & Gupta, 2024ab), the 40 years of total flow are separated into five groups based on magnitude. The $KGE_{ss}$ and its three-component metrics are calculated and summarized in Tables S4 to S7.

e) To facilitate comparison across studies, additional metrics including Root Mean Squared Error (RMSE) for single-layer networks are provided in Tables S8 and S9.
f) Table S10 provides summary information for the Multi-Layer Mass-Conserving Neural Network.
g) Table S11 details the training procedures, particularly the initialization of weights and biases for each single-layer network, while Table S12 does the same for the multi-layer network.
h) Table S13 summarizes information about the physics-based (MCP-based mass-conserving architecture in Wang & Gupta, 2024b) and the machine-learning-based upper benchmark LSTM model.
i) Table S14 presents the performance of the machine-learning-based upper benchmark LSTM model.
j) Table S15 provides an introductory summary of the pruned distributed-state networks.

In 513 selected CALUES US basins, we mainly have additional three tables to support the information in the main text. They are separated into the broad-focus and depth-focus.

Multi-Catchment Study: The performance percentiles for the 513 CAMELS US basins using the $KGE_{ss}$ and three KGE ( $\alpha^{KGE}$ : flow variability; $\beta^{KGE}$ : mass balance; and $r^{KGE}$: (Pearson correlation)—as well as the root-mean-squared error (RMSE) and mean absolute error (MAE) that the models include $MC\{O_\sigma L_{\sigma+}^{con}\}$ , $MN_{None}^{DS}(3)$ , $MN_{SALO}^{DS}(5)$ , $MN_{SALO}^{DS-R_{con}}(5)$ , and $LSTM(5)$ are included in Table S16.

Detailed Examination at One Location: We selected one more catchment the ID11473900 to examine the internal functioning of the selected $MN_{None}^{DS}(3)$ network. Where the Background Information and the Skill Metrics during testing period are included in the Table S17 and Table S18.

Text 1: The mathematical formulation of Long Short-Term Memory (LSTM) Network

As a data-driven benchmark to evaluate performance of the MCP-based networks tested in this study, we use the LSTM network, adapted from code provided by *Kratzert et al. (2019b)*. The LSTM network is a type of recurrent neural network that includes memory cells that can store information over long periods of time, and that uses three gating operations (input, forget, output) as shown in **Figure S1c**.

Given an input sequence $x = [x[1], x[2] \dots \dots, x[T]]$ with $T$ time steps, where each element $x[t]$ is a vector containing input features (model inputs) at time step $t$ $(1 \leq t \leq T)$, Equations (T1-T6) specify a single forward pass through the LSTM:

$$i[t] = \sigma(b_i x[t] + w_i h[t-1] + a_i) \quad \text{(T1)}$$
$$f[t] = \sigma\left(b_f x[t] + w_f h[t-1] + a_f\right) \quad \text{(T2)}$$
$$g[t] = tanh(b_g x[t] + w_g h[t-1] + a_g) \quad \text{(T3)}$$
$$o[t] = \sigma(b_o x[t] + w_o h[t-1] + a_o) \quad \text{(T4)}$$
$$c[t] = f[t] \odot c[t-1] + i[t] \odot g[t] \quad \text{(T5)}$$
$$h[t] = o[t] \odot \tanh\,(c[t]) \quad \text{(T6)}$$

where $i(\mathrm{t})$, $f(\mathrm{t})$, $o(\mathrm{t})$ are the input, forget and output gates respectively, $g(\mathrm{t})$ is the cell input, $x(\mathrm{t})$ is the network input at time step $t$ $(1 \leq t \leq T)$, and $h(\mathrm{t}-1)$ is the recurrent input. The terms $c(\mathrm{t})$ and $c(\mathrm{t}-1)$ indicate the cell states at the current and previous time step. At the first-time step, the hidden and cell states are initialized as vectors of zeros. The terms $\mathrm{a}$, $w$ and $\mathrm{b}$ are learnable parameters for each gate, with subscripts referring to which gate the particular weight matrix, or bias vector corresponds to. The sigmoid activation function $\sigma\,(\cdot)$ outputs a value between 0 and 1, while the hyperbolic tangent activation function $\tanh\,(\cdot)$ outputs a value between -1 and 1. The symbol $\odot$ indicates element-wise multiplication.

The values of the cell states can be modified by the forget gate $f(\mathrm{t})$, which can delete states. The cell update $g(\mathrm{t})$ can be interpreted as information that is added, while the input gate $i(\mathrm{t})$ controls into which cells new information is added. The output gate $o(\mathrm{t})$ controls which of the information, stored in the cell states, is output. Note that the cell states $c(\mathrm{t})$ characterize the memory of the system, and its simple linear interaction with the remaining LSTM cell helps to prevent the issue of exploding or vanishing gradients during the back-propagation step of network training (i.e., unstable training due to gradients becoming too large or too small; *Hochreiter & Schmidhuber, 1997*). The output of the final LSTM layer $h(\mathrm{t})$ is connected through a dense layer to a single output neuron, which computes the final output $y(\mathrm{t})$ prediction, as indicated by Equation (T7):

$$y = b_d h_n + a_d \quad \text{(T7)}$$

where $b_d$ and $a_d$ are learnable weights and bias terms of a dense output layer.

Text 2: Summary of the Training Procedure

***Summary of the Procedure for the Leaf River Basin***

***Data-splitting***

As summarized in section 3.2.1 of the main text, the data splitting algorithm proposed by *Zheng et al. (2022)* helps ensure that model performance remains relatively consistent across the three independent seta (training, selection and testing/evaluation; *Chen et al., 2022; Maier et al., 2023*). This is because the procedure causes their distributions to be more aligned, compared to random splitting based on continuous years. In the latter case, wet/dry year representation in specific subsets can become unbalanced, so that k-fold cross-validation becomes necessary during model development to verify that the procedure is sufficiently robust to identify optimal models. The *Zheng et al. (2022)* data-splitting algorithm adopted in this study reduces the need for such procedures.

To maintain consistency, for all experiments conducted in the Leaf River basin we followed the procedure used in our earlier study (*Wang & Gupta, 2024ab*). We set the $\mathcal{D}_{\text{train}}$: $\mathcal{D}_{\text{select}}$ : $\mathcal{D}_{\text{test}}$ partitioning ratio to be 2:1:1 respectively. The data splitting procedure first sorts the streamflow data based on magnitude. Next, the timestep associated with the largest streamflow magnitude was paired with the timestep associated with the smallest streamflow value, continuing with the next largest and smallest values and so on, until all time steps have been paired. These pairs were then sequentially allocated, in the abovementioned ratio, to the three independent sets (following the sequence of $\mathcal{D}_{\text{train}} \rightarrow \mathcal{D}_{\text{test}} \rightarrow \mathcal{D}_{\text{select}} \rightarrow \mathcal{D}_{\text{train}}$ etc.) until all pairs were assigned. Overall, given 14,610 time-steps/days in the 40-year LRB dataset, this resulted in a training subset consisting of 7,306 timesteps, and selection and testing subsets consisting of 3,652 time-steps each. This temporal splitting approach has been shown to maintain natural temporal continuity, where the dynamics are adequately captured and interpreted, and it does not affect model development during training. Please see *Wang & Gupta (2024a)* for further justification.

***Training Procedures, Algorithm and Hyperparameter Selection***

The training procedures, algorithm and hyperparameter selection are briefly summarized in section 3.2.3 of the main text. More detail information is presented here. A consistent learning rate was used across all model experiments in this study: $2.5 \times 10^{-2}$ for the first 300 epochs, followed by a reduction to $1.25 \times 10^{-2}$. The single-layer MN networks introduced in Sections 2.3.2 to Sections 2.3.6 and 2.4 were trained for 3000 epochs, except for the machine-learning benchmark case, which was trained for 5000 epochs.

The multi-layer MN networks were trained for 2000 epochs, due to differences in the training approach relative to the single-layer case (see below section of *Information Flows and Network Training and Model Initialization*). For the benchmark models in Section 3.3, both single-layer and multi-layer LSTM networks were trained for 2000 epochs. The mass-conserving architectures adopted from *Wang & Gupta (2024b),* modified to allow cell-state information sharing, were trained for 3000 epochs.

***Information Flows, Network Training and Model Initialization***

In all of the single-layer cases (see **Table 1** in the main text) the networks were fed with the same area-averaged catchment precipitation as input. However, they differed in: (i) the manner by which that input mass was routed through the system, and (ii) the manner by which "information" flowed between the gating components of that system.

For training the single-hidden-layer MCP networks, we initialized all nodes to the same parameter (weight) values that were obtained for the single node MCP architecture ($MC\{O_\sigma L_{\sigma+}^{con}\}$) and then added different levels of Gaussian random noise to all of the weights. In other words, the trained parameters of the single node architecture were treated as "mean" values, and the magnitude (standard deviation) of the noise was chosen to be between 2.5% and 20% of that mean value (the actual percentage was selected based visually on how notable the gating functions changed when altering the associated parameters). Further, for the cases with information sharing, the $N_\ell \times 1$ vectors of weights associated with the output or loss gates become $N_\ell \times N_\ell$ matrices, where the diagonal values control information provided by the same node. Accordingly, following the same initialization rule mentioned above, the off-diagonal values were randomly initialized with a zero mean and 0.025 standard deviation (very close to zero). Notice that each output layer weight was also initialized to be very close to $1/N_\ell$. This approach aims to keep the parameter values within a manageable range compared to the optimal value for a single node and we found this training approach to converge more efficiently than a stagewise approach.

However, the approach did not perform well for training multi-layer networks, likely because the hidden layer nodes function more as "routing" tanks than as "storage" tanks, leading to improper parameter initialization. So, we instead followed the *Wang & Gupta (2024a,b)* approach and progressively increased the numbers of nodes in a stagewise manner (on top of the trained single layer network) so that only the weights associated with newly added nodes were initialized to random values (between $-1$ to 1), while the weights associated with previously trained nodes were initialized to their optimal values previously obtained.

***Brief Summary of Training Procedure and Information Summarized***

Each network architecture was trained 10 times, from different random initializations of the weights, for 1000 to 5000 epochs using $KGE$ as the objective function. All single-layer cases were trained for 3000 epochs, except for the MLB cases, which were trained for 5000 epochs. Most multi-layer cases were trained for 1000 epochs, with a few trained for 2000 epochs. Results are reported for the model that obtained the highest selection-period $KGE_{ss}$.

A summary of the single-layer $KGE_{ss}$ scores at various percentiles is provided in Table 2 of the main text. The corresponding results for $\alpha_*^{KGE}$, $\beta_*^{KGE}$, and $r^{KGE}$ are presented in Tables S1 to S3 of supplementary materials. Also, the results based on flow separation into five groups for $KGE_{ss}$, $\alpha^{KGE}$, $\beta^{KGE}$ , and $r^{KGE}$ are shown in Tables S4 to S7 of the supplementary materials. Additionally, we provide the annual root mean square error ($RMSE$) and mean absolute error ($MAE$) at various percentiles in Table S8 and Table S9 to encompass all information needed to demonstrate the model's performance of supplementary materials. The architectural hypotheses for the single-layer and multi-layer models are summarized in Table 1 and Table S10 of the supplementary materials, respectively. Detailed information on the training setups for both single-layer and multi-layer cases is provided in Table S11 and Table S12 of the supplementary materials.

***Summary of the Procedure used for the 500+ US CAMELS Basins***

***Summary of Data Splitting Algorithm and Model Training Setup***

For all of the MN networks experiments conducted for CAMELS US, we adopted the same training procedure as was used for the Leaf River basin. Data was partitioned into training, selection, and testing ($\mathcal{D}_{train}$: $\mathcal{D}_{select}$ : $\mathcal{D}_{test}$) portions using the ratios 2:1:1, respectively. Accordingly, the total of 9,862 daily data points resulted in a training subset of 4,930 time steps, and selection and testing subsets of 2,466 time steps each. For state initialization, WY 1982 was repeated three times at the beginning, resulting in a total simulation period of 10,957 time steps.

The LSTM models were developed separately for each basin, and were trained against the $\mathrm{KGE}$ metric. The training procedure and data splitting algorithm are largely consistent with those described in the above sections used for leaf river basin, but with 2000 epochs and a constant learning rate of $1.25 \times 10^{-2}$ used for training. We tested several hyperparameter sets, including batch sizes of 256 and 512, and sequence lengths of 270, 300, 330, 360, and 390 days. Each case was run with 10 random seeds, and the model with the highest $\mathrm{KGE}_{ss}$ score on the selection set was selected from among those combinations.

Text 3: Working examples of MCP-based model

***Summary of the working example of the single node case***

In this supplementary material, we aim to provide very simple illustrative examples based on the single-node computational unit $MN\{O_\sigma L_{\sigma+}^{con}\}$, so that the complete computational workflow of rainfall-runoff dynamics can be more easily understood. Here, we utilize the example from the Leaf River basin (see main text) for which 40 years of precipitation, potential evapotranspiration (PET), and streamflow data are used.

The rainfall-runoff mapping is achieved using the mass conservation equation as:

$$\frac{dX}{dt} = U - O - L \tag{T8}$$

Here, *X* represents the internal state of the system, while *U*, *O*, and *L* denote the input, output, and loss, respectively. We use the Explicit Euler approach to discretize equation (T8) to obtain equation (T9):

$$X_{t+1} = X_t + U_t - O_t - L_t \tag{T9}$$

In equation (T9), the internal state is updated for every daily timestep, where $U_t$ is precipitation, and the output streamflow $O_t$ and system loss $L_t$ are calculated based on the equations (T10) and (T11) respectively.

$$O_t = G_t^O \times X_t \tag{T10}$$

$$L_t = G_t^L \times X_t \tag{T11}$$

Here, $G_t^O$ and $G_t^L$ denote the output and loss conductivity functions respectively. Equation (T9) then can be written as equation (T12) after plugging in equations (T10) and (T11):

$$\begin{aligned} X_{t+1} &= X_t + U_t - G_t^O \times X_t - G_t^L \times X_t \\ &= U_t + (1 - G_t^O - G_t^L) \times X_t \\ &= U_t + G_t^R \times X_t \end{aligned} \tag{T12}$$

where $G_t^R$ denotes the remember gate computed as $G_t^R$=$1 - G_t^O - G_t^L$ to ensure conservation of mass. In this work, we assume $G_t^O = \kappa_O \cdot \sigma(S_t^O)$ and $G_t^L = \kappa_L \cdot \sigma(S_t^L)$ respectively, where $S_t^O = a_O + b_O\, X_t$ and $S_t^L = a_L + b_L^*\, X_t + b_L\, PET_t$, and where $\sigma$ can be any appropriate ML activation function (here chosen to be the sigmoid activation function $\sigma(S) = 1/(1 + \exp(-S))$), and $\kappa_O$, $a_O$, $b_O$, $\kappa_L$, $a_L$, $b_L^*$ and $b_L$ are trainable parameters including the $\kappa_R$ for the remember gate. Hence, all three gates can vary on the range [0,1].

We assume the loss is of the evapotranspirative type, and depends on both PET and the internal state at the current timestep, where the relationship between the loss gate $G_t^L$ and these variables follows a monotonic non-decreasing sigmoid function. We implement one more constraint, that the loss $L_t$ should be constrained to be smaller than or equal to actual $PET_t$ ($D_t$) at each timestep across all 40 years. Namely, $D_t = G_t^{L^{con}} \cdot X_t$, where:

$$G_t^{L^{con}} = G_t^L - ReLU(G_t^L - \frac{D_t}{X_t}) \tag{T13}$$

The trajectory of the loss is then updated as equation (T14):

$$L_t^{con} = G_t^{L^{con}} \times X_t \tag{T14}$$

Notice that the $L_t^{con}$ is only different from $L_t$ when $L_t > D_t$. Through this, the remember gate is updated as equation (T15)

$$G_t^R = 1 - G_t^O - G_t^{L^{con}} \tag{T15}$$

In summary, the rainfall-runoff mapping is calculated through equation (T12) across all 40 years. The internal state at each time step is updated for the purpose of calculating the output streamflow ($O_t$) using equation (T10) and the constrained system loss ($L_t^{con}$) using equation (T14).

Below **Figure T1** shows the Hydrograph of WY 1976 simulated by the $MN\{O_\sigma L_{\sigma+}^{con}\}$ model which has the annual $KGE_{ss}$skill of 0.91. Overall, we see the model can accurately simulate both high and low flows.

**Figure T2** shows the internal functioning of the $MN\{O_\sigma L_{\sigma+}^{con}\}$ model. **Figure T2a** shows that the internal cell state must reach 500–600 mm before the gate starts to activate, thereby generating streamflow outflow (releasing water from the catchment). Since the loss gate depends on both the PET and the cell state at the current timestep, we can plot the gate as the 2D surface as shown in **Figure T2b.** For larger values of both the cell state and PET, more loss is observed (indicated in yellow). **Figure T2c** also shows the 2D remember gate, which retains more water (higher values) when both PET and cell state are small (yellow color), and vice versa. These plots are consistent with physical understanding and provide a visualization of how the relationship between the gating function and the hypothesized independent covariate actually behaves.

**Figure T2d-f** show the time series of output gate, loss gate and remember gate values during WY 1976. Given that the output gate has a non-decreasing relationship to the cell state , **Figure T2d** demonstrates how the output gate values change with different magnitudes of the cell state over time.

**Figure T2e** shows the gate series (black solid: $G_t^{L^{con}}$; blue dash: $G_t^L$) and PET time series (gray color). Note that that the original blue dashed line is greater than the adjusted black solid line at most time steps, except for a few instances around late August and early September where they are equal. This means that $G_t^L$ tends to generate more loss before the PET constraint described in Equation (13) is applied. In general, we still observe a certain degree of correlation between the magnitude of PET and $G_t^{L^{con}}$, since PET is one of the independent variables that control the magnitude of the loss gate.

Lastly, the remember gate time series is shown in **Figure T2f**. We also plot the cell state time series to demonstrate that the internal functioning of the $MN\{O_\sigma L_{\sigma+}^{con}\}$ model is consistent with physical understanding. Specifically, the remember gate value is low when the cell state value increases. This indicates that the system tends to lose water (i.e., retains less) when more water is present within the catchment.

To conclude, we have presented the mass-conservation formulation for the example of a single-node $MN\{O_\sigma L_{\sigma+}^{con}\}$ architecture, and we have outlined all equations associated with the gating functions, thereby illuminating the entire computational workflow and explaining how the MCP approach models precipitation–runoff dynamics at the catchment scale.

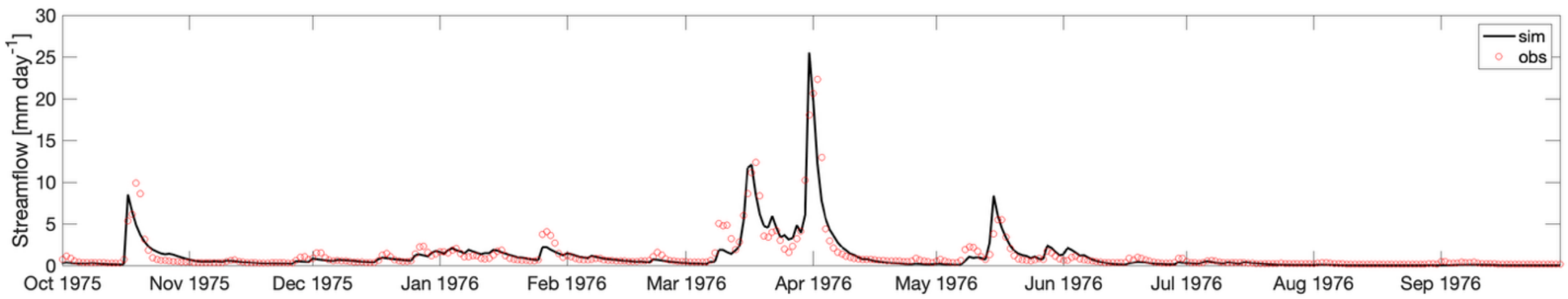

**Figure T1.** Hydrograph of WY 1976 simulated by the $MN\{O_{\sigma}L_{\sigma+}^{con}\}$ model

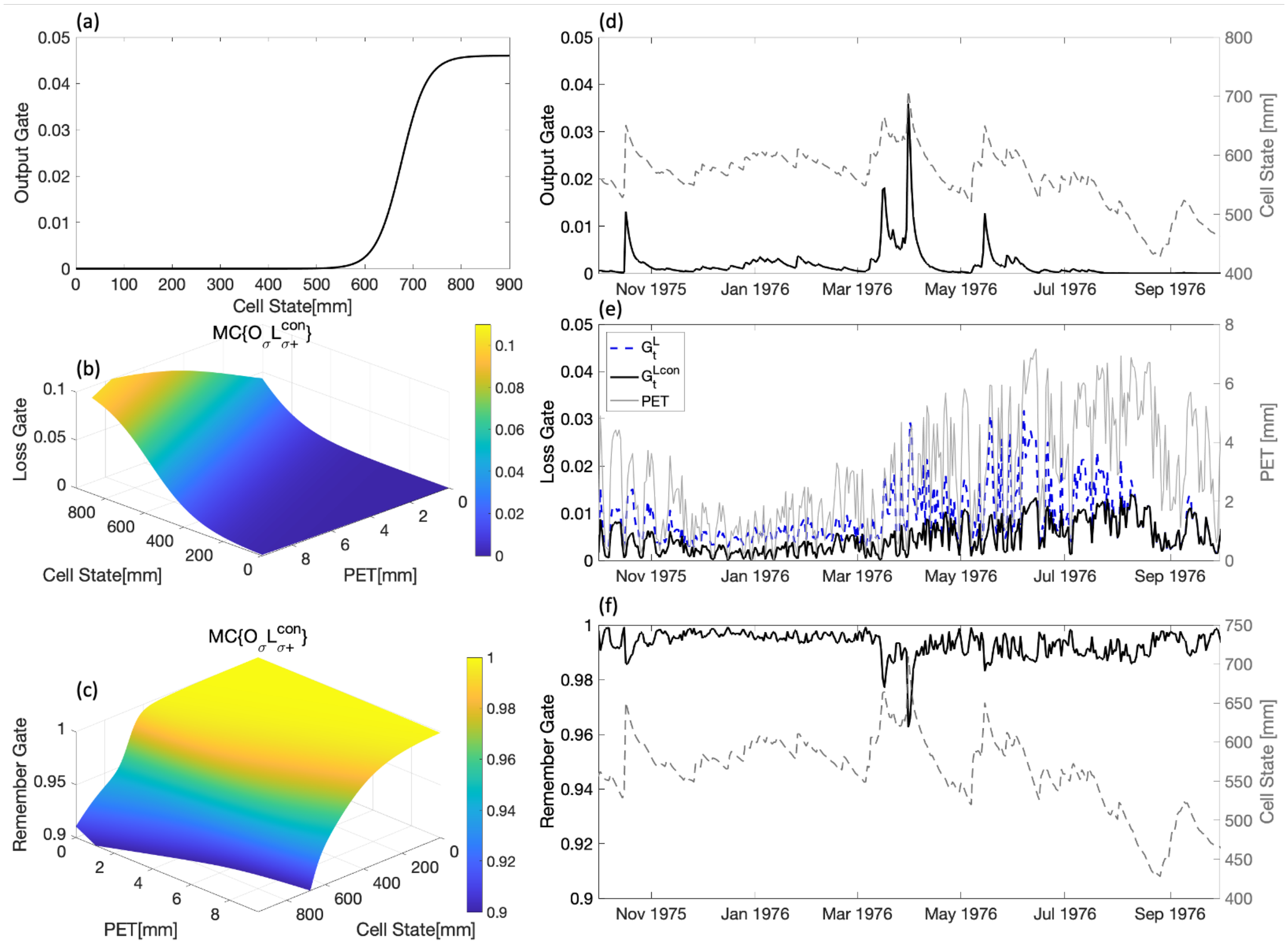

**Figure T2.** Subplots (a) through (f) show results for the $MN\{O_{\sigma}L_{\sigma+}^{con}\}$ model architecture only. Subplots (a) through (c) show the learned forms of the Output, Loss and Remember gating functions respectively. The time series plot of subplot (d) to (f) shows the results from WY 1976. Subplot (d) shows time-evolution plots of the Output gate (solid black line) and corresponding cell state (dashed gray line). Subplot (e) shows time-evolution plots of the PET-constrained-Loss gate (solid black line), unconstrained-Loss gate (dashed blue line), and corresponding Potential Loss (solid gray line). Subplot (f) shows time-evolution plots of the Remember gate (solid black line) and corresponding cell state (dashed gray line).

***Brief note on the single-layer multi-node cases and the multilayers cases***

We have demonstrated the entire calculation workflow and its physical interpretability, as well as visualized the relationship between the gating functions and the independent variables for the single-node $MN\{O_{\sigma}L_{\sigma+}^{con}\}$ case. Here, we explain how the single-layer multi-node cases and the multi-layer cases work in our manuscript.

We provide four examples of using $MN\{O_{\sigma}L_{\sigma+}^{con}\}$ as the basic unit to form the neural network. Examples 8 to 10 are single-layer cases (**Figure T3**), whereas Example 4 uses a three-layer architecture (**Figure T4**).

***Example 1:*** *Distributed-State (DS) and Distributed-State-Relaxed (DSR) Single-Layer Multi-Node Case*

Assume we have a distributed-state, single-layer configuration with two parallel nodes, denoted as $MN_{None}^{DS}(2)$ or $MN_{None}^{DSR}(2)$, that depends on the constraints at the linear output layers. Each node produces its own flow output using Equations (T16) and update the internal state at each nodal level using (T10). Those two equations are extending from Equations (10) and (12) and the constraint from Equation (13) to the multi-node case.

$$h_t^i = G_t^{O^i} \times X_t^i \tag{T16}$$

$$\begin{aligned} X_{t+1}^i &= X_t^i + U_t - G_t^{O^i} \times X_t^i - G_t^{L_i^{con}} \times X_t^i \\ &= U_t + \left(1 - G_t^{O^i} - G_t^{L_i^{con}}\right) \times X_t^i \\ &= U_t + G_t^{R^i} \times X_t^i \end{aligned} \tag{T17}$$

The superscript i denotes the node index, which in this case ranges from 1 to 2. Note that input precipitation $U_t$has the same value for both nodes, meaning the outflow generated by each node represents the simulated streamflow. Each node is initialized using the optimal single-node parameters with added noise for variability (see main text for more details).

Since there are two nodes in parallel, each node contains its own set of parameters from the $MN\{O_{\sigma}L_{\sigma+}^{con}\}$ architecture, as previously mentioned. Those include the parameters in the output gate: $\kappa_O^i, a_O^i, b_O^i$, loss gate: $\kappa_L^i, a_L^i, b_L^i, b_L^{i*}$, and the remember gate: $\kappa_R^i$. Similarly, the superscript i denotes the node index, which in this case ranges from 1 to 2.

Assume that we allow the gate to share cell-state information from another node (denoted as $MN_{SALO}^{DS}$, or $MN_{SALO}^{DSR}$ if both output and loss gates are allowed to share the cell-state information inside the gate). Then, the output gate and loss gate for the 1st and 2nd nodes are:

1st Node: $G_t^{O^1} = \kappa_O^1 \cdot \sigma(a_O^1 + b_O^{11}X_t^1 + b_O^{12}X_t^2)$; $G_t^{L^1} = \kappa_L^1 \cdot \sigma(a_L^1 + b_L^{11*}X_t^1 + b_L^{12*}X_t^2 + b_L^1 PET_t)$

$$G_t^{L_1^{con}} = G_t^{L^1} - ReLU(G_t^{L^1} - \frac{PET_t}{X_t^1})$$

2nd Node: $G_t^{O^2} = \kappa_O^2 \cdot \sigma(a_O^2 + b_O^{21}X_t^1 + b_O^{22}X_t^2)$; $G_t^{L^2} = \kappa_L^2 \cdot \sigma(a_L^2 + b_L^{21*}X_t^1 + b_L^{22*}X_t^2 + b_L^2 PET_t)$

$$G_t^{L_2^{con}} = G_t^{L^2} - ReLU(G_t^{L^2} - \frac{PET_t}{X_t^2})$$

In any case, the output $h_t^i$ needs to go through an output linear transformation layer to obtain the streamflow prediction $h_t^i \cdot w_k^{out} = O_t$. The subscript k is the order of node and so here ranges from 1 to 2

Using the non-information-sharing case as example, in the distributed state $MN_{None}^{DS}$ network case we have $\sum_k w_k^{out} = 1$, $w_k^{out} \geq 0$ for all $k$, so that mass is conserved at the network level. In the distributed state relaxed $MN_{None}^{DSR}$ network case, we have $\sum_k w_k^{out} \neq 1$, $w_k^{out} \geq 0$ for all k, so that mass is relaxed at the network level.

For the DS and DSR cases, the final flow prediction is the weighed sum of flow generated from each path (node) at the linear output layer, the weights in the linear input (distributed) layer in this case all equal to 1 ($w_j^{in} = 1$). Therefore, the output streamflow can be derived from equation (T18).

$$O_t = \sum_i w_i^{out} h_t^i \tag{T18}$$

Again, we can use a similar approach for architectures that allow cell-state information sharing within the gates (i.,e $MN_{SALO}^{DI}$ $and$ $MN_{SALO}^{DIR}$ if the sharing is allowed at both output and loss gates) .

***Example 2:*** *Distributed-Input (DI) and Distributed-State-Relaxed (DIR) Single-Layer Multi-Node Case*

The main idea of the DI or DIR cases is that the catchment can be partitioned into different numbers of response units.

Assume we have a distributed-input, single-layer configuration with two parallel nodes which denoted as $MN_{None}^{DI}(2)$ or $MN_{None}^{DIR}(2)$ that depends on the constraints at the input-distributed layers. Each node produces its own flow output using the same Equation (T16) and updates the internal state at each nodal level using Equation (T19), which extends Equations (10) and (12) to the multi-node case.

$$\begin{aligned} X_{t+1}^i &= X_t^i + U_t - G_t^{O^i} \times X_t^i - G_t^{L_i^{con}} \times X_t^i \\ &= U_t^i + \left(1 - G_t^{O^i} - G_t^{L_i^{con}}\right) \times X_t^i \\ &= U_t^i + G_t^{R^i} \times X_t^i \end{aligned} \tag{T19}$$

Here, the superscript i denotes the node index, which in this case ranges from 1 to 2. The difference from the DS case is that the input total precipitation ($U_t$) is distributed into each node with a linear-distributed gate (function). It is a level-wise mass-conserving system if the $\sum_j w_j^{in} = 1$, $w_j^{in} \geq 0$, we denote this case as distributed-input case $MN_{None}^{DI}$. Conversely, if the mass in the network level is not conserved in the sense that $\sum_j w_j^{in} \neq 1$, $w_j^{in} \geq 0$, we denote this case as distributed-input-relaxed case $MN_{None}^{DIR}$.

For the DI and DIR cases, the possible mass adjustment is hypothesized occurring at the input-distributed layer, the weight in the linear output layer in this case all equals to 1($w_k^{out} = 1$). Therefore, the output streamflow is the summation of flow derived from each node in equation (T20).

$$O_t = \sum_i h_t^i = \sum_i G_t^{O^i} \times X_t^i \tag{T20}$$

Again, we can use a similar approach for the architecture that allows cell-state information sharing within the gates (i.,e $MN^{DI}_{SALO}$ $and$ $MN^{DIR}_{SALO}$ if the sharing is allowed at both output and loss gates) .

***Example 3:*** *Machine-Learning Benchmark (MLB) Single-Layer Multi-Node Case*

The last type of single-layer network is the Machine-Learning Benchmark case, which can be understood as a hybrid of the Distributed-Input and Distributed-State cases. The goal is to provide a benchmark that allows greater flexibility within the network architecture, thereby offering an upper-bound benchmark for predictive accuracy.

Assume we have the MLB single-layer configuration with two parallel nodes which denoted as $MN^{MLB}_{None}(2)$. Each node produces its own flow output using the same above Equation (T16) and update the internal state at each nodal level using the Equation (T19). In other words, we allow the total precipitation mass to be adjusted at the input-distribution gate and the outflow at the linear output gate.

Notice however that we still only allow the $w^{in}_j \geq 0$, $\sum_j w^{in}_j \neq 1$ so that there will have no possible negative value used as precipitation input to the node. On the contrary, we relax the constraints for the output weights and add one extra bias terms to enhance the flexibility for the network to combine the information. Hence, the weights $w^{out}_k \in R$ and the bias $w^{out}_o \in R$.

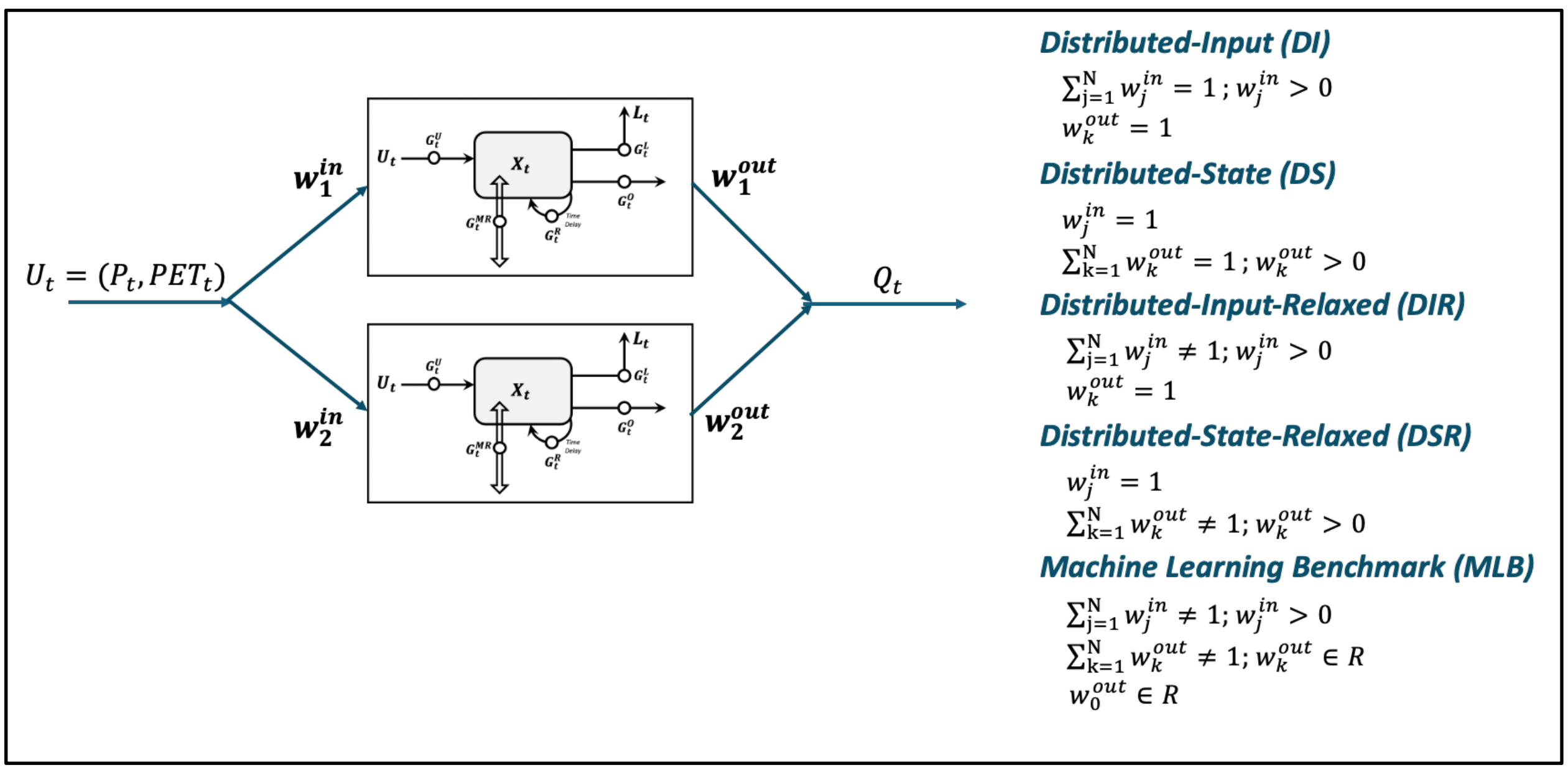

**Figure T3.** The figure 3b in the main text showing the mathematical formulation, illustrated using a single-layer two-node configuration, summarizes the five different cases tested. The differences among these cases stem from the use of linear weights at both the input ($w^{in}_j$) and output ($w^{out}_k$) layers. These include the mass-conserving configurations—distributed-input (DI) and distributed-state (DS). The distributed-input-relaxed (DIR) and distributed-state-relaxed (DSR) cases allow mass relaxation by relaxing the values of weights at the input and output layers, respectively. The machine learning benchmark (MLB) case combines both DIR and DSR, with additional bias term $w^{out}_0$ allowing mass relaxation at both the input (distribution) and output layer. Note that the $P_t$, and $Q_t$ denote the precipitation and outflow respectively while the input $U_t$ includes both $P_t$ and evapotranspiration ($PET_t$), only $P_t$ is used as mass input whereas the $PET_t$ is used as contextual

variable feed in to the loss gate. The $w_j^{in}$ and $w_k^{out}$denote the weights of the input-distribution and the linear output layer, respectively, with subscripts j and k indicating the dimension, both ranging from 1 to 2 in this case.

***Example 4:*** *Distributed-State Multi-Layer Multi-Node Case*

We only test the distributed-state (DS) architecture for the multi-layer cases. Overall, we evaluate three different transformation layers between adjacent layers, as well as from the last layer to the output. Here, we use a network with 2 nodes in the first and second layers, and 1 node in the final (third) layer as an illustrative example shown in **Figure T4**. The information regarding the meaning of each state variable as well as the weights and bias in each layer is summarized in **Table T1**.

Three cases with different uses of the linear transformation layer are named **distributed-state (DS)**, **distributed-state-relaxation (DSR)**, and **distributed-state machine-learning benchmark (DS-MLB)**. The main difference among these three architectures is that only the first one tries to preserve mass from the input to the final output, while the last two cases allow varying degrees of mass relaxation between layers.

Note that one can certainly activate the features for cell-state information sharing in the output and loss gates. Here, we mainly use the version of network without information sharing at each layer as the example for below illustration and these three models denote as $MN_{None}^{DS}(2,1,1)$, $MN_{None}^{DSR}(2,1,1)$, $MN_{None}^{MLB}(2,1,1)$.

For the nodes in first layer, the detail mathematical formulation for the calculation from the input to first layer is shown in equation (T21) to (T24):

$$h_t^{11} = G_t^{O^{11}} \times X_t^{11} \quad \text{(T21)}$$

$$X_{t+1}^{11} = U_t + \left(1 - G_t^{O^{11}} - G_t^{L_{11}^{con}}\right) \times X_t^{11} \quad \text{(T22)}$$

$$h_t^{12} = G_t^{O^{12}} \times X_t^{12} \quad \text{(T23)}$$

$$X_{t+1}^{12} = U_t + \left(1 - G_t^{O^{12}} - G_t^{L_{12}^{con}}\right) \times X_t^{12} \quad \text{(T24)}$$

$h_t^{1i}$, $G_t^{O^{1i}}$, $G_t^{L_{1i}^{con}}$, and $X_t^{1i}$ represent the outflow, output gate, loss gate and the internal state in the i[th] node in the first layer. $U_t$ is the mass input precipitation.

The linear transformation between first and second layer is shown in equation (T25) to (T26):

$$h_t^{21(in)} = h_t^{11} \times w_{11}^{L12} + h_t^{12} \times w_{21}^{L12} + w_0^{L12} \quad \text{(T25)}$$

$$h_t^{22(in)} = h_t^{11} \times w_{12}^{L12} + h_t^{12} \times w_{22}^{L12} + w_0^{L12} \quad \text{(T26)}$$

$h_t^{2i(in)}$ is the inflow to the 2nd layer of the i[th] node. $w_{jk}^{L12}$ is the weight of linear transformation layer between 1st and 2nd layer corresponds to the j[th] node in the 1st layer and kth node in the 2nd layer. $w_0^{L12}$ is the bias for the linear transformation layer between 1st and 2nd layer.

The constraint associated with weights and bias for different cases including:

**Distributed-State (DS):**

$$w_{11}^{L12} + w_{21}^{L12} = 1;\ w_{i1}^{L12} > 0 \quad \text{(T27)}$$

$$w_{12}^{L12} + w_{22}^{L12} = 1;\ w_{i2}^{L12} > 0 \quad \text{(T28)}$$

The constraint of equation (T27) and (T28) ensures the flux (mass) generated through the 1st layer is mass conserved when feeding into the node as input at the subsequent layer (2nd layer). In this case, another criterion needs to follow is that the bias term $w_0^{L12} = 0$.

**Distributed-State-Relaxed (DSR):**

$$w_{11}^{L12} + w_{21}^{L12} \neq 1;\ w_{i1}^{L12} > 0 \quad \text{(T29)}$$

$$w_{12}^{L12} + w_{22}^{L12} \neq 1;\ w_{i2}^{L12} > 0 \quad \text{(T30)}$$

The constraint in equations (T29) and (T30) ensures that the flux (mass) transferred through the first layer between layers is weighted positively, and that mass conservation is relaxed when feeding into the nodes at the subsequent (second) layer. In this case, we introduce mass relaxation using positive weights only, without allowing bias terms to play a role ($w_0^{L12} = 0$).

**Distributed-State Machine-Learning Benchmark (DS-MLB):**

$$w_{ij}^{L12} \in\ R \quad \text{(T31)}$$

$$w_0^{L12} \in\ R \quad \text{(T32)}$$

In DS-MLB, all weights and biases can take both positive and negative values as shown in equations (T31) and (T32), ensuring the greatest flexibility for the network during training and development.

For the nodes in second layer, the detail mathematical formulation for the calculation of state update and the output in the second layer is shown in equation (T33) to (T36):

$$h_t^{21(out)} = G_t^{O^{21}} \times X_t^{21} \quad \text{(T33)}$$

$$X_{t+1}^{21} = h_t^{21(in)} + \left(1 - G_t^{O^{21}} - G_t^{L_{21}^{con}}\right) \times X_t^{21} \quad \text{(T34)}$$

$$h_t^{22(out)} = G_t^{O^{22}} \times X_t^{22} \quad \text{(T35)}$$

$$X_{t+1}^{22} = h_t^{22(in)} + \left(1 - G_t^{O^{22}} - G_t^{L_{22}^{con}}\right) \times X_t^{22} \quad \text{(T36)}$$

$h_t^{2i(out)}$, $G_t^{O^{2i}}$, $G_t^{L_{2i}^{con}}$, and $X_t^{2i}$ represent the outflow, output gate, loss gate and the internal state in the i$^{th}$ node in the second layer. $h_t^{2i(in)}$ is the inflow (input) to the i$^{th}$ node in the second layer

The linear transformation between second and third layer is shown in equation (T37):

$$h_t^{31(in)} = h_t^{21(out)} \times w_{11}^{L23} + h_t^{22(out)} \times w_{21}^{L23} + w_0^{L23} \quad \text{(T37)}$$

$h_t^{3i(in)}$ is the inflow to the 3rd layer of the i$^{th}$ node. $w_{jk}^{L23}$ is the weight of linear transformation layer between 2nd and 3rd layer corresponds to the j$^{th}$ node in the 2nd layer and k$^{th}$ node in the 3rd layer. $w_0^{L23}$ is the bias for the linear transformation layer between 2nd and 3rd layer.

The constraint associated with weights and bias for different cases including:

**Distributed-State (DS):**

$$w_{11}^{L23} + w_{21}^{L23} = 1;\ w_{i1}^{L23} > 0 \quad \text{(T38)}$$

The constraint of equation (T38) ensures the flux (mass) generated through the 2nd layer between layers is mass conserved when feeding into the node as input at the subsequent layer (3rd layer). In this case, another criterion needs to follow is that the bias term $w_0^{L23} = 0$.

**Distributed-State-Relaxed (DSR):**

$$w_{11}^{L23} + w_{21}^{L23} \neq 1;\ w_{i1}^{L23} > 0 \tag{T39}$$

The constraint in equations (T39) ensures that the flux (mass) transferred through the second layer is weighted positively, and that mass conservation is relaxed when feeding into the nodes at the subsequent (third) layer. In this case, we introduce mass relaxation using positive weights only, without allowing bias terms to play a role ($w_0^{L23} = 0$).

**Distributed-State Machine-Learning Benchmark (DS-MLB):**

$$w_{ij}^{L13} \in\ R \tag{T40}$$

$$w_0^{L23} \in\ R \tag{T41}$$

In DS-MLB, all weights and biases can take both positive and negative values, as shown in equations (T40) and (T41), ensuring the greatest flexibility for the network during training and development.

For the node in third layer, the detail mathematical formulation for the calculation of state update and the output in the third layer is shown in equation (T42) to (T43):

$$h_t^{31(out)} = G_t^{O^{31}} \times X_t^{31} \tag{T42}$$

$$X_{t+1}^{31} = h_t^{31(in)} + \left(1 - G_t^{O^{31}} - G_t^{L_{31}^{con}}\right) \times X_t^{31} \tag{T43}$$

$h_t^{3i(out)}$, $G_t^{O^{3i}}$, $G_t^{L_{3i}^{con}}$, and $X_t^{3i}$ represent the outflow, output gate, loss gate and the internal state in the i^th^ node in the second layer. $h_t^{3i(in)}$ is the inflow (input) to the i^th^ node in the second layer.

The linear transformation between third layer and the output is shown in equation (T44):

$$Q_t = h_t^{31(out)} \times w_{11}^{L3out} + w_0^{L3out} \tag{T44}$$

The constraint associated with weights and bias for different cases including:

**Distributed-State (DS):**

$$w_{11}^{L3out} = 1 \tag{T45}$$

The constraint of equation (T45) ensures the flux (mass) generated through the third layer between layers is mass conserved when feeding into the node as input at the subsequent layer (3rd layer). In this case, another criterion needs to follow is that the bias term $w_0^{L3\text{out}} = 0$.

**Distributed-State-Relaxed (DSR):**

$$w_{11}^{L3out} > 0 \tag{T46}$$

The constraint in equations (T46) ensures that the flux (mass) transferred through the third layer is weighted positively, and that mass conservation is relaxed when feeding into the nodes at the subsequent (third) layer. In this case, we introduce mass relaxation using positive weights only, without allowing bias terms to play a role ($w_0^{L3out} = 0$).

**Distributed-State Machine-Learning Benchmark (DS-MLB):**

$$w_{ij}^{L3out} \in\ R \tag{T47}$$

$$w_0^{L3out} \in\ R \tag{T48}$$

In DS-MLB, again all weights and biases can take both positive and negative values as shown in equations (T47) and (T48), ensuring the greatest flexibility for the network during training and development.

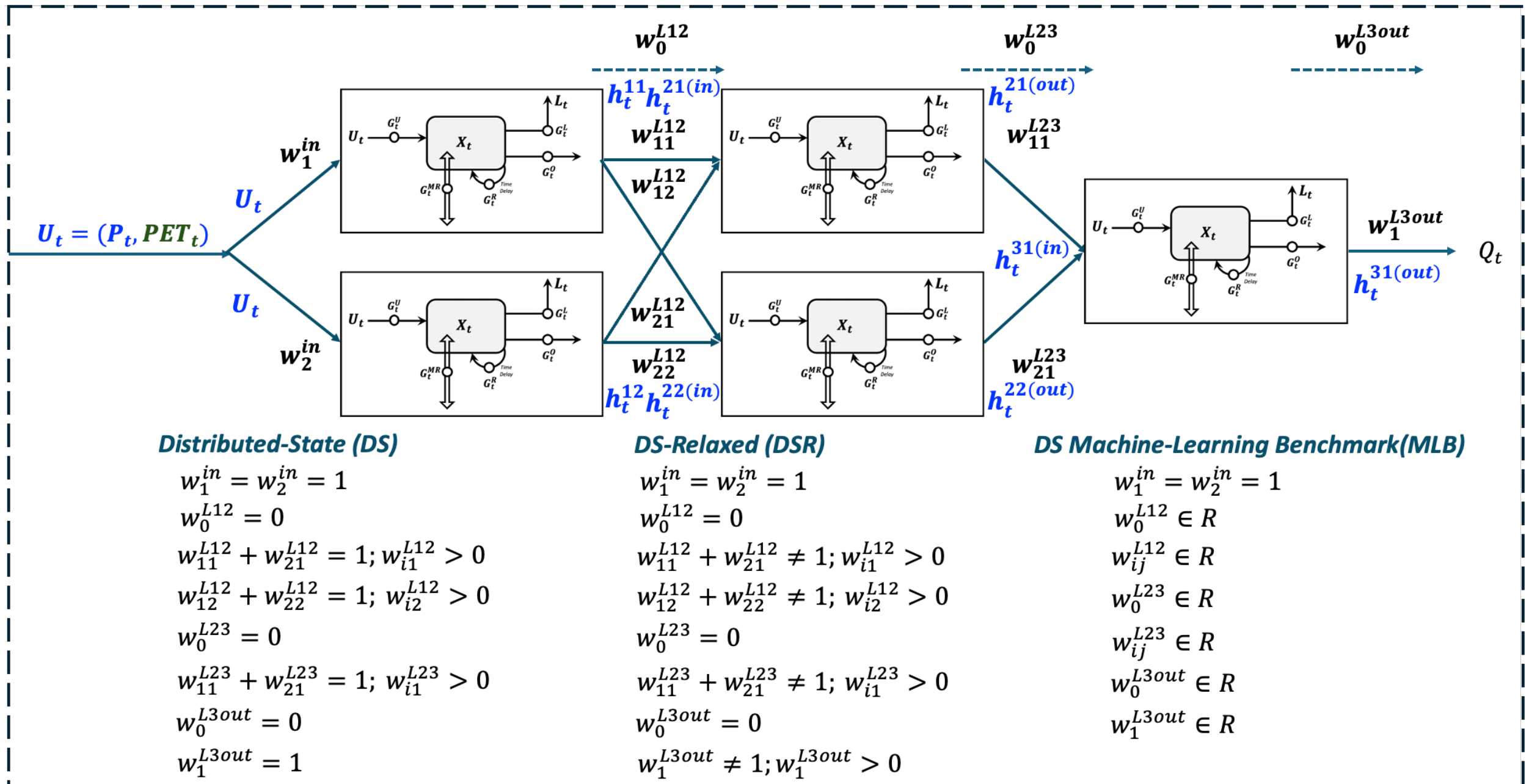

**Figure T4** The mathematical formulation, illustrated using a three-layer configuration with 2, 2, and 1 nodes respectively, summarizes the three different cases tested including distributed-state (DS), distributed-state-relaxed (DSR), and distributed-state machine-learning benchmark (DS-MLB). All w related symbols refer to the weights and bias within the architectures. The symbol in blue text denotes the internal state variables. We refer the readers to **Table T1** for the detailed explanation of each parameter and state variable.

**Table T1.** Detailed description of weights, biases, and state variables for the illustrative multi-layer, multi-node example

| Weights/Bias | Explanation |
|---|---|
| $w_1^{in}$ | The weight of the input layer for the 1st node |
| $w_2^{in}$ | The weight of the input layer for the 2nd node |
| $w_0^{L12}$ | The bias of the linear transformation layer between 1st and 2nd layer |
| $w_{11}^{L12}$ | The weight in the linear transformation layer between 1st and 2nd layer correspond to the 1st node in the 1st layer and 1st node in the 2nd layer |
| $w_{12}^{L12}$ | The weight in the linear transformation layer between 1st and 2nd layer correspond to the 1st node in the 1st layer and 2nd node in the 2nd layer |
| $w_{21}^{L12}$ | The weight in the linear transformation layer between 1st and 2nd layer correspond to the 2nd node in the 1st layer and 1st node in the 2nd layer |
| $w_{22}^{L12}$ | The weight in the linear transformation layer between 1st and 2nd layer correspond to the 2nd node in the 1st layer and 2nd node in the 2nd layer |
| $w_0^{L23}$ | The bias of the linear transformation layer between 2nd and 3rd layer |
| $w_{11}^{L23}$ | The weight in the linear transformation layer between 2nd and 3rd layer correspond to the 1st node in the 2nd layer and 1st node in the 3rd layer |
| $w_{21}^{L23}$ | The weight in the linear transformation layer between 2nd and 3rd layer correspond to the 2nd node in the 2nd layer and 1st node in the 3rd layer |
| $w_0^{L3out}$ | The bias of the linear transformation layer between 3rd and the output |
| $w_1^{L3out}$ | The weight in the linear transformation layer between 3rd layer and the output |
| **State Variables** | **Explanation** |
| $P_t$ | Precipitation at the given daily timestep |
| $PET_t$ | Potential Evapotranspiration at the given daily timestep |
| $U_t$ | Input includes Precipitation ($P_t$) and Potential Evapotranspiration ($PET_t$) at the given daily timestep. The $P_t$ is treated as mass input whereas the $PET_t$ is used as contextual feeding variables to the loss gate |
| $h_t^{11}$ | Outflow from the 1st node of the 1st layer |
| $h_t^{12}$ | Outflow from the 2nd node of the 1st layer |
| $h_t^{21(in)}$ | Inflow from the 1st node of the 2nd layer |
| $h_t^{22(in)}$ | Inflow from the 2nd node of the 2nd layer |
| $h_t^{21(out)}$ | Outflow from the 1st node of the 2nd layer |
| $h_t^{22(out)}$ | Outflow from the 2nd node of the 2nd layer |
| $h_t^{31(in)}$ | Inflow from the 1st node of the 3rd layer |
| $h_t^{31(out)}$ | Outflow from the 1st node of the 3rd layer |
| $O_t$ | Output Streamflow |

Table S1. $1-|1-\alpha^{KGE}|$ ($\alpha_*^{KGE}$) scores for the Single-Layer Mass-Conserving Neural Networks (MNs)

**1-Node**

| | $MN_{none}^{DI}$ | $MN_{none}^{DS}$ | $MN_{none}^{DIR}$ | $MN_{none}^{DSR}$ | $MN_{none}^{MLB}$ | | $MN_{SAL}^{DI}$ | $MN_{SAL}^{DS}$ | $MN_{SAL}^{DIR}$ | $MN_{SAL}^{DSR}$ | $MN_{SAL}^{MLB}$ | | $MN_{SAO}^{DI}$ | $MN_{SAO}^{DS}$ | $MN_{SAO}^{DIR}$ | $MN_{SAO}^{DSR}$ | $MN_{SAO}^{MLB}$ | | $MN_{SALO}^{DI}$ | $MN_{SALO}^{DS}$ | $MN_{SALO}^{DIR}$ | $MN_{SALO}^{DSR}$ | $MN_{SALO}^{MLB}$ |
|---|---|---|---|---|---|---|---|---|---|---|---|---|---|---|---|---|---|---|---|---|---|---|---|
| $KGE_{ss}^{min}$ | 0.60 | 0.60 | 0.60 | 0.57 | 0.40 | $KGE_{ss}^{min}$ | | | | | | $KGE_{ss}^{min}$ | | | | | | $KGE_{ss}^{min}$ | | | | | |
| $KGE_{ss}^{5\%}$ | 0.73 | 0.72 | 0.73 | 0.70 | 0.51 | $KGE_{ss}^{5\%}$ | | | | | | $KGE_{ss}^{5\%}$ | | | | | | $KGE_{ss}^{5\%}$ | | | | | |
| $KGE_{ss}^{25\%}$ | 0.83 | 0.82 | 0.83 | 0.83 | 0.74 | $KGE_{ss}^{25\%}$ | | | | | | $KGE_{ss}^{25\%}$ | | | | | | $KGE_{ss}^{25\%}$ | | | | | |
| $KGE_{ss}^{50\%}$ | 0.89 | 0.88 | 0.89 | 0.90 | 0.91 | $KGE_{ss}^{50\%}$ | | | | | | $KGE_{ss}^{50\%}$ | | | | | | $KGE_{ss}^{50\%}$ | | | | | |
| $KGE_{ss}^{75\%}$ | 0.93 | 0.93 | 0.93 | 0.94 | 0.97 | $KGE_{ss}^{75\%}$ | | | | | | $KGE_{ss}^{75\%}$ | | | | | | $KGE_{ss}^{75\%}$ | | | | | |
| $KGE_{ss}^{95\%}$ | 0.98 | 0.99 | 0.98 | 0.98 | 0.99 | $KGE_{ss}^{95\%}$ | | | | | | $KGE_{ss}^{95\%}$ | | | | | | $KGE_{ss}^{95\%}$ | | | | | |
| PNs | 8 | 8 | 9 | 9 | 11 | PNs | | | | | | PNs | | | | | | PNs | | | | | |

**2-Node**

| | $MN_{none}^{DI}$ | $MN_{none}^{DS}$ | $MN_{none}^{DIR}$ | $MN_{none}^{DSR}$ | $MN_{none}^{MLB}$ | | $MN_{SAL}^{DI}$ | $MN_{SAL}^{DS}$ | $MN_{SAL}^{DIR}$ | $MN_{SAL}^{DSR}$ | $MN_{SAL}^{MLB}$ | | $MN_{SAO}^{DI}$ | $MN_{SAO}^{DS}$ | $MN_{SAO}^{DIR}$ | $MN_{SAO}^{DSR}$ | $MN_{SAO}^{MLB}$ | | $MN_{SALO}^{DI}$ | $MN_{SALO}^{DS}$ | $MN_{SALO}^{DIR}$ | $MN_{SALO}^{DSR}$ | $MN_{SALO}^{MLB}$ |
|---|---|---|---|---|---|---|---|---|---|---|---|---|---|---|---|---|---|---|---|---|---|---|---|
| $KGE_{ss}^{min}$ | 0.58 | 0.61 | 0.56 | 0.59 | 0.67 | $KGE_{ss}^{min}$ | 0.59 | 0.65 | 0.58 | 0.61 | 0.69 | $KGE_{ss}^{min}$ | 0.60 | 0.62 | 0.61 | 0.70 | 0.65 | $KGE_{ss}^{min}$ | 0.63 | 0.60 | 0.69 | 0.68 | 0.73 |
| $KGE_{ss}^{5\%}$ | 0.66 | 0.70 | 0.67 | 0.69 | 0.71 | $KGE_{ss}^{5\%}$ | 0.59 | 0.68 | 0.71 | 0.68 | 0.73 | $KGE_{ss}^{5\%}$ | 0.73 | 0.66 | 0.69 | 0.72 | 0.70 | $KGE_{ss}^{5\%}$ | 0.70 | 0.67 | 0.70 | 0.71 | 0.74 |
| $KGE_{ss}^{25\%}$ | 0.79 | 0.82 | 0.82 | 0.79 | 0.85 | $KGE_{ss}^{25\%}$ | 0.79 | 0.82 | 0.82 | 0.77 | 0.84 | $KGE_{ss}^{25\%}$ | 0.82 | 0.80 | 0.84 | 0.85 | 0.83 | $KGE_{ss}^{25\%}$ | 0.83 | 0.80 | 0.85 | 0.85 | 0.85 |
| $KGE_{ss}^{50\%}$ | 0.89 | 0.91 | 0.90 | 0.89 | 0.92 | $KGE_{ss}^{50\%}$ | 0.89 | 0.88 | 0.89 | 0.89 | 0.92 | $KGE_{ss}^{50\%}$ | 0.91 | 0.91 | 0.90 | 0.91 | 0.92 | $KGE_{ss}^{50\%}$ | 0.91 | 0.91 | 0.90 | 0.91 | 0.92 |
| $KGE_{ss}^{75\%}$ | 0.94 | 0.95 | 0.96 | 0.95 | 0.95 | $KGE_{ss}^{75\%}$ | 0.94 | 0.95 | 0.96 | 0.95 | 0.95 | $KGE_{ss}^{75\%}$ | 0.94 | 0.97 | 0.96 | 0.94 | 0.96 | $KGE_{ss}^{75\%}$ | 0.95 | 0.96 | 0.97 | 0.94 | 0.97 |
| $KGE_{ss}^{95\%}$ | 0.98 | 0.99 | 0.99 | 0.98 | 0.99 | $KGE_{ss}^{95\%}$ | 0.98 | 0.98 | 0.99 | 0.98 | 0.99 | $KGE_{ss}^{95\%}$ | 0.97 | 0.99 | 0.99 | 0.99 | 0.99 | $KGE_{ss}^{95\%}$ | 0.99 | 0.99 | 0.99 | 0.98 | 0.99 |
| PNs | 18 | 18 | 18 | 18 | 21 | PNs | 20 | 20 | 20 | 20 | 23 | PNs | 20 | 20 | 20 | 20 | 23 | PNs | 22 | 22 | 22 | 22 | 25 |

**3-Node**

| | $MN_{none}^{DI}$ | $MN_{none}^{DS}$ | $MN_{none}^{DIR}$ | $MN_{none}^{DSR}$ | $MN_{none}^{MLB}$ | | $MN_{SAL}^{DI}$ | $MN_{SAL}^{DS}$ | $MN_{SAL}^{DIR}$ | $MN_{SAL}^{DSR}$ | $MN_{SAL}^{MLB}$ | | $MN_{SAO}^{DI}$ | $MN_{SAO}^{DS}$ | $MN_{SAO}^{DIR}$ | $MN_{SAO}^{DSR}$ | $MN_{SAO}^{MLB}$ | | $MN_{SALO}^{DI}$ | $MN_{SALO}^{DS}$ | $MN_{SALO}^{DIR}$ | $MN_{SALO}^{DSR}$ | $MN_{SALO}^{MLB}$ |
|---|---|---|---|---|---|---|---|---|---|---|---|---|---|---|---|---|---|---|---|---|---|---|---|
| $KGE_{ss}^{min}$ | 0.58 | 0.65 | 0.56 | 0.62 | 0.67 | $KGE_{ss}^{min}$ | 0.58 | 0.57 | 0.55 | 0.55 | 0.69 | $KGE_{ss}^{min}$ | 0.64 | 0.70 | 0.61 | 0.73 | 0.64 | $KGE_{ss}^{min}$ | 0.70 | 0.48 | 0.71 | 0.75 | 0.64 |
| $KGE_{ss}^{5\%}$ | 0.64 | 0.69 | 0.67 | 0.65 | 0.73 | $KGE_{ss}^{5\%}$ | 0.59 | 0.64 | 0.63 | 0.68 | 0.72 | $KGE_{ss}^{5\%}$ | 0.69 | 0.76 | 0.66 | 0.76 | 0.73 | $KGE_{ss}^{5\%}$ | 0.74 | 0.65 | 0.77 | 0.78 | 0.77 |
| $KGE_{ss}^{25\%}$ | 0.79 | 0.82 | 0.82 | 0.78 | 0.83 | $KGE_{ss}^{25\%}$ | 0.80 | 0.84 | 0.80 | 0.80 | 0.83 | $KGE_{ss}^{25\%}$ | 0.83 | 0.89 | 0.80 | 0.88 | 0.86 | $KGE_{ss}^{25\%}$ | 0.86 | 0.83 | 0.87 | 0.86 | 0.87 |
| $KGE_{ss}^{50\%}$ | 0.89 | 0.89 | 0.90 | 0.88 | 0.90 | $KGE_{ss}^{50\%}$ | 0.89 | 0.90 | 0.91 | 0.89 | 0.92 | $KGE_{ss}^{50\%}$ | 0.88 | 0.93 | 0.90 | 0.93 | 0.93 | $KGE_{ss}^{50\%}$ | 0.91 | 0.91 | 0.94 | 0.93 | 0.94 |
| $KGE_{ss}^{75\%}$ | 0.94 | 0.95 | 0.95 | 0.94 | 0.97 | $KGE_{ss}^{75\%}$ | 0.94 | 0.95 | 0.94 | 0.95 | 0.98 | $KGE_{ss}^{75\%}$ | 0.96 | 0.98 | 0.97 | 0.97 | 0.96 | $KGE_{ss}^{75\%}$ | 0.97 | 0.96 | 0.96 | 0.96 | 0.98 |
| $KGE_{ss}^{95\%}$ | 0.99 | 0.98 | 0.99 | 0.98 | 1.00 | $KGE_{ss}^{95\%}$ | 0.98 | 0.98 | 0.99 | 0.99 | 1.00 | $KGE_{ss}^{95\%}$ | 1.00 | 1.00 | 1.00 | 1.00 | 0.99 | $KGE_{ss}^{95\%}$ | 0.99 | 0.99 | 1.00 | 0.99 | 1.00 |
| PNs | 27 | 27 | 27 | 27 | 31 | PNs | 33 | 33 | 33 | 33 | 37 | PNs | 33 | 33 | 33 | 33 | 37 | PNs | 39 | 39 | 39 | 39 | 43 |

**4-Node**

| | $MN_{none}^{DI}$ | $MN_{none}^{DS}$ | $MN_{none}^{DIR}$ | $MN_{none}^{DSR}$ | $MN_{none}^{MLB}$ | | $MN_{SAL}^{DI}$ | $MN_{SAL}^{DS}$ | $MN_{SAL}^{DIR}$ | $MN_{SAL}^{DSR}$ | $MN_{SAL}^{MLB}$ | | $MN_{SAO}^{DI}$ | $MN_{SAO}^{DS}$ | $MN_{SAO}^{DIR}$ | $MN_{SAO}^{DSR}$ | $MN_{SAO}^{MLB}$ | | $MN_{SALO}^{DI}$ | $MN_{SALO}^{DS}$ | $MN_{SALO}^{DIR}$ | $MN_{SALO}^{DSR}$ | $MN_{SALO}^{MLB}$ |
|---|---|---|---|---|---|---|---|---|---|---|---|---|---|---|---|---|---|---|---|---|---|---|---|
| $KGE_{ss}^{min}$ | 0.54 | 0.63 | 0.56 | 0.61 | 0.66 | $KGE_{ss}^{min}$ | 0.60 | 0.64 | 0.59 | 0.52 | 0.69 | $KGE_{ss}^{min}$ | 0.62 | 0.74 | 0.59 | 0.69 | 0.70 | $KGE_{ss}^{min}$ | 0.75 | 0.54 | 0.75 | 0.76 | 0.61 |
| $KGE_{ss}^{5\%}$ | 0.66 | 0.67 | 0.68 | 0.65 | 0.71 | $KGE_{ss}^{5\%}$ | 0.61 | 0.72 | 0.63 | 0.66 | 0.74 | $KGE_{ss}^{5\%}$ | 0.68 | 0.76 | 0.64 | 0.76 | 0.79 | $KGE_{ss}^{5\%}$ | 0.79 | 0.71 | 0.79 | 0.80 | 0.71 |
| $KGE_{ss}^{25\%}$ | 0.82 | 0.82 | 0.82 | 0.80 | 0.82 | $KGE_{ss}^{25\%}$ | 0.81 | 0.84 | 0.81 | 0.84 | 0.83 | $KGE_{ss}^{25\%}$ | 0.83 | 0.89 | 0.85 | 0.90 | 0.88 | $KGE_{ss}^{25\%}$ | 0.88 | 0.89 | 0.88 | 0.91 | 0.86 |
| $KGE_{ss}^{50\%}$ | 0.88 | 0.88 | 0.90 | 0.89 | 0.91 | $KGE_{ss}^{50\%}$ | 0.89 | 0.91 | 0.90 | 0.90 | 0.92 | $KGE_{ss}^{50\%}$ | 0.90 | 0.94 | 0.91 | 0.94 | 0.94 | $KGE_{ss}^{50\%}$ | 0.92 | 0.94 | 0.92 | 0.94 | 0.92 |
| $KGE_{ss}^{75\%}$ | 0.93 | 0.94 | 0.94 | 0.94 | 0.96 | $KGE_{ss}^{75\%}$ | 0.95 | 0.95 | 0.96 | 0.95 | 0.95 | $KGE_{ss}^{75\%}$ | 0.96 | 0.97 | 0.96 | 0.97 | 0.97 | $KGE_{ss}^{75\%}$ | 0.96 | 0.97 | 0.96 | 0.97 | 0.95 |
| $KGE_{ss}^{95\%}$ | 0.98 | 0.98 | 0.99 | 0.98 | 0.99 | $KGE_{ss}^{95\%}$ | 0.98 | 0.99 | 1.00 | 0.99 | 0.98 | $KGE_{ss}^{95\%}$ | 0.99 | 0.99 | 0.99 | 0.99 | 0.99 | $KGE_{ss}^{95\%}$ | 0.98 | 0.99 | 0.98 | 1.00 | 1.00 |
| PNs | 36 | 36 | 36 | 36 | 41 | PNs | 48 | 48 | 48 | 48 | 53 | PNs | 48 | 48 | 48 | 48 | 53 | PNs | 60 | 60 | 60 | 60 | 65 |

**5-Node**

| | $MN_{none}^{DI}$ | $MN_{none}^{DS}$ | $MN_{none}^{DIR}$ | $MN_{none}^{DSR}$ | $MN_{none}^{MLB}$ | | $MN_{SAL}^{DI}$ | $MN_{SAL}^{DS}$ | $MN_{SAL}^{DIR}$ | $MN_{SAL}^{DSR}$ | $MN_{SAL}^{MLB}$ | | $MN_{SAO}^{DI}$ | $MN_{SAO}^{DS}$ | $MN_{SAO}^{DIR}$ | $MN_{SAO}^{DSR}$ | $MN_{SAO}^{MLB}$ | | $MN_{SALO}^{DI}$ | $MN_{SALO}^{DS}$ | $MN_{SALO}^{DIR}$ | $MN_{SALO}^{DSR}$ | $MN_{SALO}^{MLB}$ |
|---|---|---|---|---|---|---|---|---|---|---|---|---|---|---|---|---|---|---|---|---|---|---|---|
| $KGE_{ss}^{min}$ | 0.58 | 0.63 | 0.60 | 0.61 | 0.66 | $KGE_{ss}^{min}$ | 0.62 | 0.62 | 0.77 | 0.61 | 0.65 | $KGE_{ss}^{min}$ | 0.62 | 0.62 | 0.66 | 0.69 | 0.76 | $KGE_{ss}^{min}$ | 0.62 | 0.68 | 0.71 | 0.72 | 0.73 |
| $KGE_{ss}^{5\%}$ | 0.65 | 0.67 | 0.65 | 0.65 | 0.69 | $KGE_{ss}^{5\%}$ | 0.72 | 0.71 | 0.81 | 0.72 | 0.72 | $KGE_{ss}^{5\%}$ | 0.66 | 0.71 | 0.68 | 0.74 | 0.80 | $KGE_{ss}^{5\%}$ | 0.72 | 0.75 | 0.81 | 0.79 | 0.76 |
| $KGE_{ss}^{25\%}$ | 0.79 | 0.82 | 0.80 | 0.79 | 0.81 | $KGE_{ss}^{25\%}$ | 0.83 | 0.85 | 0.90 | 0.81 | 0.86 | $KGE_{ss}^{25\%}$ | 0.83 | 0.89 | 0.84 | 0.87 | 0.91 | $KGE_{ss}^{25\%}$ | 0.87 | 0.89 | 0.88 | 0.90 | 0.89 |
| $KGE_{ss}^{50\%}$ | 0.89 | 0.88 | 0.92 | 0.89 | 0.92 | $KGE_{ss}^{50\%}$ | 0.87 | 0.89 | 0.93 | 0.89 | 0.93 | $KGE_{ss}^{50\%}$ | 0.91 | 0.94 | 0.92 | 0.96 | 0.94 | $KGE_{ss}^{50\%}$ | 0.92 | 0.96 | 0.92 | 0.95 | 0.95 |
| $KGE_{ss}^{75\%}$ | 0.94 | 0.94 | 0.95 | 0.95 | 0.97 | $KGE_{ss}^{75\%}$ | 0.91 | 0.95 | 0.94 | 0.94 | 0.96 | $KGE_{ss}^{75\%}$ | 0.96 | 0.98 | 0.96 | 0.97 | 0.98 | $KGE_{ss}^{75\%}$ | 0.97 | 0.98 | 0.97 | 0.97 | 0.97 |
| $KGE_{ss}^{95\%}$ | 0.99 | 0.98 | 0.99 | 0.99 | 0.99 | $KGE_{ss}^{95\%}$ | 0.97 | 0.97 | 0.99 | 0.99 | 0.99 | $KGE_{ss}^{95\%}$ | 0.99 | 0.99 | 0.99 | 0.99 | 0.99 | $KGE_{ss}^{95\%}$ | 1.00 | 0.99 | 1.00 | 0.99 | 1.00 |
| PNs | 45 | 45 | 45 | 45 | 51 | PNs | 65 | 65 | 65 | 65 | 71 | PNs | 65 | 65 | 65 | 65 | 71 | PNs | 85 | 85 | 85 | 85 | 91 |

*Note that $\alpha^{KGE}$ measures the consistency of flow variability between simulation and observation, with an optimal value of 1 and a range from 0 to $\infty$. The use of the transformed metric $\alpha_*^{KGE}$ $(1-|1-\alpha^{KGE}|)$ maps the range to between $-\infty$ and 1, which facilitates model evaluation since higher values indicate better performance given we are listing various value from percentiles. The cases include single-layer Distributed-Input (DI), Distributed-State (DS), Distributed-Input Relaxed (DIR), Distributed-State Relaxed (DSR), and Machine-Learning Benchmark (MLB) networks, with the number of nodes varying from 1 to 5. “None” refers to the case without cell-information sharing among different nodes in the gates, whereas “SAL,” “SAO,” and “SALO” represent cases where sharing occurs only in the loss gate, only in the output gate, and in both gates, respectively. We report values to two decimal places; therefore, any difference between two metrics that is greater than or equal to 0.01 is noteworthy. PNs=Parameter Numbers

Table S2. $1-|1-\beta^{KGE}|$ ($\beta_*^{KGE}$) scores for the Single-Layer Mass-Conserving Neural Networks (MNs)

**1-Node**

| | $MN_{none}^{DI}$ | $MN_{none}^{DS}$ | $MN_{none}^{DIR}$ | $MN_{none}^{DSR}$ | $MN_{none}^{MLB}$ | | $MN_{SAL}^{DI}$ | $MN_{SAL}^{DS}$ | $MN_{SAL}^{DIR}$ | $MN_{SAL}^{DSR}$ | $MN_{SAL}^{MLB}$ | | $MN_{SAO}^{DI}$ | $MN_{SAO}^{DS}$ | $MN_{SAO}^{DIR}$ | $MN_{SAO}^{DSR}$ | $MN_{SAO}^{MLB}$ | | $MN_{SALO}^{DI}$ | $MN_{SALO}^{DS}$ | $MN_{SALO}^{DIR}$ | $MN_{SALO}^{DSR}$ | $MN_{SALO}^{MLB}$ |
|---|---|---|---|---|---|---|---|---|---|---|---|---|---|---|---|---|---|---|---|---|---|---|---|
| $KGE_{ss}^{min}$ | 0.62 | 0.62 | 0.62 | 0.60 | 0.36 | $KGE_{ss}^{min}$ | | | | | | $KGE_{ss}^{min}$ | | | | | | $KGE_{ss}^{min}$ | | | | | |
| $KGE_{ss}^{5\%}$ | 0.74 | 0.72 | 0.74 | 0.70 | 0.41 | $KGE_{ss}^{5\%}$ | | | | | | $KGE_{ss}^{5\%}$ | | | | | | $KGE_{ss}^{5\%}$ | | | | | |
| $KGE_{ss}^{25\%}$ | 0.85 | 0.84 | 0.85 | 0.84 | 0.64 | $KGE_{ss}^{25\%}$ | | | | | | $KGE_{ss}^{25\%}$ | | | | | | $KGE_{ss}^{25\%}$ | | | | | |
| $KGE_{ss}^{50\%}$ | 0.90 | 0.90 | 0.90 | 0.90 | 0.82 | $KGE_{ss}^{50\%}$ | | | | | | $KGE_{ss}^{50\%}$ | | | | | | $KGE_{ss}^{50\%}$ | | | | | |
| $KGE_{ss}^{75\%}$ | 0.97 | 0.96 | 0.97 | 0.96 | 0.93 | $KGE_{ss}^{75\%}$ | | | | | | $KGE_{ss}^{75\%}$ | | | | | | $KGE_{ss}^{75\%}$ | | | | | |
| $KGE_{ss}^{95\%}$ | 0.99 | 0.99 | 0.99 | 1.00 | 0.98 | $KGE_{ss}^{95\%}$ | | | | | | $KGE_{ss}^{95\%}$ | | | | | | $KGE_{ss}^{95\%}$ | | | | | |
| PNs | 8 | 8 | 9 | 9 | 11 | PNs | | | | | | PNs | | | | | | PNs | | | | | |

**2-Node**

| | $MN_{none}^{DI}$ | $MN_{none}^{DS}$ | $MN_{none}^{DIR}$ | $MN_{none}^{DSR}$ | $MN_{none}^{MLB}$ | | $MN_{SAL}^{DI}$ | $MN_{SAL}^{DS}$ | $MN_{SAL}^{DIR}$ | $MN_{SAL}^{DSR}$ | $MN_{SAL}^{MLB}$ | | $MN_{SAO}^{DI}$ | $MN_{SAO}^{DS}$ | $MN_{SAO}^{DIR}$ | $MN_{SAO}^{DSR}$ | $MN_{SAO}^{MLB}$ | | $MN_{SALO}^{DI}$ | $MN_{SALO}^{DS}$ | $MN_{SALO}^{DIR}$ | $MN_{SALO}^{DSR}$ | $MN_{SALO}^{MLB}$ |
|---|---|---|---|---|---|---|---|---|---|---|---|---|---|---|---|---|---|---|---|---|---|---|---|
| $KGE_{ss}^{min}$ | 0.50 | 0.61 | 0.56 | 0.57 | 0.62 | $KGE_{ss}^{min}$ | 0.44 | 0.63 | 0.57 | 0.50 | 0.63 | $KGE_{ss}^{min}$ | 0.64 | 0.60 | 0.60 | 0.67 | 0.60 | $KGE_{ss}^{min}$ | 0.57 | 0.60 | 0.60 | 0.62 | 0.66 |
| $KGE_{ss}^{5\%}$ | 0.58 | 0.67 | 0.60 | 0.62 | 0.69 | $KGE_{ss}^{5\%}$ | 0.53 | 0.64 | 0.64 | 0.59 | 0.66 | $KGE_{ss}^{5\%}$ | 0.69 | 0.66 | 0.63 | 0.69 | 0.70 | $KGE_{ss}^{5\%}$ | 0.61 | 0.63 | 0.63 | 0.68 | 0.70 |
| $KGE_{ss}^{25\%}$ | 0.81 | 0.83 | 0.84 | 0.81 | 0.82 | $KGE_{ss}^{25\%}$ | 0.81 | 0.82 | 0.85 | 0.80 | 0.81 | $KGE_{ss}^{25\%}$ | 0.84 | 0.83 | 0.83 | 0.84 | 0.82 | $KGE_{ss}^{25\%}$ | 0.83 | 0.82 | 0.83 | 0.83 | 0.82 |
| $KGE_{ss}^{50\%}$ | 0.88 | 0.90 | 0.90 | 0.89 | 0.90 | $KGE_{ss}^{50\%}$ | 0.89 | 0.90 | 0.91 | 0.87 | 0.91 | $KGE_{ss}^{50\%}$ | 0.91 | 0.89 | 0.90 | 0.90 | 0.90 | $KGE_{ss}^{50\%}$ | 0.89 | 0.90 | 0.91 | 0.89 | 0.91 |
| $KGE_{ss}^{75\%}$ | 0.95 | 0.97 | 0.95 | 0.96 | 0.96 | $KGE_{ss}^{75\%}$ | 0.94 | 0.96 | 0.96 | 0.97 | 0.96 | $KGE_{ss}^{75\%}$ | 0.95 | 0.96 | 0.96 | 0.96 | 0.96 | $KGE_{ss}^{75\%}$ | 0.95 | 0.96 | 0.96 | 0.96 | 0.97 |
| $KGE_{ss}^{95\%}$ | 0.99 | 0.99 | 0.99 | 0.99 | 0.99 | $KGE_{ss}^{95\%}$ | 0.98 | 0.99 | 1.00 | 0.99 | 0.99 | $KGE_{ss}^{95\%}$ | 0.99 | 0.97 | 1.00 | 0.99 | 1.00 | $KGE_{ss}^{95\%}$ | 0.99 | 0.98 | 1.00 | 0.99 | 0.99 |
| PNs | 18 | 18 | 18 | 18 | 21 | PNs | 20 | 20 | 20 | 20 | 23 | PNs | 20 | 20 | 20 | 20 | 23 | PNs | 22 | 22 | 22 | 22 | 25 |

**3-Node**

| | $MN_{none}^{DI}$ | $MN_{none}^{DS}$ | $MN_{none}^{DIR}$ | $MN_{none}^{DSR}$ | $MN_{none}^{MLB}$ | | $MN_{SAL}^{DI}$ | $MN_{SAL}^{DS}$ | $MN_{SAL}^{DIR}$ | $MN_{SAL}^{DSR}$ | $MN_{SAL}^{MLB}$ | | $MN_{SAO}^{DI}$ | $MN_{SAO}^{DS}$ | $MN_{SAO}^{DIR}$ | $MN_{SAO}^{DSR}$ | $MN_{SAO}^{MLB}$ | | $MN_{SALO}^{DI}$ | $MN_{SALO}^{DS}$ | $MN_{SALO}^{DIR}$ | $MN_{SALO}^{DSR}$ | $MN_{SALO}^{MLB}$ |
|---|---|---|---|---|---|---|---|---|---|---|---|---|---|---|---|---|---|---|---|---|---|---|---|
| $KGE_{ss}^{min}$ | 0.49 | 0.62 | 0.56 | 0.58 | 0.66 | $KGE_{ss}^{min}$ | 0.43 | 0.46 | 0.53 | 0.46 | 0.62 | $KGE_{ss}^{min}$ | 0.64 | 0.67 | 0.54 | 0.68 | 0.65 | $KGE_{ss}^{min}$ | 0.72 | 0.56 | 0.71 | 0.67 | 0.64 |
| $KGE_{ss}^{5\%}$ | 0.57 | 0.69 | 0.60 | 0.60 | 0.69 | $KGE_{ss}^{5\%}$ | 0.53 | 0.68 | 0.55 | 0.51 | 0.66 | $KGE_{ss}^{5\%}$ | 0.67 | 0.74 | 0.62 | 0.73 | 0.66 | $KGE_{ss}^{5\%}$ | 0.78 | 0.68 | 0.74 | 0.71 | 0.76 |
| $KGE_{ss}^{25\%}$ | 0.81 | 0.83 | 0.84 | 0.81 | 0.82 | $KGE_{ss}^{25\%}$ | 0.81 | 0.82 | 0.79 | 0.81 | 0.81 | $KGE_{ss}^{25\%}$ | 0.81 | 0.85 | 0.84 | 0.83 | 0.80 | $KGE_{ss}^{25\%}$ | 0.85 | 0.83 | 0.85 | 0.83 | 0.84 |
| $KGE_{ss}^{50\%}$ | 0.88 | 0.90 | 0.90 | 0.89 | 0.91 | $KGE_{ss}^{50\%}$ | 0.89 | 0.92 | 0.90 | 0.88 | 0.89 | $KGE_{ss}^{50\%}$ | 0.89 | 0.91 | 0.91 | 0.93 | 0.90 | $KGE_{ss}^{50\%}$ | 0.92 | 0.90 | 0.91 | 0.89 | 0.91 |
| $KGE_{ss}^{75\%}$ | 0.95 | 0.97 | 0.95 | 0.96 | 0.96 | $KGE_{ss}^{75\%}$ | 0.94 | 0.95 | 0.96 | 0.96 | 0.96 | $KGE_{ss}^{75\%}$ | 0.97 | 0.96 | 0.95 | 0.98 | 0.94 | $KGE_{ss}^{75\%}$ | 0.96 | 0.95 | 0.96 | 0.97 | 0.95 |
| $KGE_{ss}^{95\%}$ | 0.99 | 0.99 | 0.99 | 0.99 | 0.99 | $KGE_{ss}^{95\%}$ | 0.98 | 0.99 | 0.99 | 1.00 | 1.00 | $KGE_{ss}^{95\%}$ | 0.99 | 0.99 | 1.00 | 1.00 | 0.98 | $KGE_{ss}^{95\%}$ | 1.00 | 0.99 | 0.99 | 0.99 | 0.99 |
| PNs | 27 | 27 | 27 | 27 | 31 | PNs | 33 | 33 | 33 | 33 | 37 | PNs | 33 | 33 | 33 | 33 | 37 | PNs | 39 | 39 | 39 | 39 | 43 |

**4-Node**

| | $MN_{none}^{DI}$ | $MN_{none}^{DS}$ | $MN_{none}^{DIR}$ | $MN_{none}^{DSR}$ | $MN_{none}^{MLB}$ | | $MN_{SAL}^{DI}$ | $MN_{SAL}^{DS}$ | $MN_{SAL}^{DIR}$ | $MN_{SAL}^{DSR}$ | $MN_{SAL}^{MLB}$ | | $MN_{SAO}^{DI}$ | $MN_{SAO}^{DS}$ | $MN_{SAO}^{DIR}$ | $MN_{SAO}^{DSR}$ | $MN_{SAO}^{MLB}$ | | $MN_{SALO}^{DI}$ | $MN_{SALO}^{DS}$ | $MN_{SALO}^{DIR}$ | $MN_{SALO}^{DSR}$ | $MN_{SALO}^{MLB}$ |
|---|---|---|---|---|---|---|---|---|---|---|---|---|---|---|---|---|---|---|---|---|---|---|---|
| $KGE_{ss}^{min}$ | 0.57 | 0.62 | 0.56 | 0.57 | 0.63 | $KGE_{ss}^{min}$ | 0.50 | 0.69 | 0.57 | 0.43 | 0.61 | $KGE_{ss}^{min}$ | 0.63 | 0.71 | 0.57 | 0.69 | 0.76 | $KGE_{ss}^{min}$ | 0.67 | 0.50 | 0.67 | 0.72 | 0.65 |
| $KGE_{ss}^{5\%}$ | 0.63 | 0.67 | 0.61 | 0.61 | 0.69 | $KGE_{ss}^{5\%}$ | 0.53 | 0.72 | 0.59 | 0.62 | 0.71 | $KGE_{ss}^{5\%}$ | 0.65 | 0.76 | 0.66 | 0.71 | 0.79 | $KGE_{ss}^{5\%}$ | 0.71 | 0.71 | 0.71 | 0.80 | 0.67 |
| $KGE_{ss}^{25\%}$ | 0.84 | 0.83 | 0.84 | 0.80 | 0.83 | $KGE_{ss}^{25\%}$ | 0.83 | 0.84 | 0.80 | 0.83 | 0.85 | $KGE_{ss}^{25\%}$ | 0.83 | 0.86 | 0.84 | 0.87 | 0.88 | $KGE_{ss}^{25\%}$ | 0.85 | 0.85 | 0.85 | 0.88 | 0.85 |
| $KGE_{ss}^{50\%}$ | 0.89 | 0.90 | 0.90 | 0.89 | 0.90 | $KGE_{ss}^{50\%}$ | 0.89 | 0.92 | 0.90 | 0.90 | 0.91 | $KGE_{ss}^{50\%}$ | 0.90 | 0.93 | 0.90 | 0.93 | 0.92 | $KGE_{ss}^{50\%}$ | 0.92 | 0.90 | 0.92 | 0.93 | 0.91 |
| $KGE_{ss}^{75\%}$ | 0.94 | 0.97 | 0.95 | 0.96 | 0.96 | $KGE_{ss}^{75\%}$ | 0.95 | 0.94 | 0.95 | 0.96 | 0.98 | $KGE_{ss}^{75\%}$ | 0.94 | 0.97 | 0.95 | 0.97 | 0.97 | $KGE_{ss}^{75\%}$ | 0.95 | 0.96 | 0.95 | 0.97 | 0.96 |
| $KGE_{ss}^{95\%}$ | 0.99 | 0.99 | 0.99 | 0.99 | 1.00 | $KGE_{ss}^{95\%}$ | 0.98 | 0.98 | 1.00 | 0.99 | 1.00 | $KGE_{ss}^{95\%}$ | 0.99 | 0.99 | 0.98 | 0.99 | 0.99 | $KGE_{ss}^{95\%}$ | 0.98 | 0.99 | 0.98 | 0.99 | 1.00 |
| PNs | 36 | 36 | 36 | 36 | 41 | PNs | 48 | 48 | 48 | 48 | 53 | PNs | 48 | 48 | 48 | 48 | 53 | PNs | 60 | 60 | 60 | 60 | 65 |

**5-Node**

| | $MN_{none}^{DI}$ | $MN_{none}^{DS}$ | $MN_{none}^{DIR}$ | $MN_{none}^{DSR}$ | $MN_{none}^{MLB}$ | | $MN_{SAL}^{DI}$ | $MN_{SAL}^{DS}$ | $MN_{SAL}^{DIR}$ | $MN_{SAL}^{DSR}$ | $MN_{SAL}^{MLB}$ | | $MN_{SAO}^{DI}$ | $MN_{SAO}^{DS}$ | $MN_{SAO}^{DIR}$ | $MN_{SAO}^{DSR}$ | $MN_{SAO}^{MLB}$ | | $MN_{SALO}^{DI}$ | $MN_{SALO}^{DS}$ | $MN_{SALO}^{DIR}$ | $MN_{SALO}^{DSR}$ | $MN_{SALO}^{MLB}$ |
|---|---|---|---|---|---|---|---|---|---|---|---|---|---|---|---|---|---|---|---|---|---|---|---|
| $KGE_{ss}^{min}$ | 0.49 | 0.62 | 0.55 | 0.57 | 0.59 | $KGE_{ss}^{min}$ | 0.54 | 0.64 | 0.78 | 0.54 | 0.68 | $KGE_{ss}^{min}$ | 0.64 | 0.65 | 0.54 | 0.73 | 0.75 | $KGE_{ss}^{min}$ | 0.72 | 0.80 | 0.73 | 0.79 | 0.69 |
| $KGE_{ss}^{5\%}$ | 0.57 | 0.66 | 0.58 | 0.61 | 0.71 | $KGE_{ss}^{5\%}$ | 0.60 | 0.73 | 0.79 | 0.63 | 0.70 | $KGE_{ss}^{5\%}$ | 0.67 | 0.69 | 0.67 | 0.78 | 0.77 | $KGE_{ss}^{5\%}$ | 0.74 | 0.82 | 0.81 | 0.81 | 0.81 |
| $KGE_{ss}^{25\%}$ | 0.81 | 0.83 | 0.82 | 0.80 | 0.85 | $KGE_{ss}^{25\%}$ | 0.82 | 0.82 | 0.87 | 0.82 | 0.82 | $KGE_{ss}^{25\%}$ | 0.83 | 0.84 | 0.84 | 0.87 | 0.87 | $KGE_{ss}^{25\%}$ | 0.86 | 0.88 | 0.88 | 0.89 | 0.87 |
| $KGE_{ss}^{50\%}$ | 0.89 | 0.90 | 0.91 | 0.89 | 0.92 | $KGE_{ss}^{50\%}$ | 0.89 | 0.90 | 0.94 | 0.87 | 0.91 | $KGE_{ss}^{50\%}$ | 0.91 | 0.91 | 0.90 | 0.93 | 0.92 | $KGE_{ss}^{50\%}$ | 0.92 | 0.93 | 0.93 | 0.94 | 0.91 |
| $KGE_{ss}^{75\%}$ | 0.95 | 0.97 | 0.96 | 0.96 | 0.95 | $KGE_{ss}^{75\%}$ | 0.94 | 0.95 | 0.97 | 0.95 | 0.96 | $KGE_{ss}^{75\%}$ | 0.95 | 0.96 | 0.94 | 0.97 | 0.98 | $KGE_{ss}^{75\%}$ | 0.96 | 0.96 | 0.96 | 0.98 | 0.97 |
| $KGE_{ss}^{95\%}$ | 0.98 | 0.99 | 0.99 | 0.99 | 1.00 | $KGE_{ss}^{95\%}$ | 0.98 | 0.98 | 1.00 | 0.99 | 0.99 | $KGE_{ss}^{95\%}$ | 0.99 | 1.00 | 0.98 | 0.99 | 1.00 | $KGE_{ss}^{95\%}$ | 0.99 | 0.99 | 1.00 | 1.00 | 0.99 |
| PNs | 45 | 45 | 45 | 45 | 51 | PNs | 65 | 65 | 65 | 65 | 71 | PNs | 65 | 65 | 65 | 65 | 71 | PNs | 85 | 85 | 85 | 85 | 91 |

*Note that $\beta^{KGE}$ measures the consistency of mass balance between simulation and observation, with an optimal value of 1 and a range from 0 to ∞. The use of the transformed metric $\beta_*^{KGE}$ ($1-|1-\beta^{KGE}|$) maps the range to between $-\infty$ and 1, which facilitates model evaluation since higher values indicate better performance given we are listing various value from percentiles. The cases include single-layer Distributed-Input (DI), Distributed-State (DS), Distributed-Input Relaxed (DIR), Distributed-State Relaxed (DSR), and Machine-Learning Benchmark (MLB) networks, with the number of nodes varying from 1 to 5. "None" refers to the case without cell-information sharing among different nodes in the gates, whereas "SAL," "SAO," and "SALO" represent cases where sharing occurs only in the loss gate, only in the output gate, and in both gates, respectively. We report values to two decimal places; therefore, any difference between two metrics that is greater than or equal to 0.01 is noteworthy. PNs=Parameter Numbers

Table S3. $\gamma^{KGE}$ scores for the Single-Layer Mass-Conserving Neural Networks (MNs)

**1-Node**

| | $MN_{none}^{DI}$ | $MN_{none}^{DS}$ | $MN_{none}^{DIR}$ | $MN_{none}^{DSR}$ | $MN_{none}^{MLB}$ | | $MN_{SAL}^{DI}$ | $MN_{SAL}^{DS}$ | $MN_{SAL}^{DIR}$ | $MN_{SAL}^{DSR}$ | $MN_{SAL}^{MLB}$ | | $MN_{SAO}^{DI}$ | $MN_{SAO}^{DS}$ | $MN_{SAO}^{DIR}$ | $MN_{SAO}^{DSR}$ | $MN_{SAO}^{MLB}$ | | $MN_{SALO}^{DI}$ | $MN_{SALO}^{DS}$ | $MN_{SALO}^{DIR}$ | $MN_{SALO}^{DSR}$ | $MN_{SALO}^{MLB}$ |
|---|---|---|---|---|---|---|---|---|---|---|---|---|---|---|---|---|---|---|---|---|---|---|---|
| $KGE_{ss}^{min}$ | 0.71 | 0.70 | 0.71 | 0.69 | 0.56 | $KGE_{ss}^{min}$ | | | | | | $KGE_{ss}^{min}$ | | | | | | $KGE_{ss}^{min}$ | | | | | |
| $KGE_{ss}^{5\%}$ | 0.72 | 0.71 | 0.72 | 0.71 | 0.69 | $KGE_{ss}^{5\%}$ | | | | | | $KGE_{ss}^{5\%}$ | | | | | | $KGE_{ss}^{5\%}$ | | | | | |
| $KGE_{ss}^{25\%}$ | 0.81 | 0.81 | 0.81 | 0.81 | 0.78 | $KGE_{ss}^{25\%}$ | | | | | | $KGE_{ss}^{25\%}$ | | | | | | $KGE_{ss}^{25\%}$ | | | | | |
| $KGE_{ss}^{50\%}$ | 0.85 | 0.85 | 0.85 | 0.85 | 0.83 | $KGE_{ss}^{50\%}$ | | | | | | $KGE_{ss}^{50\%}$ | | | | | | $KGE_{ss}^{50\%}$ | | | | | |
| $KGE_{ss}^{75\%}$ | 0.90 | 0.90 | 0.90 | 0.90 | 0.90 | $KGE_{ss}^{75\%}$ | | | | | | $KGE_{ss}^{75\%}$ | | | | | | $KGE_{ss}^{75\%}$ | | | | | |
| $KGE_{ss}^{95\%}$ | 0.93 | 0.92 | 0.93 | 0.92 | 0.92 | $KGE_{ss}^{95\%}$ | | | | | | $KGE_{ss}^{95\%}$ | | | | | | $KGE_{ss}^{95\%}$ | | | | | |
| PNs | 8 | 8 | 9 | 9 | 11 | PNs | | | | | | PNs | | | | | | PNs | | | | | |

**2-Node**

| | $MN_{none}^{DI}$ | $MN_{none}^{DS}$ | $MN_{none}^{DIR}$ | $MN_{none}^{DSR}$ | $MN_{none}^{MLB}$ | | $MN_{SAL}^{DI}$ | $MN_{SAL}^{DS}$ | $MN_{SAL}^{DIR}$ | $MN_{SAL}^{DSR}$ | $MN_{SAL}^{MLB}$ | | $MN_{SAO}^{DI}$ | $MN_{SAO}^{DS}$ | $MN_{SAO}^{DIR}$ | $MN_{SAO}^{DSR}$ | $MN_{SAO}^{MLB}$ | | $MN_{SALO}^{DI}$ | $MN_{SALO}^{DS}$ | $MN_{SALO}^{DIR}$ | $MN_{SALO}^{DSR}$ | $MN_{SALO}^{MLB}$ |
|---|---|---|---|---|---|---|---|---|---|---|---|---|---|---|---|---|---|---|---|---|---|---|---|
| $KGE_{ss}^{min}$ | 0.75 | 0.80 | 0.77 | 0.78 | 0.80 | $KGE_{ss}^{min}$ | 0.75 | 0.79 | 0.80 | 0.79 | 0.82 | $KGE_{ss}^{min}$ | 0.79 | 0.73 | 0.76 | 0.75 | 0.81 | $KGE_{ss}^{min}$ | 0.70 | 0.75 | 0.76 | 0.77 | 0.82 |
| $KGE_{ss}^{5\%}$ | 0.81 | 0.80 | 0.82 | 0.80 | 0.82 | $KGE_{ss}^{5\%}$ | 0.79 | 0.80 | 0.81 | 0.81 | 0.83 | $KGE_{ss}^{5\%}$ | 0.83 | 0.75 | 0.82 | 0.79 | 0.83 | $KGE_{ss}^{5\%}$ | 0.79 | 0.76 | 0.81 | 0.79 | 0.84 |
| $KGE_{ss}^{25\%}$ | 0.84 | 0.84 | 0.84 | 0.84 | 0.85 | $KGE_{ss}^{25\%}$ | 0.85 | 0.84 | 0.85 | 0.85 | 0.86 | $KGE_{ss}^{25\%}$ | 0.86 | 0.82 | 0.84 | 0.83 | 0.85 | $KGE_{ss}^{25\%}$ | 0.86 | 0.83 | 0.85 | 0.84 | 0.87 |
| $KGE_{ss}^{50\%}$ | 0.87 | 0.87 | 0.87 | 0.87 | 0.88 | $KGE_{ss}^{50\%}$ | 0.88 | 0.87 | 0.87 | 0.87 | 0.88 | $KGE_{ss}^{50\%}$ | 0.88 | 0.87 | 0.88 | 0.89 | 0.88 | $KGE_{ss}^{50\%}$ | 0.89 | 0.88 | 0.88 | 0.90 | 0.89 |
| $KGE_{ss}^{75\%}$ | 0.90 | 0.90 | 0.91 | 0.91 | 0.91 | $KGE_{ss}^{75\%}$ | 0.90 | 0.91 | 0.91 | 0.91 | 0.91 | $KGE_{ss}^{75\%}$ | 0.92 | 0.93 | 0.91 | 0.93 | 0.91 | $KGE_{ss}^{75\%}$ | 0.92 | 0.93 | 0.92 | 0.93 | 0.92 |
| $KGE_{ss}^{95\%}$ | 0.93 | 0.93 | 0.93 | 0.93 | 0.93 | $KGE_{ss}^{95\%}$ | 0.93 | 0.93 | 0.93 | 0.93 | 0.93 | $KGE_{ss}^{95\%}$ | 0.94 | 0.96 | 0.94 | 0.96 | 0.94 | $KGE_{ss}^{95\%}$ | 0.94 | 0.96 | 0.94 | 0.96 | 0.94 |
| PNs | 18 | 18 | 18 | 18 | 21 | PNs | 20 | 20 | 20 | 20 | 23 | PNs | 20 | 20 | 20 | 20 | 23 | PNs | 22 | 22 | 22 | 22 | 25 |

**3-Node**

| | $MN_{none}^{DI}$ | $MN_{none}^{DS}$ | $MN_{none}^{DIR}$ | $MN_{none}^{DSR}$ | $MN_{none}^{MLB}$ | | $MN_{SAL}^{DI}$ | $MN_{SAL}^{DS}$ | $MN_{SAL}^{DIR}$ | $MN_{SAL}^{DSR}$ | $MN_{SAL}^{MLB}$ | | $MN_{SAO}^{DI}$ | $MN_{SAO}^{DS}$ | $MN_{SAO}^{DIR}$ | $MN_{SAO}^{DSR}$ | $MN_{SAO}^{MLB}$ | | $MN_{SALO}^{DI}$ | $MN_{SALO}^{DS}$ | $MN_{SALO}^{DIR}$ | $MN_{SALO}^{DSR}$ | $MN_{SALO}^{MLB}$ |
|---|---|---|---|---|---|---|---|---|---|---|---|---|---|---|---|---|---|---|---|---|---|---|---|
| $KGE_{ss}^{min}$ | 0.75 | 0.78 | 0.77 | 0.79 | 0.76 | $KGE_{ss}^{min}$ | 0.75 | 0.79 | 0.77 | 0.78 | 0.76 | $KGE_{ss}^{min}$ | 0.73 | 0.82 | 0.81 | 0.82 | 0.73 | $KGE_{ss}^{min}$ | 0.77 | 0.78 | 0.80 | 0.78 | 0.78 |
| $KGE_{ss}^{5\%}$ | 0.82 | 0.81 | 0.82 | 0.81 | 0.84 | $KGE_{ss}^{5\%}$ | 0.79 | 0.81 | 0.82 | 0.81 | 0.84 | $KGE_{ss}^{5\%}$ | 0.78 | 0.85 | 0.83 | 0.83 | 0.84 | $KGE_{ss}^{5\%}$ | 0.82 | 0.80 | 0.83 | 0.82 | 0.84 |
| $KGE_{ss}^{25\%}$ | 0.84 | 0.84 | 0.85 | 0.85 | 0.86 | $KGE_{ss}^{25\%}$ | 0.85 | 0.85 | 0.84 | 0.85 | 0.86 | $KGE_{ss}^{25\%}$ | 0.85 | 0.89 | 0.86 | 0.88 | 0.88 | $KGE_{ss}^{25\%}$ | 0.88 | 0.87 | 0.87 | 0.87 | 0.88 |
| $KGE_{ss}^{50\%}$ | 0.87 | 0.87 | 0.87 | 0.88 | 0.89 | $KGE_{ss}^{50\%}$ | 0.88 | 0.88 | 0.87 | 0.88 | 0.89 | $KGE_{ss}^{50\%}$ | 0.89 | 0.94 | 0.89 | 0.93 | 0.91 | $KGE_{ss}^{50\%}$ | 0.90 | 0.92 | 0.90 | 0.90 | 0.93 |
| $KGE_{ss}^{75\%}$ | 0.90 | 0.91 | 0.91 | 0.91 | 0.91 | $KGE_{ss}^{75\%}$ | 0.90 | 0.91 | 0.91 | 0.91 | 0.92 | $KGE_{ss}^{75\%}$ | 0.93 | 0.96 | 0.92 | 0.95 | 0.94 | $KGE_{ss}^{75\%}$ | 0.93 | 0.94 | 0.93 | 0.93 | 0.95 |
| $KGE_{ss}^{95\%}$ | 0.93 | 0.93 | 0.93 | 0.93 | 0.93 | $KGE_{ss}^{95\%}$ | 0.93 | 0.93 | 0.93 | 0.93 | 0.93 | $KGE_{ss}^{95\%}$ | 0.95 | 0.97 | 0.94 | 0.98 | 0.97 | $KGE_{ss}^{95\%}$ | 0.95 | 0.96 | 0.95 | 0.97 | 0.98 |
| PNs | 27 | 27 | 27 | 27 | 31 | PNs | 33 | 33 | 33 | 33 | 37 | PNs | 33 | 33 | 33 | 33 | 37 | PNs | 39 | 39 | 39 | 39 | 43 |

**4-Node**

| | $MN_{none}^{DI}$ | $MN_{none}^{DS}$ | $MN_{none}^{DIR}$ | $MN_{none}^{DSR}$ | $MN_{none}^{MLB}$ | | $MN_{SAL}^{DI}$ | $MN_{SAL}^{DS}$ | $MN_{SAL}^{DIR}$ | $MN_{SAL}^{DSR}$ | $MN_{SAL}^{MLB}$ | | $MN_{SAO}^{DI}$ | $MN_{SAO}^{DS}$ | $MN_{SAO}^{DIR}$ | $MN_{SAO}^{DSR}$ | $MN_{SAO}^{MLB}$ | | $MN_{SALO}^{DI}$ | $MN_{SALO}^{DS}$ | $MN_{SALO}^{DIR}$ | $MN_{SALO}^{DSR}$ | $MN_{SALO}^{MLB}$ |
|---|---|---|---|---|---|---|---|---|---|---|---|---|---|---|---|---|---|---|---|---|---|---|---|
| $KGE_{ss}^{min}$ | 0.64 | 0.80 | 0.77 | 0.79 | 0.78 | $KGE_{ss}^{min}$ | 0.75 | 0.77 | 0.78 | 0.76 | 0.82 | $KGE_{ss}^{min}$ | 0.72 | 0.86 | 0.79 | 0.82 | 0.82 | $KGE_{ss}^{min}$ | 0.78 | 0.76 | 0.78 | 0.85 | 0.82 |
| $KGE_{ss}^{5\%}$ | 0.70 | 0.81 | 0.82 | 0.81 | 0.84 | $KGE_{ss}^{5\%}$ | 0.79 | 0.79 | 0.82 | 0.80 | 0.84 | $KGE_{ss}^{5\%}$ | 0.77 | 0.87 | 0.82 | 0.86 | 0.86 | $KGE_{ss}^{5\%}$ | 0.81 | 0.79 | 0.81 | 0.88 | 0.86 |
| $KGE_{ss}^{25\%}$ | 0.82 | 0.85 | 0.85 | 0.85 | 0.86 | $KGE_{ss}^{25\%}$ | 0.84 | 0.86 | 0.85 | 0.85 | 0.87 | $KGE_{ss}^{25\%}$ | 0.85 | 0.90 | 0.87 | 0.90 | 0.91 | $KGE_{ss}^{25\%}$ | 0.87 | 0.89 | 0.87 | 0.92 | 0.89 |
| $KGE_{ss}^{50\%}$ | 0.86 | 0.88 | 0.87 | 0.88 | 0.89 | $KGE_{ss}^{50\%}$ | 0.88 | 0.89 | 0.88 | 0.88 | 0.89 | $KGE_{ss}^{50\%}$ | 0.90 | 0.95 | 0.90 | 0.94 | 0.94 | $KGE_{ss}^{50\%}$ | 0.90 | 0.94 | 0.90 | 0.95 | 0.92 |
| $KGE_{ss}^{75\%}$ | 0.90 | 0.91 | 0.91 | 0.91 | 0.92 | $KGE_{ss}^{75\%}$ | 0.90 | 0.91 | 0.91 | 0.91 | 0.92 | $KGE_{ss}^{75\%}$ | 0.93 | 0.96 | 0.93 | 0.96 | 0.96 | $KGE_{ss}^{75\%}$ | 0.92 | 0.95 | 0.92 | 0.96 | 0.94 |
| $KGE_{ss}^{95\%}$ | 0.93 | 0.93 | 0.93 | 0.93 | 0.93 | $KGE_{ss}^{95\%}$ | 0.93 | 0.93 | 0.93 | 0.93 | 0.94 | $KGE_{ss}^{95\%}$ | 0.96 | 0.98 | 0.95 | 0.98 | 0.98 | $KGE_{ss}^{95\%}$ | 0.94 | 0.98 | 0.94 | 0.98 | 0.96 |
| PNs | 36 | 36 | 36 | 36 | 41 | PNs | 48 | 48 | 48 | 48 | 53 | PNs | 48 | 48 | 48 | 48 | 53 | PNs | 60 | 60 | 60 | 60 | 65 |

**5-Node**

| | $MN_{none}^{DI}$ | $MN_{none}^{DS}$ | $MN_{none}^{DIR}$ | $MN_{none}^{DSR}$ | $MN_{none}^{MLB}$ | | $MN_{SAL}^{DI}$ | $MN_{SAL}^{DS}$ | $MN_{SAL}^{DIR}$ | $MN_{SAL}^{DSR}$ | $MN_{SAL}^{MLB}$ | | $MN_{SAO}^{DI}$ | $MN_{SAO}^{DS}$ | $MN_{SAO}^{DIR}$ | $MN_{SAO}^{DSR}$ | $MN_{SAO}^{MLB}$ | | $MN_{SALO}^{DI}$ | $MN_{SALO}^{DS}$ | $MN_{SALO}^{DIR}$ | $MN_{SALO}^{DSR}$ | $MN_{SALO}^{MLB}$ |
|---|---|---|---|---|---|---|---|---|---|---|---|---|---|---|---|---|---|---|---|---|---|---|---|
| $KGE_{ss}^{min}$ | 0.75 | 0.80 | 0.80 | 0.79 | 0.78 | $KGE_{ss}^{min}$ | 0.77 | 0.76 | 0.79 | 0.79 | 0.72 | $KGE_{ss}^{min}$ | 0.71 | 0.82 | 0.80 | 0.85 | 0.80 | $KGE_{ss}^{min}$ | 0.72 | 0.86 | 0.80 | 0.88 | 0.83 |
| $KGE_{ss}^{5\%}$ | 0.81 | 0.81 | 0.83 | 0.81 | 0.83 | $KGE_{ss}^{5\%}$ | 0.81 | 0.81 | 0.83 | 0.81 | 0.83 | $KGE_{ss}^{5\%}$ | 0.77 | 0.85 | 0.83 | 0.86 | 0.86 | $KGE_{ss}^{5\%}$ | 0.81 | 0.88 | 0.85 | 0.90 | 0.85 |
| $KGE_{ss}^{25\%}$ | 0.84 | 0.85 | 0.85 | 0.85 | 0.86 | $KGE_{ss}^{25\%}$ | 0.84 | 0.86 | 0.86 | 0.85 | 0.86 | $KGE_{ss}^{25\%}$ | 0.86 | 0.91 | 0.88 | 0.92 | 0.89 | $KGE_{ss}^{25\%}$ | 0.86 | 0.92 | 0.89 | 0.92 | 0.91 |
| $KGE_{ss}^{50\%}$ | 0.87 | 0.88 | 0.88 | 0.88 | 0.89 | $KGE_{ss}^{50\%}$ | 0.88 | 0.88 | 0.89 | 0.88 | 0.88 | $KGE_{ss}^{50\%}$ | 0.89 | 0.95 | 0.90 | 0.95 | 0.94 | $KGE_{ss}^{50\%}$ | 0.90 | 0.95 | 0.90 | 0.95 | 0.94 |
| $KGE_{ss}^{75\%}$ | 0.90 | 0.91 | 0.91 | 0.91 | 0.92 | $KGE_{ss}^{75\%}$ | 0.90 | 0.92 | 0.91 | 0.91 | 0.91 | $KGE_{ss}^{75\%}$ | 0.93 | 0.96 | 0.93 | 0.96 | 0.96 | $KGE_{ss}^{75\%}$ | 0.93 | 0.97 | 0.93 | 0.96 | 0.95 |
| $KGE_{ss}^{95\%}$ | 0.93 | 0.93 | 0.93 | 0.93 | 0.94 | $KGE_{ss}^{95\%}$ | 0.93 | 0.93 | 0.93 | 0.93 | 0.94 | $KGE_{ss}^{95\%}$ | 0.96 | 0.98 | 0.96 | 0.98 | 0.98 | $KGE_{ss}^{95\%}$ | 0.97 | 0.98 | 0.95 | 0.98 | 0.97 |
| PNs | 45 | 45 | 45 | 45 | 51 | PNs | 65 | 65 | 65 | 65 | 71 | PNs | 65 | 65 | 65 | 65 | 71 | PNs | 85 | 85 | 85 | 85 | 91 |

*Note also that $r^{KGE}$ measures the linear correlation between simulation and observation, with a positive 1 for perfect positive linear correlation, a zero (0) for no linear correlation, a negative 1 for negative linear correlation, and a range from -1 to 1. The cases include single-layer Distributed-Input (DI), Distributed-State (DS), Distributed-Input Relaxed (DIR), Distributed-State Relaxed (DSR), and Machine-Learning Benchmark (MLB) networks, with the number of nodes varying from 1 to 5. The cases include single-layer Distributed-Input (DI), Distributed-State (DS), Distributed-Input Relaxed (DIR), Distributed-State Relaxed (DSR), and Machine-Learning Benchmark (MLB) networks, with the number of nodes

Table S4. $KGE_{ss}$ scores of flow magnitude for the Single-Layer Mass-Conserving Neural Networks (MNs)

**1-Node**

| | $MN_{none}^{DI}$ | $MN_{none}^{DS}$ | $MN_{none}^{DIR}$ | $MN_{none}^{DSR}$ | $MN_{none}^{MLB}$ | $MN_{SAL}^{DI}$ | $MN_{SAL}^{DS}$ | $MN_{SAL}^{DIR}$ | $MN_{SAL}^{DSR}$ | $MN_{SAL}^{MLB}$ | $MN_{SAO}^{DI}$ | $MN_{SAO}^{DS}$ | $MN_{SAO}^{DIR}$ | $MN_{SAO}^{DSR}$ | $MN_{SAO}^{MLB}$ | $MN_{SALO}^{DI}$ | $MN_{SALO}^{DS}$ | $MN_{SALO}^{DIR}$ | $MN_{SALO}^{DSR}$ | $MN_{SALO}^{MLB}$ |
|---|---|---|---|---|---|---|---|---|---|---|---|---|---|---|---|---|---|---|---|---|
| $1_{group}^{st}$ | 0.20 | 0.20 | 0.19 | 0.20 | 0.06 | | | | | | | | | | | | | | | |
| $2_{group}^{nd}$ | -1.04 | -1.04 | -0.99 | -0.93 | -0.18 | | | | | | | | | | | | | | | |
| $3_{group}^{rd}$ | -1.58 | -1.58 | -1.58 | -1.52 | -0.79 | | | | | | | | | | | | | | | |
| $4_{group}^{th}$ | -0.63 | -0.63 | -0.66 | -0.61 | -0.41 | | | | | | | | | | | | | | | |
| $5_{group}^{th}$ | 0.87 | 0.87 | 0.87 | 0.86 | 0.83 | | | | | | | | | | | | | | | |

**2-Node**

| | $MN_{none}^{DI}$ | $MN_{none}^{DS}$ | $MN_{none}^{DIR}$ | $MN_{none}^{DSR}$ | $MN_{none}^{MLB}$ | $MN_{SAL}^{DI}$ | $MN_{SAL}^{DS}$ | $MN_{SAL}^{DIR}$ | $MN_{SAL}^{DSR}$ | $MN_{SAL}^{MLB}$ | $MN_{SAO}^{DI}$ | $MN_{SAO}^{DS}$ | $MN_{SAO}^{DIR}$ | $MN_{SAO}^{DSR}$ | $MN_{SAO}^{MLB}$ | $MN_{SALO}^{DI}$ | $MN_{SALO}^{DS}$ | $MN_{SALO}^{DIR}$ | $MN_{SALO}^{DSR}$ | $MN_{SALO}^{MLB}$ |
|---|---|---|---|---|---|---|---|---|---|---|---|---|---|---|---|---|---|---|---|---|
| $1_{group}^{st}$ | -0.07 | 0.02 | -0.13 | 0.17 | -0.48 | -0.04 | 0.16 | 0.09 | 0.18 | -0.13 | 0.09 | 0.11 | 0.19 | 0.00 | 0.03 | -0.06 | 0.17 | 0.18 | 0.01 | -0.11 |
| $2_{group}^{nd}$ | -1.55 | -1.46 | -1.42 | -1.11 | -2.44 | -1.38 | -1.13 | -0.80 | -1.05 | -1.99 | -1.24 | -1.40 | -0.94 | -1.79 | -1.84 | -1.39 | -1.28 | -0.85 | -1.62 | -2.06 |
| $3_{group}^{rd}$ | -1.22 | -1.08 | -0.96 | -1.08 | -1.40 | -1.14 | -1.09 | -0.66 | -1.04 | -1.26 | -1.09 | -1.38 | -1.01 | -1.47 | -1.29 | -1.16 | -1.37 | -0.99 | -1.44 | -1.25 |
| $4_{group}^{th}$ | -0.36 | -0.35 | -0.32 | -0.43 | -0.38 | -0.34 | -0.40 | -0.24 | -0.42 | -0.32 | -0.28 | -0.55 | -0.32 | -0.50 | -0.33 | -0.27 | -0.53 | -0.30 | -0.48 | -0.22 |
| $5_{group}^{th}$ | 0.87 | 0.88 | 0.88 | 0.88 | 0.88 | 0.87 | 0.88 | 0.88 | 0.88 | 0.88 | 0.89 | 0.90 | 0.89 | 0.91 | 0.89 | 0.89 | 0.91 | 0.89 | 0.91 | 0.89 |

**3-Node**

| | $MN_{none}^{DI}$ | $MN_{none}^{DS}$ | $MN_{none}^{DIR}$ | $MN_{none}^{DSR}$ | $MN_{none}^{MLB}$ | $MN_{SAL}^{DI}$ | $MN_{SAL}^{DS}$ | $MN_{SAL}^{DIR}$ | $MN_{SAL}^{DSR}$ | $MN_{SAL}^{MLB}$ | $MN_{SAO}^{DI}$ | $MN_{SAO}^{DS}$ | $MN_{SAO}^{DIR}$ | $MN_{SAO}^{DSR}$ | $MN_{SAO}^{MLB}$ | $MN_{SALO}^{DI}$ | $MN_{SALO}^{DS}$ | $MN_{SALO}^{DIR}$ | $MN_{SALO}^{DSR}$ | $MN_{SALO}^{MLB}$ |
|---|---|---|---|---|---|---|---|---|---|---|---|---|---|---|---|---|---|---|---|---|
| $1_{group}^{st}$ | -0.07 | 0.13 | -0.12 | 0.18 | -0.94 | -0.03 | 0.15 | 0.09 | 0.22 | -0.45 | -1.14 | -0.20 | 0.00 | -0.65 | -1.78 | -0.42 | -0.12 | 0.13 | -0.12 | -0.85 |
| $2_{group}^{nd}$ | -1.56 | -1.33 | -1.41 | -0.95 | -2.55 | -1.36 | -1.06 | -0.99 | -1.00 | -2.26 | -2.29 | -1.21 | -1.44 | -1.51 | -2.50 | -1.31 | -1.18 | -1.02 | -1.51 | -1.98 |
| $3_{group}^{rd}$ | -1.22 | -1.14 | -0.94 | -0.98 | -1.30 | -1.13 | -1.02 | -0.83 | -1.00 | -1.25 | -1.08 | -0.64 | -1.07 | -0.58 | -0.88 | -0.66 | -0.74 | -0.85 | -1.11 | -1.00 |
| $4_{group}^{th}$ | -0.35 | -0.40 | -0.31 | -0.40 | -0.27 | -0.34 | -0.36 | -0.28 | -0.39 | -0.27 | -0.10 | -0.01 | -0.24 | 0.03 | 0.02 | -0.06 | -0.11 | -0.18 | -0.35 | -0.08 |
| $5_{group}^{th}$ | 0.87 | 0.88 | 0.88 | 0.88 | 0.89 | 0.87 | 0.88 | 0.88 | 0.88 | 0.88 | 0.90 | 0.94 | 0.90 | 0.93 | 0.93 | 0.91 | 0.91 | 0.91 | 0.92 | 0.94 |

**4-Node**

| | $MN_{none}^{DI}$ | $MN_{none}^{DS}$ | $MN_{none}^{DIR}$ | $MN_{none}^{DSR}$ | $MN_{none}^{MLB}$ | $MN_{SAL}^{DI}$ | $MN_{SAL}^{DS}$ | $MN_{SAL}^{DIR}$ | $MN_{SAL}^{DSR}$ | $MN_{SAL}^{MLB}$ | $MN_{SAO}^{DI}$ | $MN_{SAO}^{DS}$ | $MN_{SAO}^{DIR}$ | $MN_{SAO}^{DSR}$ | $MN_{SAO}^{MLB}$ | $MN_{SALO}^{DI}$ | $MN_{SALO}^{DS}$ | $MN_{SALO}^{DIR}$ | $MN_{SALO}^{DSR}$ | $MN_{SALO}^{MLB}$ |
|---|---|---|---|---|---|---|---|---|---|---|---|---|---|---|---|---|---|---|---|---|
| $1_{group}^{st}$ | 0.11 | 0.18 | -0.11 | 0.19 | -0.90 | -0.05 | 0.09 | 0.09 | 0.15 | 0.09 | -1.99 | -0.20 | -0.01 | -0.92 | -1.20 | 0.02 | -0.03 | 0.02 | -0.09 | -0.92 |
| $2_{group}^{nd}$ | -1.23 | -1.05 | -1.41 | -0.95 | -2.55 | -1.40 | -1.14 | -0.98 | -1.13 | -1.29 | -2.61 | -1.14 | -0.77 | -1.27 | -1.96 | -1.22 | -1.62 | -1.22 | -0.82 | -1.99 |
| $3_{group}^{rd}$ | -1.59 | -0.97 | -0.94 | -0.97 | -1.33 | -1.14 | -1.04 | -0.83 | -1.02 | -0.98 | -1.04 | -0.46 | -0.51 | -0.45 | -0.82 | -0.89 | -1.38 | -0.89 | -0.44 | -0.79 |
| $4_{group}^{th}$ | -0.61 | -0.36 | -0.31 | -0.40 | -0.25 | -0.32 | -0.36 | -0.27 | -0.39 | -0.24 | -0.08 | 0.18 | -0.05 | 0.10 | 0.03 | -0.20 | -0.30 | -0.20 | 0.11 | 0.02 |
| $5_{group}^{th}$ | 0.87 | 0.88 | 0.88 | 0.88 | 0.88 | 0.87 | 0.88 | 0.88 | 0.88 | 0.89 | 0.90 | 0.94 | 0.91 | 0.94 | 0.94 | 0.90 | 0.92 | 0.90 | 0.94 | 0.93 |

**5-Node**

| | $MN_{none}^{DI}$ | $MN_{none}^{DS}$ | $MN_{none}^{DIR}$ | $MN_{none}^{DSR}$ | $MN_{none}^{MLB}$ | $MN_{SAL}^{DI}$ | $MN_{SAL}^{DS}$ | $MN_{SAL}^{DIR}$ | $MN_{SAL}^{DSR}$ | $MN_{SAL}^{MLB}$ | $MN_{SAO}^{DI}$ | $MN_{SAO}^{DS}$ | $MN_{SAO}^{DIR}$ | $MN_{SAO}^{DSR}$ | $MN_{SAO}^{MLB}$ | $MN_{SALO}^{DI}$ | $MN_{SALO}^{DS}$ | $MN_{SALO}^{DIR}$ | $MN_{SALO}^{DSR}$ | $MN_{SALO}^{MLB}$ |
|---|---|---|---|---|---|---|---|---|---|---|---|---|---|---|---|---|---|---|---|---|
| $1_{group}^{st}$ | -0.06 | 0.15 | 0.05 | 0.18 | -1.35 | -0.05 | 0.15 | 0.22 | 0.19 | -1.11 | -1.41 | -0.22 | 0.21 | -0.25 | -2.16 | -1.14 | 0.27 | -0.51 | -0.55 | -1.68 |
| $2_{group}^{nd}$ | -1.55 | -1.13 | -1.29 | -0.96 | -2.36 | -1.35 | -1.01 | -0.75 | -0.95 | -2.89 | -2.15 | -1.08 | -0.77 | -0.93 | -2.33 | -2.02 | -0.45 | -1.13 | -1.12 | -1.46 |
| $3_{group}^{rd}$ | -1.21 | -0.97 | -0.94 | -0.96 | -1.18 | -1.17 | -0.98 | -0.65 | -0.99 | -1.47 | -0.97 | -0.58 | -0.55 | -0.41 | -0.92 | -1.02 | -0.30 | -0.75 | -0.46 | -0.57 |
| $4_{group}^{th}$ | -0.35 | -0.36 | -0.27 | -0.39 | -0.21 | -0.33 | -0.36 | -0.15 | -0.37 | -0.33 | -0.12 | 0.16 | -0.08 | 0.16 | 0.05 | -0.19 | 0.24 | -0.18 | 0.11 | -0.02 |
| $5_{group}^{th}$ | 0.87 | 0.88 | 0.88 | 0.88 | 0.90 | 0.87 | 0.88 | 0.89 | 0.88 | 0.88 | 0.90 | 0.94 | 0.91 | 0.95 | 0.94 | 0.91 | 0.95 | 0.92 | 0.95 | 0.94 |

*Note that the 1st group represents the flows within the lowest 20th percentile and the 5th group represents the flows within the highest 20th percentile.

*Note also that $KGE_{ss}$ measures the linear correlation between simulation and observation, with an optimal value of 1, and a range from $-\infty$ to 1. The cases include single-layer Distributed-Input (DI), Distributed-State (DS), Distributed-Input Relaxed (DIR), Distributed-State Relaxed (DSR), and Machine-Learning Benchmark (MLB) networks, with the number of nodes varying from 1 to 5. “None” refers to the case without cell-information sharing among different nodes in the gates, whereas “SAL,” “SAO,” and “SALO” represent cases where sharing occurs only in the loss gate, only in the output gate, and in both gates, respectively. We report values to two decimal places; therefore, any difference between two metrics that is greater than or equal to 0.01 is noteworthy. PNs=Parameter Numbers

Table S5. $\alpha^{KGE}$ scores of flow magnitude for the Single-Layer Mass-Conserving Neural Networks (MNs)

**1-Node**

| | $MN_{none}^{DI}$ | $MN_{none}^{DS}$ | $MN_{none}^{DIR}$ | $MN_{none}^{DSR}$ | $MN_{none}^{MLB}$ | | $MN_{SAL}^{DI}$ | $MN_{SAL}^{DS}$ | $MN_{SAL}^{DIR}$ | $MN_{SAL}^{DSR}$ | $MN_{SAL}^{MLB}$ | | $MN_{SAO}^{DI}$ | $MN_{SAO}^{DS}$ | $MN_{SAO}^{DIR}$ | $MN_{SAO}^{DSR}$ | $MN_{SAO}^{MLB}$ | | $MN_{SALO}^{DI}$ | $MN_{SALO}^{DS}$ | $MN_{SALO}^{DIR}$ | $MN_{SALO}^{DSR}$ | $MN_{SALO}^{MLB}$ |
|---|---|---|---|---|---|---|---|---|---|---|---|---|---|---|---|---|---|---|---|---|---|---|---|
| $1_{group}^{st}$ | 1.11 | 1.11 | 1.09 | 1.03 | 0.50 | $1_{group}^{st}$ | | | | | | $1_{group}^{st}$ | | | | | | $1_{group}^{st}$ | | | | | |
| $2_{group}^{nd}$ | 3.74 | 3.74 | 3.67 | 3.58 | 2.27 | $2_{group}^{nd}$ | | | | | | $2_{group}^{nd}$ | | | | | | $2_{group}^{nd}$ | | | | | |
| $3_{group}^{rd}$ | 4.60 | 4.60 | 4.60 | 4.51 | 3.43 | $3_{group}^{rd}$ | | | | | | $3_{group}^{rd}$ | | | | | | $3_{group}^{rd}$ | | | | | |
| $4_{group}^{th}$ | 3.19 | 3.19 | 3.24 | 3.16 | 2.88 | $4_{group}^{th}$ | | | | | | $4_{group}^{th}$ | | | | | | $4_{group}^{th}$ | | | | | |
| $5_{group}^{th}$ | 0.96 | 0.96 | 0.96 | 0.94 | 0.93 | $5_{group}^{th}$ | | | | | | $5_{group}^{th}$ | | | | | | $5_{group}^{th}$ | | | | | |

**2-Node**

| | $MN_{none}^{DI}$ | $MN_{none}^{DS}$ | $MN_{none}^{DIR}$ | $MN_{none}^{DSR}$ | $MN_{none}^{MLB}$ | | $MN_{SAL}^{DI}$ | $MN_{SAL}^{DS}$ | $MN_{SAL}^{DIR}$ | $MN_{SAL}^{DSR}$ | $MN_{SAL}^{MLB}$ | | $MN_{SAO}^{DI}$ | $MN_{SAO}^{DS}$ | $MN_{SAO}^{DIR}$ | $MN_{SAO}^{DSR}$ | $MN_{SAO}^{MLB}$ | | $MN_{SALO}^{DI}$ | $MN_{SALO}^{DS}$ | $MN_{SALO}^{DIR}$ | $MN_{SALO}^{DSR}$ | $MN_{SALO}^{MLB}$ |
|---|---|---|---|---|---|---|---|---|---|---|---|---|---|---|---|---|---|---|---|---|---|---|---|
| $1_{group}^{st}$ | 2.09 | 1.96 | 2.23 | 1.41 | 2.72 | $1_{group}^{st}$ | 2.02 | 1.48 | 1.81 | 1.41 | 2.00 | $1_{group}^{st}$ | 1.68 | 1.74 | 1.35 | 1.94 | 1.60 | $1_{group}^{st}$ | 2.02 | 1.53 | 1.31 | 1.90 | 1.97 |
| $2_{group}^{nd}$ | 4.53 | 4.40 | 4.33 | 3.87 | 5.79 | $2_{group}^{nd}$ | 4.28 | 3.91 | 3.44 | 3.79 | 5.15 | $2_{group}^{nd}$ | 4.07 | 4.30 | 3.62 | 4.88 | 4.92 | $2_{group}^{nd}$ | 4.29 | 4.13 | 3.48 | 4.63 | 5.26 |
| $3_{group}^{rd}$ | 4.09 | 3.89 | 3.71 | 3.89 | 4.34 | $3_{group}^{rd}$ | 3.96 | 3.89 | 3.29 | 3.83 | 4.15 | $3_{group}^{rd}$ | 3.90 | 4.31 | 3.79 | 4.43 | 4.19 | $3_{group}^{rd}$ | 4.00 | 4.30 | 3.76 | 4.39 | 4.13 |
| $4_{group}^{th}$ | 2.80 | 2.79 | 2.76 | 2.92 | 2.83 | $4_{group}^{th}$ | 2.78 | 2.87 | 2.65 | 2.90 | 2.75 | $4_{group}^{th}$ | 2.70 | 3.07 | 2.76 | 3.00 | 2.77 | $4_{group}^{th}$ | 2.68 | 3.04 | 2.73 | 2.97 | 2.60 |
| $5_{group}^{th}$ | 0.97 | 0.97 | 0.99 | 0.97 | 0.95 | $5_{group}^{th}$ | 0.97 | 0.97 | 0.98 | 0.97 | 0.95 | $5_{group}^{th}$ | 0.98 | 1.01 | 0.98 | 1.01 | 0.97 | $5_{group}^{th}$ | 0.99 | 1.00 | 0.98 | 1.00 | 0.96 |

**3-Node**

| | $MN_{none}^{DI}$ | $MN_{none}^{DS}$ | $MN_{none}^{DIR}$ | $MN_{none}^{DSR}$ | $MN_{none}^{MLB}$ | | $MN_{SAL}^{DI}$ | $MN_{SAL}^{DS}$ | $MN_{SAL}^{DIR}$ | $MN_{SAL}^{DSR}$ | $MN_{SAL}^{MLB}$ | | $MN_{SAO}^{DI}$ | $MN_{SAO}^{DS}$ | $MN_{SAO}^{DIR}$ | $MN_{SAO}^{DSR}$ | $MN_{SAO}^{MLB}$ | | $MN_{SALO}^{DI}$ | $MN_{SALO}^{DS}$ | $MN_{SALO}^{DIR}$ | $MN_{SALO}^{DSR}$ | $MN_{SALO}^{MLB}$ |
|---|---|---|---|---|---|---|---|---|---|---|---|---|---|---|---|---|---|---|---|---|---|---|---|
| $1_{group}^{st}$ | 2.08 | 1.57 | 2.22 | 1.29 | 3.51 | $1_{group}^{st}$ | 2.00 | 1.55 | 1.76 | 1.29 | 2.68 | $1_{group}^{st}$ | 3.91 | 2.52 | 1.94 | 3.22 | 4.84 | $1_{group}^{st}$ | 2.86 | 2.37 | 1.53 | 2.19 | 3.43 |
| $2_{group}^{nd}$ | 4.54 | 4.19 | 4.31 | 3.63 | 5.95 | $2_{group}^{nd}$ | 4.25 | 3.81 | 3.71 | 3.72 | 5.54 | $2_{group}^{nd}$ | 5.61 | 4.07 | 4.37 | 4.50 | 5.91 | $2_{group}^{nd}$ | 4.19 | 4.03 | 3.75 | 4.48 | 5.16 |
| $3_{group}^{rd}$ | 4.08 | 3.96 | 3.69 | 3.74 | 4.20 | $3_{group}^{rd}$ | 3.96 | 3.80 | 3.52 | 3.77 | 4.12 | $3_{group}^{rd}$ | 3.87 | 3.24 | 3.87 | 3.16 | 3.58 | $3_{group}^{rd}$ | 3.27 | 3.40 | 3.56 | 3.92 | 3.76 |
| $4_{group}^{th}$ | 2.80 | 2.88 | 2.75 | 2.88 | 2.69 | $4_{group}^{th}$ | 2.78 | 2.81 | 2.70 | 2.86 | 2.68 | $4_{group}^{th}$ | 2.43 | 2.28 | 2.65 | 2.22 | 2.25 | $4_{group}^{th}$ | 2.37 | 2.43 | 2.57 | 2.79 | 2.39 |
| $5_{group}^{th}$ | 0.97 | 0.97 | 0.99 | 0.96 | 0.95 | $5_{group}^{th}$ | 0.97 | 0.97 | 0.98 | 0.97 | 0.95 | $5_{group}^{th}$ | 1.05 | 1.02 | 0.99 | 1.05 | 1.01 | $5_{group}^{th}$ | 1.03 | 1.02 | 0.99 | 1.00 | 1.02 |

**4-Node**

| | $MN_{none}^{DI}$ | $MN_{none}^{DS}$ | $MN_{none}^{DIR}$ | $MN_{none}^{DSR}$ | $MN_{none}^{MLB}$ | | $MN_{SAL}^{DI}$ | $MN_{SAL}^{DS}$ | $MN_{SAL}^{DIR}$ | $MN_{SAL}^{DSR}$ | $MN_{SAL}^{MLB}$ | | $MN_{SAO}^{DI}$ | $MN_{SAO}^{DS}$ | $MN_{SAO}^{DIR}$ | $MN_{SAO}^{DSR}$ | $MN_{SAO}^{MLB}$ | | $MN_{SALO}^{DI}$ | $MN_{SALO}^{DS}$ | $MN_{SALO}^{DIR}$ | $MN_{SALO}^{DSR}$ | $MN_{SALO}^{MLB}$ |
|---|---|---|---|---|---|---|---|---|---|---|---|---|---|---|---|---|---|---|---|---|---|---|---|
| $1_{group}^{st}$ | 1.55 | 1.48 | 2.20 | 1.29 | 3.44 | $1_{group}^{st}$ | 2.06 | 1.72 | 1.72 | 1.58 | 1.40 | $1_{group}^{st}$ | 5.17 | 2.53 | 2.28 | 3.64 | 3.99 | $1_{group}^{st}$ | 1.90 | 2.13 | 1.90 | 2.41 | 3.55 |
| $2_{group}^{nd}$ | 4.04 | 3.80 | 4.32 | 3.64 | 5.95 | $2_{group}^{nd}$ | 4.31 | 3.93 | 3.69 | 3.91 | 4.13 | $2_{group}^{nd}$ | 6.05 | 3.97 | 3.42 | 4.15 | 5.13 | $2_{group}^{nd}$ | 4.05 | 4.12 | 4.05 | 3.49 | 5.18 |
| $3_{group}^{rd}$ | 4.61 | 3.73 | 3.68 | 3.73 | 4.24 | $3_{group}^{rd}$ | 3.97 | 3.82 | 3.54 | 3.80 | 3.75 | $3_{group}^{rd}$ | 3.80 | 3.00 | 3.06 | 2.96 | 3.49 | $3_{group}^{rd}$ | 3.61 | 3.51 | 3.61 | 2.96 | 3.47 |
| $4_{group}^{th}$ | 3.17 | 2.81 | 2.75 | 2.87 | 2.66 | $4_{group}^{th}$ | 2.76 | 2.81 | 2.69 | 2.86 | 2.63 | $4_{group}^{th}$ | 2.40 | 2.03 | 2.36 | 2.14 | 2.23 | $4_{group}^{th}$ | 2.59 | 2.36 | 2.59 | 2.11 | 2.26 |
| $5_{group}^{th}$ | 0.97 | 0.97 | 0.99 | 0.96 | 0.95 | $5_{group}^{th}$ | 0.97 | 0.97 | 0.98 | 0.96 | 0.97 | $5_{group}^{th}$ | 1.04 | 1.02 | 1.03 | 1.04 | 1.01 | $5_{group}^{th}$ | 0.99 | 1.03 | 0.99 | 1.03 | 0.98 |

**5-Node**

| | $MN_{none}^{DI}$ | $MN_{none}^{DS}$ | $MN_{none}^{DIR}$ | $MN_{none}^{DSR}$ | $MN_{none}^{MLB}$ | | $MN_{SAL}^{DI}$ | $MN_{SAL}^{DS}$ | $MN_{SAL}^{DIR}$ | $MN_{SAL}^{DSR}$ | $MN_{SAL}^{MLB}$ | | $MN_{SAO}^{DI}$ | $MN_{SAO}^{DS}$ | $MN_{SAO}^{DIR}$ | $MN_{SAO}^{DSR}$ | $MN_{SAO}^{MLB}$ | | $MN_{SALO}^{DI}$ | $MN_{SALO}^{DS}$ | $MN_{SALO}^{DIR}$ | $MN_{SALO}^{DSR}$ | $MN_{SALO}^{MLB}$ |
|---|---|---|---|---|---|---|---|---|---|---|---|---|---|---|---|---|---|---|---|---|---|---|---|
| $1_{group}^{st}$ | 2.08 | 1.53 | 1.87 | 1.31 | 4.18 | $1_{group}^{st}$ | 2.06 | 1.49 | 1.55 | 1.37 | 3.78 | $1_{group}^{st}$ | 4.34 | 2.53 | 1.85 | 2.61 | 5.37 | $1_{group}^{st}$ | 3.92 | 1.83 | 2.96 | 3.12 | 4.74 |
| $2_{group}^{nd}$ | 4.51 | 3.90 | 4.14 | 3.65 | 5.69 | $2_{group}^{nd}$ | 4.24 | 3.73 | 3.39 | 3.64 | 6.45 | $2_{group}^{nd}$ | 5.40 | 3.89 | 3.43 | 3.67 | 5.65 | $2_{group}^{nd}$ | 5.22 | 2.99 | 3.91 | 3.92 | 4.41 |
| $3_{group}^{rd}$ | 4.07 | 3.73 | 3.69 | 3.71 | 4.02 | $3_{group}^{rd}$ | 4.02 | 3.74 | 3.27 | 3.75 | 4.44 | $3_{group}^{rd}$ | 3.71 | 3.16 | 3.12 | 2.92 | 3.65 | $3_{group}^{rd}$ | 3.77 | 2.76 | 3.41 | 2.98 | 3.14 |
| $4_{group}^{th}$ | 2.79 | 2.82 | 2.69 | 2.86 | 2.60 | $4_{group}^{th}$ | 2.77 | 2.82 | 2.51 | 2.83 | 2.76 | $4_{group}^{th}$ | 2.45 | 2.06 | 2.41 | 2.06 | 2.20 | $4_{group}^{th}$ | 2.55 | 1.93 | 2.56 | 2.14 | 2.32 |
| $5_{group}^{th}$ | 0.97 | 0.96 | 0.98 | 0.96 | 0.98 | $5_{group}^{th}$ | 0.97 | 0.96 | 0.98 | 0.96 | 0.94 | $5_{group}^{th}$ | 1.05 | 1.02 | 1.01 | 1.03 | 1.02 | $5_{group}^{th}$ | 1.05 | 1.03 | 1.03 | 1.02 | 1.01 |

*Note that the 1st group represents the flows within the lowest 20th percentile and the 5th group represents the flows within the highest 20th percentile.

*Note also that $\alpha^{KGE}$ measures the consistency of flow variability between simulation and observation, with an optimal value of 1 and a range from 0 to ∞. The cases include single-layer Distributed-Input (DI), Distributed-State (DS), Distributed-Input Relaxed (DIR), Distributed-State Relaxed (DSR), and Machine-Learning Benchmark (MLB) networks, with the number of nodes varying from 1 to 5. "None" refers to the case without cell-information sharing among different nodes in the gates, whereas "SAL," "SAO," and "SALO" represent cases where sharing occurs only in the loss gate, only in the output gate, and in both gates, respectively. We report values to two decimal places; therefore, any difference between two metrics that is greater than or equal to 0.01 is noteworthy. PNs=Parameter Numbers

Table S6. $\beta^{KGE}$ scores of flow magnitude for the Single-Layer Mass-Conserving Neural Networks (MNs)

1-Node

| | $MN_{none}^{DI}$ | $MN_{none}^{DS}$ | $MN_{none}^{DIR}$ | $MN_{none}^{DSR}$ | $MN_{none}^{MLB}$ | | $MN_{SAL}^{DI}$ | $MN_{SAL}^{DS}$ | $MN_{SAL}^{DIR}$ | $MN_{SAL}^{DSR}$ | $MN_{SAL}^{MLB}$ | | $MN_{SAO}^{DI}$ | $MN_{SAO}^{DS}$ | $MN_{SAO}^{DIR}$ | $MN_{SAO}^{DSR}$ | $MN_{SAO}^{MLB}$ | | $MN_{SALO}^{DI}$ | $MN_{SALO}^{DS}$ | $MN_{SALO}^{DIR}$ | $MN_{SALO}^{DSR}$ | $MN_{SALO}^{MLB}$ |
|---|---|---|---|---|---|---|---|---|---|---|---|---|---|---|---|---|---|---|---|---|---|---|---|
| $1_{group}^{st}$ | 0.12 | 0.12 | 0.10 | 0.10 | 0.03 | $1_{group}^{st}$ | | | | | | $1_{group}^{st}$ | | | | | | $1_{group}^{st}$ | | | | | |
| $2_{group}^{nd}$ | 0.43 | 0.43 | 0.41 | 0.40 | 0.19 | $2_{group}^{nd}$ | | | | | | $2_{group}^{nd}$ | | | | | | $2_{group}^{nd}$ | | | | | |
| $3_{group}^{rd}$ | 0.97 | 0.97 | 0.97 | 0.95 | 0.62 | $3_{group}^{rd}$ | | | | | | $3_{group}^{rd}$ | | | | | | $3_{group}^{rd}$ | | | | | |
| $4_{group}^{th}$ | 1.25 | 1.25 | 1.27 | 1.24 | 0.97 | $4_{group}^{th}$ | | | | | | $4_{group}^{th}$ | | | | | | $4_{group}^{th}$ | | | | | |
| $5_{group}^{th}$ | 0.98 | 0.98 | 0.99 | 0.96 | 0.87 | $5_{group}^{th}$ | | | | | | $5_{group}^{th}$ | | | | | | $5_{group}^{th}$ | | | | | |

2-Node

| | $MN_{none}^{DI}$ | $MN_{none}^{DS}$ | $MN_{none}^{DIR}$ | $MN_{none}^{DSR}$ | $MN_{none}^{MLB}$ | | $MN_{SAL}^{DI}$ | $MN_{SAL}^{DS}$ | $MN_{SAL}^{DIR}$ | $MN_{SAL}^{DSR}$ | $MN_{SAL}^{MLB}$ | | $MN_{SAO}^{DI}$ | $MN_{SAO}^{DS}$ | $MN_{SAO}^{DIR}$ | $MN_{SAO}^{DSR}$ | $MN_{SAO}^{MLB}$ | | $MN_{SALO}^{DI}$ | $MN_{SALO}^{DS}$ | $MN_{SALO}^{DIR}$ | $MN_{SALO}^{DSR}$ | $MN_{SALO}^{MLB}$ |
|---|---|---|---|---|---|---|---|---|---|---|---|---|---|---|---|---|---|---|---|---|---|---|---|
| $1_{group}^{st}$ | 0.23 | 0.24 | 0.26 | 0.14 | 0.14 | $1_{group}^{st}$ | 0.22 | 0.16 | 0.24 | 0.15 | 0.08 | $1_{group}^{st}$ | 0.17 | 0.22 | 0.15 | 0.20 | 0.07 | $1_{group}^{st}$ | 0.19 | 0.18 | 0.13 | 0.18 | 0.09 |
| $2_{group}^{nd}$ | 0.60 | 0.65 | 0.62 | 0.51 | 0.61 | $2_{group}^{nd}$ | 0.59 | 0.53 | 0.60 | 0.52 | 0.54 | $2_{group}^{nd}$ | 0.54 | 0.61 | 0.50 | 0.66 | 0.52 | $2_{group}^{nd}$ | 0.56 | 0.57 | 0.45 | 0.62 | 0.56 |
| $3_{group}^{rd}$ | 1.03 | 1.06 | 0.99 | 0.98 | 1.05 | $3_{group}^{rd}$ | 1.02 | 1.00 | 0.99 | 0.99 | 1.08 | $3_{group}^{rd}$ | 1.00 | 1.09 | 0.98 | 1.14 | 1.04 | $3_{group}^{rd}$ | 1.03 | 1.08 | 0.95 | 1.13 | 1.10 |
| $4_{group}^{th}$ | 1.24 | 1.26 | 1.23 | 1.25 | 1.34 | $4_{group}^{th}$ | 1.24 | 1.25 | 1.22 | 1.25 | 1.34 | $4_{group}^{th}$ | 1.23 | 1.32 | 1.24 | 1.30 | 1.31 | $4_{group}^{th}$ | 1.24 | 1.31 | 1.23 | 1.30 | 1.32 |
| $5_{group}^{th}$ | 0.99 | 0.97 | 0.99 | 0.99 | 1.01 | $5_{group}^{th}$ | 0.99 | 0.99 | 0.99 | 0.99 | 1.01 | $5_{group}^{th}$ | 1.00 | 0.96 | 1.00 | 0.96 | 1.00 | $5_{group}^{th}$ | 0.99 | 0.96 | 1.00 | 0.96 | 1.01 |

3-Node

| | $MN_{none}^{DI}$ | $MN_{none}^{DS}$ | $MN_{none}^{DIR}$ | $MN_{none}^{DSR}$ | $MN_{none}^{MLB}$ | | $MN_{SAL}^{DI}$ | $MN_{SAL}^{DS}$ | $MN_{SAL}^{DIR}$ | $MN_{SAL}^{DSR}$ | $MN_{SAL}^{MLB}$ | | $MN_{SAO}^{DI}$ | $MN_{SAO}^{DS}$ | $MN_{SAO}^{DIR}$ | $MN_{SAO}^{DSR}$ | $MN_{SAO}^{MLB}$ | | $MN_{SALO}^{DI}$ | $MN_{SALO}^{DS}$ | $MN_{SALO}^{DIR}$ | $MN_{SALO}^{DSR}$ | $MN_{SALO}^{MLB}$ |
|---|---|---|---|---|---|---|---|---|---|---|---|---|---|---|---|---|---|---|---|---|---|---|---|
| $1_{group}^{st}$ | 0.22 | 0.16 | 0.26 | 0.13 | 0.23 | $1_{group}^{st}$ | 0.22 | 0.18 | 0.20 | 0.15 | 0.14 | $1_{group}^{st}$ | 0.56 | 0.52 | 0.22 | 0.61 | 0.49 | $1_{group}^{st}$ | 0.87 | 0.46 | 0.16 | 0.21 | 0.36 |
| $2_{group}^{nd}$ | 0.60 | 0.54 | 0.62 | 0.47 | 0.64 | $2_{group}^{nd}$ | 0.59 | 0.55 | 0.56 | 0.51 | 0.59 | $2_{group}^{nd}$ | 0.99 | 0.90 | 0.60 | 1.01 | 1.10 | $2_{group}^{nd}$ | 1.00 | 0.86 | 0.51 | 0.67 | 0.83 |
| $3_{group}^{rd}$ | 1.03 | 1.00 | 0.99 | 0.95 | 1.02 | $3_{group}^{rd}$ | 1.02 | 1.01 | 0.98 | 1.00 | 1.04 | $3_{group}^{rd}$ | 1.19 | 1.14 | 0.99 | 1.19 | 1.25 | $3_{group}^{rd}$ | 1.13 | 1.19 | 0.94 | 1.12 | 1.20 |
| $4_{group}^{th}$ | 1.24 | 1.24 | 1.23 | 1.24 | 1.30 | $4_{group}^{th}$ | 1.24 | 1.24 | 1.23 | 1.25 | 1.33 | $4_{group}^{th}$ | 1.20 | 1.17 | 1.22 | 1.19 | 1.25 | $4_{group}^{th}$ | 1.15 | 1.27 | 1.18 | 1.26 | 1.25 |
| $5_{group}^{th}$ | 0.99 | 0.99 | 0.99 | 1.00 | 1.01 | $5_{group}^{th}$ | 0.99 | 0.99 | 0.99 | 0.99 | 1.01 | $5_{group}^{th}$ | 0.95 | 0.96 | 0.99 | 0.96 | 0.96 | $5_{group}^{th}$ | 0.96 | 0.94 | 1.00 | 0.97 | 0.97 |

4-Node

| | $MN_{none}^{DI}$ | $MN_{none}^{DS}$ | $MN_{none}^{DIR}$ | $MN_{none}^{DSR}$ | $MN_{none}^{MLB}$ | | $MN_{SAL}^{DI}$ | $MN_{SAL}^{DS}$ | $MN_{SAL}^{DIR}$ | $MN_{SAL}^{DSR}$ | $MN_{SAL}^{MLB}$ | | $MN_{SAO}^{DI}$ | $MN_{SAO}^{DS}$ | $MN_{SAO}^{DIR}$ | $MN_{SAO}^{DSR}$ | $MN_{SAO}^{MLB}$ | | $MN_{SALO}^{DI}$ | $MN_{SALO}^{DS}$ | $MN_{SALO}^{DIR}$ | $MN_{SALO}^{DSR}$ | $MN_{SALO}^{MLB}$ |
|---|---|---|---|---|---|---|---|---|---|---|---|---|---|---|---|---|---|---|---|---|---|---|---|
| $1_{group}^{st}$ | 0.14 | 0.19 | 0.26 | 0.13 | 0.22 | $1_{group}^{st}$ | 0.23 | 0.18 | 0.19 | 0.20 | 0.07 | $1_{group}^{st}$ | 0.88 | 0.54 | 0.79 | 0.75 | 0.44 | $1_{group}^{st}$ | 0.21 | 0.29 | 0.21 | 0.98 | 0.40 |
| $2_{group}^{nd}$ | 0.46 | 0.54 | 0.62 | 0.48 | 0.66 | $2_{group}^{nd}$ | 0.61 | 0.56 | 0.54 | 0.55 | 0.43 | $2_{group}^{nd}$ | 1.21 | 0.91 | 0.95 | 1.03 | 0.94 | $2_{group}^{nd}$ | 0.59 | 0.76 | 0.59 | 1.01 | 0.90 |
| $3_{group}^{rd}$ | 1.01 | 0.99 | 0.99 | 0.96 | 1.05 | $3_{group}^{rd}$ | 1.03 | 1.01 | 0.97 | 1.00 | 0.99 | $3_{group}^{rd}$ | 1.24 | 1.16 | 1.11 | 1.12 | 1.22 | $3_{group}^{rd}$ | 0.98 | 1.18 | 0.98 | 1.12 | 1.16 |
| $4_{group}^{th}$ | 1.28 | 1.24 | 1.23 | 1.24 | 1.30 | $4_{group}^{th}$ | 1.24 | 1.24 | 1.23 | 1.24 | 1.27 | $4_{group}^{th}$ | 1.20 | 1.17 | 1.19 | 1.13 | 1.23 | $4_{group}^{th}$ | 1.18 | 1.26 | 1.18 | 1.13 | 1.26 |
| $5_{group}^{th}$ | 0.99 | 0.99 | 0.99 | 1.00 | 1.01 | $5_{group}^{th}$ | 0.99 | 0.99 | 1.00 | 0.99 | 1.01 | $5_{group}^{th}$ | 0.94 | 0.97 | 0.95 | 0.96 | 0.97 | $5_{group}^{th}$ | 0.98 | 0.96 | 0.98 | 0.96 | 0.98 |

5-Node

| | $MN_{none}^{DI}$ | $MN_{none}^{DS}$ | $MN_{none}^{DIR}$ | $MN_{none}^{DSR}$ | $MN_{none}^{MLB}$ | | $MN_{SAL}^{DI}$ | $MN_{SAL}^{DS}$ | $MN_{SAL}^{DIR}$ | $MN_{SAL}^{DSR}$ | $MN_{SAL}^{MLB}$ | | $MN_{SAO}^{DI}$ | $MN_{SAO}^{DS}$ | $MN_{SAO}^{DIR}$ | $MN_{SAO}^{DSR}$ | $MN_{SAO}^{MLB}$ | | $MN_{SALO}^{DI}$ | $MN_{SALO}^{DS}$ | $MN_{SALO}^{DIR}$ | $MN_{SALO}^{DSR}$ | $MN_{SALO}^{MLB}$ |
|---|---|---|---|---|---|---|---|---|---|---|---|---|---|---|---|---|---|---|---|---|---|---|---|
| $1_{group}^{st}$ | 0.22 | 0.16 | 0.24 | 0.13 | 0.41 | $1_{group}^{st}$ | 0.23 | 0.15 | 0.23 | 0.15 | 0.25 | $1_{group}^{st}$ | 0.82 | 0.43 | 0.57 | 0.74 | 0.75 | $1_{group}^{st}$ | 0.78 | 0.72 | 0.84 | 0.90 | 0.86 |
| $2_{group}^{nd}$ | 0.60 | 0.54 | 0.62 | 0.48 | 0.73 | $2_{group}^{nd}$ | 0.62 | 0.50 | 0.63 | 0.50 | 0.68 | $2_{group}^{nd}$ | 1.08 | 0.86 | 0.83 | 0.95 | 0.89 | $2_{group}^{nd}$ | 1.04 | 0.93 | 0.84 | 1.03 | 1.06 |
| $3_{group}^{rd}$ | 1.02 | 0.99 | 1.00 | 0.96 | 0.98 | $3_{group}^{rd}$ | 1.05 | 0.97 | 1.04 | 0.99 | 1.04 | $3_{group}^{rd}$ | 1.15 | 1.14 | 1.12 | 1.11 | 1.14 | $3_{group}^{rd}$ | 1.12 | 1.13 | 1.03 | 1.08 | 1.11 |
| $4_{group}^{th}$ | 1.24 | 1.24 | 1.23 | 1.24 | 1.21 | $4_{group}^{th}$ | 1.25 | 1.23 | 1.21 | 1.25 | 1.34 | $4_{group}^{th}$ | 1.18 | 1.19 | 1.22 | 1.15 | 1.23 | $4_{group}^{th}$ | 1.17 | 1.15 | 1.18 | 1.12 | 1.19 |
| $5_{group}^{th}$ | 0.98 | 0.99 | 0.98 | 1.00 | 1.00 | $5_{group}^{th}$ | 0.99 | 0.99 | 0.99 | 0.99 | 1.01 | $5_{group}^{th}$ | 0.94 | 0.97 | 0.96 | 0.97 | 0.97 | $5_{group}^{th}$ | 0.96 | 0.97 | 0.97 | 0.97 | 0.96 |

*Note that the 1st group represents the flows within the lowest 20th percentile and the 5th group represents the flows within the highest 20th percentile.

*Note also that $\beta^{KGE}$ measures the consistency of flow variability between simulation and observation, with an optimal value of 1 and a range from 0 to $\infty$. The cases include single-layer Distributed-Input (DI), Distributed-State (DS), Distributed-Input Relaxed (DIR), Distributed-State Relaxed (DSR), and Machine-Learning Benchmark (MLB) networks, with the number of nodes varying from 1 to 5. "None" refers to the case without cell-information sharing among different nodes in the gates, whereas "SAL," "SAO," and "SALO" represent cases where sharing occurs only in the loss gate, only in the output gate, and in both gates, respectively. We report values to two decimal places; therefore, any difference between two metrics that is greater than or equal to 0.01 is noteworthy. PNs=Parameter Numbers

Table S7. $r^{KGE}$ scores of flow magnitude for the Single-Layer Mass-Conserving Neural Networks (MNs)

**1-Node**

| | $MN_{none}^{DI}$ | $MN_{none}^{DS}$ | $MN_{none}^{DIR}$ | $MN_{none}^{DSR}$ | $MN_{none}^{MLB}$ | | $MN_{SAL}^{DI}$ | $MN_{SAL}^{DS}$ | $MN_{SAL}^{DIR}$ | $MN_{SAL}^{DSR}$ | $MN_{SAL}^{MLB}$ | | $MN_{SAO}^{DI}$ | $MN_{SAO}^{DS}$ | $MN_{SAO}^{DIR}$ | $MN_{SAO}^{DSR}$ | $MN_{SAO}^{MLB}$ | | $MN_{SALO}^{DI}$ | $MN_{SALO}^{DS}$ | $MN_{SALO}^{DIR}$ | $MN_{SALO}^{DSR}$ | $MN_{SALO}^{MLB}$ |
|---|---|---|---|---|---|---|---|---|---|---|---|---|---|---|---|---|---|---|---|---|---|---|---|
| $1_{group}^{st}$ | 0.29 | 0.33 | 0.30 | 0.31 | 0.19 | $1_{group}^{st}$ | | | | | | $1_{group}^{st}$ | | | | | | $1_{group}^{st}$ | | | | | |
| $2_{group}^{nd}$ | 0.32 | 0.35 | 0.32 | 0.35 | 0.28 | $2_{group}^{nd}$ | | | | | | $2_{group}^{nd}$ | | | | | | $2_{group}^{nd}$ | | | | | |
| $3_{group}^{rd}$ | 0.40 | 0.41 | 0.42 | 0.42 | 0.41 | $3_{group}^{rd}$ | | | | | | $3_{group}^{rd}$ | | | | | | $3_{group}^{rd}$ | | | | | |
| $4_{group}^{th}$ | 0.39 | 0.41 | 0.42 | 0.40 | 0.44 | $4_{group}^{th}$ | | | | | | $4_{group}^{th}$ | | | | | | $4_{group}^{th}$ | | | | | |
| $5_{group}^{th}$ | 0.82 | 0.83 | 0.83 | 0.83 | 0.84 | $5_{group}^{th}$ | | | | | | $5_{group}^{th}$ | | | | | | $5_{group}^{th}$ | | | | | |

**2-Node**

| | $MN_{none}^{DI}$ | $MN_{none}^{DS}$ | $MN_{none}^{DIR}$ | $MN_{none}^{DSR}$ | $MN_{none}^{MLB}$ | | $MN_{SAL}^{DI}$ | $MN_{SAL}^{DS}$ | $MN_{SAL}^{DIR}$ | $MN_{SAL}^{DSR}$ | $MN_{SAL}^{MLB}$ | | $MN_{SAO}^{DI}$ | $MN_{SAO}^{DS}$ | $MN_{SAO}^{DIR}$ | $MN_{SAO}^{DSR}$ | $MN_{SAO}^{MLB}$ | | $MN_{SALO}^{DI}$ | $MN_{SALO}^{DS}$ | $MN_{SALO}^{DIR}$ | $MN_{SALO}^{DSR}$ | $MN_{SALO}^{MLB}$ |
|---|---|---|---|---|---|---|---|---|---|---|---|---|---|---|---|---|---|---|---|---|---|---|---|
| $1_{group}^{st}$ | 0.29 | 0.33 | 0.30 | 0.31 | 0.19 | $1_{group}^{st}$ | 0.29 | 0.32 | 0.35 | 0.33 | 0.16 | $1_{group}^{st}$ | 0.29 | 0.34 | 0.33 | 0.33 | 0.18 | $1_{group}^{st}$ | 0.35 | 0.31 | 0.27 | 0.29 | 0.17 |
| $2_{group}^{nd}$ | 0.32 | 0.35 | 0.32 | 0.35 | 0.28 | $2_{group}^{nd}$ | 0.34 | 0.36 | 0.41 | 0.36 | 0.34 | $2_{group}^{nd}$ | 0.36 | 0.34 | 0.38 | 0.35 | 0.33 | $2_{group}^{nd}$ | 0.35 | 0.36 | 0.35 | 0.37 | 0.35 |
| $3_{group}^{rd}$ | 0.40 | 0.41 | 0.42 | 0.42 | 0.41 | $3_{group}^{rd}$ | 0.41 | 0.42 | 0.48 | 0.42 | 0.43 | $3_{group}^{rd}$ | 0.42 | 0.42 | 0.44 | 0.38 | 0.42 | $3_{group}^{rd}$ | 0.41 | 0.38 | 0.42 | 0.45 | 0.42 |
| $4_{group}^{th}$ | 0.39 | 0.41 | 0.42 | 0.40 | 0.44 | $4_{group}^{th}$ | 0.40 | 0.40 | 0.43 | 0.40 | 0.45 | $4_{group}^{th}$ | 0.41 | 0.36 | 0.42 | 0.35 | 0.43 | $4_{group}^{th}$ | 0.36 | 0.36 | 0.40 | 0.42 | 0.46 |
| $5_{group}^{th}$ | 0.82 | 0.83 | 0.83 | 0.83 | 0.84 | $5_{group}^{th}$ | 0.83 | 0.83 | 0.84 | 0.83 | 0.84 | $5_{group}^{th}$ | 0.85 | 0.87 | 0.84 | 0.88 | 0.85 | $5_{group}^{th}$ | 0.87 | 0.88 | 0.85 | 0.85 | 0.85 |

**3-Node**

| | $MN_{none}^{DI}$ | $MN_{none}^{DS}$ | $MN_{none}^{DIR}$ | $MN_{none}^{DSR}$ | $MN_{none}^{MLB}$ | | $MN_{SAL}^{DI}$ | $MN_{SAL}^{DS}$ | $MN_{SAL}^{DIR}$ | $MN_{SAL}^{DSR}$ | $MN_{SAL}^{MLB}$ | | $MN_{SAO}^{DI}$ | $MN_{SAO}^{DS}$ | $MN_{SAO}^{DIR}$ | $MN_{SAO}^{DSR}$ | $MN_{SAO}^{MLB}$ | | $MN_{SALO}^{DI}$ | $MN_{SALO}^{DS}$ | $MN_{SALO}^{DIR}$ | $MN_{SALO}^{DSR}$ | $MN_{SALO}^{MLB}$ |
|---|---|---|---|---|---|---|---|---|---|---|---|---|---|---|---|---|---|---|---|---|---|---|---|
| $1_{group}^{st}$ | 0.29 | 0.31 | 0.31 | 0.31 | 0.19 | $1_{group}^{st}$ | 0.30 | 0.33 | 0.32 | 0.36 | 0.19 | $1_{group}^{st}$ | 0.33 | 0.43 | 0.30 | 0.39 | 0.34 | $1_{group}^{st}$ | 0.42 | 0.31 | 0.25 | 0.27 | 0.28 |
| $2_{group}^{nd}$ | 0.32 | 0.34 | 0.33 | 0.36 | 0.27 | $2_{group}^{nd}$ | 0.34 | 0.37 | 0.38 | 0.37 | 0.31 | $2_{group}^{nd}$ | 0.32 | 0.42 | 0.34 | 0.40 | 0.36 | $2_{group}^{nd}$ | 0.42 | 0.37 | 0.34 | 0.36 | 0.34 |
| $3_{group}^{rd}$ | 0.40 | 0.41 | 0.42 | 0.44 | 0.39 | $3_{group}^{rd}$ | 0.41 | 0.43 | 0.45 | 0.43 | 0.42 | $3_{group}^{rd}$ | 0.36 | 0.42 | 0.41 | 0.42 | 0.38 | $3_{group}^{rd}$ | 0.47 | 0.41 | 0.42 | 0.45 | 0.44 |
| $4_{group}^{th}$ | 0.39 | 0.40 | 0.42 | 0.41 | 0.46 | $4_{group}^{th}$ | 0.40 | 0.40 | 0.43 | 0.41 | 0.47 | $4_{group}^{th}$ | 0.41 | 0.41 | 0.43 | 0.41 | 0.44 | $4_{group}^{th}$ | 0.40 | 0.37 | 0.42 | 0.43 | 0.40 |
| $5_{group}^{th}$ | 0.82 | 0.83 | 0.84 | 0.83 | 0.84 | $5_{group}^{th}$ | 0.83 | 0.84 | 0.83 | 0.83 | 0.85 | $5_{group}^{th}$ | 0.88 | 0.92 | 0.85 | 0.92 | 0.90 | $5_{group}^{th}$ | 0.90 | 0.89 | 0.89 | 0.87 | 0.92 |

**4-Node**

| | $MN_{none}^{DI}$ | $MN_{none}^{DS}$ | $MN_{none}^{DIR}$ | $MN_{none}^{DSR}$ | $MN_{none}^{MLB}$ | | $MN_{SAL}^{DI}$ | $MN_{SAL}^{DS}$ | $MN_{SAL}^{DIR}$ | $MN_{SAL}^{DSR}$ | $MN_{SAL}^{MLB}$ | | $MN_{SAO}^{DI}$ | $MN_{SAO}^{DS}$ | $MN_{SAO}^{DIR}$ | $MN_{SAO}^{DSR}$ | $MN_{SAO}^{MLB}$ | | $MN_{SALO}^{DI}$ | $MN_{SALO}^{DS}$ | $MN_{SALO}^{DIR}$ | $MN_{SALO}^{DSR}$ | $MN_{SALO}^{MLB}$ |
|---|---|---|---|---|---|---|---|---|---|---|---|---|---|---|---|---|---|---|---|---|---|---|---|
| $1_{group}^{st}$ | 0.26 | 0.33 | 0.31 | 0.31 | 0.19 | $1_{group}^{st}$ | 0.30 | 0.32 | 0.31 | 0.32 | 0.19 | $1_{group}^{st}$ | 0.32 | 0.42 | 0.41 | 0.43 | 0.36 | $1_{group}^{st}$ | 0.30 | 0.37 | 0.30 | 0.30 | 0.29 |
| $2_{group}^{nd}$ | 0.32 | 0.36 | 0.33 | 0.36 | 0.29 | $2_{group}^{nd}$ | 0.35 | 0.36 | 0.37 | 0.36 | 0.36 | $2_{group}^{nd}$ | 0.29 | 0.44 | 0.40 | 0.36 | 0.36 | $2_{group}^{nd}$ | 0.32 | 0.36 | 0.37 | 0.37 | 0.37 |
| $3_{group}^{rd}$ | 0.40 | 0.44 | 0.43 | 0.44 | 0.39 | $3_{group}^{rd}$ | 0.41 | 0.43 | 0.46 | 0.43 | 0.48 | $3_{group}^{rd}$ | 0.34 | 0.47 | 0.46 | 0.42 | 0.43 | $3_{group}^{rd}$ | 0.41 | 0.46 | 0.42 | 0.42 | 0.44 |
| $4_{group}^{th}$ | 0.35 | 0.41 | 0.42 | 0.41 | 0.47 | $4_{group}^{th}$ | 0.40 | 0.40 | 0.43 | 0.40 | 0.44 | $4_{group}^{th}$ | 0.42 | 0.47 | 0.44 | 0.44 | 0.46 | $4_{group}^{th}$ | 0.39 | 0.44 | 0.42 | 0.42 | 0.50 |
| $5_{group}^{th}$ | 0.82 | 0.83 | 0.84 | 0.83 | 0.84 | $5_{group}^{th}$ | 0.83 | 0.84 | 0.83 | 0.83 | 0.85 | $5_{group}^{th}$ | 0.88 | 0.93 | 0.89 | 0.93 | 0.93 | $5_{group}^{th}$ | 0.89 | 0.93 | 0.86 | 0.86 | 0.90 |

**5-Node**

| | $MN_{none}^{DI}$ | $MN_{none}^{DS}$ | $MN_{none}^{DIR}$ | $MN_{none}^{DSR}$ | $MN_{none}^{MLB}$ | | $MN_{SAL}^{DI}$ | $MN_{SAL}^{DS}$ | $MN_{SAL}^{DIR}$ | $MN_{SAL}^{DSR}$ | $MN_{SAL}^{MLB}$ | | $MN_{SAO}^{DI}$ | $MN_{SAO}^{DS}$ | $MN_{SAO}^{DIR}$ | $MN_{SAO}^{DSR}$ | $MN_{SAO}^{MLB}$ | | $MN_{SALO}^{DI}$ | $MN_{SALO}^{DS}$ | $MN_{SALO}^{DIR}$ | $MN_{SALO}^{DSR}$ | $MN_{SALO}^{MLB}$ |
|---|---|---|---|---|---|---|---|---|---|---|---|---|---|---|---|---|---|---|---|---|---|---|---|
| $1_{group}^{st}$ | 0.29 | 0.31 | 0.33 | 0.31 | 0.22 | $1_{group}^{st}$ | 0.30 | 0.30 | 0.44 | 0.34 | 0.21 | $1_{group}^{st}$ | 0.33 | 0.43 | 0.43 | 0.33 | 0.08 | $1_{group}^{st}$ | 0.43 | 0.33 | 0.33 | 0.43 | 0.08 |
| $2_{group}^{nd}$ | 0.32 | 0.36 | 0.35 | 0.37 | 0.27 | $2_{group}^{nd}$ | 0.37 | 0.36 | 0.44 | 0.36 | 0.24 | $2_{group}^{nd}$ | 0.29 | 0.44 | 0.44 | 0.40 | 0.31 | $2_{group}^{nd}$ | 0.44 | 0.40 | 0.29 | 0.44 | 0.31 |
| $3_{group}^{rd}$ | 0.40 | 0.44 | 0.42 | 0.44 | 0.36 | $3_{group}^{rd}$ | 0.41 | 0.43 | 0.49 | 0.44 | 0.39 | $3_{group}^{rd}$ | 0.36 | 0.46 | 0.48 | 0.48 | 0.41 | $3_{group}^{rd}$ | 0.46 | 0.48 | 0.36 | 0.48 | 0.41 |
| $4_{group}^{th}$ | 0.39 | 0.41 | 0.43 | 0.41 | 0.46 | $4_{group}^{th}$ | 0.40 | 0.40 | 0.43 | 0.41 | 0.45 | $4_{group}^{th}$ | 0.41 | 0.48 | 0.45 | 0.48 | 0.45 | $4_{group}^{th}$ | 0.48 | 0.48 | 0.41 | 0.45 | 0.45 |
| $5_{group}^{th}$ | 0.82 | 0.83 | 0.83 | 0.83 | 0.85 | $5_{group}^{th}$ | 0.83 | 0.84 | 0.85 | 0.84 | 0.85 | $5_{group}^{th}$ | 0.88 | 0.93 | 0.89 | 0.93 | 0.93 | $5_{group}^{th}$ | 0.93 | 0.93 | 0.88 | 0.89 | 0.93 |

*Note that the 1st group represents the flows within the lowest 20th percentile and the 5th group represents the flows within the highest 20th percentile.

*Note also that $r^{KGE}$ measures the linear correlation between simulation and observation, with a positive 1 for perfect positive linear correlation, a zero (0) for no linear correlation, a negative 1 for negative linear correlation, and a range from -1 to 1. The cases include single-layer Distributed-Input (DI), Distributed-State (DS), Distributed-Input Relaxed (DIR), Distributed-State Relaxed (DSR), and Machine-Learning Benchmark (MLB) networks, with the number of nodes varying from 1 to 5. "None" refers to the case without cell-information sharing among different nodes in the gates, whereas "SAL," "SAO," and "SALO" represent cases where sharing occurs only in the loss gate, only in the output gate, and in both gates, respectively. We report values to two decimal places; therefore, any difference between two metrics that is greater than or equal to 0.01 is noteworthy. PNs=Parameter Numbers

Table S8. Root Mean Squared Error $(RMSE)$ scores for the Single-Layer Mass-Conserving Neural Networks (MNs)

**1-Node**

| | $MN_{none}^{DI}$ | $MN_{none}^{DS}$ | $MN_{none}^{DIR}$ | $MN_{none}^{DSR}$ | $MN_{none}^{MLB}$ | $MN_{SAL}^{DI}$ | $MN_{SAL}^{DS}$ | $MN_{SAL}^{DIR}$ | $MN_{SAL}^{DSR}$ | $MN_{SAL}^{MLB}$ | $MN_{SAO}^{DI}$ | $MN_{SAO}^{DS}$ | $MN_{SAO}^{DIR}$ | $MN_{SAO}^{DSR}$ | $MN_{SAO}^{MLB}$ | $MN_{SALO}^{DI}$ | $MN_{SALO}^{DS}$ | $MN_{SALO}^{DIR}$ | $MN_{SALO}^{DSR}$ | $MN_{SALO}^{MLB}$ |
|---|---|---|---|---|---|---|---|---|---|---|---|---|---|---|---|---|---|---|---|---|
| $RMSE^{min}$ | 2.33 | 2.45 | 2.33 | 2.40 | 2.52 | | | | | | | | | | | | | | | |
| $RMSE^{5\%}$ | 2.25 | 2.25 | 2.25 | 2.22 | 2.24 | | | | | | | | | | | | | | | |
| $RMSE^{25\%}$ | 1.63 | 1.60 | 1.63 | 1.59 | 1.66 | | | | | | | | | | | | | | | |
| $RMSE^{50\%}$ | 1.18 | 1.17 | 1.18 | 1.15 | 1.15 | | | | | | | | | | | | | | | |
| $RMSE^{75\%}$ | 0.94 | 0.97 | 0.94 | 0.95 | 0.99 | | | | | | | | | | | | | | | |
| $RMSE^{95\%}$ | 0.66 | 0.67 | 0.66 | 0.66 | 0.71 | | | | | | | | | | | | | | | |
| PNs | 8 | 8 | 9 | 9 | 11 | | | | | | | | | | | | | | | |

**2-Node**

| | $MN_{none}^{DI}$ | $MN_{none}^{DS}$ | $MN_{none}^{DIR}$ | $MN_{none}^{DSR}$ | $MN_{none}^{MLB}$ | $MN_{SAL}^{DI}$ | $MN_{SAL}^{DS}$ | $MN_{SAL}^{DIR}$ | $MN_{SAL}^{DSR}$ | $MN_{SAL}^{MLB}$ | $MN_{SAO}^{DI}$ | $MN_{SAO}^{DS}$ | $MN_{SAO}^{DIR}$ | $MN_{SAO}^{DSR}$ | $MN_{SAO}^{MLB}$ | $MN_{SALO}^{DI}$ | $MN_{SALO}^{DS}$ | $MN_{SALO}^{DIR}$ | $MN_{SALO}^{DSR}$ | $MN_{SALO}^{MLB}$ |
|---|---|---|---|---|---|---|---|---|---|---|---|---|---|---|---|---|---|---|---|---|
| $RMSE^{min}$ | 2.29 | 2.21 | 2.29 | 2.15 | 2.15 | 2.28 | 2.17 | 2.29 | 2.17 | 2.19 | 2.17 | 2.09 | 2.15 | 2.02 | 2.14 | 2.14 | 2.17 | 2.15 | 2.01 | 2.15 |
| $RMSE^{5\%}$ | 2.20 | 2.18 | 2.14 | 2.13 | 2.10 | 2.20 | 2.16 | 2.14 | 2.15 | 2.11 | 2.10 | 2.00 | 2.12 | 1.99 | 2.11 | 2.06 | 1.96 | 2.08 | 1.98 | 2.09 |
| $RMSE^{25\%}$ | 1.56 | 1.52 | 1.49 | 1.49 | 1.48 | 1.55 | 1.50 | 1.48 | 1.49 | 1.47 | 1.51 | 1.39 | 1.51 | 1.31 | 1.48 | 1.45 | 1.38 | 1.48 | 1.30 | 1.43 |
| $RMSE^{50\%}$ | 1.18 | 1.13 | 1.09 | 1.13 | 1.06 | 1.17 | 1.14 | 1.07 | 1.14 | 1.04 | 1.08 | 1.08 | 1.03 | 1.05 | 1.03 | 1.04 | 1.08 | 1.00 | 1.05 | 1.00 |
| $RMSE^{75\%}$ | 0.86 | 0.86 | 0.86 | 0.87 | 0.81 | 0.87 | 0.87 | 0.85 | 0.87 | 0.81 | 0.82 | 0.88 | 0.84 | 0.87 | 0.83 | 0.81 | 0.87 | 0.82 | 0.85 | 0.77 |
| $RMSE^{95\%}$ | 0.57 | 0.52 | 0.55 | 0.54 | 0.55 | 0.58 | 0.52 | 0.53 | 0.55 | 0.54 | 0.53 | 0.62 | 0.56 | 0.56 | 0.51 | 0.57 | 0.60 | 0.56 | 0.56 | 0.52 |
| PNs | 18 | 18 | 18 | 18 | 21 | 20 | 20 | 20 | 20 | 23 | 20 | 20 | 20 | 20 | 23 | 22 | 22 | 22 | 22 | 25 |

**3-Node**

| | $MN_{none}^{DI}$ | $MN_{none}^{DS}$ | $MN_{none}^{DIR}$ | $MN_{none}^{DSR}$ | $MN_{none}^{MLB}$ | $MN_{SAL}^{DI}$ | $MN_{SAL}^{DS}$ | $MN_{SAL}^{DIR}$ | $MN_{SAL}^{DSR}$ | $MN_{SAL}^{MLB}$ | $MN_{SAO}^{DI}$ | $MN_{SAO}^{DS}$ | $MN_{SAO}^{DIR}$ | $MN_{SAO}^{DSR}$ | $MN_{SAO}^{MLB}$ | $MN_{SALO}^{DI}$ | $MN_{SALO}^{DS}$ | $MN_{SALO}^{DIR}$ | $MN_{SALO}^{DSR}$ | $MN_{SALO}^{MLB}$ |
|---|---|---|---|---|---|---|---|---|---|---|---|---|---|---|---|---|---|---|---|---|
| $RMSE^{min}$ | 2.30 | 2.17 | 2.29 | 2.17 | 2.15 | 2.29 | 2.19 | 2.25 | 2.16 | 2.17 | 1.95 | 1.89 | 2.08 | 1.82 | 1.87 | 2.10 | 2.01 | 2.07 | 1.89 | 1.84 |
| $RMSE^{5\%}$ | 2.20 | 2.15 | 2.14 | 2.15 | 2.09 | 2.20 | 2.14 | 2.15 | 2.16 | 2.09 | 1.89 | 1.57 | 2.03 | 1.65 | 1.68 | 1.85 | 1.87 | 1.94 | 1.71 | 1.53 |
| $RMSE^{25\%}$ | 1.56 | 1.50 | 1.49 | 1.48 | 1.46 | 1.55 | 1.50 | 1.50 | 1.48 | 1.46 | 1.36 | 1.14 | 1.44 | 1.03 | 1.10 | 1.27 | 1.20 | 1.41 | 1.21 | 1.04 |
| $RMSE^{50\%}$ | 1.18 | 1.15 | 1.09 | 1.12 | 1.06 | 1.17 | 1.13 | 1.08 | 1.13 | 1.04 | 1.02 | 0.79 | 1.01 | 0.81 | 0.97 | 0.94 | 0.93 | 0.94 | 0.95 | 0.84 |
| $RMSE^{75\%}$ | 0.86 | 0.89 | 0.86 | 0.86 | 0.83 | 0.87 | 0.81 | 0.87 | 0.85 | 0.81 | 0.82 | 0.65 | 0.83 | 0.67 | 0.77 | 0.77 | 0.71 | 0.76 | 0.79 | 0.70 |
| $RMSE^{95\%}$ | 0.57 | 0.52 | 0.55 | 0.54 | 0.53 | 0.59 | 0.52 | 0.57 | 0.56 | 0.53 | 0.58 | 0.48 | 0.55 | 0.49 | 0.53 | 0.53 | 0.52 | 0.54 | 0.56 | 0.50 |
| PNs | 27 | 27 | 27 | 27 | 31 | 33 | 33 | 33 | 33 | 37 | 33 | 33 | 33 | 33 | 37 | 39 | 39 | 39 | 39 | 43 |

**4-Node**

| | $MN_{none}^{DI}$ | $MN_{none}^{DS}$ | $MN_{none}^{DIR}$ | $MN_{none}^{DSR}$ | $MN_{none}^{MLB}$ | $MN_{SAL}^{DI}$ | $MN_{SAL}^{DS}$ | $MN_{SAL}^{DIR}$ | $MN_{SAL}^{DSR}$ | $MN_{SAL}^{MLB}$ | $MN_{SAO}^{DI}$ | $MN_{SAO}^{DS}$ | $MN_{SAO}^{DIR}$ | $MN_{SAO}^{DSR}$ | $MN_{SAO}^{MLB}$ | $MN_{SALO}^{DI}$ | $MN_{SALO}^{DS}$ | $MN_{SALO}^{DIR}$ | $MN_{SALO}^{DSR}$ | $MN_{SALO}^{MLB}$ |
|---|---|---|---|---|---|---|---|---|---|---|---|---|---|---|---|---|---|---|---|---|
| $RMSE^{min}$ | 2.20 | 2.16 | 2.29 | 2.16 | 2.26 | 2.28 | 2.17 | 2.27 | 2.25 | 2.09 | 1.94 | 1.69 | 1.88 | 1.82 | 1.69 | 2.04 | 1.84 | 2.04 | 1.71 | 1.78 |
| $RMSE^{5\%}$ | 2.19 | 2.16 | 2.14 | 2.14 | 2.10 | 2.20 | 2.12 | 2.15 | 2.14 | 2.03 | 1.85 | 1.54 | 1.86 | 1.51 | 1.50 | 1.97 | 1.57 | 1.97 | 1.56 | 1.69 |
| $RMSE^{25\%}$ | 1.58 | 1.50 | 1.49 | 1.48 | 1.54 | 1.56 | 1.50 | 1.50 | 1.47 | 1.46 | 1.42 | 0.95 | 1.26 | 0.91 | 0.98 | 1.46 | 1.06 | 1.46 | 1.01 | 1.14 |
| $RMSE^{50\%}$ | 1.24 | 1.13 | 1.09 | 1.12 | 1.03 | 1.15 | 1.12 | 1.07 | 1.13 | 1.03 | 1.00 | 0.75 | 0.97 | 0.75 | 0.83 | 0.90 | 0.85 | 0.90 | 0.76 | 0.94 |
| $RMSE^{75\%}$ | 0.93 | 0.87 | 0.85 | 0.87 | 0.80 | 0.87 | 0.79 | 0.86 | 0.82 | 0.79 | 0.81 | 0.58 | 0.78 | 0.59 | 0.62 | 0.79 | 0.65 | 0.79 | 0.57 | 0.73 |
| $RMSE^{95\%}$ | 0.64 | 0.52 | 0.55 | 0.54 | 0.53 | 0.56 | 0.54 | 0.55 | 0.55 | 0.50 | 0.59 | 0.45 | 0.56 | 0.47 | 0.44 | 0.56 | 0.53 | 0.56 | 0.41 | 0.53 |
| PNs | 36 | 36 | 36 | 36 | 41 | 48 | 48 | 48 | 48 | 53 | 48 | 48 | 48 | 48 | 53 | 60 | 60 | 60 | 60 | 65 |

**5-Node**

| | $MN_{none}^{DI}$ | $MN_{none}^{DS}$ | $MN_{none}^{DIR}$ | $MN_{none}^{DSR}$ | $MN_{none}^{MLB}$ | $MN_{SAL}^{DI}$ | $MN_{SAL}^{DS}$ | $MN_{SAL}^{DIR}$ | $MN_{SAL}^{DSR}$ | $MN_{SAL}^{MLB}$ | $MN_{SAO}^{DI}$ | $MN_{SAO}^{DS}$ | $MN_{SAO}^{DIR}$ | $MN_{SAO}^{DSR}$ | $MN_{SAO}^{MLB}$ | $MN_{SALO}^{DI}$ | $MN_{SALO}^{DS}$ | $MN_{SALO}^{DIR}$ | $MN_{SALO}^{DSR}$ | $MN_{SALO}^{MLB}$ |
|---|---|---|---|---|---|---|---|---|---|---|---|---|---|---|---|---|---|---|---|---|
| $RMSE^{min}$ | 2.29 | 2.17 | 2.37 | 2.15 | 2.08 | 2.29 | 2.19 | 2.38 | 2.19 | 2.16 | 1.93 | 1.61 | 2.26 | 1.60 | 1.67 | 2.25 | 1.55 | 2.01 | 1.66 | 1.67 |
| $RMSE^{5\%}$ | 2.20 | 2.16 | 2.10 | 2.14 | 2.02 | 2.20 | 2.14 | 2.04 | 2.15 | 2.05 | 1.89 | 1.51 | 1.80 | 1.49 | 1.52 | 1.88 | 1.42 | 1.93 | 1.40 | 1.52 |
| $RMSE^{25\%}$ | 1.56 | 1.49 | 1.49 | 1.48 | 1.43 | 1.60 | 1.49 | 1.45 | 1.48 | 1.50 | 1.39 | 0.98 | 1.23 | 0.96 | 1.00 | 1.20 | 1.01 | 1.28 | 0.95 | 1.02 |
| $RMSE^{50\%}$ | 1.18 | 1.13 | 1.07 | 1.11 | 1.04 | 1.13 | 1.13 | 1.01 | 1.11 | 1.07 | 1.01 | 0.80 | 0.96 | 0.77 | 0.80 | 0.97 | 0.72 | 0.87 | 0.74 | 0.77 |
| $RMSE^{75\%}$ | 0.86 | 0.87 | 0.84 | 0.86 | 0.81 | 0.88 | 0.79 | 0.78 | 0.81 | 0.79 | 0.81 | 0.61 | 0.76 | 0.58 | 0.63 | 0.80 | 0.54 | 0.71 | 0.56 | 0.63 |
| $RMSE^{95\%}$ | 0.57 | 0.52 | 0.55 | 0.54 | 0.53 | 0.56 | 0.51 | 0.53 | 0.52 | 0.53 | 0.58 | 0.40 | 0.55 | 0.41 | 0.43 | 0.51 | 0.37 | 0.49 | 0.41 | 0.45 |
| PNs | 45 | 45 | 45 | 45 | 51 | 65 | 65 | 65 | 65 | 71 | 65 | 65 | 65 | 65 | 71 | 85 | 85 | 85 | 85 | 91 |

*Note that the cases include single-layer Distributed-Input (DI), Distributed-State (DS), Distributed-Input Relaxed (DIR), Distributed-State Relaxed (DSR), and Machine-Learning Benchmark (MLB) networks, with the number of nodes varying from 1 to 5. "None" refers to the case without cell-information sharing among different nodes in the gates, whereas "SAL," "SAO," and "SALO" represent cases where sharing occurs only in the loss gate, only in the output gate, and in both gates, respectively. We report values to two decimal places; therefore, any difference between two metrics that is greater than or equal to 0.01 is noteworthy. PNs=Parameter Numbers

Table S9. Mean Absolute Error ($MAE$) scores for the Single-Layer Mass-Conserving Neural Networks (MNs)

**1-Node**

| | $MN^{DI}_{none}$ | $MN^{DS}_{none}$ | $MN^{DIR}_{none}$ | $MN^{DSR}_{none}$ | $MN^{MLB}_{none}$ | | $MN^{DI}_{SAL}$ | $MN^{DS}_{SAL}$ | $MN^{DIR}_{SAL}$ | $MN^{DSR}_{SAL}$ | $MN^{MLB}_{SAL}$ | | $MN^{DI}_{SAO}$ | $MN^{DS}_{SAO}$ | $MN^{DIR}_{SAO}$ | $MN^{DSR}_{SAO}$ | $MN^{MLB}_{SAO}$ | | $MN^{DI}_{SALO}$ | $MN^{DS}_{SALO}$ | $MN^{DIR}_{SALO}$ | $MN^{DSR}_{SALO}$ | $MN^{MLB}_{SALO}$ |
|---|---|---|---|---|---|---|---|---|---|---|---|---|---|---|---|---|---|---|---|---|---|---|---|
| $MAE^{min}$ | 1.08 | 1.11 | 1.08 | 1.08 | 1.08 | $MAE^{min}$ | | | | | | $MAE^{min}$ | | | | | | $MAE^{min}$ | | | | | |
| $MAE^{5\%}$ | 1.07 | 1.08 | 1.07 | 1.07 | 1.07 | $MAE^{5\%}$ | | | | | | $MAE^{5\%}$ | | | | | | $MAE^{5\%}$ | | | | | |
| $MAE^{25\%}$ | 0.74 | 0.76 | 0.74 | 0.75 | 0.76 | $MAE^{25\%}$ | | | | | | $MAE^{25\%}$ | | | | | | $MAE^{25\%}$ | | | | | |
| $MAE^{50\%}$ | 0.58 | 0.59 | 0.58 | 0.59 | 0.59 | $MAE^{50\%}$ | | | | | | $MAE^{50\%}$ | | | | | | $MAE^{50\%}$ | | | | | |
| $MAE^{75\%}$ | 0.46 | 0.47 | 0.46 | 0.46 | 0.49 | $MAE^{75\%}$ | | | | | | $MAE^{75\%}$ | | | | | | $MAE^{75\%}$ | | | | | |
| $MAE^{95\%}$ | 0.36 | 0.35 | 0.36 | 0.35 | 0.38 | $MAE^{95\%}$ | | | | | | $MAE^{95\%}$ | | | | | | $MAE^{95\%}$ | | | | | |
| PNs | 8 | 8 | 9 | 9 | 11 | PNs | | | | | | PNs | | | | | | PNs | | | | | |

**2-Node**

| | $MN^{DI}_{none}$ | $MN^{DS}_{none}$ | $MN^{DIR}_{none}$ | $MN^{DSR}_{none}$ | $MN^{MLB}_{none}$ | | $MN^{DI}_{SAL}$ | $MN^{DS}_{SAL}$ | $MN^{DIR}_{SAL}$ | $MN^{DSR}_{SAL}$ | $MN^{MLB}_{SAL}$ | | $MN^{DI}_{SAO}$ | $MN^{DS}_{SAO}$ | $MN^{DIR}_{SAO}$ | $MN^{DSR}_{SAO}$ | $MN^{MLB}_{SAO}$ | | $MN^{DI}_{SALO}$ | $MN^{DS}_{SALO}$ | $MN^{DIR}_{SALO}$ | $MN^{DSR}_{SALO}$ | $MN^{MLB}_{SALO}$ |
|---|---|---|---|---|---|---|---|---|---|---|---|---|---|---|---|---|---|---|---|---|---|---|---|
| $MAE^{min}$ | 1.04 | 1.08 | 1.09 | 1.12 | 1.10 | $MAE^{min}$ | 1.03 | 1.07 | 1.05 | 1.11 | 1.10 | $MAE^{min}$ | 1.01 | 1.09 | 1.06 | 1.06 | 1.09 | $MAE^{min}$ | 1.02 | 1.07 | 1.06 | 1.05 | 1.07 |
| $MAE^{5\%}$ | 1.04 | 1.04 | 1.07 | 1.07 | 1.05 | $MAE^{5\%}$ | 1.02 | 1.04 | 1.03 | 1.07 | 1.03 | $MAE^{5\%}$ | 0.98 | 1.02 | 1.03 | 0.99 | 1.05 | $MAE^{5\%}$ | 0.96 | 1.01 | 0.99 | 0.97 | 0.99 |
| $MAE^{25\%}$ | 0.74 | 0.71 | 0.73 | 0.74 | 0.77 | $MAE^{25\%}$ | 0.72 | 0.73 | 0.72 | 0.74 | 0.77 | $MAE^{25\%}$ | 0.71 | 0.70 | 0.73 | 0.71 | 0.74 | $MAE^{25\%}$ | 0.68 | 0.70 | 0.72 | 0.70 | 0.73 |
| $MAE^{50\%}$ | 0.55 | 0.52 | 0.52 | 0.53 | 0.55 | $MAE^{50\%}$ | 0.55 | 0.52 | 0.50 | 0.52 | 0.53 | $MAE^{50\%}$ | 0.50 | 0.54 | 0.49 | 0.53 | 0.50 | $MAE^{50\%}$ | 0.50 | 0.54 | 0.50 | 0.52 | 0.51 |
| $MAE^{75\%}$ | 0.43 | 0.40 | 0.40 | 0.41 | 0.43 | $MAE^{75\%}$ | 0.44 | 0.41 | 0.37 | 0.40 | 0.42 | $MAE^{75\%}$ | 0.41 | 0.42 | 0.40 | 0.43 | 0.42 | $MAE^{75\%}$ | 0.42 | 0.43 | 0.40 | 0.41 | 0.40 |
| $MAE^{95\%}$ | 0.31 | 0.26 | 0.30 | 0.27 | 0.30 | $MAE^{95\%}$ | 0.30 | 0.27 | 0.27 | 0.29 | 0.30 | $MAE^{95\%}$ | 0.29 | 0.29 | 0.30 | 0.29 | 0.27 | $MAE^{95\%}$ | 0.31 | 0.29 | 0.30 | 0.29 | 0.29 |
| PNs | 18 | 18 | 18 | 18 | 21 | PNs | 20 | 20 | 20 | 20 | 23 | PNs | 20 | 20 | 20 | 20 | 23 | PNs | 22 | 22 | 22 | 22 | 25 |

**3-Node**

| | $MN^{DI}_{none}$ | $MN^{DS}_{none}$ | $MN^{DIR}_{none}$ | $MN^{DSR}_{none}$ | $MN^{MLB}_{none}$ | | $MN^{DI}_{SAL}$ | $MN^{DS}_{SAL}$ | $MN^{DIR}_{SAL}$ | $MN^{DSR}_{SAL}$ | $MN^{MLB}_{SAL}$ | | $MN^{DI}_{SAO}$ | $MN^{DS}_{SAO}$ | $MN^{DIR}_{SAO}$ | $MN^{DSR}_{SAO}$ | $MN^{MLB}_{SAO}$ | | $MN^{DI}_{SALO}$ | $MN^{DS}_{SALO}$ | $MN^{DIR}_{SALO}$ | $MN^{DSR}_{SALO}$ | $MN^{MLB}_{SALO}$ |
|---|---|---|---|---|---|---|---|---|---|---|---|---|---|---|---|---|---|---|---|---|---|---|---|
| $MAE^{min}$ | 1.04 | 1.08 | 1.09 | 1.11 | 1.10 | $MAE^{min}$ | 1.03 | 1.07 | 1.12 | 1.10 | 1.09 | $MAE^{min}$ | 1.00 | 0.90 | 1.03 | 0.88 | 0.86 | $MAE^{min}$ | 0.95 | 0.97 | 1.00 | 0.93 | 0.89 |
| $MAE^{5\%}$ | 1.04 | 1.04 | 1.07 | 1.06 | 1.02 | $MAE^{5\%}$ | 1.02 | 1.03 | 1.06 | 1.06 | 1.02 | $MAE^{5\%}$ | 0.93 | 0.83 | 0.99 | 0.81 | 0.81 | $MAE^{5\%}$ | 0.87 | 0.92 | 0.94 | 0.85 | 0.78 |
| $MAE^{25\%}$ | 0.74 | 0.74 | 0.73 | 0.74 | 0.77 | $MAE^{25\%}$ | 0.72 | 0.70 | 0.73 | 0.72 | 0.76 | $MAE^{25\%}$ | 0.61 | 0.51 | 0.70 | 0.54 | 0.54 | $MAE^{25\%}$ | 0.59 | 0.58 | 0.66 | 0.59 | 0.52 |
| $MAE^{50\%}$ | 0.55 | 0.52 | 0.52 | 0.51 | 0.52 | $MAE^{50\%}$ | 0.55 | 0.52 | 0.51 | 0.51 | 0.52 | $MAE^{50\%}$ | 0.48 | 0.38 | 0.50 | 0.41 | 0.47 | $MAE^{50\%}$ | 0.42 | 0.43 | 0.47 | 0.49 | 0.44 |
| $MAE^{75\%}$ | 0.43 | 0.41 | 0.40 | 0.40 | 0.43 | $MAE^{75\%}$ | 0.44 | 0.37 | 0.39 | 0.40 | 0.42 | $MAE^{75\%}$ | 0.38 | 0.32 | 0.40 | 0.33 | 0.39 | $MAE^{75\%}$ | 0.35 | 0.35 | 0.39 | 0.40 | 0.37 |
| $MAE^{95\%}$ | 0.30 | 0.28 | 0.30 | 0.27 | 0.29 | $MAE^{95\%}$ | 0.30 | 0.27 | 0.29 | 0.29 | 0.29 | $MAE^{95\%}$ | 0.29 | 0.23 | 0.29 | 0.22 | 0.28 | $MAE^{95\%}$ | 0.26 | 0.25 | 0.27 | 0.28 | 0.27 |
| PNs | 27 | 27 | 27 | 27 | 31 | PNs | 33 | 33 | 33 | 33 | 37 | PNs | 33 | 33 | 33 | 33 | 37 | PNs | 39 | 39 | 39 | 39 | 43 |

**4-Node**

| | $MN^{DI}_{none}$ | $MN^{DS}_{none}$ | $MN^{DIR}_{none}$ | $MN^{DSR}_{none}$ | $MN^{MLB}_{none}$ | | $MN^{DI}_{SAL}$ | $MN^{DS}_{SAL}$ | $MN^{DIR}_{SAL}$ | $MN^{DSR}_{SAL}$ | $MN^{MLB}_{SAL}$ | | $MN^{DI}_{SAO}$ | $MN^{DS}_{SAO}$ | $MN^{DIR}_{SAO}$ | $MN^{DSR}_{SAO}$ | $MN^{MLB}_{SAO}$ | | $MN^{DI}_{SALO}$ | $MN^{DS}_{SALO}$ | $MN^{DIR}_{SALO}$ | $MN^{DSR}_{SALO}$ | $MN^{MLB}_{SALO}$ |
|---|---|---|---|---|---|---|---|---|---|---|---|---|---|---|---|---|---|---|---|---|---|---|---|
| $MAE^{min}$ | 1.09 | 1.07 | 1.09 | 1.11 | 1.09 | $MAE^{min}$ | 1.05 | 1.06 | 1.11 | 1.11 | 1.05 | $MAE^{min}$ | 0.97 | 0.81 | 0.93 | 0.80 | 0.81 | $MAE^{min}$ | 0.96 | 0.87 | 0.96 | 0.78 | 0.85 |
| $MAE^{5\%}$ | 1.07 | 1.03 | 1.07 | 1.06 | 1.03 | $MAE^{5\%}$ | 1.02 | 1.03 | 1.06 | 1.06 | 0.99 | $MAE^{5\%}$ | 0.90 | 0.75 | 0.88 | 0.72 | 0.72 | $MAE^{5\%}$ | 0.94 | 0.80 | 0.94 | 0.67 | 0.82 |
| $MAE^{25\%}$ | 0.77 | 0.73 | 0.73 | 0.74 | 0.76 | $MAE^{25\%}$ | 0.71 | 0.72 | 0.72 | 0.71 | 0.71 | $MAE^{25\%}$ | 0.59 | 0.46 | 0.57 | 0.44 | 0.52 | $MAE^{25\%}$ | 0.70 | 0.52 | 0.70 | 0.48 | 0.54 |
| $MAE^{50\%}$ | 0.59 | 0.51 | 0.52 | 0.51 | 0.51 | $MAE^{50\%}$ | 0.54 | 0.52 | 0.51 | 0.51 | 0.51 | $MAE^{50\%}$ | 0.47 | 0.35 | 0.42 | 0.37 | 0.39 | $MAE^{50\%}$ | 0.48 | 0.43 | 0.48 | 0.35 | 0.43 |
| $MAE^{75\%}$ | 0.46 | 0.39 | 0.40 | 0.39 | 0.43 | $MAE^{75\%}$ | 0.44 | 0.38 | 0.39 | 0.38 | 0.39 | $MAE^{75\%}$ | 0.38 | 0.30 | 0.34 | 0.30 | 0.33 | $MAE^{75\%}$ | 0.38 | 0.33 | 0.38 | 0.29 | 0.37 |
| $MAE^{95\%}$ | 0.35 | 0.26 | 0.30 | 0.27 | 0.29 | $MAE^{95\%}$ | 0.30 | 0.28 | 0.28 | 0.29 | 0.28 | $MAE^{95\%}$ | 0.29 | 0.22 | 0.26 | 0.24 | 0.25 | $MAE^{95\%}$ | 0.29 | 0.28 | 0.29 | 0.20 | 0.28 |
| PNs | 36 | 36 | 36 | 36 | 41 | PNs | 48 | 48 | 48 | 48 | 53 | PNs | 48 | 48 | 48 | 48 | 53 | PNs | 60 | 60 | 60 | 60 | 65 |

**5-Node**

| | $MN^{DI}_{none}$ | $MN^{DS}_{none}$ | $MN^{DIR}_{none}$ | $MN^{DSR}_{none}$ | $MN^{MLB}_{none}$ | | $MN^{DI}_{SAL}$ | $MN^{DS}_{SAL}$ | $MN^{DIR}_{SAL}$ | $MN^{DSR}_{SAL}$ | $MN^{MLB}_{SAL}$ | | $MN^{DI}_{SAO}$ | $MN^{DS}_{SAO}$ | $MN^{DIR}_{SAO}$ | $MN^{DSR}_{SAO}$ | $MN^{MLB}_{SAO}$ | | $MN^{DI}_{SALO}$ | $MN^{DS}_{SALO}$ | $MN^{DIR}_{SALO}$ | $MN^{DSR}_{SALO}$ | $MN^{MLB}_{SALO}$ |
|---|---|---|---|---|---|---|---|---|---|---|---|---|---|---|---|---|---|---|---|---|---|---|---|
| $MAE^{min}$ | 1.04 | 1.07 | 1.17 | 1.11 | 1.02 | $MAE^{min}$ | 1.09 | 1.07 | 1.01 | 1.10 | 1.09 | $MAE^{min}$ | 0.98 | 0.81 | 0.94 | 0.74 | 0.85 | $MAE^{min}$ | 0.93 | 0.73 | 0.91 | 0.78 | 0.83 |
| $MAE^{5\%}$ | 1.03 | 1.03 | 1.05 | 1.06 | 0.96 | $MAE^{5\%}$ | 1.04 | 1.03 | 0.98 | 1.06 | 1.01 | $MAE^{5\%}$ | 0.90 | 0.77 | 0.87 | 0.67 | 0.73 | $MAE^{5\%}$ | 0.90 | 0.67 | 0.88 | 0.64 | 0.77 |
| $MAE^{25\%}$ | 0.73 | 0.73 | 0.69 | 0.74 | 0.71 | $MAE^{25\%}$ | 0.71 | 0.70 | 0.65 | 0.72 | 0.79 | $MAE^{25\%}$ | 0.60 | 0.48 | 0.55 | 0.46 | 0.50 | $MAE^{25\%}$ | 0.62 | 0.43 | 0.58 | 0.44 | 0.51 |
| $MAE^{50\%}$ | 0.55 | 0.51 | 0.52 | 0.51 | 0.50 | $MAE^{50\%}$ | 0.54 | 0.52 | 0.47 | 0.52 | 0.55 | $MAE^{50\%}$ | 0.47 | 0.38 | 0.42 | 0.36 | 0.42 | $MAE^{50\%}$ | 0.44 | 0.34 | 0.43 | 0.35 | 0.40 |
| $MAE^{75\%}$ | 0.43 | 0.39 | 0.41 | 0.39 | 0.41 | $MAE^{75\%}$ | 0.42 | 0.37 | 0.35 | 0.38 | 0.42 | $MAE^{75\%}$ | 0.37 | 0.31 | 0.34 | 0.29 | 0.36 | $MAE^{75\%}$ | 0.38 | 0.27 | 0.32 | 0.29 | 0.31 |
| $MAE^{95\%}$ | 0.30 | 0.27 | 0.29 | 0.27 | 0.28 | $MAE^{95\%}$ | 0.28 | 0.27 | 0.27 | 0.27 | 0.30 | $MAE^{95\%}$ | 0.28 | 0.22 | 0.25 | 0.20 | 0.26 | $MAE^{95\%}$ | 0.25 | 0.18 | 0.25 | 0.21 | 0.23 |
| PNs | 45 | 45 | 45 | 45 | 51 | PNs | 65 | 65 | 65 | 65 | 71 | PNs | 65 | 65 | 65 | 65 | 71 | PNs | 85 | 85 | 85 | 85 | 91 |

*Note that the cases include single-layer Distributed-Input (DI), Distributed-State (DS), Distributed-Input Relaxed (DIR), Distributed-State Relaxed (DSR), and Machine-Learning Benchmark (MLB) networks, with the number of nodes varying from 1 to 5. “None” refers to the case without cell-information sharing among different nodes in the gates, whereas “SAL,” “SAO,” and “SALO” represent cases where sharing occurs only in the loss gate, only in the output gate, and in both gates, respectively. We report values to two decimal places; therefore, any difference between two metrics that is greater than or equal to 0.01 is noteworthy. PNs=Parameter Numbers

Table S10. Summary of Multi-Layer Mass-Conserving Neural Network ($MN_s$) architectural hypothesis tested in this study

| Model Name | Input Distribution Gate | Linear Transformation Layer | Information (Internal State) Sharing | Node Number in Each Layer |
|---|---|---|---|---|
| $MN_{None}^{DS}(N_1, N_2, N_3)$ | Unity | Weight sum to 1.0 | No | $(1 \leq N_1 \leq 3; 0 \leq N_2 \leq 3; 0 \leq N_3 \leq 3$ |
| $MN_{None}^{DSR}(N_1, N_2, N_3)$ | Unity | Positive Weight | No | $1 \leq N_1 \leq 3; 0 \leq N_2 \leq 3; 0 \leq N_3 \leq 3$ |
| $MN_{None}^{MLB}(N_1, N_2, N_3)$ | Unity | Real values weight & bias | No | $1 \leq N_1 \leq 3; 0 \leq N_2 \leq 3; 0 \leq N_3 \leq 3$ |
| $MN_{SALO}^{DS}(N_1, N_2, N_3)$ | Unity | Weight sum to 1.0 | Yes | $1 \leq N_1 \leq 3; 0 \leq N_2 \leq 3; 0 \leq N_3 \leq 3$ |
| $MN_{SALO}^{DSR}(N_1, N_2, N_3)$ | Unity | Positive Weight | Yes | $1 \leq N_1 \leq 3; 0 \leq N_2 \leq 3; 0 \leq N_3 \leq 3$ |
| $MN_{SALO}^{MLB}(N_1, N_2, N_3)$ | Unity | Real values weight & bias | Yes | $1 \leq N_1 \leq 3; 0 \leq N_2 \leq 3; 0 \leq N_3 \leq 3$ |
| $MN_{None}^{DS}(N_1, 0, 0)$ | Unity | Weight sum to 1.0 | No | $4 \leq N_1 \leq 5$ |
| $MN_{None}^{DSR}(N_1, 0, 0)$ | Unity | Positive Weight | No | $4 \leq N_1 \leq 5$ |
| $MN_{None}^{MLB}(N_1, 0, 0)$ | Unity | Real values weight & bias | No | $4 \leq N_1 \leq 5$ |
| $MN_{SALO}^{DS}(N_1, 0, 0)$ | Unity | Weight sum to 1.0 | Yes | $4 \leq N_1 \leq 5$ |
| $MN_{SALO}^{DSR}(N_1, 0, 0)$ | Unity | Positive Weight | Yes | $4 \leq N_1 \leq 5$ |
| $MN_{SALO}^{DS-MLB}(N_1, 0, 0)$ | Unity | Real values weight & bias | Yes | $4 \leq N_1 \leq 5$ |

Table S11. Detailed information on the training setup for single-layer cases

| Model Name | Model Inheritance | Parameter Inheritance and Initialization Approach | Epoch |
|---|---|---|---|
| $\boldsymbol{MN_{None}^{DI}(N_1)}$ | $\boldsymbol{MC\{O_\sigma L_{\sigma+}^{con}\}}$ | Each node is initialized based on the inheritance model with varying noise added to each parameter (see Section 3.7). The linear output weights are randomly initialized with values close to $\boldsymbol{1/N_1}$ and the bias is randomly initialized with small positive values close to zero. | 3000 |
| $\boldsymbol{MN_{None}^{DS}(N_1)}$ | | | 3000 |
| $\boldsymbol{MN_{None}^{DIR}(N_1)}$ | | | 3000 |
| $\boldsymbol{MN_{None}^{DSR}(N_1)}$ | | | 3000 |
| $\boldsymbol{MN_{None}^{MLB}(N_1)}$ | | | 5000 |
| $\boldsymbol{MN_{SAL}^{DI}(N_1)}$ | $\boldsymbol{MC\{O_\sigma L_{\sigma+}^{con}\}}$ | Each node is initialized based on the inheritance model, with varying noise added to each parameter (see Section 3.7). Note that in the cell-state information sharing cases, only the weights in the gate corresponding to the node itself (i.e., the diagonal of the matrix) are initialized in the same way as in the non-sharing case; the remaining gate weights are initialized with values very close to zero. The linear output weights are randomly initialized with values close to $1/\boldsymbol{N_1}$ and the bias is randomly initialized with small positive values close to zero. | 3000 |
| $\boldsymbol{MN_{SAL}^{DS}(N_1)}$ | | | 3000 |
| $\boldsymbol{MN_{SAL}^{DIR}(N_1)}$ | | | 3000 |
| $\boldsymbol{MN_{SAL}^{DSR}(N_1)}$ | | | 3000 |
| $\boldsymbol{MN_{SAL}^{MLB}(N_1)}$ | | | 5000 |
| $\boldsymbol{MN_{SAO}^{DI}(N_1)}$ | | | 3000 |
| $\boldsymbol{MN_{SAO}^{DS}(N_1)}$ | | | 3000 |
| $\boldsymbol{MN_{SAO}^{DIR}(N_1)}$ | | | 3000 |
| $\boldsymbol{MN_{SAO}^{DSR}(N_1)}$ | | | 3000 |
| $\boldsymbol{MN_{SAO}^{MLB}(N_1)}$ | | | 5000 |
| $\boldsymbol{MN_{SALO}^{DI}(N_1)}$ | | | 3000 |
| $\boldsymbol{MN_{SALO}^{DS}(N_1)}$ | | | 3000 |
| $\boldsymbol{MN_{SALO}^{DIR}(N_1)}$ | | | 3000 |
| $\boldsymbol{MN_{SALO}^{DSR}(N_1)}$ | | | 3000 |
| $\boldsymbol{MN_{SALO}^{MLB}(N_1)}$ | | | 5000 |

*The cases include single-layer Distributed-Input (DI), Distributed-State (DS), Distributed-Input Relaxed (DIR), Distributed-State Relaxed (DSR), and Machine-Learning Benchmark (MLB) networks, with the number of nodes varying from 1 to 5. “None” refers to the case without cell-information sharing among different nodes in the gates, whereas “SAL,” “SAO,” and “SALO” represent cases where sharing occurs only in the loss gate, only in the output gate, and in both gates, respectively. In this study, the number of nodes $\boldsymbol{N_1}$ is tested from 1 to 5 and is trained with 10 random seeds number include 2925, 9998, 2025, 2525, 3410, 9899, 5555, 2520, 2828, and 3140. Notice that the model architectures of $MN_{None}^{DI}(1)$, $MN_{SAL}^{DI}(1)$, $MN_{SAO}^{DI}(1)$, and $MN_{SALO}^{DI}(1)$ are equivalent since cell-state information sharing is not possible. The same applies to $MN_{None}^{DS}(1)$, $MN_{SAL}^{DS}(1)$, $MN_{SAO}^{DS}(1)$, and $MN_{SALO}^{DS}(1)$ . In fact, all eight models are architecturally equivalent, since the input-distributed gate for DI is set to 1, and the linear output weights are also identical. This means there is no functional distinction among these models in this case. On the contrary, although there is no architectural distinction among the single-node cases for DI, DSR, and MLB, these models are not functionally equivalent. Each introduces relaxation at different locations: DI is relaxed at the input-distributed layer, DSR at the output layer, and MLB at both layers.

Table S12. Detailed information on the training setup for multi-layer cases

| Model Name | Model Inheritance | Epoch | Parameter Inheritance and Initialization Approach |
|---|---|---|---|
| $MN_S^M(1,1,0)$ | $MN_S^M(1,0,0)$ | 2000 | In the model $MN_S^M(N_1, N_2, N_3)$, the superscript "M" denotes the model type, which can be "distributed-state" (DS), "distributed-state-relaxed" (DSR), or "machine-learning benchmark" (MLB). The subscript "S" indicates the sharing configuration, which can be either "none" (no sharing) or "SALO" (cell-state sharing in both the loss and output gates).<br><br>The model is only initialized from another model with the same model type and sharing configuration. For instance, the parameters of $MN_{None}^{DS}(1,2,0)$ are inherited from $MN_{None}^{DS}(1,1,0)$, and $MN_{SALO}^{MLB}(3,2,2)$ are inherited from $MN_{SALO}^{MLB}(3,2,1)$.<br><br>The parameters of each specific model (listed in the **Model Name** column) are inherited from its corresponding parent model (as indicated in the **Model Inheritance** column). The remaining newly added parameters are randomly initialized within the range of -1 to 1.<br><br>Note that in the DS case, no linear transformation layers are inserted between any two adjacent layers in order to ensure mass conservation. Conversely, linear transformation layers are present in both the DSR and MLB cases to adjust the total mass between layers. These transformation layers are initialized using parameters from the parent model, with newly added parameters initialized as follows: for DSR, only the weights are initialized with positive values; for MLB, both weights and biases are initialized with arbitrary values.<br><br>A linear output layer is included in all three cases (DS, DSR, and MLB) when the number of nodes in the last layer is greater than one. However, when the last layer contains only one node, the linear output layer is omitted in the DS case, whereas it remains in use for the DSR and MLB cases to allow for mass adjustment. The newly added components are randomly initialized in a manner similar to that of the linear transformation layers |
| $MN_S^M(1,2,0)$ | $MN_S^M(1,1,0)$ | 2000 | |
| $MN_S^M(2,1,0)$ | $MN_S^M(2,0,0)$ | 2000 | |
| $MN_S^M(1,3,0)$ | $MN_S^M(1,2,0)$ | 2000 | |
| $MN_S^M(2,2,0)$ | $MN_S^M(2,1,0)$ | 2000 | |
| $MN_S^M(3,1,0)$ | $MN_S^M(3,0,0)$ | 2000 | |
| $MN_S^M(2,3,0)$ | $MN_S^M(2,2,0)$ | 2000 | |
| $MN_S^M(3,2,0)$ | $MN_S^M(3,1,0)$ | 2000 | |
| $MN_S^M(3,3,0)$ | $MN_S^M(3,2,0)$ | 2000 | |
| $MN_S^M(1,1,1)$ | $MN_S^M(1,1,0)$ | 1000 | |
| $MN_S^M(1,2,1)$ | $MN_S^M(1,2,0)$ | 1000 | |
| $MN_S^M(2,1,1)$ | $MN_S^M(2,1,0)$ | 1000 | |
| $MN_S^M(1,3,1)$ | $MN_S^M(1,3,0)$ | 1000 | |
| $MN_S^M(2,2,1)$ | $MN_S^M(2,2,0)$ | 1000 | |
| $MN_S^M(3,1,1)$ | $MN_S^M(3,1,0)$ | 1000 | |
| $MN_S^M(2,3,1)$ | $MN_S^M(2,3,0)$ | 1000 | |
| $MN_S^M(3,2,1)$ | $MN_S^M(3,2,0)$ | 1000 | |
| $MN_S^M(3,3,1)$ | $MN_S^M(3,3,0)$ | 1000 | |
| $MN_S^M(1,1,2)$ | $MN_S^M(1,1,1)$ | 1000 | |
| $MN_S^M(1,2,2)$ | $MN_S^M(1,2,1)$ | 1000 | |
| $MN_S^M(2,1,1)$ | $MN_S^M(2,1,0)$ | 1000 | |
| $MN_S^M(1,3,2)$ | $MN_S^M(1,3,1)$ | 1000 | |
| $MN_S^M(2,2,2)$ | $MN_S^M(2,2,1)$ | 1000 | |
| $MN_S^M(3,1,2)$ | $MN_S^M(3,1,1)$ | 1000 | |
| $MN_S^M(2,3,3)$ | $MN_S^M(2,3,2)$ | 1000 | |
| $MN_S^M(3,2,2)$ | $MN_S^M(3,2,1)$ | 1000 | |
| $MN_S^M(3,3,2)$ | $MN_S^M(3,3,1)$ | 1000 | |
| $MN_S^M(1,1,3)$ | $MN_S^M(1,1,2)$ | 1000 | |
| $MN_S^M(1,2,3)$ | $MN_S^M(1,2,2)$ | 1000 | |
| $MN_S^M(2,1,3)$ | $MN_S^M(2,1,2)$ | 1000 | |
| $MN_S^M(1,3,3)$ | $MN_S^M(1,3,2)$ | 1000 | |
| $MN_S^M(2,2,3)$ | $MN_S^M(2,2,2)$ | 1000 | |
| $MN_S^M(3,1,3)$ | $MN_S^M(3,1,2)$ | 1000 | |
| $MN_S^M(2,3,3)$ | $MN_S^M(2,3,2)$ | 1000 | |
| $MN_S^M(3,2,3)$ | $MN_S^M(3,2,2)$ | 1000 | |
| $MN_S^M(3,3,3)$ | $MN_S^M(3,3,2)$ | 1000 | |

*Note that all the networks trained with 10 random seeds number include 2925, 9998, 2025, 2525, 3410, 9899, 5555, 2520, 2828, and 3140.

Table S13. Summary of benchmark models used in this study

| Model Name | Model Description |
| --- | --- |
| $\boldsymbol{LSTM(N_1, 0, 0)}$ | Single-Layer LSTM network with node number $N_1$ varies from 1 to 6 |
| $\boldsymbol{LSTM(N_1, N_1, 0)}$ | Two-Layer LSTM network with node number $N_1$ varies from 1 to 5 |
| $\boldsymbol{LSTM(N_1, N_1, N_1)}$ | Three-Layer LSTM network with node number $N_1$ varies from 1 to 5 |
| $\boldsymbol{MA_{3(SAO)}}$ | Two Cell-state Single Flow path Mass-Conserving-Architecture with Share Augmented Output gate |
| $\boldsymbol{MA_{4(SAO)}}$ | Two Cell-state Two path Mass-Conserving-Architecture with Share Augmented Output gate |
| $\boldsymbol{MA_{5(SAO)}}$ | Three Cell-state Two Flow path Mass-Conserving-Architecture with Share Augmented Output gate |
| $\boldsymbol{MA_{6(SAO)}}$ | Three Cell-state Three Flow path Mass-Conserving-Architecture with Share Augmented Output gate |

*Note that all the networks trained with 10 random seeds number include 2925, 9998, 2025, 2525, 3410, 9899, 5555, 2520, 2828, and 3140

Table S14. $\mathrm{KGE_{ss}}$ Statistics and Numbers of Parameter for the benchmark LSTM network used in this study

| Model Names | $LSTM(1)$ | $LSTM(2)$ | $LSTM(3)$ | $LSTM(4)$ | $LSTM(5)$ | $LSTM(6)$ | $LSTM(1,1,0)$ | $LSTM(2,2,0)$ |
|---|---|---|---|---|---|---|---|---|
| $\mathrm{KGE_{ss}^{min}}$ | 0.41 | 0.68 | 0.66 | 0.69 | 0.66 | 0.55 | 0.64 | 0.70 |
| $\mathrm{KGE_{ss}^{5\%}}$ | 0.46 | 0.72 | 0.75 | 0.76 | 0.78 | 0.68 | 0.72 | 0.75 |
| $\mathrm{KGE_{ss}^{25\%}}$ | 0.74 | 0.83 | 0.84 | 0.84 | 0.85 | 0.81 | 0.84 | 0.86 |
| $\mathrm{KGE_{ss}^{median}}$ | 0.77 | 0.87 | 0.88 | 0.89 | 0.90 | 0.86 | 0.89 | 0.90 |
| $\mathrm{KGE_{ss}^{75\%}}$ | 0.84 | 0.92 | 0.93 | 0.93 | 0.93 | 0.91 | 0.92 | 0.94 |
| $\mathrm{KGE_{ss}^{95\%}}$ | 0.91 | 0.95 | 0.96 | 0.96 | 0.98 | 0.95 | 0.96 | 0.96 |
| Par no. | 16 | 43 | 76 | 117 | 166 | 223 | 32 | 83 |
| Model Names | $LSTM(3,3,0)$ | $LSTM(4,4,0)$ | $LSTM(5,5,0)$ | $LSTM(1,1,1)$ | $LSTM(2,2,2)$ | $LSTM(3,3,3)$ | $LSTM(4,4,4)$ | $LSTM(5,5,5)$ |
| $\mathrm{KGE_{ss}^{min}}$ | 0.76 | 0.71 | 0.66 | 0.70 | 0.62 | 0.62 | 0.59 | 0.73 |
| $\mathrm{KGE_{ss}^{5\%}}$ | 0.76 | 0.72 | 0.66 | 0.72 | 0.67 | 0.69 | 0.66 | 0.78 |
| $\mathrm{KGE_{ss}^{25\%}}$ | 0.83 | 0.82 | 0.80 | 0.82 | 0.82 | 0.82 | 0.87 | 0.88 |
| $\mathrm{KGE_{ss}^{median}}$ | 0.88 | 0.90 | 0.84 | 0.89 | 0.89 | 0.90 | 0.91 | 0.92 |
| $\mathrm{KGE_{ss}^{75\%}}$ | 0.93 | 0.94 | 0.90 | 0.92 | 0.92 | 0.93 | 0.93 | 0.95 |
| $\mathrm{KGE_{ss}^{95\%}}$ | 0.95 | 0.96 | 0.95 | 0.95 | 0.95 | 0.97 | 0.97 | 0.96 |
| Par no. | 148 | 229 | 326 | 48 | 123 | 220 | 341 | 486 |

* Par no.=Number of Parameters; The single-layer $LSTM(N)$ indicates a model with N LSTM cell in a single layer. The multi-layer $LSTM(N_1, N_2, N_3)$ indicates a model with $N_1$, $N_2$, and $N_3$ LSTM cells in the first, second, and third layers, respectively.

Table S15. Summary of the pruned network models in this study

| Model Name | Parent | Effective Cell State | Effective Flow Path | Note |
|---|---|---|---|---|
| $MN_{None}^{DS}(5) - D1(P)$ | $MN_{None}^{DS}(5)$ | 4 | 4 | Pruned (one flow path) four-cell-state case to its parent |
| $MN_{None}^{DS}(5) - D2(P)$ | | 3 | 3 | Pruned (two flow path) three-cell-state case to its parent |
| $MN_{None}^{DS}(5) - D3(P)$ | | 2 | 2 | Pruned (three flow path) two-cell-state case to its parent |
| $MN_{None}^{DS}(5) - D4(P)$ | | 1 | 1 | Pruned (four flow path) one-cell-state case to its parent |
| $MN_{SALO}^{DS}(5) - D1(P)$ | $MN_{SALO}^{DS}(5)$ | 5 | 4 | Pruned (one flow path) five-cell-state case to its parent |
| $MN_{SALO}^{DS}(5) - D2(P)$ | | 5 | 3 | Pruned (two flow path) five-cell-state case to its parent |
| $MN_{SALO}^{DS}(5) - D3(P)$ | | 5 | 2 | Pruned (three flow path) five-cell-state case to its parent |
| $MN_{SALO}^{DS}(5) - D4(P)$ | | 5 | 1 | Pruned (four flow path) five-cell-state case to its parent |
| $MN_{SALO}^{DS}(5) - D1(F)$ | | 4 | 4 | Pruned (one flow path) four-cell-state case to its parent |
| $MN_{SALO}^{DS}(5) - D2(F)$ | | 3 | 3 | Pruned (two flow path) three-cell-state case to its parent |
| $MN_{SALO}^{DS}(5) - D3(F)$ | | 2 | 2 | Pruned (three flow path) two-cell-state case to its parent |
| $MN_{SALO}^{DS}(5) - D4(F)$ | | 1 | 1 | Pruned (four flow path) one-cell-state case to its parent |

*P refer to *"Path"* meaning the number of flow paths is removed. F refer to *"Full"* meaning the number of flow paths and the full effect of cell states are removed. Notice that removing the flow path (P) from $MN_{None}^{DS}(5)$ eliminates the effect of the cell state, as no information is shared through the gate. In contrast, for $MN_{SALO}^{DS}(5)$, the cell state influence remains unless a "Full" removal is performed.

Table S16. Performance percentiles across the 513 CAMELS US basins using the scaled Kling–Gupta Efficiency ($KGE_{ss}$), its three components—$\alpha^{KGE}$ (flow variability), $\beta^{KGE}$ (mass balance), and $r^{KGE}$: (Pearson correlation)—as well as the root-mean-squared error (RMSE) and mean absolute error (MAE). Note that we use $\alpha_*^{KGE}$ and $\beta_*^{KGE}$ instead of the original $\alpha^{KGE}$ and $\beta^{KGE}$ to ensure that the values are upper-bounded by 1.

| Model Name | $MC\{O_\sigma L_{\sigma+}^{con}\}$ | $MN_{None}^{DS}(3)$ | $MN_{SALO}^{DS}(5)$ | $MN_{SALO}^{DS}(5)$ | $\boldsymbol{LSTM(5)}$ | Model Name | $MC\{O_\sigma L_{\sigma+}^{con}\}$ | $MN_{None}^{DS}(3)$ | $MN_{SALO}^{DS}(5)$ | $MN_{SALO}^{DS}(5)$ | $\boldsymbol{LSTM(5)}$ |
|---|---|---|---|---|---|---|---|---|---|---|---|
| Skill Metric | | | $KGE_{ss}$ | | | Skill Metric | | | $\alpha_*^{KGE}$ | | |
| Worst | -4.21 | -3.60 | -6.00 | -0.25 | 0.14 | Worst | -1.04 | -0.97 | -2.25 | -0.69 | 0.17 |
| 5% | 0.26 | 0.28 | -0.26 | 0.55 | 0.69 | 5% | 0.45 | 0.48 | -0.06 | 0.62 | 0.71 |
| 25% | 0.59 | 0.65 | 0.63 | 0.78 | 0.83 | 25% | 0.83 | 0.85 | 0.79 | 0.88 | 0.89 |
| 50% | 0.72 | 0.78 | 0.81 | 0.86 | 0.89 | 50% | 0.93 | 0.93 | 0.92 | 0.95 | 0.95 |
| 75% | 0.80 | 0.86 | 0.88 | 0.91 | 0.93 | 75% | 0.97 | 0.97 | 0.97 | 0.98 | 0.98 |
| 95% | 0.89 | 0.92 | 0.95 | 0.95 | 0.97 | 95% | 0.99 | 0.99 | 0.99 | 1.00 | 1.00 |
| Skill Metric | | | $RMSE$ | | | Skill Metric | | | $\beta_*^{KGE}$ | | |
| Worst | 12.60 | 12.45 | 13.28 | 13.04 | 8.87 | Worst | -6.12 | -5.24 | -8.62 | -0.22 | 0.73 |
| 5% | 5.55 | 5.12 | 5.44 | 4.60 | 3.51 | 5% | 0.63 | 0.64 | 0.17 | 0.86 | 0.89 |
| 25% | 2.74 | 2.39 | 2.46 | 2.06 | 1.72 | 25% | 0.90 | 0.92 | 0.91 | 0.96 | 0.96 |
| 50% | 1.88 | 1.62 | 1.62 | 1.31 | 1.12 | 50% | 0.96 | 0.97 | 0.97 | 0.98 | 0.98 |
| 75% | 1.14 | 0.96 | 0.95 | 0.73 | 0.61 | 75% | 0.98 | 0.99 | 0.99 | 0.99 | 0.99 |
| 95% | 0.26 | 0.21 | 0.15 | 0.14 | 0.12 | 95% | 1.00 | 1.00 | 1.00 | 1.00 | 1.00 |
| Skill Metric | | | $MAE$ | | | Skill Metric | | | $r^{KGE}$ | | |
| Worst | 6.41 | 6.46 | 6.84 | 6.57 | 4.14 | Worst | -0.07 | -0.06 | -0.27 | 0.27 | 0.15 |
| 5% | 2.97 | 2.57 | 2.67 | 1.85 | 1.53 | 5% | 0.15 | 0.16 | 0.17 | 0.52 | 0.68 |
| 25% | 1.19 | 1.03 | 0.98 | 0.77 | 0.65 | 25% | 0.47 | 0.55 | 0.60 | 0.74 | 0.81 |
| 50% | 0.78 | 0.61 | 0.58 | 0.49 | 0.42 | 50% | 0.64 | 0.72 | 0.77 | 0.82 | 0.87 |
| 75% | 0.46 | 0.38 | 0.34 | 0.27 | 0.23 | 75% | 0.76 | 0.82 | 0.86 | 0.89 | 0.92 |
| 95% | 0.07 | 0.06 | 0.05 | 0.05 | 0.04 | 95% | 0.87 | 0.90 | 0.93 | 0.94 | 0.96 |

Table S17. Background Information on the selected 11473900 catchment

| Catchment ID | Name | Longitude | Latitude | Area | Category report in Jiang et al. (2022) |
|---|---|---|---|---|---|
| 11473900 | Middle Fork Eel River, near Dos Rios, CA | -123.33 | 39.71 | 1925.01 | Recent rainfall-dominated |

Table S18. Skill Metrics during testing period for the selected 11473900 catchment

| Model Architecture | $KGE_{ss}$ | $\alpha^{KGE}$ | $\beta^{KGE}$ | $r^{KGE}$ | $RMSE$ | $MAE$ |
|---|---|---|---|---|---|---|
| $MC\{O_{\sigma}L_{\sigma+}^{con}\}$ | 0.77 | 1.03 | 1.11 | 0.69 | 3.50 | 1.36 |
| $MN_{None}^{DS}(3)$ | 0.86 | 0.94 | 1.08 | 0.82 | 2.56 | 0.97 |
| $MN_{SALO}^{DS}(5)$ | 0.77 | 1.05 | 1.03 | 0.68 | 3.60 | 1.41 |
| $MN_{SALO}^{DS-R_{con}}(5)$ | 0.88 | 0.93 | 1.01 | 0.85 | 2.34 | 0.79 |

**Mass-Conserving-Perceptron (MCP)**

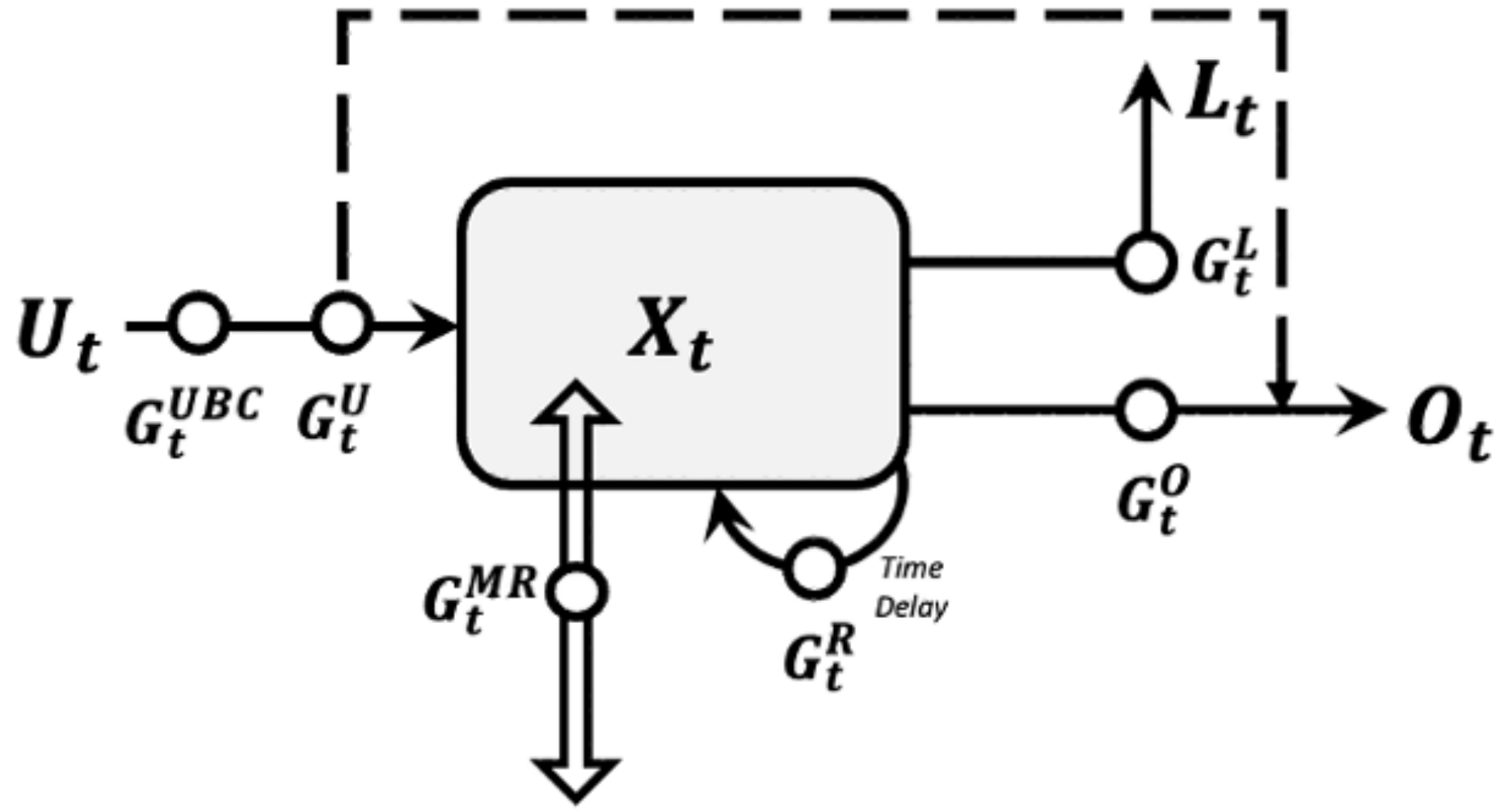

$$ib[\mathrm{t}] = G_t^{UBC} = 1.0$$
$$mr[\mathrm{t}] = G_t^{MR} = 0.0$$
$$i[\mathrm{t}] = G_t^{U} = 1.0$$
$$f[\mathrm{t}] = G_t^{R} = 1.0 - G_t^{O} - G_t^{L}$$
$$g[t] = U_t$$
$$o[t] = G_t^{O} = \kappa_O \odot \sigma(b_o c[t] + a_o)$$
$$l[t] = G_t^{L} = \kappa_L \odot \sigma(b_L PET_t + b_L^* c[t] + a_L)$$
$$c[t+1] = X_{t+1} = f[t] \odot X_t + i[t] \odot g[t]$$
$$\mathrm{O}[t] = o[t] \odot X_t$$

**Figure S1:** Directed-graph representation and mathematical formulation of the mass-conserving-perceptron (MCP) modeling network used in this study. As shown in Figure S1 of the supplementary materials, the input, forget, and output gates used in recurrent neural networks such as the LSTM— $i[t]$,$f[t]$, $o[t]$ — correspond functionally to $G_t^U$, $G_t^R$, $G_t^O$ respectively. The $l[t]$ is the newly introduced loss gate denoted as $G_t^L$. Similarly, ib$[t]$ and $mr[t]$ are the Input-bypass gate and the mass-relaxation gate denote as $G_t^{UBC}$ and $G_t^{MR}$ as proposed in *Wang & Gupta (2024a)*. These two gates are not activated in this study and are set to 1.0 and 0.0, respectively. The internal state variable $X[t]$ is analogous to the LSTM cell state $c[t]$. Similarly, the external input $u[t]$ aligns with the LSTM input $x[t]$, while the internal update $g[t]$ resembles the candidate state update. The final output $O[t]$ — corresponding to $h[t]$ in the LSTM — also serves a functionally equivalent role. Note that the remaining symbols—$\kappa_o, a_o, b_o, a_L, b_L, b_L^*, \kappa_L$—represent trainable parameters associated with the model architecture, while $\odot$ indicates element-wise multiplication.

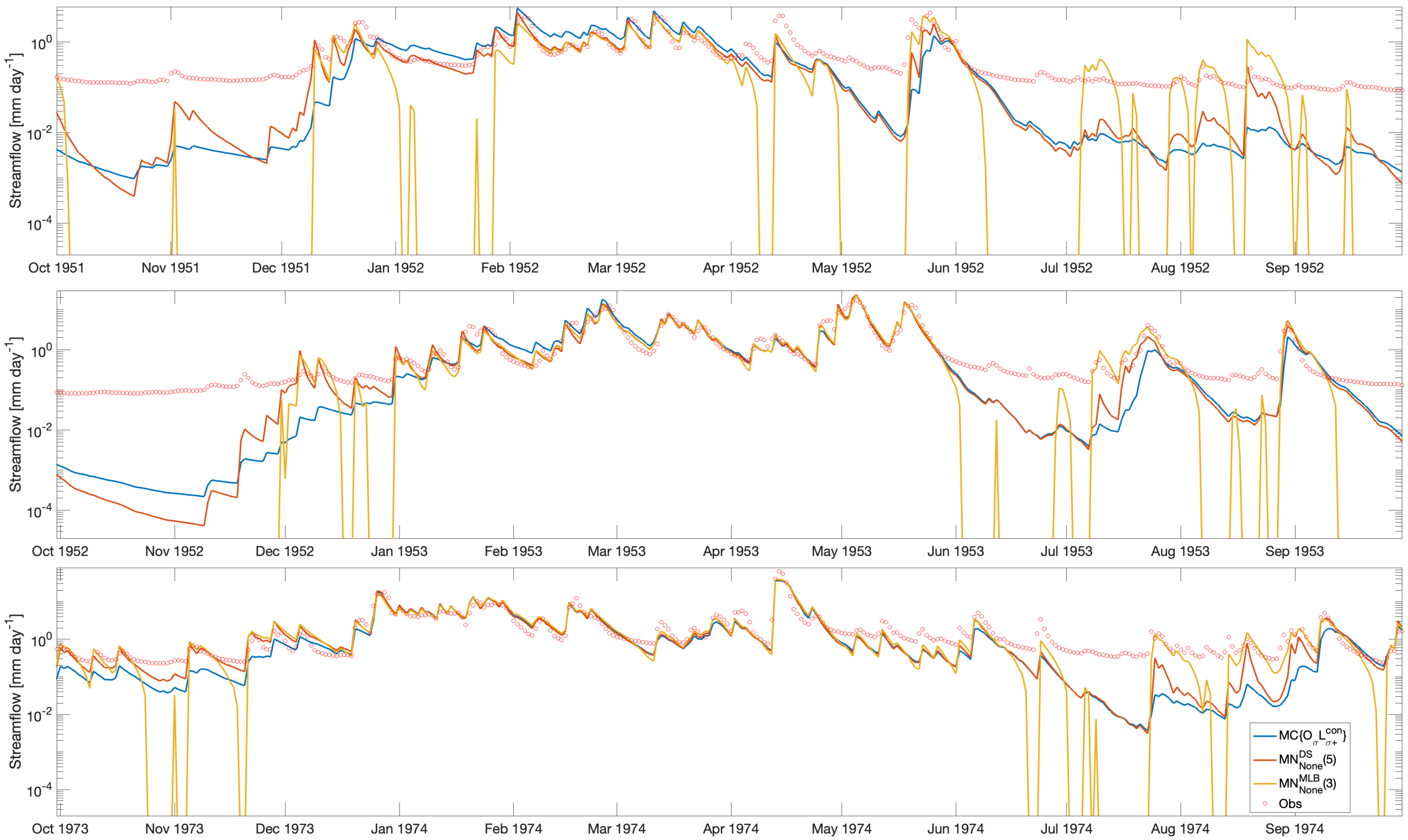

**Figure S2:** Comparison of selected single-layer Mass-Conserving Network performance using hydrographs in log-scale for selected dry, median, and wet years in subplots (a) to (c). *Notice that the results are presented in water year (WY) including dry year WY1952, median year WY1953, and wet year WY1974. The dry, median, and wet years are determined by ranking the annual flow peaks over a 40-year period and selecting the years with the smallest, the 20th highest, and the largest peaks, respectively.

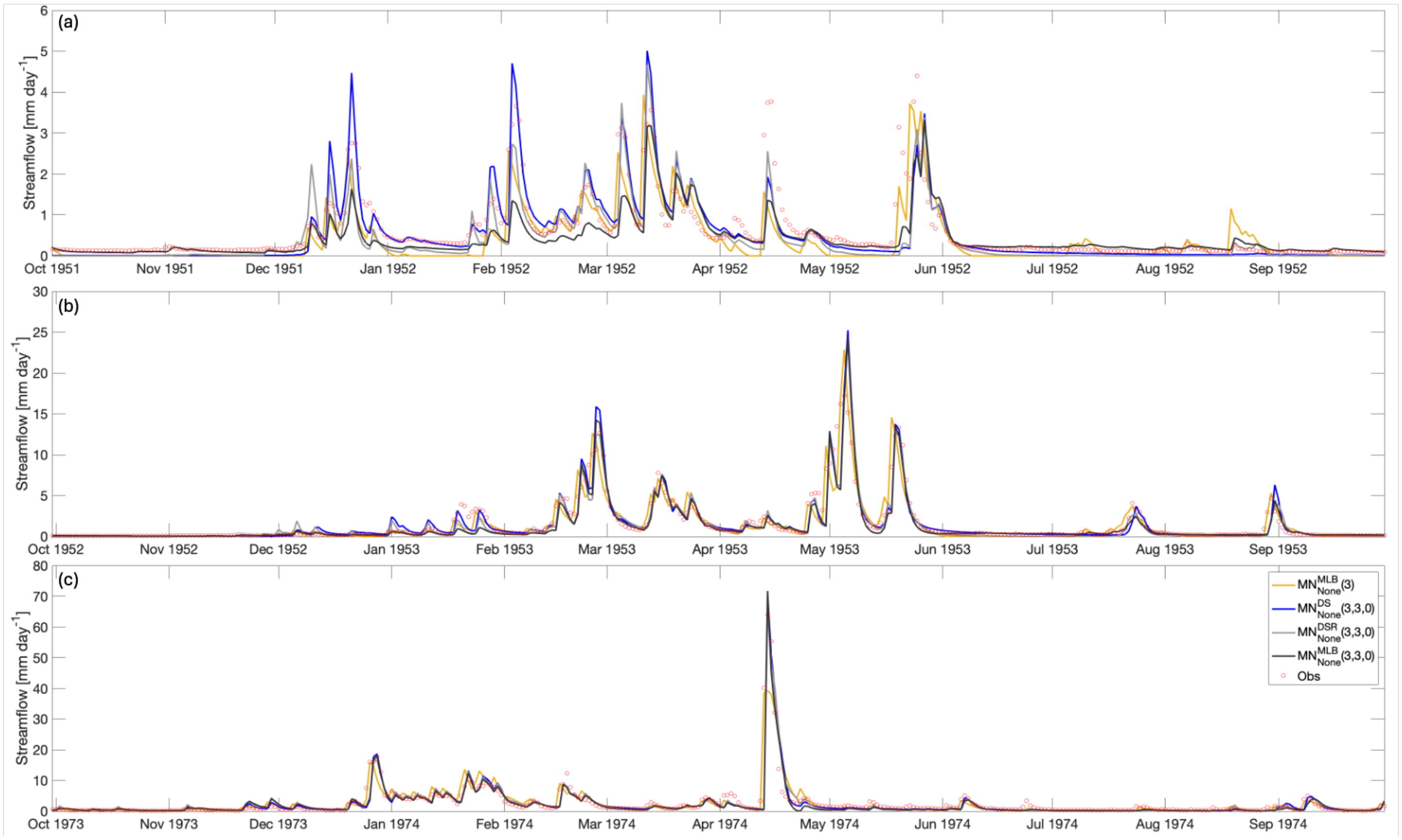

**Figure S3:** Comparison of selected single-layer Mass-Conserving Network performance using the hydrographs for selected dry, median, and wet years in subplots (a) to (c). *Notice that the results are presented in water year (WY) including dry year WY1952, median year WY1953, and wet year WY1974. The dry, median, and wet years are determined by ranking the annual flow peaks over a 40-year period and selecting the years with the smallest, the 20th highest, and the largest peaks, respectively. The y-axes differ significantly between plots due to variations in the maximum values for each water year.

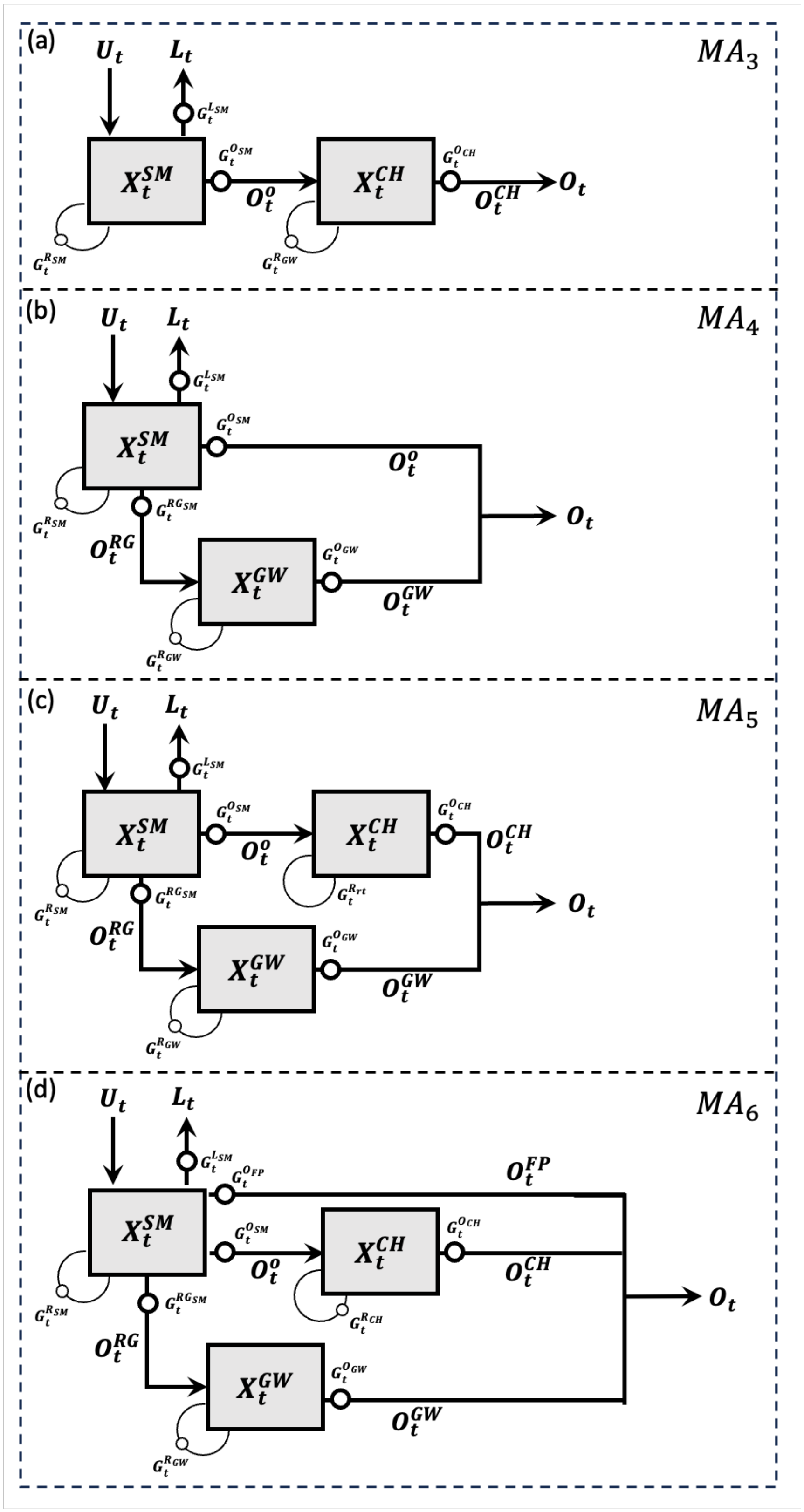

**Figure S4:** Mass-Conserving Architecture (MA) proposed by Wang & Gupta (2024b), including: (a) one-flow-path, two-state MA3 hypothesis; (b) two-flow-path, two-state MA4 hypothesis; (c) two-flow-path, three-state MA5 hypothesis; and (d) three-flow-path, three-state MA6 hypothesis. $U_t$: observed (precipitation) input; $O_t$: observed (streamflow) output; $L_t$: unobserved (evapotranspirative) output loss. $X_t^{SM}$: soil-moisture state in the Soil-Moisture (SM) tank. $G_t^{O_{SM}}$: output gate, $G_t^{L_{SM}}$: loss gate, $G_t^{R_{SM}}$: remember gate; $G_t^{RG_{SM}}$: recharge gate, $G_t^{O_{FP}}$: Quick-flow Path output in the SM tank. $X_t^{CH}$: channel routing state in the Channel Routing (CH) tank, $G_t^{O_{CH}}$: output gate; $G_t^{R_{SM}}$: remember gate in the CH tank. $X_t^{GW}$: groundwater state in the Ground Water (GW) tank, $G_t^{O_{GW}}$: output gate; $G_t^{R_{GW}}$: remember gate in the GW tank. $O_t^{O}$: Output gate flux in soil-moisture tank; $O_t^{RG}$: Recharge Gate flux; $O_t^{FP}$: Quick-Flow Path Gate flux; $O_t^{CH}$: Output gate flux in routing tank; $O_t^{GW}$: Output gate flux in groundwater tank

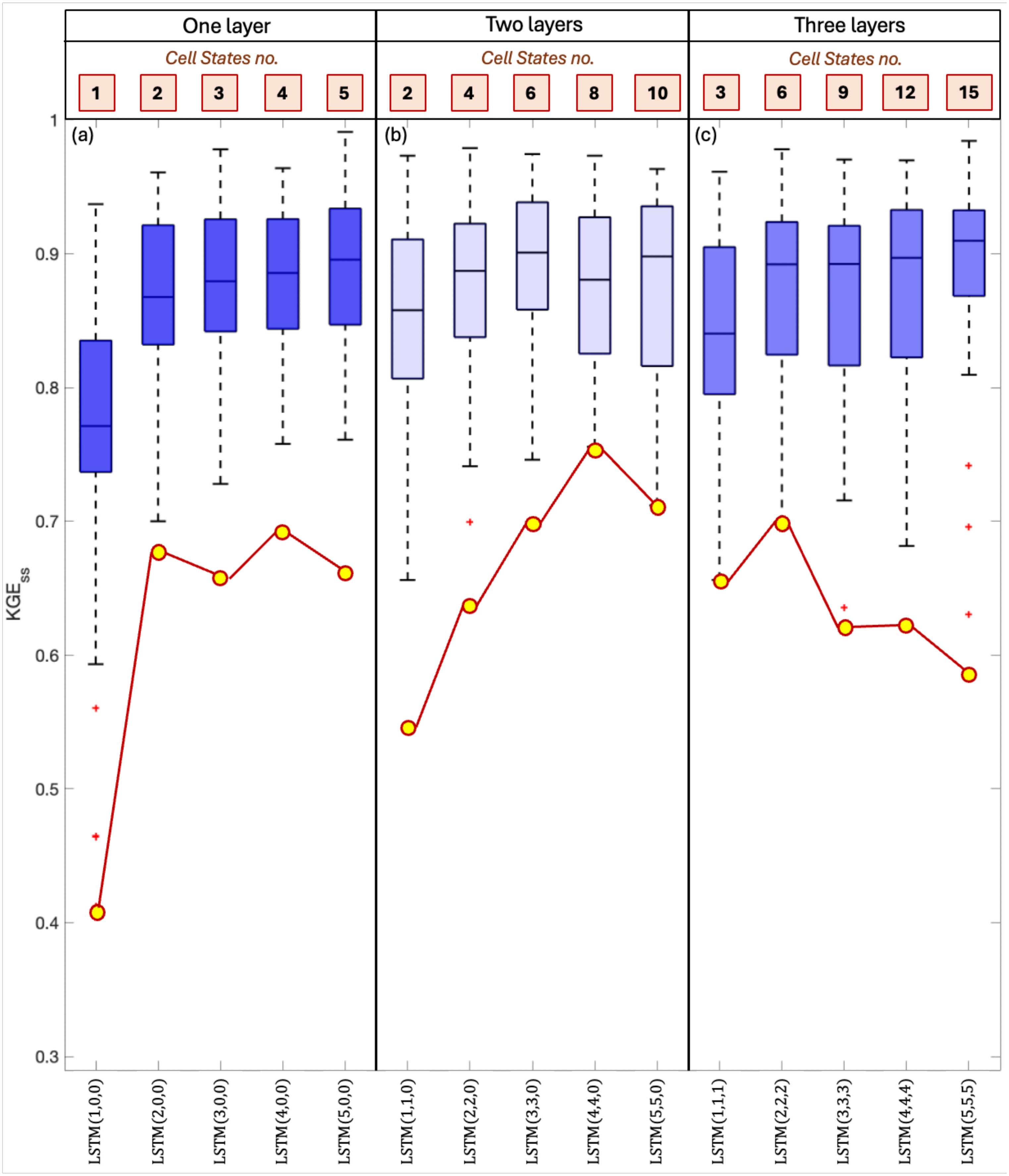

**Figure S5:** Box and whisker plots of the distributions of annual $\mathrm{KGE_{ss}}$ scores, comparing the (a) 1-layer, (b) 2-layer, and (c) 3-layer long short-term memory (LSTM) networks, with the number of nodes varying from 1 to 5. The boxplots are created based on 38 years of data, as the LSTM networks used in the experiments employ a maximum sequence length of 390 days. LSTM(5,0,0), LSTM(5,5,0), and LSTM(5,5,5) are shown in Figure 10. Yellow-Filled Red Circles: The worst-year (among all 38 years) $\mathrm{KGE_{ss}}$ score derived from each single-layer MN networks.

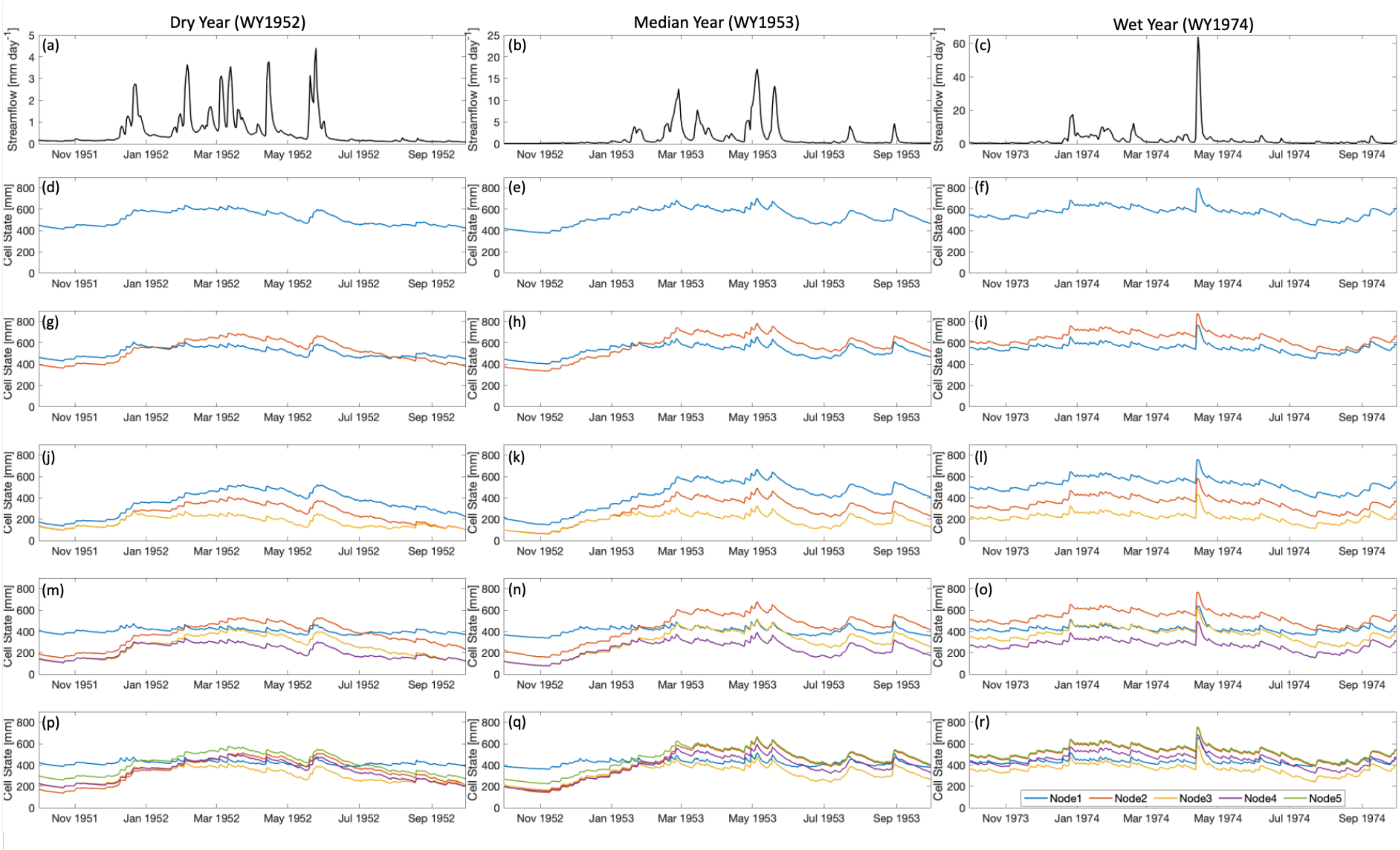

**Figure S6:** Time series plots of streamflow observations for dry, median, and wet years are shown in subplots (a) to (c). The cell states for the distributed case (DS) are presented with 1-node in subplots (d) to (f) (dry, median, and wet years), 2-node in subplots (g) to (i), 3-node in subplots (j) to (l), 4-node in subplots (m) to (o), and 5-node in subplots (p) to (r). The nodes are ordered according to the magnitude of the peak values (from small to large) of the output gate, as shown in Figure 11 of the main text.

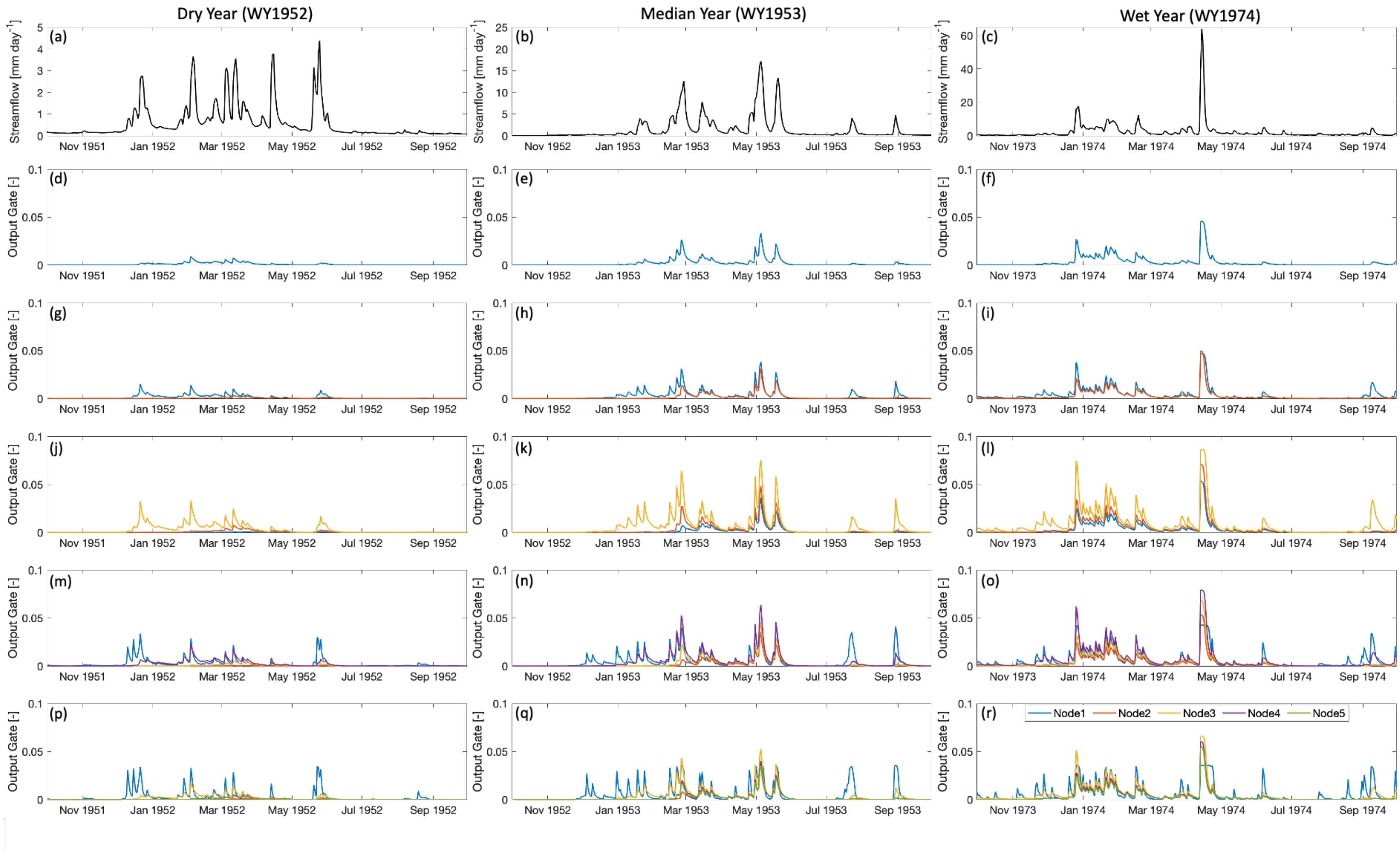

**Figure S7:** Time series plots of streamflow observations for dry, median, and wet years are shown in subplots (a) to (c). The output gate series for the distributed case (DS) are presented with 1-node in subplots (d) to (f) (dry, median, and wet years), 2-node in subplots (g) to (i), 3-node in subplots (j) to (l), 4-node in subplots (m) to (o), and 5-node in subplots (p) to (r). The output gate depends on both the cell state and the potential evapotranspiration (PET) at the current timestep. The output gate series value is passed through a sigmoid activation function, which squashes the output gate value into a range between 0 and 1. The nodes are ordered according to the magnitude of the peak values (from small to large) of the output gate, as shown in Figure 11 of the main text.

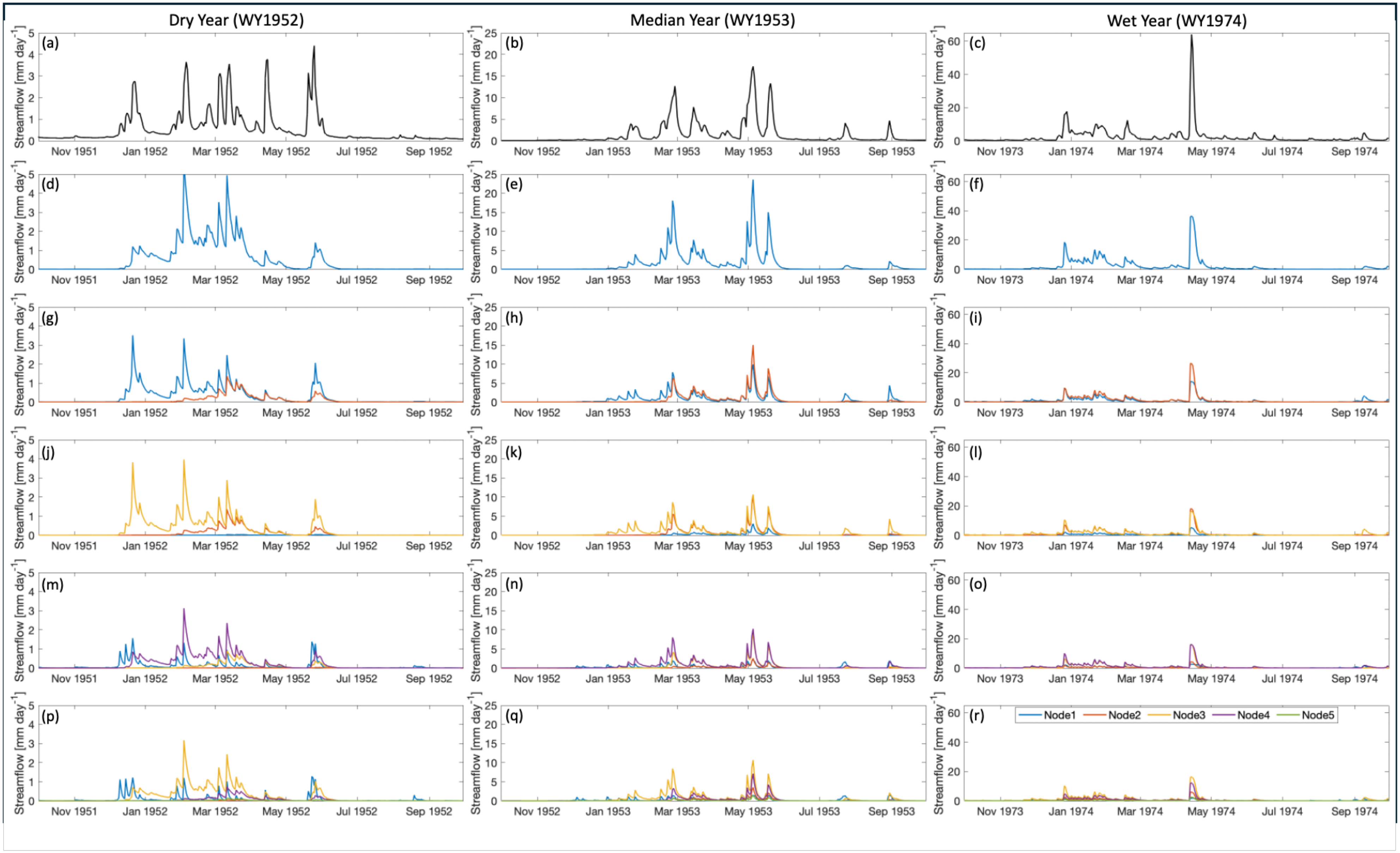

**Figure S8:** Time series plots of streamflow observations for dry, median, and wet years are shown in subplots (a) to (c). The streamflow components from each node for the distributed case (DS) are presented with 1-node in subplots (d) to (f) (dry, median, and wet years), 2-node in subplots (g) to (i), 3-node in subplots (j) to (l), 4-node in subplots (m) to (o), and 5-node in subplots (p) to (r). The y-axes are not consistent across different columns because the wetness conditions vary by year, making it more effective to use different y-axis ranges for visualization. Note that the streamflow components from each node are obtained by multiplying the node's output with its associated output weight. The nodes are ordered according to the magnitude of the peak values (from small to large) of the output gate, as shown in Figure 11 of the main text.

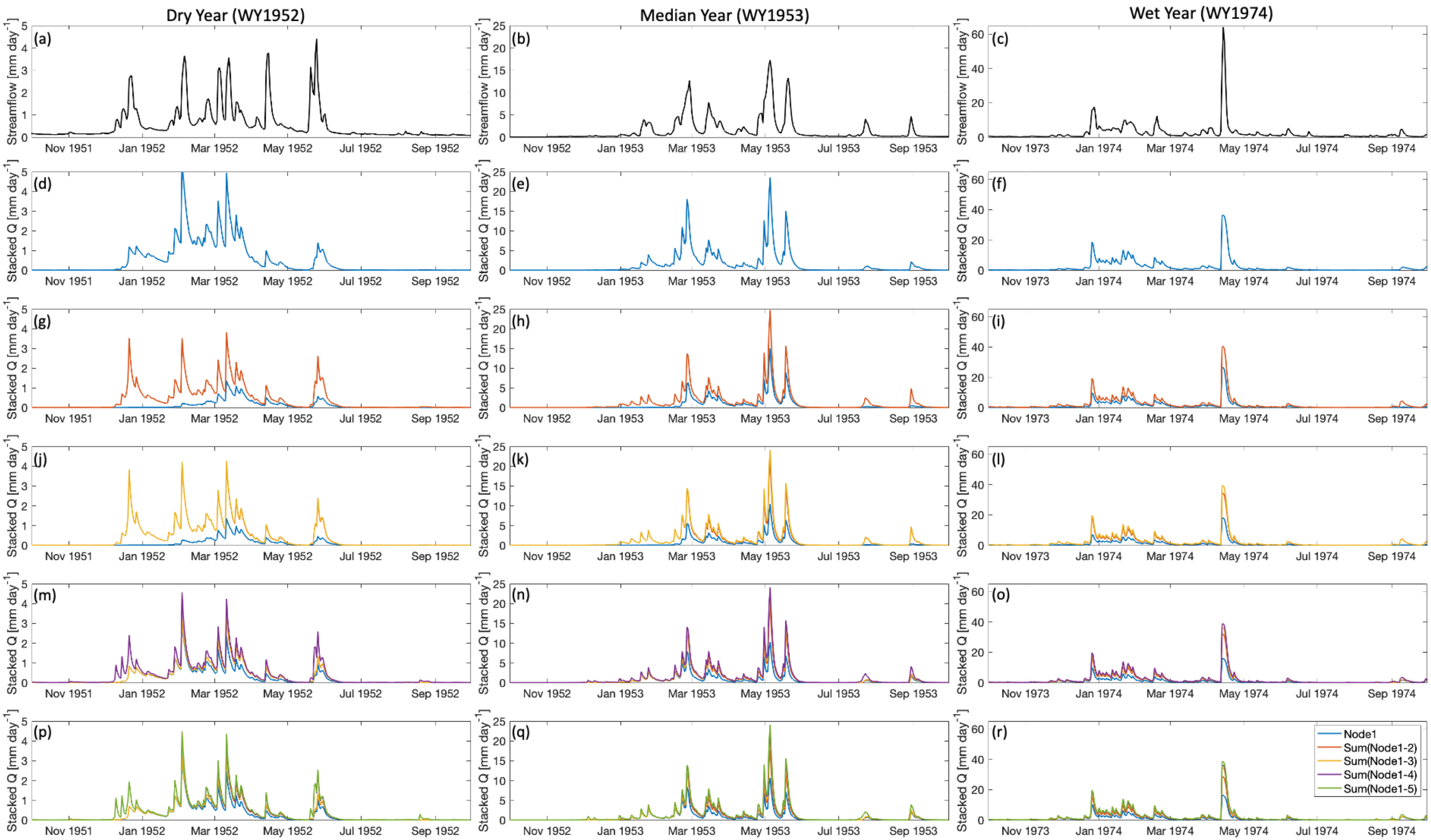

**Figure S9:** Time series plots of streamflow observations for dry, median, and wet years are shown in subplots (a) to (c). The cumulative streamflow components for the distributed case (DS) are presented with 1-node in subplots (d) to (f) (dry, median, and wet years), 2-node in subplots (g) to (i), 3-node in subplots (j) to (l), 4-node in subplots (m) to (o), and 5-node in subplots (p) to (r). Q=streamflow. The y-axes are not consistent across different columns because the wetness conditions vary by year, making it more effective to use different y-axis ranges for visualization. The nodes are ordered according to the magnitude of the peak values (from small to large) of the output gate, as shown in Figure 11 of the main text.

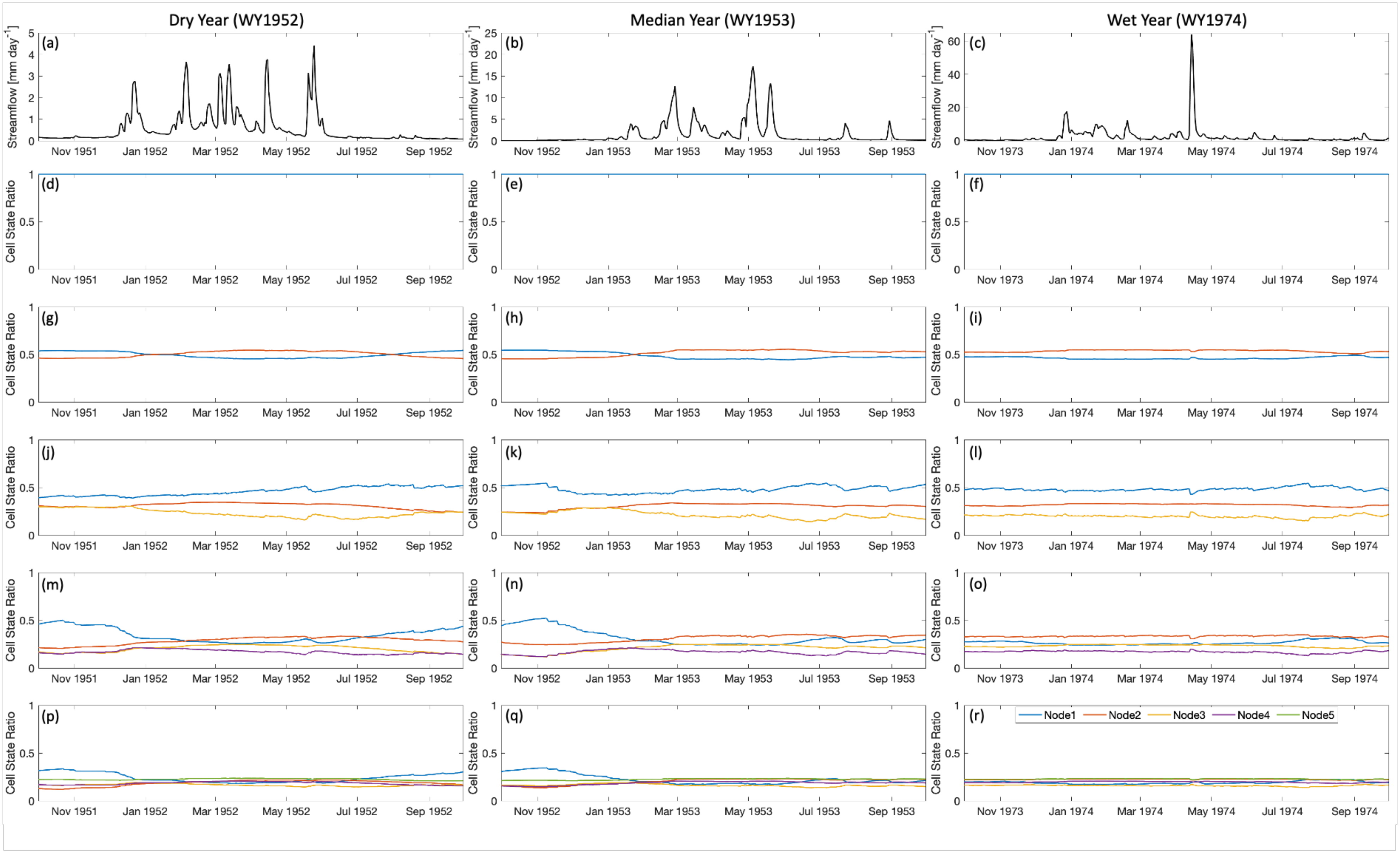

**Figure S10:** Time series plots of streamflow observations for dry, median, and wet years are shown in subplots (a) to (c). The cell states ratio for the distributed case (DS) are presented with 1-node in subplots (d) to (f) (dry, median, and wet years), 2-node in subplots (g) to (i), 3-node in subplots (j) to (l), 4-node in subplots (m) to (o), and 5-node in subplots (p) to (r). The nodes are ordered according to the magnitude of the peak values (from small to large) of the output gate, as shown in Figure 11 of the main text. Note that The large differences in cell state ratios when the total streamflow is low are likely, to some extent, caused by an unresolvable numerical issue—where a small streamflow value (derived from the product of the output gate and cell state) can result from either a very small or a very large value in one component while being compensated by the other.

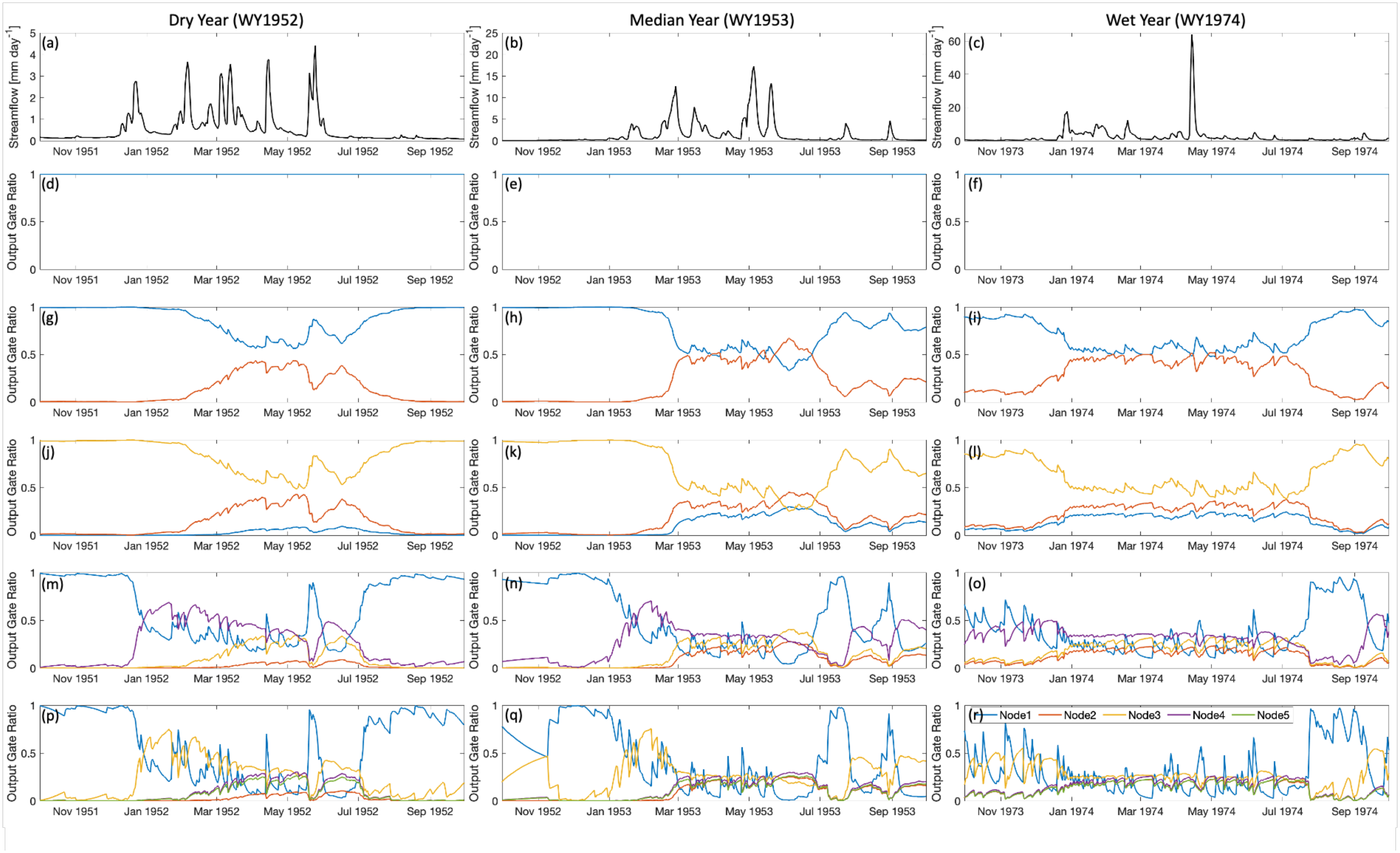

**Figure S11:** Time series plots of streamflow observations for dry, median, and wet years are shown in subplots (a) to (c). The ratio of output gate series for the distributed case (DS) are presented with 1-node in subplots (d) to (f) (dry, median, and wet years), 2-node in subplots (g) to (i), 3-node in subplots (j) to (l), 4-node in subplots (m) to (o), and 5-node in subplots (p) to (r). The nodes are ordered according to the magnitude of the peak values (from small to large) of the output gate, as shown in Figure 11 of the main text. The output gate series ratio indicates the extent to which each flow path is activated and always ranges between 0 and 1.

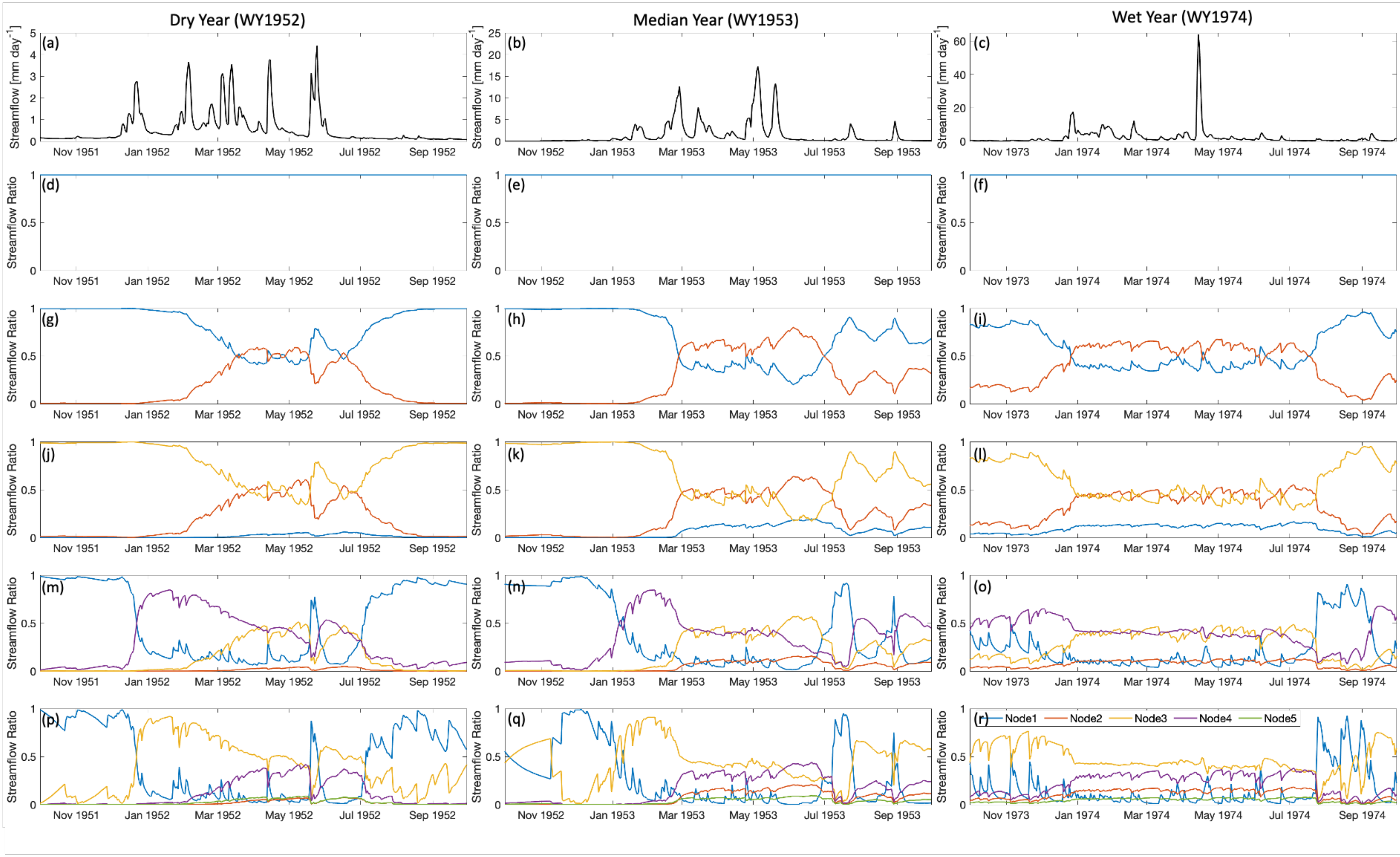

**Figure S12:** Time series plots of streamflow observations for dry, median, and wet years are shown in subplots (a) to (c). The ratio of streamflow components for the distributed case (DS) are presented with 1-node in subplots (d) to (f) (dry, median, and wet years), 2-node in subplots (g) to (i), 3-node in subplots (j) to (l), 4-node in subplots (m) to (o), and 5-node in subplots (p) to (r). The nodes are ordered according to the magnitude of the peak values (from small to large) of the output gate, as shown in Figure 11 of the main text. The streamflow ratio indicates the proportion of the total water that comes out of a flow path and always ranges between 0 and 1.

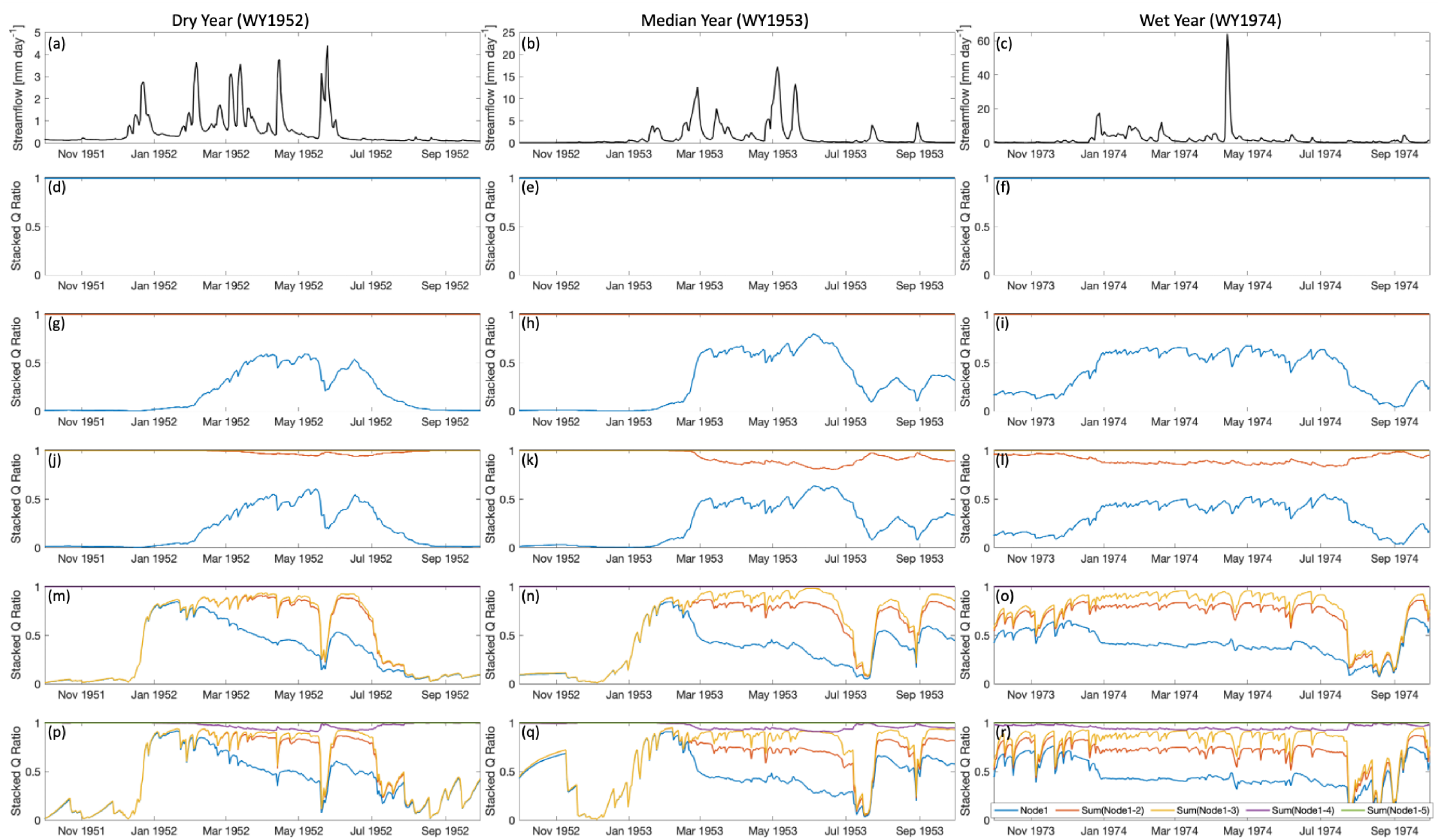

**Figure S13:** Time series plots of streamflow observations for dry, median, and wet years are shown in subplots (a) to (c). The ratio of cumulative streamflow components for the distributed case (DS) are presented with 1-node in subplots (d) to (f) (dry, median, and wet years), 2-node in subplots (g) to (i), 3-node in subplots (j) to (l), 4-node in subplots (m) to (o), and 5-node in subplots (p) to (r). Q=streamflow. The nodes are ordered according to the magnitude of the peak values (from small to large) of the output gate, as shown in Figure 11 of the main text. The stacked Q (streamflow) ratio indicates the cumulative proportion of the total water discharged from a flow path and always ranges between 0 and 1.